%% file: ms.tex
\documentclass[a4paper]{article} 

\usepackage[english]{babel} 
\usepackage[T1]{fontenc} 
\usepackage[utf8]{inputenc} 
\usepackage[babel]{microtype} 

\usepackage[margin={0.85in, 0.85in}]{geometry}

\usepackage{graphicx}
\usepackage{booktabs} 
\graphicspath{{Figures/}}
\usepackage{subfiles}
\usepackage{wrapfig}
\usepackage{amsmath,amssymb,amsfonts}
\usepackage{bbm}
\usepackage{mathtools}
\usepackage{subcaption}
\usepackage{multirow}
\usepackage{color,soul}
\usepackage{xcolor}
\setulcolor{red}
\usepackage [space]{cite}
\usepackage{amsthm}

\usepackage{tabularx}
\newcolumntype{C}{>{\centering\arraybackslash}X}

\usepackage{booktabs} 
\usepackage{diagbox}
\input{math_commands.tex}

\newcommand{\indep}{\perp \!\!\! \perp}
\newcommand{\norm}[1]{\left\lVert#1\right\rVert}
\newtheorem{theorem}{Theorem}
\usepackage{hyperref}

\newcommand{\email}[1]{\href{mailto:#1}{\nolinkurl{#1}}}

\providecommand{\keywords}[1]{\textbf{\textit{Index terms---}} #1}

\usepackage{authblk}

\date{}
\title{Disparity Between Batches \\ as a Signal for Early Stopping}
\author{Mahsa Forouzesh}
\author{Patrick Thiran }

\affil{Information and Network Dynamics Group (INDY) \protect\\
	School of Computer and Communication Sciences (IC) \protect \\
	\'{E}cole Polytechnique F\'{e}d\'{e}rale de Lausanne \protect\\
	Switzerland \protect\\
	\email{firstname.lastname@epfl.ch}}

\begin{document}

\maketitle              

\begin{abstract}
We propose a metric for evaluating the generalization ability of deep neural networks trained with mini-batch gradient descent. Our metric, called \emph{gradient disparity}, is the $\ell_2$ norm distance between the gradient vectors of two mini-batches drawn from the training set. It is derived from a probabilistic upper bound on the difference between the classification errors over a given mini-batch, when the network is trained on this mini-batch and when the network is trained on another mini-batch of points sampled from the same dataset. We empirically show that gradient disparity is a very promising early-stopping criterion (i) when data is limited, as it uses all the samples for training and (ii) when available data has noisy labels, as it signals overfitting better than the validation data. 
Furthermore, we show in a wide range of experimental settings that gradient disparity is strongly related to the generalization error between the training and test sets, and that it is also very informative about the level of label noise.  

\end{abstract}

\input{Introduction.tex}

\input{relatedwork.tex}

\input{gen_pen.tex}

\input{bound.tex}

\input{gradient_disparity.tex}

\input{experimental_results.tex}

\clearpage
\section*{Acknowledgments}
This work has been accepted at the ECML PKDD 2021 conference (\url{https://2021.ecmlpkdd.org/wp-content/uploads/2021/07/sub_1075.pdf}). We would like to thank anonymous reviewers and Tianzong Zhang for their helpful comments. 

\bibliography{main}

\newpage
\appendix
\input{proof.tex}

\input{details.tex}
\input{kfoldapp.tex}

\input{add.tex}

\input{beyond.tex}

\input{inner_product.tex}

\end{document}

%% file: math_commands.tex
\usepackage{amsmath,amsfonts,bm}

\def\eqref#1{equation~\ref{#1}}

\def\1{\bm{1}}

\DeclareMathAlphabet{\mathsfit}{\encodingdefault}{\sfdefault}{m}{sl}
\SetMathAlphabet{\mathsfit}{bold}{\encodingdefault}{\sfdefault}{bx}{n}



%% file: gradient disparity arxiv/introduction.tex
	\section{Introduction}

	Early-stopping using a separate validation set is one of the most popular techniques used to avoid under/over fitting deep neural networks trained with iterative methods, such as gradient descent \cite{prechelt1998early,yao2007early,gu2018recent}. The optimization is stopped when the performance of the model on a validation set starts to diverge from its performance on the training set. Early stopping requires an accurately labeled validation set, separated from the training set, to act as an unbiased proxy on the unseen test error.
	Obtaining such a reliable validation set can be expensive in many real-world applications as data collection is a time-consuming process that might require domain expertise.
	Furthermore, deep learning is becoming popular in applications for which there is simply not enough available data \cite{roh2019survey,ipeirotis2010quality}. Finally, inexperienced label collectors, complex tasks (e.g., distinguishing a guinea pig from a hamster),
	and corrupted labels due for instance to adversarial attacks result in datasets that contain noisy labels \cite{frenay2013classification}. Deep neural networks have the unfortunate ability to overfit to such small and/or noisy labeled datasets, an issue that cannot be completely solved by popular regularization techniques \cite{zhang2016understanding}. 
	A signal of overfitting during training is therefore particularly useful, if it does \emph{not} need a separate, accurately labeled validation set, which is the purpose of this paper.

	\begin{wrapfigure}[23]{R}{0.5\textwidth}
		\centering
		\vspace*{1em}
		\includegraphics[width=0.5\textwidth, height=0.283\textwidth]{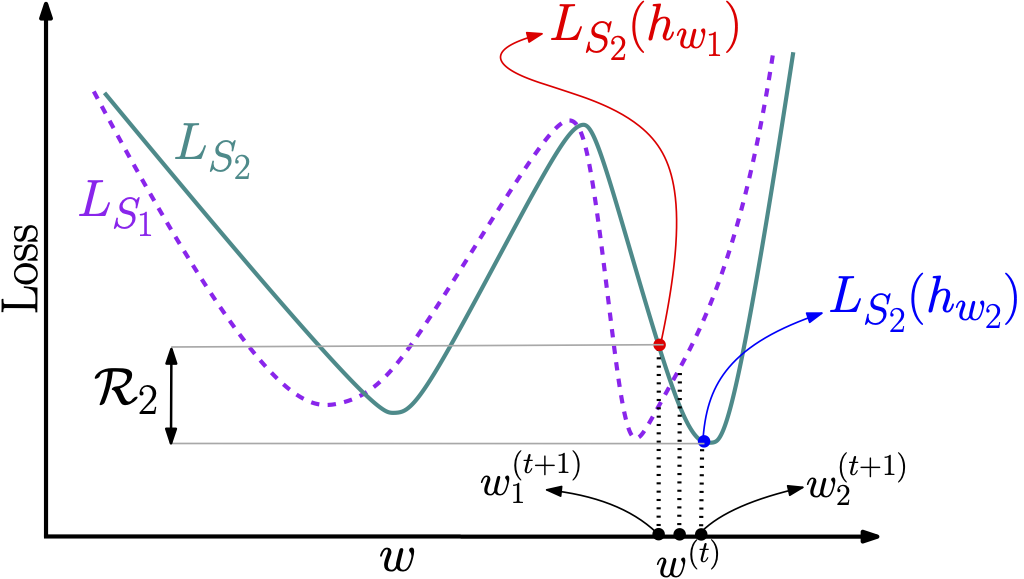}
		\caption{An illustration of the penalty term $\mathcal{R}_2$, where the y-axis is the loss, and the x-axis indicates the parameter of the model. $L_{S_1}$ and $L_{S_2}$ are the average losses over mini-batches $S_1$ and $S_2$, respectively. $w^{(t)}$ is the parameter at iteration $t$ and $w_i^{(t+1)}$ is the parameter at iteration $t+1$ if batch $S_i$ was selected for the update step at iteration $t$, with $i\in\{1,2\}$. }\label{fig:f1}
	\end{wrapfigure}

	Let $S_1$ and $S_2$ be two mini-batches of points sampled from the available (training) dataset. Suppose that $S_1$ is selected for an iteration (step) of the mini-batch gradient descent (SGD), at the end of which the parameter vector is updated to $w_1$. 
	The average loss over $S_1$ (denoted by $L_{S_1}(h_{w_1})$) 
	is in principle reduced, given a sufficiently small learning rate. The average loss $L_{S_2}(h_{w_1})$ over the other mini-batch $S_2$ is not as likely to be reduced. It is more likely to remain larger than the loss $L_{S_2}(h_{w_2})$ computed over $S_2$, if it was $S_2$ instead of $S_1$ that had been selected for this iteration.
	The difference $\mathcal{R}_2 = L_{S_2}(h_{w_1}) - L_{S_2}(h_{w_2})$ 
	is the penalty that we pay for choosing $S_1$ over $S_2$ (and similarly, $\mathcal{R}_1$ is the penalty that we would pay for choosing $S_2$ over $S_1$). $\mathcal{R}_2$ is illustrated in Fig. \ref{fig:f1} for a hypothetical non-convex loss as a function of a one dimensional parameter $w$. 
	The expected penalty measures how much, in an iteration, a model updated on one mini-batch ($S_1$) is able to generalize on average to another mini-batch ($S_2$) from the dataset. Hence, we call $\mathcal{R}$ the \emph{generalization penalty}.

	We establish a probabilistic upper bound on the sum of the expected penalties $\mathbb{E}\left[\mathcal{R}_1\right]+\mathbb{E}\left[\mathcal{R}_2\right]$ by adapting the PAC-Bayesian framework \cite{mcallester1999pac,mcallester1999some,mcallester2003simplified}, given a pair of mini-batches $S_1$ and $S_2$ sampled from the dataset (Theorem \ref{THM1}). 
	Interestingly, under some mild assumptions, this upper bound is essentially a simple expression driven by $\norm{g_1 - g_2}_2$, where $g_1$ and $g_2$ are the gradient vectors over the two mini-batches $S_1$ and $S_2$, respectively. We call it \emph{gradient disparity}: it measures how much a small gradient step on one mini-batch negatively affects the performance on the other one.
	
	\begin{table*}[t]
		\centering
		\caption{
			Test loss and area under the receiver operating characteristic curve (AUC score) of the MRNet dataset \cite{bien2018deep} when using 5-fold cross-validation (5-fold CV) and gradient disparity (GD) as early stopping criteria for detecting the presence of abnormally, ACL tears, and meniscal tears from the sagittal plane MRI scans. 
			The corresponding curves during training are shown in Fig.~\ref{fig:mrnet} (see Appendix~\ref{app:mrnet} for more details). 
			The results of early stopping are given, both when the metric (GD or validation loss) has increased for 5 epochs from the beginning of training and between parenthesis when the metric has increased for 5 consecutive epochs. Using GD outperforms 5-fold CV with either choice of the early stopping threshold. The standard deviations are obtained from 5 runs. }\label{tab:mrnet}
		\vspace{1em}
		\begin{tabular}{c|c|c|c}
			\toprule
			Task & Method    & Test Loss & Test AUC Score (in percentage)  \\ 
			
			\midrule
			\multirow{2}{*}{Abnormal}  & 5-fold CV  & $0.284_{\pm 0.016} (0.307_{\pm 0.057})$ & $71.016_{\pm 3.66}  (87.44_{\pm 1.35})$  \\
			& GD          & $\mathbf{0.274}_{\pm 0.004} (\mathbf{0.275}_{\pm 0.053})$ & $\mathbf{72.67}_{\pm 3.85} (\mathbf{88.12}_{\pm 0.35})$ \\
			\midrule
			\multirow{2}{*}{ACL}& 5-fold CV  & $0.973_{\pm 0.111} (1.246_{\pm 0.142})$ & $79.80_{\pm 1.23}  (89.32_{\pm 1.47})$  \\
			& GD          & $\mathbf{0.842}_{\pm 0.101} (\mathbf{1.136}_{\pm 0.121})$ & $\mathbf{81.81}_{\pm 1.64} (\mathbf{91.52}_{\pm 0.09})$ \\
			\midrule
			\multirow{2}{*}{Meniscal} & 5-fold CV  & $0.758_{\pm 0.04} (1.163_{\pm 0.127})$ & $73.53_{\pm 1.30}  (72.14_{\pm 0.74})$  \\
			& GD          & $\mathbf{0.726}_{\pm 0.019} (\mathbf{1.14}_{\pm 0.323})$ & $\mathbf{74.08}_{\pm 0.79} (\mathbf{73.80}_{\pm 0.24})$ \\
			\bottomrule
		\end{tabular}
	\end{table*}

	We propose gradient disparity as an effective early stopping criterion, because of its computational tractability that makes it simple to use during the course of training, and because of its strong link with generalization error, as evidenced in the experiments that we run on state-of-the-art configurations. 
	Gradient disparity is particularly well suited 
	when the available dataset has limited labeled data, because it does not require splitting the available dataset into training and validation sets: all the available data can be used during training, unlike for instance $k$-fold cross-validation. We observe that using gradient disparity, instead of an unbiased validation set, results in a predictive improvement of at least $1\%$ for classification tasks with limited and very costly available data, such as the MRNet dataset, which is a small size image-classification dataset used for detecting knee injuries (Table~\ref{tab:mrnet}).
	
	Moreover, we find that gradient disparity is a more accurate early stopping criterion than validation loss when the available dataset contains noisy labels. 
	Gradient disparity reflects the label noise level quite well throughout the training process, especially at early stages of training. 
	Finally, we observe that gradient disparity has a strong positive correlation with the test error across experimental settings that differ in training set size, batch size, and network width. Code to reproduce our results is available at \url{https://github.com/mahf93/disparity_early_stopping}.

%% file: relatedwork.tex
\section{Related Work}\label{sec:rel}

The coherent gradient hypothesis \cite{chatterjee2020coherent} states that the gradient is stronger in directions where similar examples exist and towards which the parameter update is biased. He and Su \cite{He2020The} study the local elasticity phenomenon, which measures how the prediction over one sample changes, as the network is updated on another sample. Motivated by \cite{He2020The}, reference \cite{deng2020toward} proposes generalization upper bounds using locally elastic stability. The generalization penalty introduced in our work measures how the prediction over one sample (batch) changes when the network is updated on the same sample, instead of being updated on another sample. 

Finding a practical metric that completely captures the generalization properties of deep neural networks, and in particular indicates the level of label noise and decreases with the size of the training set, is still an active research direction \cite{dziugaite2017computing,neyshabur2017exploring,nagarajan2019uniform,Chatterji2020The}. Recently, there have been a few studies that propose similarity between gradients as a generalization metric. The benefit of tracking generalization by measuring the similarity between gradient vectors is its tractability during training, and the dispensable access to unseen data. 
Sankararaman et al.~\cite{sankararaman2019impact} propose gradient confusion, which is a bound on the inner product of two gradient vectors, and shows that the larger the gradient confusion is, the slower the convergence is. 
Gradient interference (when the gradient inner product is negative)
has been studied in multi-task learning, reinforcement learning and temporal difference learning \cite{riemer2018learning,liutoward,bengio2020interference}. Yin et al. \cite{yin2017gradient} study the relation between gradient diversity, which measures the dissimilarity between gradient vectors, and the convergence performance of distributed SGD algorithms. 
Fort et al. \cite{fort2019stiffness} propose a metric called stiffness, which is the cosine similarity between two gradient vectors, and shows empirically that it is related to generalization. 
Fu et al. \cite{fu2020rethinking} study the cosine similarity between two gradient vectors for natural language processing tasks. Reference \cite{mehta2020extreme} measures the alignment between the gradient vectors within the same class (denoted by $\Omega_{c}$)
, and studies the relation between $\Omega_{c}$ and generalization as the scale of initialization (the variance of the probability distribution the network parameters are initially drawn from) is increased. These metrics are usually not meant to be used as early stopping criteria, and indeed in Table~\ref{tab:RWmain} and Table~\ref{tab:RW} in the appendix, we observe that none of them consistently outperforms $k$-fold cross-validation.

Another interesting line of work is the study of the variance of gradients in deep learning settings. Negrea et al. \cite{negrea2019information} derive mutual information generalization error bounds for stochastic gradient Langevin dynamics (SGLD) 
as a function of the sum (over the iterations) of square gradient incoherences, which is closely related to the variance of gradients. Two-sample gradient incoherences also appear in \cite{haghifam2020sharpened}, which are taken between a training sample and a ``ghost" sample that is not used during training and therefore taken from a validation set (unlike gradient disparity).
The upper bounds in \cite{negrea2019information,haghifam2020sharpened} are cumulative bounds that increase with the number of iterations and are not intended to be used as early stopping criteria.
can be used as an early stopping criterion not only for SGD with additive noise (such as SGLD), but also other adaptive optimizers.
Reference \cite{qian2020impact} shows that the variance of gradients is a decreasing function of the batch size. However, reference \cite{jastrzebski2020break} hypothesizes that gradient variance counter-intuitively increases with the batch size, by studying the effect of the learning rate on the variance of gradients, which is consistent with our results on convolutional neural networks in Section~\ref{sec:gen}. 
References \cite{jastrzebski2020break,qian2020impact} mention the connection between variance of gradients and generalization as promising future directions. 
Our study shows that variance of gradients used as an early stopping criterion outperforms $k$-fold cross-validation (see Table~\ref{tab:RW}). 

Liu et al. \cite{Liu2020Understanding} propose a relation between gradient signal-to-noise ratio (SNR), called GSNR, and the one-step generalization error, with the assumption that both the training and test sets are large. 
Mahsereci et al. \cite{mahsereci2017early} also study gradient SNR and propose an early stopping criterion called evidence-based criterion (EB) that eliminates the need for a held-out validation set. 
Reference \cite{liu2008optimized} proposes an early stopping criterion based on the signal-to-noise ratio figure, which is further studied in \cite{piotrowski2013comparison}, a study that shows the average test error achieved by standard early stopping is lower than the one obtained by this criterion. Zhang et al. \cite{zhang2021optimization} empirically show that the variance term in the bias-variance decomposition of the loss function dominates the variations of the test loss, and hence propose optimization variance (OV) as an early stopping criterion.

\paragraph{Summary of Comparison to Related Work} In Table~\ref{tab:RWmain} and Appendix~\ref{app:var}, we compare gradient disparity (GD) to EB, GSNR, gradient inner product, sign of the gradient inner product, variance of gradients, cosine similarity, $\Omega_c$, and OV. We observe that the only metrics that consistently outperform $k$-fold cross-validation as early stopping criteria across various settings (see Table~\ref{tab:RW} in the appendix), and that reflect well the label noise level  (see in Figs.~\ref{fig:eb} and \ref{fig:inner_prod} that metrics such as EB and $\text{sign}(g_i\cdot g_j)$ do not correctly detect the label noise level), are gradient disparity and variance of gradients. The two are analytically very close as discussed in Appendix~\ref{app:var2}. However, we observe that the correlation between gradient disparity and the test loss is in general larger than the correlation between variance of gradients and the test loss (see Table~\ref{tab:var} in the appendix).

\begin{table*}[t]
	\caption{Test error (TE) and test loss (TL) achieved by using various metrics as early stopping criteria for an AlexNet trained on the MNIST dataset with 50\% random labels. See Table~\ref{tab:RW} in the appendix for further details and experiments.
	}\label{tab:RWmain}  
\vspace*{1em}
	\begin{subtable}{\linewidth}{
			\resizebox{\textwidth}{!}{
				\begin{tabular}{c|c|cccccccc|c|c}
					\toprule
					& Min    & GD/Var   \hspace*{-0.7em}  & EB   \hspace*{-0.7em}   & GSNR  & $g_i\cdot g_j$ & \hspace*{-0.3em} $\text{sign}(g_i\cdot g_j)$   & $\cos(g_i\cdot g_j)$  & $\Omega_c$ \hspace*{-0.7em} & OV & $k$-fold& No ES\\ 
					\midrule
					\multicolumn{1}{l|}{TE} &  $13.76$  & \ul{$\mathbf{16.66}$} & $24.63$  & $35.68$   & $37.92$              & $24.63$ &  
					$35.68$         & $29.40$ & $34.36$ &  $17.86$ & $25.72$\\ 
					TL                      & $0.75$  & \ul{$1.08$}  & \ul{$\mathbf{0.86}$} & $1.68$  & $1.82$    & \ul{$\mathbf{0.86}$}              
					& $1.68$            & $1.46$  & $1.65$&  $1.09$ & $0.91$\\
					\bottomrule
		\end{tabular}}}
	\end{subtable}
\end{table*}

%% file: gen_pen.tex
	\section{Generalization Penalty}\label{sec:pre}

	Consider a classification task with input $x \in \mathcal{X} \coloneqq \mathbb{R}^n$ and ground truth label $y\in \{1, 2, \cdots, k \}$, where $k$ is the number of classes. Let $h_w \in \mathcal{H}: \mathcal{X} \rightarrow \mathcal{Y} \coloneqq \mathbb{R}^k$ be a predictor (classifier) parameterized by the parameter vector $w \in \mathbb{R}^d$, and $l(\cdot, \cdot)$ be the 0-1 loss function
	$
	{l\left(h_w(x), y\right) = \mathbbm{1}\left[h_w(x)[y] < \max_{j\neq y}h_w(x)[j]\right] }
	$
	for all $h_w \in \mathcal{H}$ and $(x,y)\in \mathcal{X} \times \{1, 2, \cdots, k \}$. The expected loss and the empirical loss over the training set $S$ of size $m$ are respectively defined as\\
	\begin{align}
		L(h_w) &= \mathbb{E}_{(x,y)\sim D} \left[l\left(h_w(x),y\right)\right], 
	\end{align}\label{eq:exp_loss}
	and
	\begin{align}
		 L_{S}(h_w) = \frac{1}{m} \sum_{i=1}^{m} l(h_w(x_i),y_i) , \label{eq:train_loss}
	\end{align}
	where $D$ is the probability distribution of the data points and $S=\{(x_i,y_i)\}^m$ is a collection of $m$ i.i.d. samples drawn from $D$.
	Similar to the notation used in \cite{dziugaite2017computing}, distributions on the hypotheses space $\mathcal{H}$ are simply distributions on the underlying parameterization. With some abuse of notation, $\nabla L_{S_i}$ refers to the gradient with respect to the surrogate differentiable loss function, which in our experiments is cross entropy\footnote{We have also studied networks trained with the mean square error in Appendix~\ref{app:mse}, and we observe that there is a strong positive correlation between the test error/loss and gradient disparity for this choice of the surrogate loss function as well (see Fig.~\ref{fig:train_size_mse}).}. 
	
	In a mini-batch gradient descent (SGD) setting, let mini-batches $S_1$ and $S_2$ have sizes $m_1$ and $m_2$, respectively, with ${m_1 + m_2 \leq m}$.
	Let ${w = w^{(t)}}$ be the parameter vector at the beginning of an iteration $t$. If $S_1$ is selected for the next iteration, $w$ gets updated to ${w_1 = w^{(t+1)}}$ with
	\begin{equation}\label{eq:w1}
	 w_1 = w - \gamma \nabla L_{S_1}\left(h_{w}\right) ,
	\end{equation}
	where $\gamma$ is the learning rate. 
	The generalization penalty $\mathcal{R}_2$ 
	is defined as
	the gap between the loss over $S_2$, $L_{S_2}\left(h_{w_1}\right)$, and its target value, $L_{S_2}\left(h_{w_2}\right)$, at the end of iteration $t$.

	When selecting $S_1$ for the parameter update, Eq.~(\ref{eq:w1})
	makes a step towards learning the input-output relations of mini-batch $S_1$. If this negatively affects the performance on mini-batch~$S_2$,~$\mathcal{R}_2$ will be large; the model is learning the data structures that are unique to $S_1$ and that do not appear in~$S_2$.
	Because~$S_1$ and $S_2$ are mini-batches of points sampled from the same distribution $D$, they have data structures in common. If, throughout the learning process, we consistently observe that, in each update step, the model learns structures unique to only one mini-batch, then it is very likely that the model is memorizing the labels instead of learning the common data-structures. This is captured by the generalization penalty~$\mathcal{R}$.

%% file: bound.tex
	We adapt the PAC-Bayesian framework \cite{mcallester1999pac,mcallester1999some} to account for the trajectory of the learning algorithm; For each learning iteration $t$ we define a prior, and two possible posteriors depending on the choice of the batch selection. Let $w\sim P$ follow a prior distribution~$P$, which is a $\mathcal{F}_t$-measurable function, where $\mathcal{F}_t$ denotes the filtration of the available information at the beginning of iteration $t$. 
	Let $h_{w_1}, h_{w_2}$ be the two learned single predictors, at the end of iteration $t$, from $S_1$ and $S_2$, respectively. 
	In this framework, for $i \in \{1,2\}$, each predictor $h_{w_i}$ is randomized and becomes $h_{\nu_i}$ with $\nu_i = w_i + u_i$, where $u_i$ is a random variable whose distribution might depend on $S_i$. 
	Let $Q_i$ be the distribution of~$\nu_i$, which is a distribution over the predictor space $\mathcal{H}$ that depends on $S_i$ via $w_i$ and possibly $u_i$. Let $\mathcal{G}_i$ be a $\sigma$-field such that $\sigma(S_i) \cup \mathcal{F}_t \subset \mathcal{G}_i$ and such that the posterior distribution $Q_i$ is $\mathcal{G}_i$-measurable for $i \in \{1,2\}$. We further assume that 
	the random variable $\nu_1\sim Q_1$ is statistically independent from the draw of the mini-batch $S_2$ 
	and, vice versa, that $\nu_2\sim Q_2$ is independent from the batch $S_1$\footnote{Mini-batches $S_1$ and $S_2$ are drawn without replacement, and the random selection of indices of mini-batches $S_1$ and $S_2$ is independent from the dataset $S$. Hence, similarly to \cite{negrea2019information,dziugaite2020role}, we have $\sigma(S_1) \indep \sigma(S_2)$.},
	i.e., $\mathcal{G}_1 \indep \sigma(S_2)$ and $\mathcal{G}_2 \indep \sigma(S_1)$.

	\begin{theorem}\label{THM1}
		For any $\delta \in (0,1]$, with probability at least $1-\delta$ over the sampling of sets $S_1$ and $S_2$, the sum of the expected penalties conditional on $S_1$, and $S_2$, respectively, satisfies
		\begin{align}\label{eq:mybound}
		\mathbb{E} \left[\mathcal{R}_1\right] + \mathbb{E} \left[\mathcal{R}_2\right]  \leq  \sqrt{\frac{2\text{KL}(Q_2||Q_1) + 2\ln{\frac{2m_2}{\delta}}}{{m_2}-2}}  + \sqrt{\frac{2\text{KL}(Q_1||Q_2) + 2\ln{\frac{2m_1}{\delta}}}{{m_1}-2}} .
		\end{align}
	\end{theorem}
	In this paper, the goal is to get a signal of overfitting that indicates at the beginning of each iteration $t$ whether to stop or to continue training. This signal should track the performance of the model at the end of iteration~$t$ by investigating its evolution over all the possible outcomes of the batch sampling process during this iteration. For simplicity, we consider two possible outcomes: either mini-batch $S_1$ or mini-batch $S_2$ is chosen for this iteration (we later in the next section extend to more pairs of batches). If we were to use bounds such as the ones in \cite{mcallester2003simplified,neyshabur2017pac} for one iteration at a time, the generalization error at the end of that iteration can be bounded by a function of either $\text{KL}(Q_1||P)$ or $\text{KL}(Q_2||P)$, depending on the selected batch. Therefore, as each of the two batches is equally likely to be sampled, we should track $\text{KL}(Q_1||P)$ and $\text{KL}(Q_2||P)$ for a signal of overfitting at the end of the iteration, which requires in turn access to the three distributions $P$, $Q_1$ and $Q_2$. 
	In contrast, the upper bound on the generalization penalty given in Theorem~\ref{THM1} only requires the two distributions $Q_1$ and $Q_2$, which is a first step towards a simpler metric since, loosely speaking, the symmetry between the random choices for $S_1$ and $S_2$ should carry over these two distributions, leading us to assume the random perturbations $u_1$ and $u_2$ to be identically distributed. If furthermore we assume them to be Gaussian, then we show in the next section that $\text{KL}(Q_2||Q_1)$ and $\text{KL}(Q_1||Q_2)$ are equal and boil down to a very tractable generalization metric, which we call gradient disparity.

%% file: gradient_disparity.tex
	\section{Gradient Disparity}\label{sec:BGP}
	In Section~\ref{sec:pre}, the randomness modeled by the additional perturbation $u_i$, conditioned on the current mini-batch $S_i$, comes from (i) the parameter vector at the beginning of the iteration $w$, which itself comes from the random parameter initialization and the stochasticity of the parameter updates until that iteration, and (ii) the gradient vector $\nabla L_{S_i}$ (simply denoted by $g_i$), 
	which may also be random because of the possible additional randomness in the network structure due for instance to dropout \cite{srivastava2014dropout}. 
	A common assumption made in the literature is that the random perturbation $u_i$ follows a normal distribution \cite{bellido1993backpropagation,neyshabur2017pac}. 
	The upper bound in Theorem~\ref{THM1} takes a particularly simple form if we assume that for $i \in \{1,2\}$, $u_i$ are zero mean i.i.d. normal variables (${u_i \sim \mathcal{N} (0, \sigma^2 I)} $), and that $w_i$ is fixed, as in the setting of \cite{dziugaite2017computing}.

	As ${w_i=w - \gamma g_i}$ for $i \in \{1,2\}$, the KL-divergence between $Q_1 = \mathcal{N}(w_1, \sigma^2 I)$ and ${Q_2 = \mathcal{N}(w_2, \sigma^2 I)}$ (Lemma~\hyperref[lem:1]{1} in Appendix~\ref{app:add}) is simply
	\begin{align}\label{eq:kl}
	\text{KL}(Q_1 || Q_2) = \frac{1}{2} \frac{\gamma^2}{\sigma^2} \norm{g_1 - g_2}_2^2 
	= \text{KL}(Q_2 || Q_1) ,
	\end{align}
	which shows that, keeping a constant step size $\gamma$ and assuming the same variance for the random perturbations $\sigma^2$ in all the steps of the training, the bound in Theorem \ref{THM1} is driven by $\norm{g_1 - g_2}_2 $. This indicates that the smaller the $\ell_2$ distance between gradient vectors is, the lower the upper bound on the generalization penalty is, and therefore the closer the performance of a model trained on one batch is to a model trained on another batch. 
	
	For two mini-batches of points $S_i$ and $S_j$, with respective gradient vectors~$g_i$ and $g_j$, we define the \emph{gradient disparity} (GD) between $S_i$ and $S_j$ as
	\begin{equation}\label{eq:bgd}
	\mathcal{D}_{i,j} = \norm{g_i - g_j}_2 .
	\end{equation}

\begin{figure}[h]
		\centering
		\begin{subfigure}[b]{1\textwidth} 
			\centering           
			\includegraphics[width=0.60\textwidth, height=0.336\textwidth]{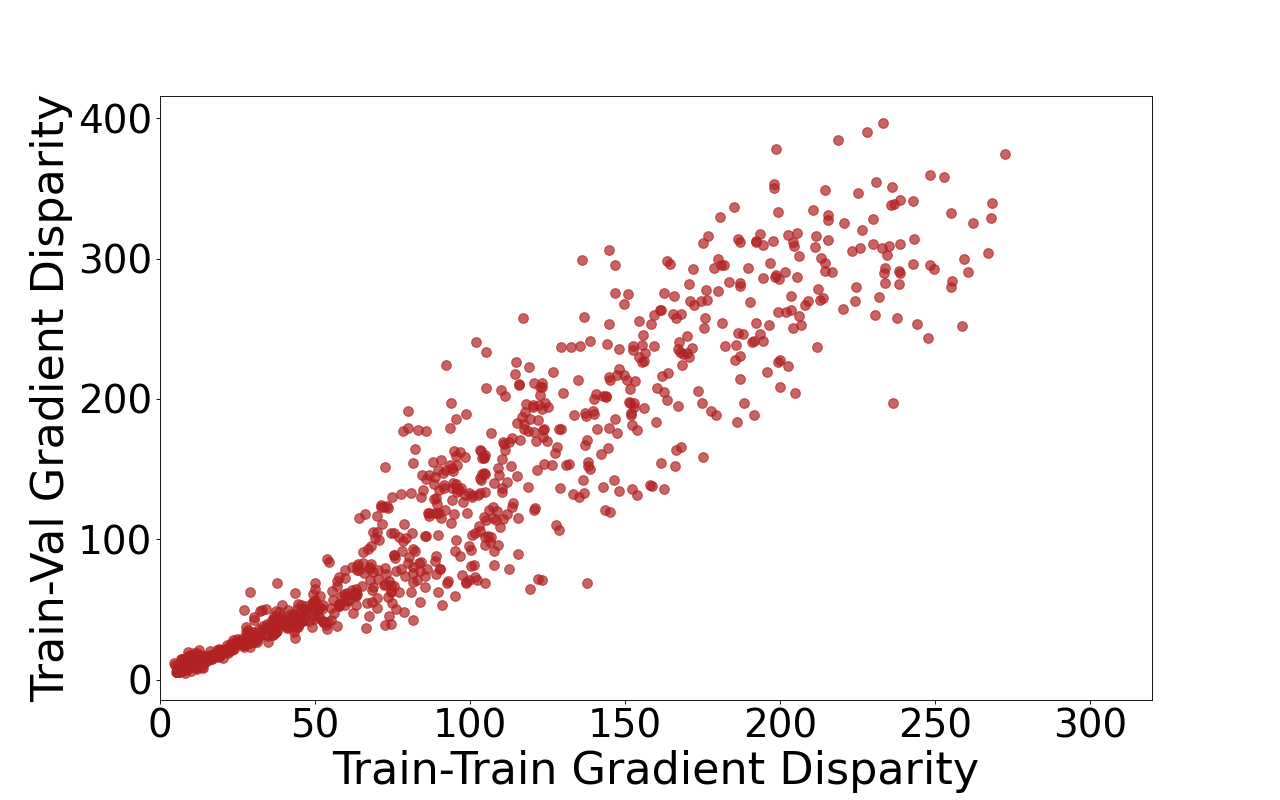}%
		\end{subfigure}
		\caption{``Train-val" gradient disparity versus ``train-train" gradient disparity for 220 experimental settings that vary in architecture, dataset, training set size, label noise level and initial random seed. 
			Pearson's correlation coefficient is $\rho=0.957$.
		} \label{fig:valtrain}
	\end{figure}

	To compute $\mathcal{D}_{i,j}$, a first option is to sample $S_i$ from the training set and $S_j$ from the held-out validation set, which we refer to as the ``train-val" setting, following~\cite{fort2019stiffness}. The generalization penalty $\mathcal{R}_j$ in this setting measures how much, during the course of an iteration, a model updated on a training set is able to generalize to a validation set, making the resulting (``train-val”) gradient disparity $\mathcal{D}_{i,j}$ a natural candidate for tracking overfitting. But it requires access to a validation set to sample $S_j$, which we want to avoid. The second option is to sample both $S_i$ and $S_j$ from the training set, as proposed in this paper, to yield now a value of $\mathcal{D}_{i,j}$ that we could call ``train-train" gradient disparity (GD) by analogy. Importantly, we observe a strong positive correlation between the two types of gradient disparities ($\rho=0.957$) in Fig.~\ref{fig:valtrain}. 
	Therefore, we can expect that both of them do (almost) equally well in detecting overfitting, with the advantage that the latter does not require to set data aside, contrary to the former. We will therefore consider GD when both batches are sampled from the training set and evaluate it in this paper. 
	
	To track the upper bound of the generalization penalty for more pairs of mini-batches, we can compute an average gradient disparity over $B$ mini-batches, which requires all the $B$ gradient vectors at each iteration, which is computationally expensive if $B$ is large. We approximate it by computing GD over only a much smaller subset of the mini-batches, of size $s \ll B$,
	\begin{equation*}
	\overline{\mathcal{D}} =  \sum_{i=1}^{s} \sum_{\substack{j=1 , j\neq i }}^{s} \frac{\mathcal{D}_{i,j}}{s(s-1)} .
	\end{equation*}
	In our experiments, $s=5$; we observe that such a small subset is already sufficient (see Appendix~\ref{app:s} for an experimental comparison of different values of $s$).

	Consider two training iterations $t_1$ and $t_2$ where $t_1 \ll t_2$. At  earlier stages of the training (iteration $t_1$), the parameter vector ($w^{(t_1)}$) is likely to be located in a steep region of the training loss landscape, where the gradient vector of training batches, $g_i$, and the training loss $L_{S_i}(h_{w^{(t_1)}})$ take large values. At later stages of training (iteration $t_2$), the parameter vector ($w^{(t_2)}$) is more likely in a flatter region of the training loss landscape where $g_i$ and $L_{S_i}(h_{w^{(t_2)}})$ take small values. To compensate for this scale mismatch when comparing the distance between gradient vectors at different stages of training, we re-scale the loss values within each batch before computing $\overline{\mathcal{D}}$ (see Appendix~\ref{app:norm} for more details). 
	Note that this re-scaling is only done for 
	the purpose of using GD as a metric, and therefore does not have any effect on the training process itself.

We focus on the vanilla SGD optimizer. In Appendix~\ref{app:opt}, we extend the analysis to other stochastic optimization algorithms: SGD with momentum, Adagrad, Adadelta, and Adam. In all these optimizers, we observe that GD (Eq.~(\ref{eq:bgd})) appears in $\text{KL}(Q_1 || Q_2) $ with other factors that depend on a decaying average of past gradient vectors. Experimental results support the use of GD as an early stopping metric also for these popular optimizers (see Fig.~\ref{fig:opt} in Appendix~\ref{app:opt}). For vanilla SGD optimizer, we also provide an alternative and simpler derivation leading to gradient disparity from the linearization of the loss function in Appendix~\ref{app:simple}.

%% file: experimental_results.tex
	\section{Early Stopping Criterion}\label{sec:early}

\input{kfold.tex}

	\section{Discussion and Final Remarks}\label{sec:gen}

	We propose gradient disparity (GD), as a simple to compute early stopping criterion that is particularly well-suited when the dataset is limited and/or noisy. Beyond indicating the early stopping time, GD is well aligned with factors that contribute to improve or degrade the generalization performance of a model, which have an often strikingly similar effect on the value of GD as well. We briefly discuss in this section some of these observations that further validate the use of GD as an effective early stopping criterion; more details are provided in the appendix.

	\textbf{Label Noise Level.} We observe that GD reflects well the label noise level throughout the training process, even at early stages of training, where the generalization gap fails to do so (see Fig.~\ref{fig:random}, and Figs.~\ref{fig:fc_mnist}, \ref{fig:fc_cifar10}, and~\ref{fig:resnet_cifar100} in Appendix~\ref{app:more}).
	
	\begin{figure*}[h]
		\centering
		\begin{subfigure}[b]{0.32\textwidth}
			\centering
			\includegraphics[width=\textwidth, height=0.5625\textwidth]{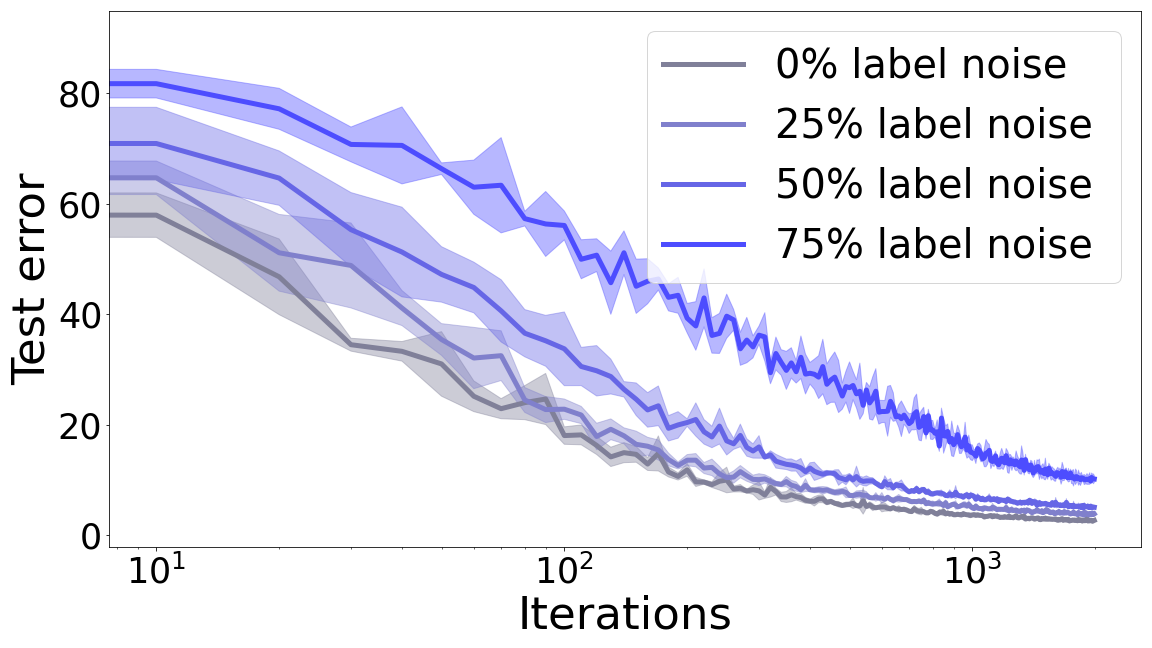}
			\caption{Test error}
		\end{subfigure}\hspace{0.4em}%
		\begin{subfigure}[b]{0.32\textwidth}            
			\includegraphics[width=\textwidth, height=0.575\textwidth]{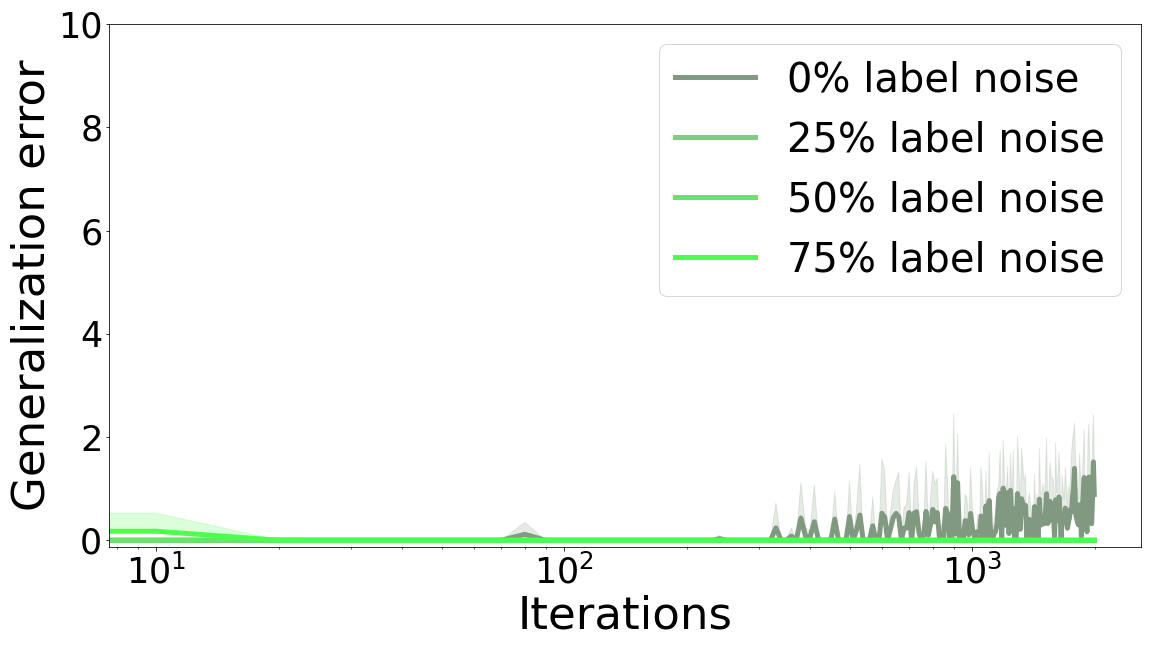}
			\caption{Generalization error }
		\end{subfigure}\hspace{0.4em}%
		\begin{subfigure}[b]{0.32\textwidth}
			\centering
			\includegraphics[width=\textwidth, height=0.5625\textwidth]{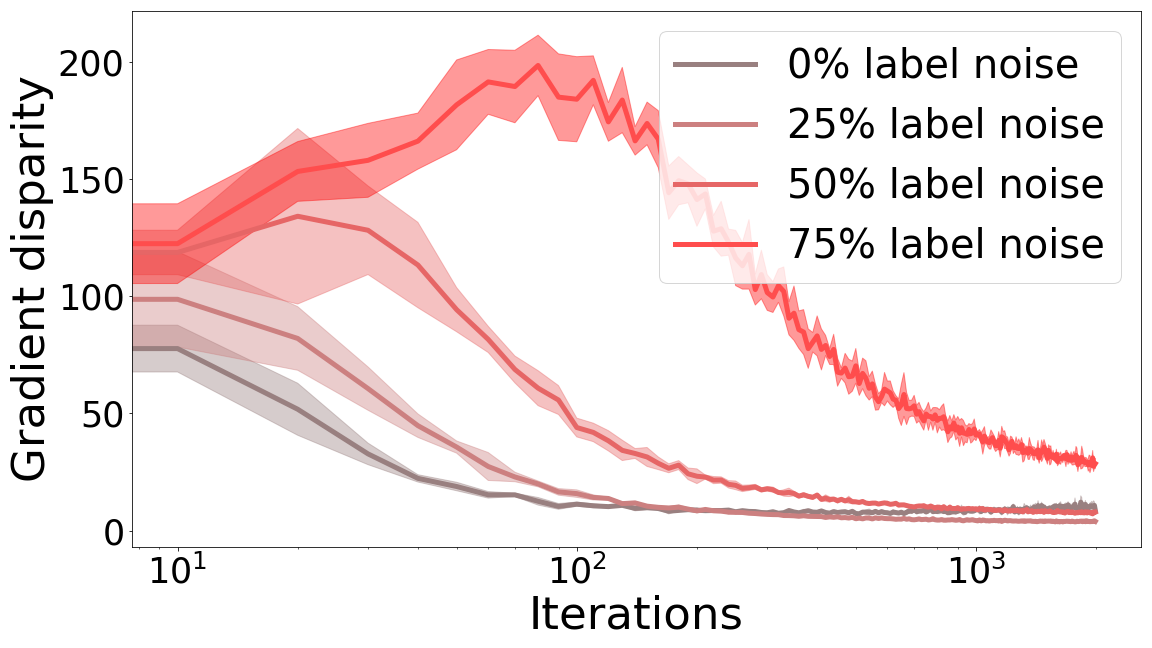}
			\caption{ $\overline{\mathcal{D}}$}
		\end{subfigure}
		\caption{ The error percentage and $\overline{\mathcal{D}}$ during training for 
			an AlexNet trained on a subset of 12.8 k points of the MNIST training dataset with different amounts of label noise. Pearson's correlation coefficient between gradient disparity and test error (TE)/test loss (TL) over all iterations and  label noise levels are $\rho_{\overline{\mathcal{D}}, \text{TE}} = 0.861$ and $\rho_{\overline{\mathcal{D}}, \text{TL}} = 0.802$. The generalization error (gap) is the difference between the train and test errors and for this experiment fails to account for the label noise level, unlike gradient disparity.}\label{fig:random}
	\end{figure*}
	
	\textbf{Training Set Size.} We observe that GD, similarly to the test error, decreases with training set size, unlike many previous metrics as shown by \cite{neyshabur2017exploring,nagarajan2019uniform}. 
	Moreover, we observe that applying data augmentation decreases the values of both GD and the test error (see Fig.~\ref{fig:DA} and Fig.~\ref{fig:train_size2} in Appendix~\ref{app:more}).
	
	\begin{figure*}[h]
		\centering
		\begin{subfigure}[b]{0.32\textwidth}            
			\includegraphics[width=0.9\textwidth, height=0.5062\textwidth]{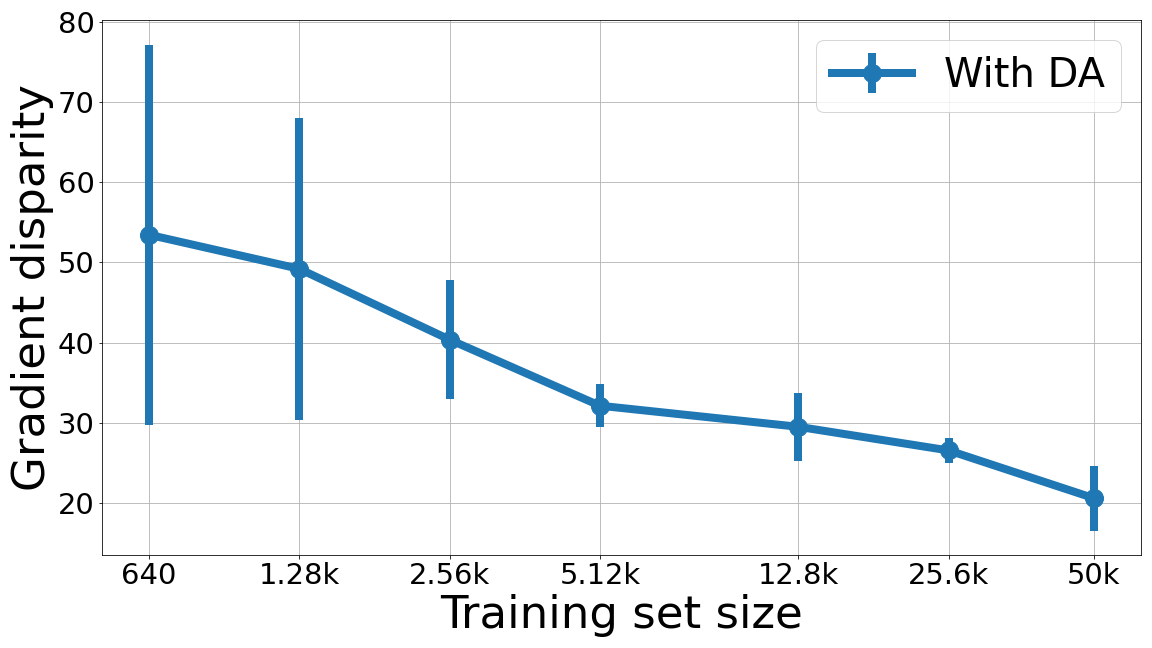}\\
			\includegraphics[width=0.9\textwidth, height=0.5062\textwidth]{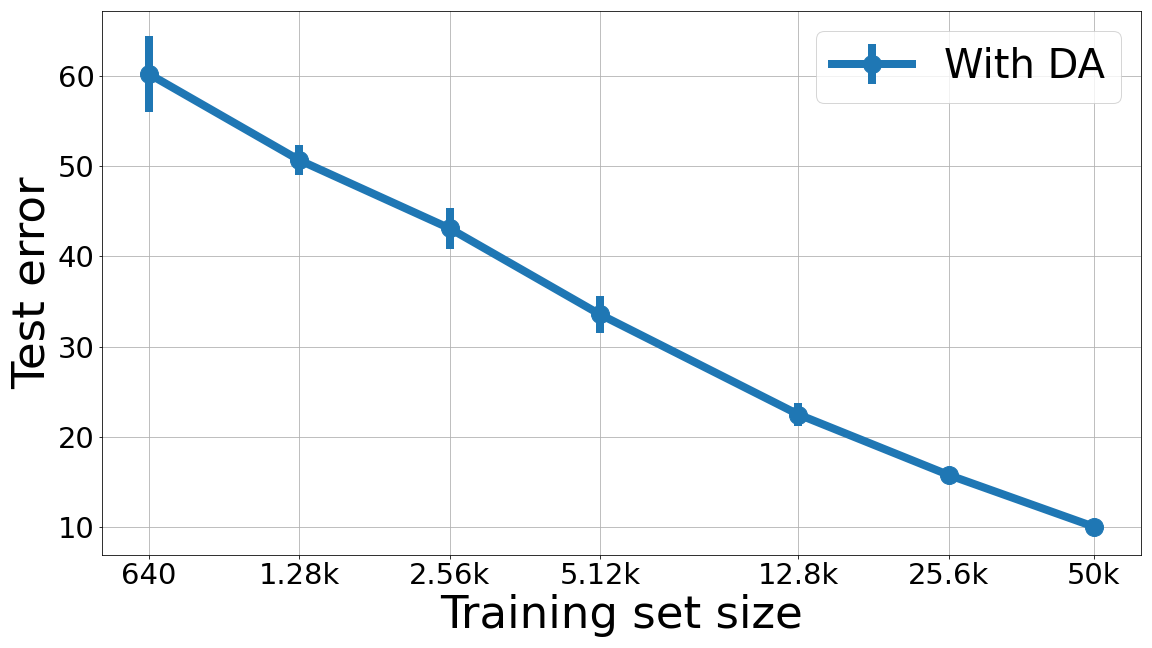}
			\caption{$\rho_{\overline{\mathcal{D}}, \text{TE}}=0.979$}
		\end{subfigure}%
		\begin{subfigure}[b]{0.32\textwidth}            
			\includegraphics[width=0.9\textwidth, height=0.5062\textwidth]{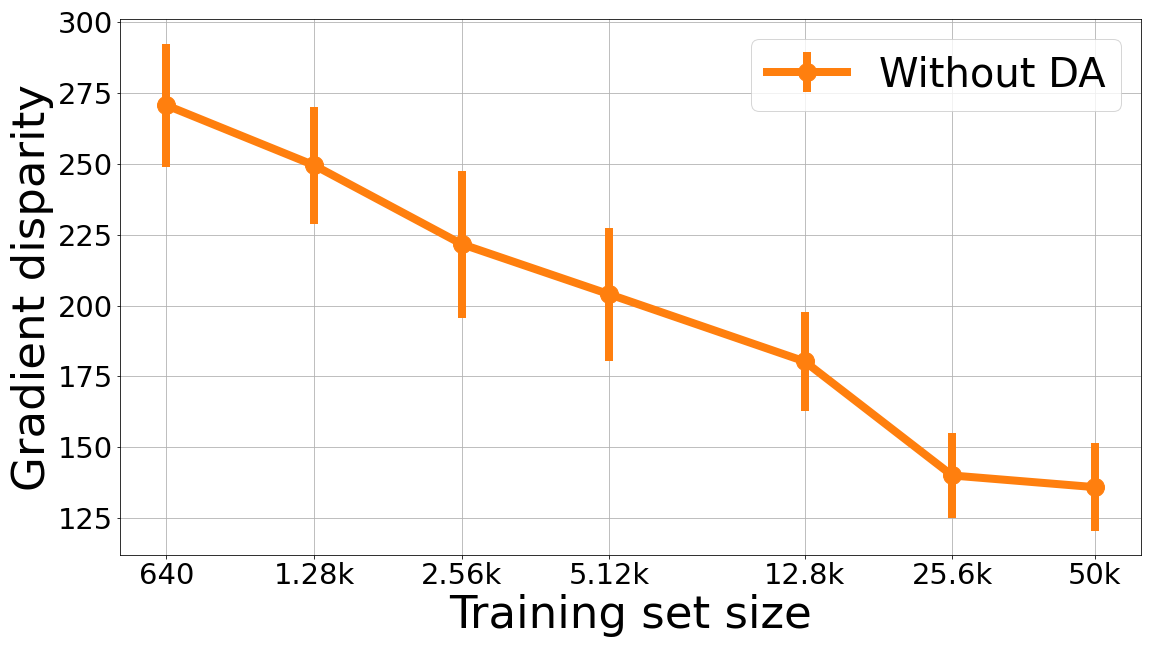}\\
			\includegraphics[width=0.9\textwidth, height=0.5062\textwidth]{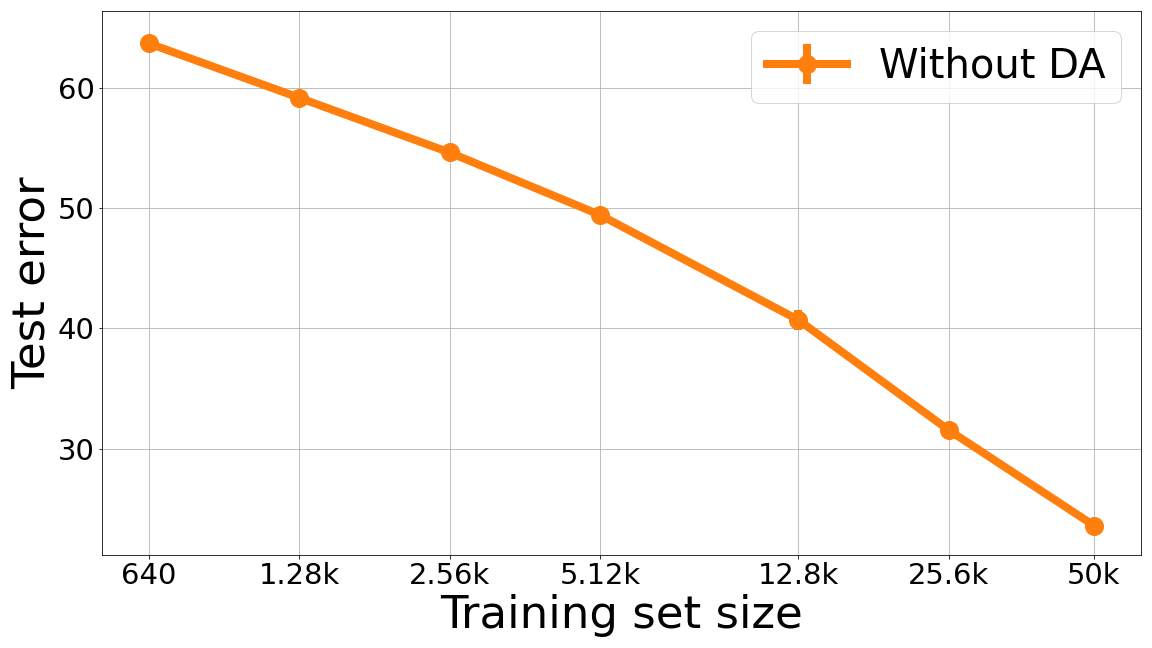}
			\caption{$\rho_{\overline{\mathcal{D}}, \text{TE}}=0.984$}
		\end{subfigure}%
		\begin{subfigure}[b]{0.32\textwidth}            
			\includegraphics[width=0.9\textwidth, height=0.5062\textwidth]{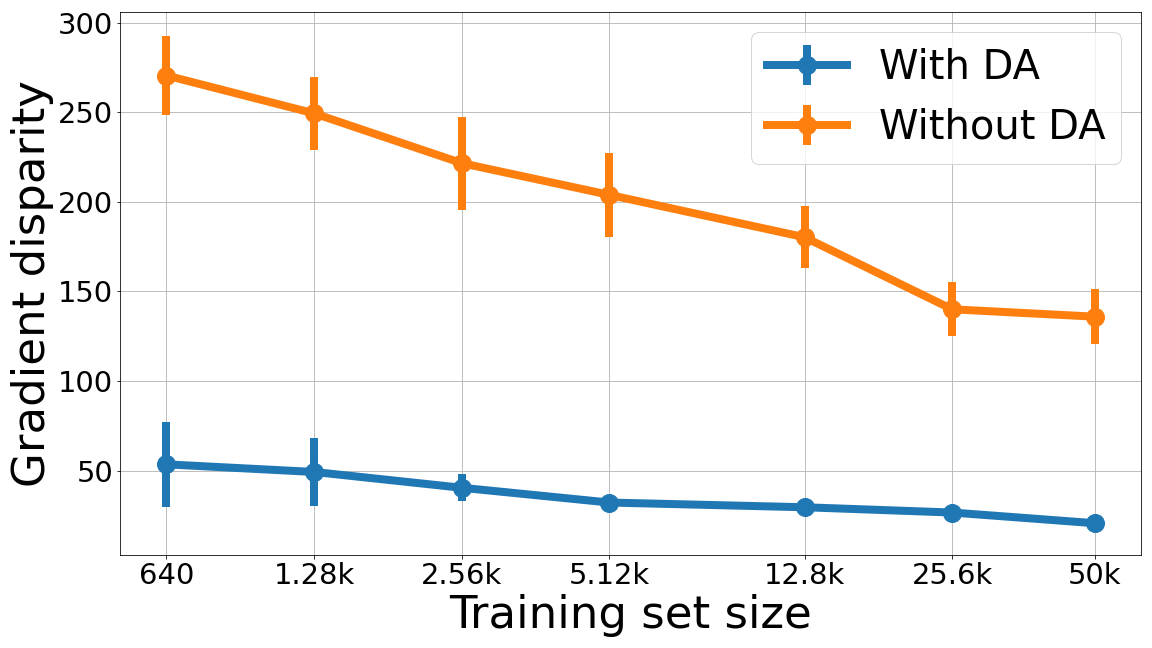}\\
			\includegraphics[width=0.9\textwidth, height=0.5062\textwidth]{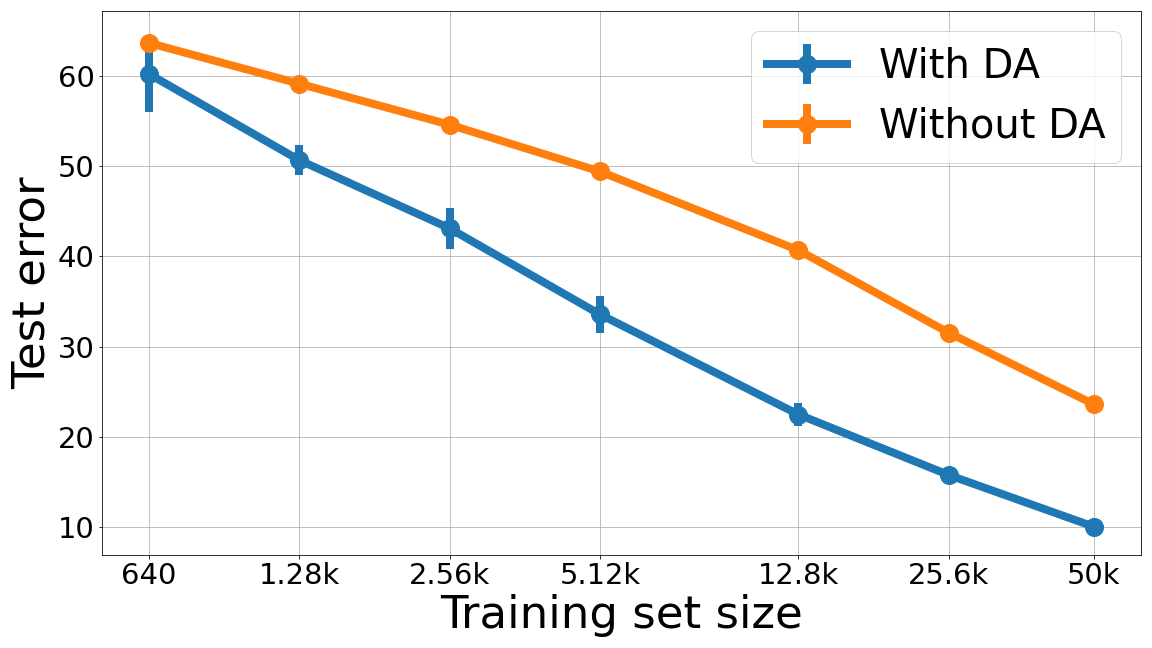}
			\caption{$\rho_{\overline{\mathcal{D}}, \text{TE}}=0.619$}
		\end{subfigure}
		\caption{Test error (TE) (bottom row) and gradient disparity ($\overline{\mathcal{D}}$) (top row) for a ResNet-18 trained 
			on the CIFAR-10 dataset with different training set sizes. (a) Results with data augmentation (DA) 
			. (b) Results without using any data augmentation technique. (c) Combined results of (a) and (b). We observe a strong positive correlation between $\overline{\mathcal{D}}$ and TE regardless of using data augmentation. We also observe that using data augmentation decreases the values of both gradient disparity and the test error.}\label{fig:DA}
	\end{figure*} 

	\textbf{Batch Size.} We observe that both the test error and GD increase with batch size.
	This observation is counter-intuitive because one might expect that gradient vectors get more similar when they are averaged over a larger batch.
	GD matches the ranking of test errors for different networks, trained with different batch sizes, as long as the batch sizes are not too large 
	(see Fig.~\ref{fig:batch_size2} in Appendix~\ref{app:more}).
	\begin{figure*}[h]
		\begin{subfigure}[b]{1\textwidth}
			\hspace{3em}%
			\includegraphics[width=0.4\textwidth, height=0.224\textwidth]{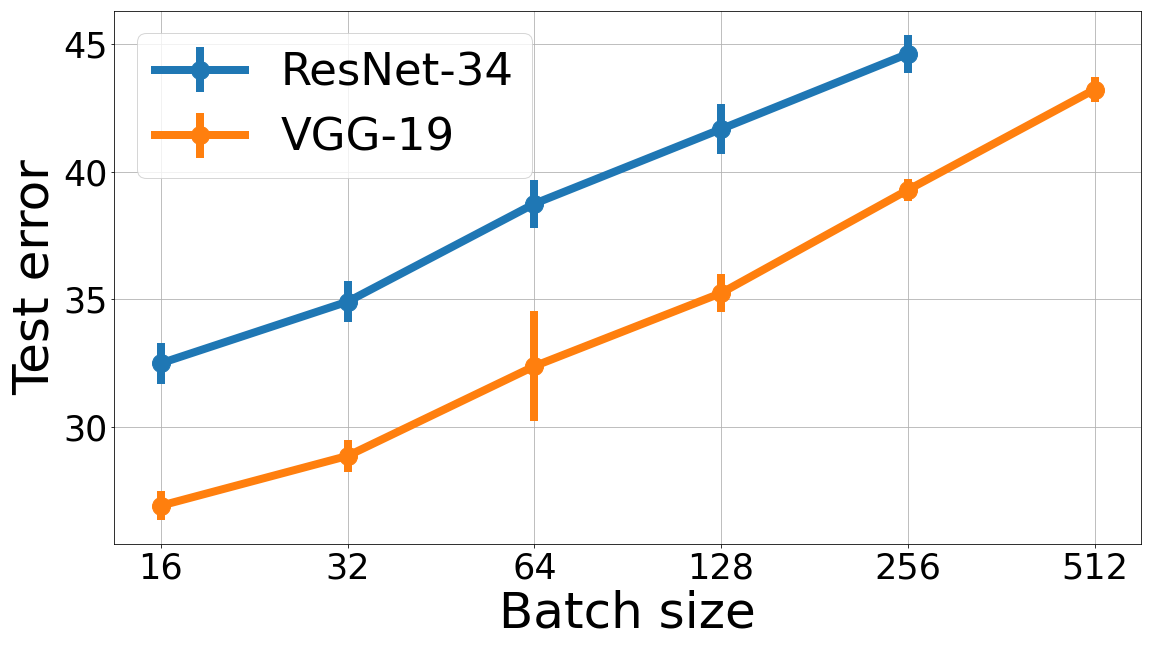}%
			\hspace{5em}%
			\includegraphics[width=0.4\textwidth, height=0.224\textwidth]{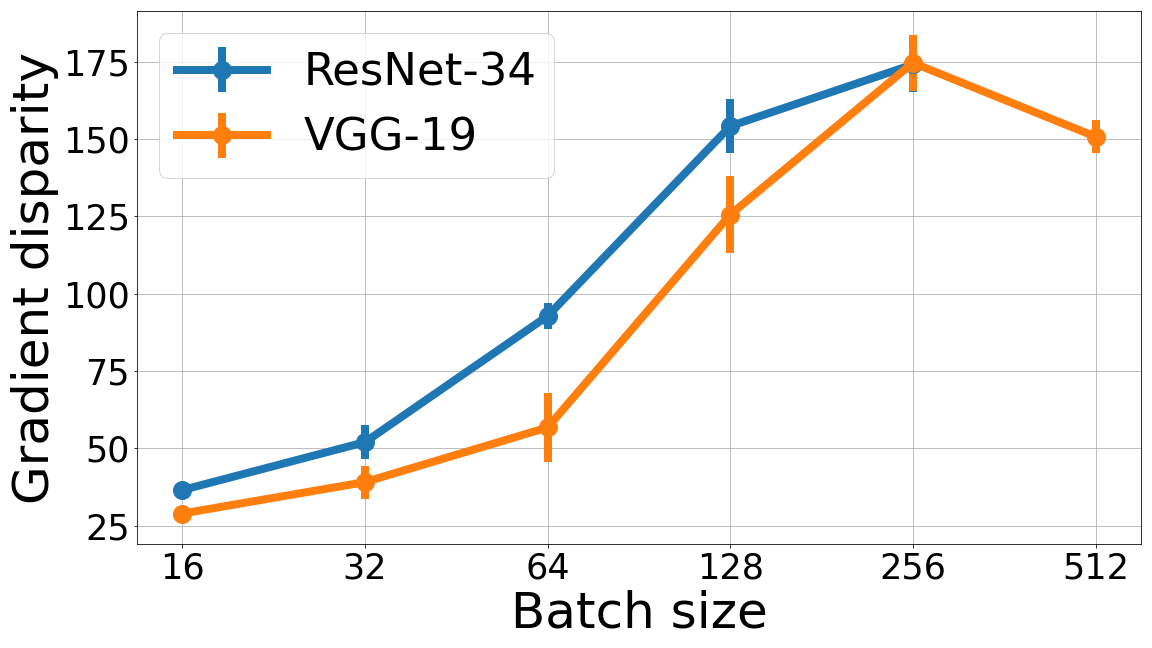}
		\end{subfigure}
		\caption{Test error and gradient disparity for networks that are trained with different batch sizes trained on 12.8 k points of the CIFAR-10 dataset. The training is stopped when the training loss is below $0.01$. The correlation between normalized gradient disparity and test error is ${\rho_{\tilde{\mathcal{D}}, \text{TE}} = 0.893}$.}\label{fig:batch_size}
	\end{figure*}
	
	\textbf{Width.} We observe that both the test error and GD (normalized with respect to the number of parameters) decrease with the network width 
	for ResNet, VGG and fully connected neural networks (see Fig.~\ref{fig:width} and Fig.~\ref{fig:width2} in Appendix~\ref{app:more}).
	
	\begin{figure*}[h]
		\begin{subfigure}[b]{1\textwidth} 
			\hspace{2.2em}%
			\includegraphics[width=0.46\textwidth, height=0.252\textwidth]{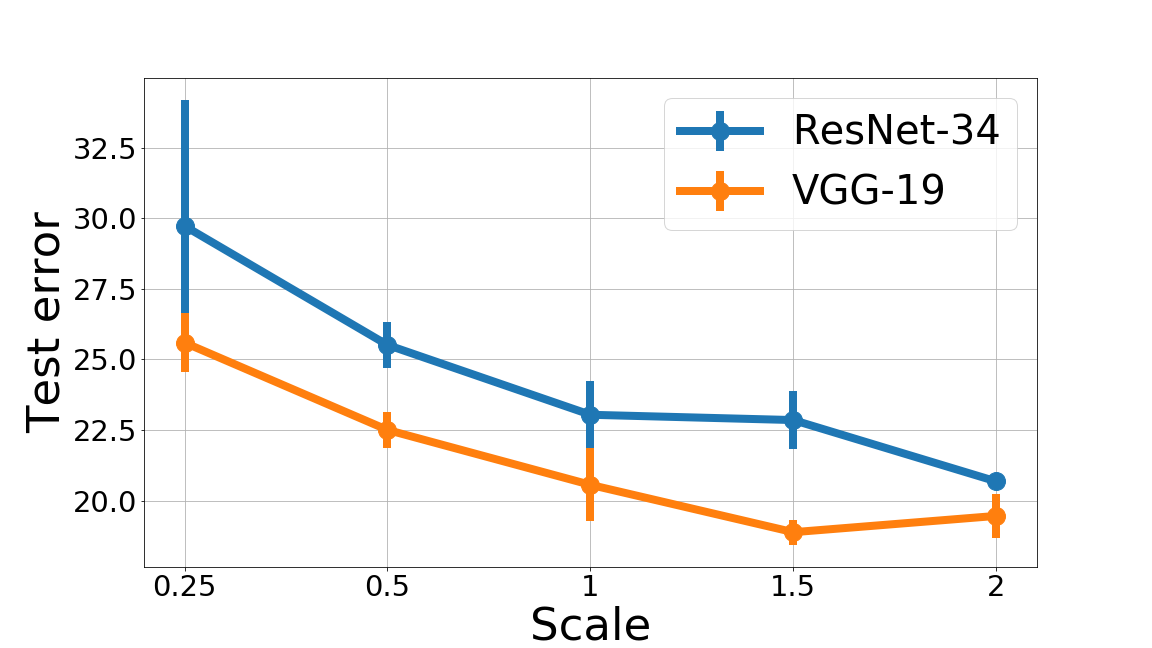}%
			\hspace{2.59em}%
			\includegraphics[width=0.45\textwidth, height=0.252\textwidth]{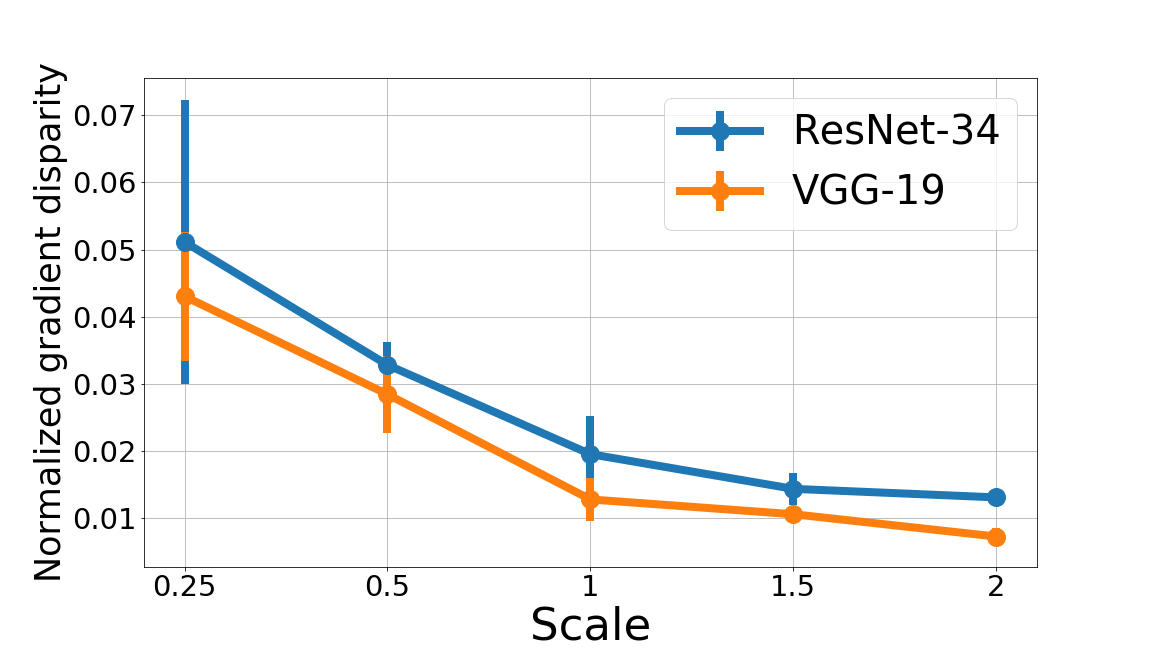}
		\end{subfigure}
		\caption{Test error and normalized gradient disparity for networks trained 
			on the CIFAR-10 dataset with different number of channels and hidden units for convolutional neural networks (CNN) (scale = 1 recovers the original configurations). The correlation between normalized gradient disparity and test loss $\rho_{\tilde{\mathcal{D}}, \text{TL}}$ and between normalized gradient disparity and test error $\rho_{\tilde{\mathcal{D}}, \text{TE}}$ are ${\rho_{\tilde{\mathcal{D}}, \text{TL}} = 0.970}$ and ${\rho_{\tilde{\mathcal{D}}, \text{TE}} = 0.939}$, respectively.}\label{fig:width}
	\end{figure*}

Gradient disparity belongs to the same class of metrics based on the similarity between two gradient vectors \cite{sankararaman2019impact,fort2019stiffness,fu2020rethinking,mehta2020extreme,jastrzebski2020break}. A common drawback of all these metrics  
is that they are not informative when the gradient vectors are very small. In practice however, we observe (see for instance Figure~\ref{fig:fc_mnist0} in the appendix) that the time at which the test and training losses start to diverge, which is the time when overfitting kicks in, does not only coincide with the time at 
which gradient disparity increases, but also occurs much before the training loss becomes infinitesimal. This drawback is therefore unlikely to cause a problem for gradient disparity when it is used as an early stopping criterion. 
Nevertheless, as a future direction, it would be interesting to explore this further especially for scenarios such as epoch-wise double-descent \cite{heckel2020early}.

%% file: kfold.tex
	In the presence of \emph{large} amounts of \emph{reliable} data, it is affordable to split the available dataset into a training and a validation set and to perform early stopping by evaluating the performance of the model on the held-out validation set. However, if the dataset is \emph{limited}, this approach makes an inefficient use of the data because the model never learns the information that is still present in the validation set. Moreover, if the dataset is \emph{noisy}, held-out validation might poorly estimate the performance of the model as the validation set might contain a high percentage of noisy samples. 
	To avoid these issues, $k$-fold cross-validation~\cite{stone1974cross} is a solution that makes an efficient usage of the available data while providing an unbiased estimate of the performance, at the expense of a high computational overhead and of a possibly underestimated variance \cite{bengio2004no}. 
	While each of its $k$ rounds is itself a setting with a held-out validation set, $k$-fold cross-validation (as opposed to held-out validation) would be therefore advantageous to use in the presence of limited and/or noisy data. It
	extracts more information from the dataset as it uses all the data samples for both training and validation, and it is less dependent on how the data is split into training and validation sets.

	The baseline to beat is therefore $k$-fold cross-validation (CV). We compare gradient disparity to CV in the two target settings: (i) when the available dataset is limited and (ii) when the available dataset has corrupted labels. Medical applications are one of the practical examples of setting (i), where datasets are costly because they require the collection of patient data, and the medical staff's expertise to label the data. An example of such an application is the MRNet dataset \cite{bien2018deep}, which contains a limited number of MRI scans to study the presence of abnormally, ACL tears and meniscal tears in knee injuries. This dataset is by nature limited and we use the entire available data for both early stopping methods GD and $k$-fold CV. In addition, to further simulate setting (i), we use small subsets of three image classification benchmark datasets: MNIST, \mbox{CIFAR-10} and CIFAR-100. Performing early stopping in the presence of label noise (setting (ii)) is also practically very important, because it has been empirically observed that deep neural networks trained on noisy datasets overfit to noisy labeled samples at later stages of training. A good early stopping signal can therefore prevent such an overfitting \cite{li2020gradient,song2020prestopping,xia2021robust}. To simulate setting~(ii), we use a corrupted version of these image classification benchmark datasets, where for a fraction of the samples (the amount of noise), we choose the labels at random. 

	\begin{table*}[t]
		\centering
		\caption{Test loss and accuracy when using gradient disparity (GD) and $k$-fold cross-validation (CV) ($k=$5) as early stopping criteria when the available dataset is limited: (top) VGG-13 trained on 1.28 k samples of the CIFAR-10 dataset, and (bottom) AlexNet trained on 256 samples of the MNIST dataset. The corresponding curves during training are presented in Fig.~\ref{fig:kfoldl}. The results below are obtained by stopping the optimization when the metric (either validation loss or GD) has increased for 5 epochs from the beginning of training.}\label{tab:kfoldlmain}
		\vspace*{1em}
		\begin{tabular}{c|c|c|c}
			\toprule
			Setting & Method & Test loss & Test accuracy  \\ 
			\midrule
			\multirow{2}{*}{CIFAR-10, VGG-13} & 5-fold CV& $1.846_{\pm 0.016}$ & $35.982_{\pm 0.393}$  \\
			& GD &  $\mathbf{1.793}_{\pm 0.016}$ & $\mathbf{36.96}_{\pm 0.861}$  \\ 
			\midrule
			\multirow{2}{*}{MNIST, AlexNet} & 5-fold CV &  $1.123_{\pm 0.25}$ & $62.62_{\pm 6.36}$  \\ 
			& GD & $\mathbf{0.656}_{\pm 0.080}$ & $\mathbf{79.12}_{\pm 3.04}$  \\ 
			\bottomrule
		\end{tabular}

	\end{table*}
	(i) 
	We observe that using gradient disparity instead of a validation loss in $k$-fold CV results in an improvement of more than  $1\%$ (on average over all three tasks) in the test AUC score of the MRNet dataset, and therefore adds a correct detection for more than one patient for each task (see Table~\ref{tab:mrnet}). 
	Furthermore, we observe that gradient disparity performs better than $k$-fold CV as an early stopping criterion for image-classification benchmark datasets as well (see Table~\ref{tab:kfoldlmain}). 
	A plausible explanation for the better peformance of GD over $k$-fold CV is that, although CV uses the entire set of samples over the $k$ rounds for both training and validation, it trains the model only on a $(1-1/k)$ portion of the dataset in each individual round. In contrast, GD allows to train the model over the entire dataset in a single run, which therefore results in a better performance on the final unseen (test) data when data is limited. For more experimental results refer to Table~\ref{tab:kfoldl} and Figs.~\ref{fig:kfoldl} and~\ref{fig:mrnet} in Appendix~\ref{app:kfold}.

	(ii) We observe that gradient disparity performs better than $k$-fold cross-validation as an early stopping criterion when data is noisy (see Table~\ref{tab:kfoldnmain}). 
	When the labels of the available data are noisy, the validation set is no longer a reliable estimate of the test set. 
	Nevertheless, and although it is computed over the noisy training set, gradient disparity reflects the performance on the test set quite well\footnote{See for example Fig.~\ref{fig:kfoldn} (left column) where the validation loss fails to estimate the test loss, but where GD (Fig.~\ref{fig:kfoldn} (middle left column)) does signal overfitting correctly.}
	For more experimental results refer to Table~\ref{tab:kfoldn} and Fig.~\ref{fig:kfoldn} in Appendix~\ref{app:kfold}.

	Quite surprisingly, we observe that GD performs better in terms of accuracy than an extension of $k$-fold CV, which we call $k^+$-fold CV, which uses the entire dataset for training with the early stopping signal found by $k$-fold CV (see Table~\ref{tab:kfoldnmain}, where $k=10$ for these settings). More precisely, $k^+$-fold CV is done in 3 steps: (1) perform $k$-fold CV, (2) compute the stopping epoch by tracking the validation loss found in step (1), and (3) retrain the model on the entire dataset and stop at the epoch obtained in step (2). $k^+$-fold CV uses therefore $k+1$ rounds because of step (3), thus one more round than $k$-fold CV, but unlike $k$-fold CV (and similarly to GD), $k^+$-fold CV produces models that are trained on the entire dataset. It is therefore interesting to note that using GD still outperforms $k^+$-fold CV in terms of accuracy (although not in terms of loss).

	\begin{table*}[h]
		\centering
		\caption{Test loss and accuracy when using gradient disparity (GD) and $k$-fold cross-validation (CV) ($k=$10)  as early stopping criteria when the available dataset is noisy: $50\%$ of the available data has random labels. The corresponding curves during training are shown in Fig.~\ref{fig:kfoldn}. The results below are obtained by stopping the optimization when the metric (either validation loss or GD) has increased for 5 epochs from the beginning of training. The last row in each setting, which we call 10$^{+}$-fold CV, refers to the test loss and accuracy reached at the epoch suggested by 10-fold CV, for a network trained on the entire set. 
			Notice that for the CIFAR-100 experiments (the top rows), for computational reasons, the models are trained on only 1.28 k samples of the dataset which explains the very low test accuracy for this experiment. However, for the MNIST experiments (the bottom rows), the models are trained on the entire dataset, and we observe rather high test accuracies. 
		}\label{tab:kfoldnmain}
		\vspace*{1em}
		\begin{tabular}{c|c|c|c}
			\toprule
			Setting & Method    & Test loss & Test accuracy  \\ 
			\midrule
			\multirow{3}{*}{CIFAR-100, ResNet-18} & 10-fold CV  & $5.023_{\pm 0.083}$ & $1.59_{\pm 0.15}$  (top-5: $6.47_{\pm 0.52}$) \\ 
			& GD          & $\mathbf{4.463}_{\pm 0.038}$ & $\mathbf{3.68}_{\pm 0.52}$  (top-5: $\mathbf{15.22}_{\pm 1.24}$) \\
			& 10$^{+}$-fold CV          & $4.964_{\pm 0.057}$ & $1.68_{\pm 0.24}$  (top-5: $7.05_{\pm 0.71}$) \\
			\midrule
			\multirow{3}{*}{MNIST, AlexNet} & 10-fold CV & $0.656_{\pm 0.034}$ & $97.28_{\pm 0.20}$  \\
			& GD  & $0.654_{\pm 0.031}$ & $\mathbf{97.32}_{\pm 0.27}$  \\
			& 10$^{+}$-fold CV  & $\mathbf{0.639}_{\pm 0.029}$ & $97.31_{\pm 0.15}$  \\
			\bottomrule
		\end{tabular}
		
	\end{table*}
	
	\begin{figure}[h]
		\centering
		\begin{subfigure}[b]{1\textwidth}       
			\centering     
			\includegraphics[width=0.5\textwidth, height=0.28\textwidth]{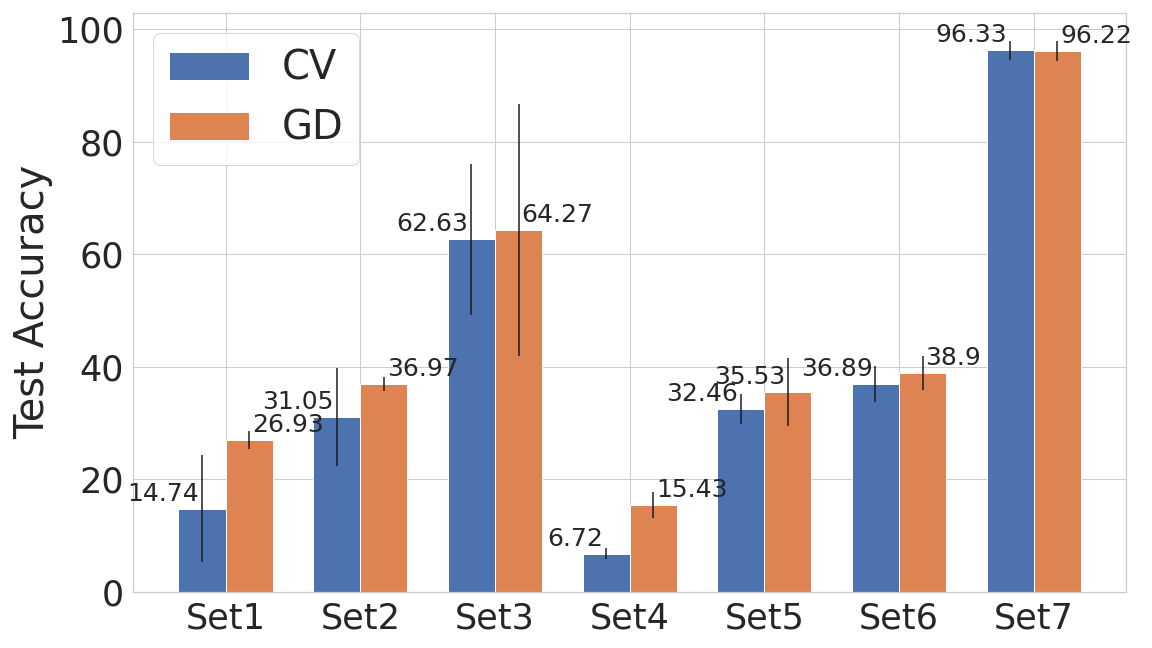}%
		\end{subfigure}
		\caption{Test Accuracy achieved by using GD and $k$-fold CV as early stopping methods in 7 experimental settings (indicated in the x-axis by Set1-7). The result is averaged over 20 choices of the early stopping threshold. The complete set of results are reported in Tables~\ref{tab:robustness} and~\ref{tab:robustnessn}. For the CIFAR-100 experiments (Set1, 4, and 5) the top-$5$ accuracy is reported.
		} \label{fig:senmain}
	\end{figure}

	The metrics used as early stopping criteria, whether they are the validation loss or gradient disparity, are measured on signals that are subject to random fluctuations. As a result, they rely on a pre-defined threshold $p$ (sometimes called \emph{patience} by practitioners) that sets the number of iterations during which the metric increases before the algorithm is stopped. We use two popular thresholds: (t1) the first one is to stop the algorithm when the metric (GD or validation loss) has increased for $p=5$ (possibly non consecutive) epochs from the beginning of training, and (t2) the second is the same as (t1) but the $p=5$ epochs must be consecutive. Both GD and $k$-fold CV might be sensitive to the choice of (t1) or (t2), or even to the value of $p$ itself. It is therefore important to study the sensitivity of an early stopping metric to the choice of the threshold $p$, which is done in Appendix~\ref{app:patthre} for both GD and $k$-fold CV for ten different values of $p \in \{1,\cdots,10\}$ and the two thresholds (t1) and (t2). We observe that GD always gives similar or higher test accuracy than $k$-fold CV for all 20 possible thresholds (see Fig.~\ref{fig:senmain}). More importantly, GD is much more robust to the choice of the early stopping threshold (see Table~\ref{tab:senmain}).

	\begin{table}[h]
		\centering
		\caption{Sensitivity of each method to the choice of the early stopping threshold. The sensitivity is computed from the reported values of Tables~\ref{tab:robustness} and~\ref{tab:robustnessn} according to Eq.~\ref{eq:sen} in the appendix.}\label{tab:senmain}
		\vspace*{1em}
		\begin{tabular}{c|c|c}
			\toprule
			Method & Sensitivity of the Test Accuracy & Sensitivity of the Test Loss  \\ 
			\midrule
			GD & $\mathbf{0.916}$ & $\mathbf{0.886}$\\
			\midrule
			CV & $1.613$ & $1.019$ \\
			\bottomrule
		\end{tabular}
	\end{table}
	
	\begin{figure*}
		\centering
		\begin{subfigure}[b]{0.5\textwidth}     
			\hspace{2em}       
			\includegraphics[width=0.8\textwidth, height=0.4499\textwidth]{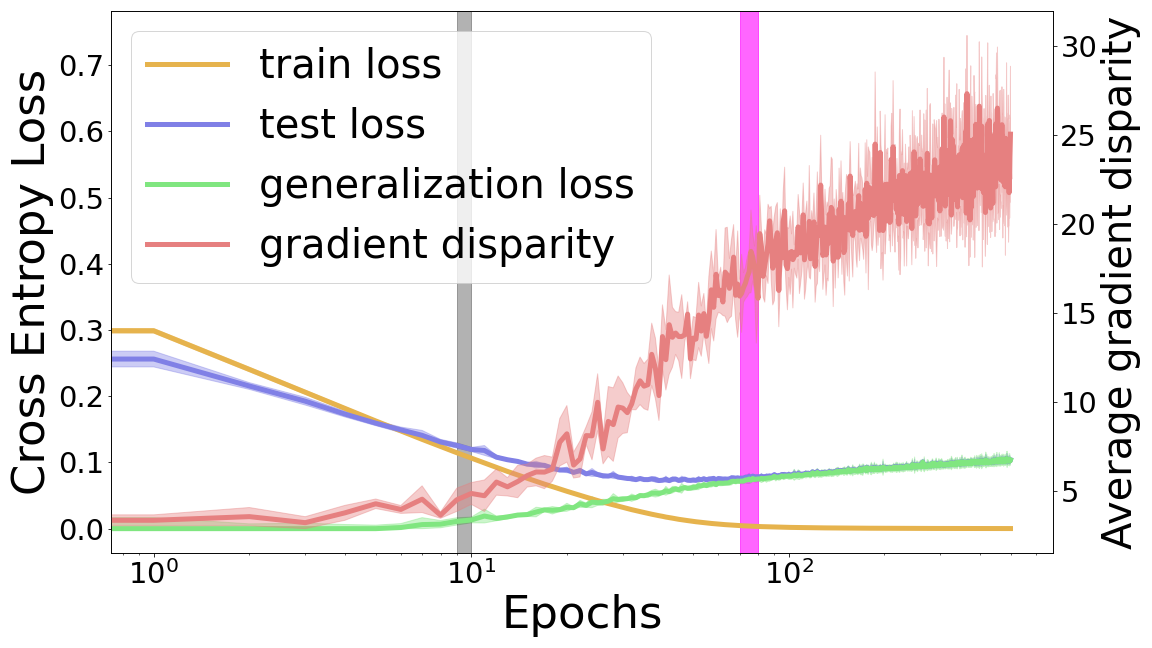}
			\caption{Loss vs gradient disparity}
		\end{subfigure}%
		\begin{subfigure}[b]{0.5\textwidth}
			\centering
			\includegraphics[width=0.8\textwidth, height=0.4499\textwidth]{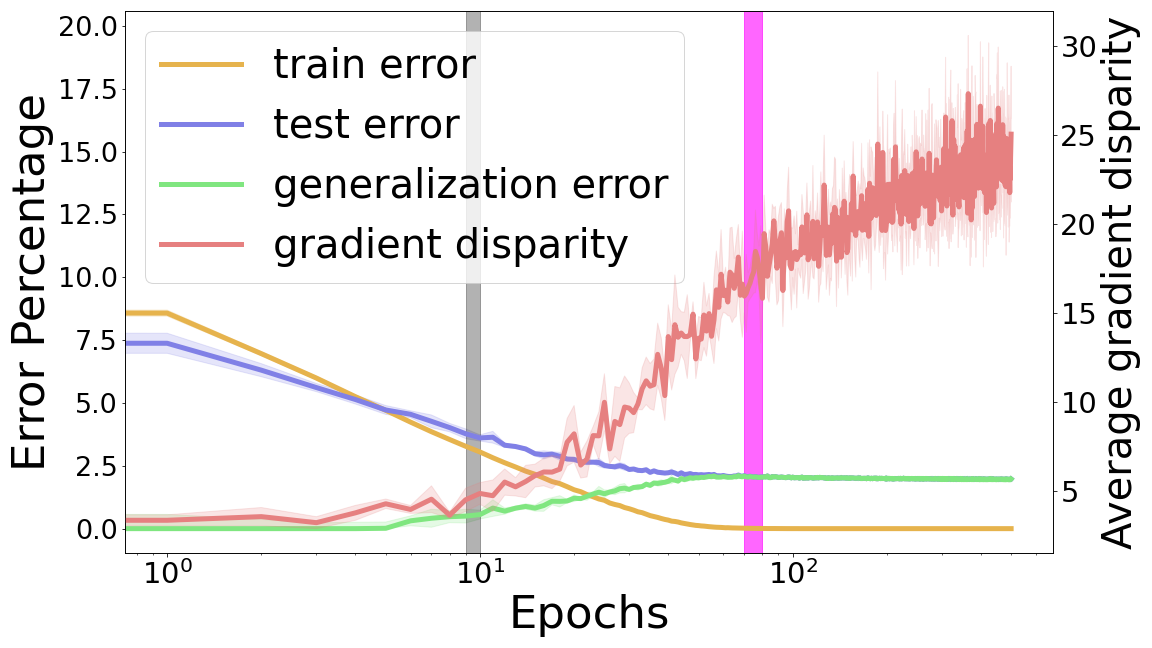}
			\caption{Error vs gradient disparity}
		\end{subfigure}%
		\caption{The cross entropy loss, error percentage, and average gradient disparity during training for a 4-layer fully connected neural network with 500 hidden units trained on the entire MNIST dataset with 0\% label noise. The parameter initialization is the He initialization with normal distribution. Pearson's correlation coefficient $\rho$ between $\overline{\mathcal{D}}$ and generalization loss/error over all the training iterations are $\rho_{\overline{\mathcal{D}}, \text{gen loss}} = 0.967$ and $\rho_{\overline{\mathcal{D}}, \text{gen error}} = 0.734$. The gray vertical bar indicates when GD increases for 5 epochs from the beginning of training. The magenta vertical bar indicates when GD increases for 5 \emph{consecutive} epochs. We observe that the gray bar signals when overfitting is starting, which is when the training and testing curves are starting to diverge. The magenta bar would be a good stopping time, because if we train beyond this point, although the test error remains the same, the test loss would increase, which would result in overconfidence on wrong predictions.}\label{fig:fc_mnist0_main}
	\end{figure*}
	
	When data is abundant and clean, the validation loss is affordable and trustworthy to use as an early stopping signal. GD does also correctly signal overfitting in this case (see for example Fig.~\ref{fig:fc_mnist0_main}). 
	However, when data is limited and/or noisy (which is also when early stopping is particularly important), we observe that the validation loss is costly and unreliable to use as an early stopping signal. In contrast, in these settings, GD does not cost a separate held-out validation set and is a reliable signal of overfitting even in the presence of label noise.
	In practice, the label noise level of a given dataset is in general not known a priori and we do not know whether the size of the dataset is large enough to afford sacrificing a subset for validation. We often do not know whether we are in the former setting, with abundant and clean data, or in the later setting, with limited and/or noisy data. It is therefore important to have a good early stopping criterion that works for both settings. Unlike the validation loss, GD is such a signal.

%% file: proof.tex
	\section{Organization of the Appendix}

This appendix includes sections that are provided here for the sake of completeness and reproducibility (such as Sections~\ref{app:det} and \ref{app:more}) and/or for lack of space in the main paper (such as Sections~\ref{app:opt} and \ref{app:var}). It is structured as follows.

\begin{itemize}
	
	\item Appendix~\ref{app:proof} gives the proof of Theorem~\ref{THM1}, which also uses Hoeffding’s bound recalled in Appendix~\ref{app:add}.
	
	\item Appendix~\ref{app:simple} provides a simple relation between gradient disparity and generalization penalty from linearization.
	
	\item A number of details common to all experiments are provided in Appendix~\ref{app:det}, which discusses in particular how the loss is re-scaled before computing gradient disparity (Appendix~\ref{app:norm}) and how gradient disparity can also be applied to networks trained with the mean square error (Appendix~\ref{app:mse}).
	
	\item A detailed comparison of gradient disparity to $k$-fold cross validation as early stopping criteria is given in Appendix~\ref{app:kfold}, which includes a study on the robustness to the early stopping threshold in Appendix~\ref{app:patthre} (Tables~\ref{tab:robustness},~\ref{tab:robustnessn} and~\ref{tab:patience}) and additional experiments on four image classification datasets (Figs.~\ref{fig:kfoldl},~\ref{fig:kfoldn}, and \ref{fig:mrnet} together with Tables~\ref{tab:kfoldl}, and \ref{tab:kfoldn}).
	
	\item Additional experiments on benchmark datasets are provided in Appendix~\ref{app:more} to study the effect of label noise level, training set size, batch size and network width on the value of gradient disparity. The results, which support the claims made in the main paper, are displayed in Figs.~\ref{fig:random} to \ref{fig:resnet_cifar100}.
	
	\item Besides the vanilla SGD algorithm adopted in the main paper, gradient disparity can be extended to other stochastic optimization algorithms (SGD with momentum, Adagrad, Adadelta, and Adam) as shown in Appendix~\ref{app:opt}.
	
	\item Finally, a detailed comparison with related work is presented in Appendix~\ref{app:var} (Tables~\ref{tab:RW} and \ref{tab:var} together with Figs.~\ref{fig:eb} and \ref{fig:inner_prod}).
	
\end{itemize}

	\normalsize
	\section{Additional Theorem}\label{app:add}
	Hoeffding's bound is used in the proof of Theorem~\ref{THM1}, and Lemma~\hyperref[lem:1]{1} is used in Section~\ref{sec:BGP}.
	\begin{theorem}[Hoeffding's Bound]\label{thm:2}
		Let $Z_1,\cdots,Z_n$ be independent bounded random variables on $\left[a,b\right]$ (i.e., $Z_i \in [a,b]$ for all $1 \leq i \leq n$ with $ -\infty < a \leq b < \infty$). Then
		\begin{equation*}
		\mathbb{P} \left(\frac{1}{n} \sum_{i=1}^{n} \left(Z_i - \mathbb{E}[Z_i]\right) \geq t\right) \leq \exp \left(-\frac{2nt^2}{(b-a)^2}\right) 
		\end{equation*}
		and 
		\begin{equation*}
		\mathbb{P} \left(\frac{1}{n} \sum_{i=1}^{n} \left(Z_i - \mathbb{E}[Z_i]\right) \leq -t\right) \leq \exp \left(-\frac{2nt^2}{(b-a)^2}\right) 
		\end{equation*}
		for all $t \geq 0$.
	\end{theorem}

		\textbf{Lemma 1\label{lem:1}} If $N_1 = \mathcal{N} (\mu_1, \Sigma_1)$ and $N_2 = \mathcal{N} (\mu_2, \Sigma_2)$ are two multivariate normal distributions in $\mathbb{R}^d$, where $\Sigma_1$ and $\Sigma_2$ are positive definite,
		\begin{align*}
		\text{KL}(N_1 || N_2) = \frac{1}{2} \Bigg( \text{tr} \left( \Sigma_2^{-1} \Sigma_1 \right) - d  \Bigg. + (\mu_2 - \mu_1)^T \Sigma_2^{-1} (\mu_2 - \mu_1) +  \Bigg. \ln \left(\frac{\det \Sigma_2}{\det \Sigma_1}\right) \Bigg) .
		\end{align*}
	\section{Proof of Theorem \ref{THM1}}\label{app:proof}
	\begin{proof}
		We compute the upper bound in Eq.~(\ref{eq:mybound}) using a similar approach as in \cite{mcallester2003simplified}. The main challenge in the proof is the definition of a function $X_{S_2}$ of the variables and parameters of the problem, which can then be bounded using similar techniques as in \cite{mcallester2003simplified}. $S_1$ is a batch of points (with size $m_1$) that is randomly drawn from the available set $S$ at the beginning of iteration $t$, and $S_2$ is a batch of points (with size $m_2$) that is randomly drawn from the remaining set $S \setminus S_1$. Hence, $S_1$ and $S_2$ are drawn from the set $S$ without replacement ($S_1 \cap S_2 = \emptyset$). Similar to the setting of \cite{negrea2019information,dziugaite2020role}, as the random selection of indices of $S_1$ and $S_2$ is independent from the dataset $S$, $\sigma(S_1) \indep \sigma(S_2)$, and as a result, $\mathcal{G}_1 \indep \sigma(S_2)$ and $\mathcal{G}_2 \indep \sigma(S_1)$.
		Recall that $\nu_i$ is the random parameter vector at the end of iteration $t$ that depends on $S_i$, for $i \in \{1,2\}$. For a given sample set $S_i$, denote the conditional probability distribution of $\nu_i$ by $Q_{S_i}$. For ease of notation, we represent $Q_{S_i}$ by $Q_i$.

		Let us denote
		\begin{equation}\label{eq:delta}
		\Delta\left(h_{\nu_1}, h_{\nu_2}\right) \triangleq  \left(L_{S_2}(h_{\nu_1}) - L(h_{\nu_1}) \right) 
		- \left(L_{S_2}(h_{\nu_2}) - L(h_{\nu_2}) \right) ,
		\end{equation}
		and 
		\begin{equation}\label{eq:def_f_s}
		X_{S_2} \triangleq   \sup_{Q_1,Q_2} \;\left(\frac{m_2}{2}-1\right)  \mathbb{E}_{\nu_1\sim Q_1} \left[ \mathbb{E}_{\nu_2\sim Q_2} \left[ \left(\Delta\left(h_{\nu_1}, h_{\nu_2}\right)\right)^2\right] \right]   - \text{KL}(Q_2||Q_1) .
		\end{equation}
		Note that $X_{S_2}$ is a random function of the batch $S_2$. Expanding the KL-divergence, we find that
		\begin{align*}
		\left(\frac{m_2}{2}-1\right) \; & \mathbb{E}_{\nu_1\sim Q_1} \left[ \mathbb{E}_{\nu_2\sim Q_2} \left[ \left(\Delta\left(h_{\nu_1}, h_{\nu_2}\right)\right)^2\right] \right] - \text{KL}(Q_2||Q_1) \\ 
		&= \mathbb{E}_{\nu_1\sim Q_1} \left[ \left(\frac{m_2}{2}-1\right) \; \mathbb{E}_{\nu_2\sim Q_2} \left[ \left(\Delta\left(h_{\nu_1}, h_{\nu_2}\right)\right)^2\right] \right.  + \left. \mathbb{E}_{\nu_2\sim Q_2} \left[ \ln{\frac{Q_1(\nu_2)}{Q_2(\nu_2)}} \right] \right]\\
		&\leq \mathbb{E}_{\nu_1\sim Q_1} \left[ \ln{\mathbb{E}_{\nu_2\sim Q_2} \left[e^{(\frac{m_2}{2}-1)\left(\Delta\left(h_{\nu_1}, h_{\nu_2}\right)\right)^2} \frac{Q_1(\nu_2)}{Q_2(\nu_2)}\right]} \right] 
		\\ &=   \mathbb{E}_{\nu_1\sim Q_1} \left[ \ln \mathbb{E}_{\nu_1^\prime\sim Q_1} \left[e^{(\frac{m_2}{2}-1)\left(\Delta\left(h_{\nu_1}, h_{\nu_1^\prime}\right)\right)^2}\right] \right],
		\end{align*}
		where the inequality above follows from Jensen's inequality as logarithm is a concave function. Therefore, again by applying Jensen's inequality
		\begin{equation*}
		X_{S_2} \leq \ln \mathbb{E}_{\nu_1\sim Q_1}  \mathbb{E}_{\nu_1^\prime\sim Q_1} \left[e^{(\frac{m_2}{2}-1)\left(\Delta(h_{\nu_1}, h_{\nu_1^\prime})\right)^2}\right].
		\end{equation*}
		Taking expectations over $S_2$, we have that
		\begin{align}\label{eq:22}
		\mathbb{E}_{S_2}\left[e^{X_{S_2}}\right]  & \leq \mathbb{E}_{S_2} \mathbb{E}_{\nu_1\sim Q_1} \mathbb{E}_{\nu_1^\prime\sim Q_1} \left[e^{(\frac{m_2}{2}-1)\left(\Delta(h_{\nu_1}, h_{\nu_1^\prime})\right)^2}\right] \nonumber\\ 
		&= \mathbb{E}_{\nu_1\sim Q_1} \mathbb{E}_{\nu_1^\prime\sim Q_1} \mathbb{E}_{S_2} \left[e^{(\frac{m_2}{2}-1)\left(\Delta(h_{\nu_1}, h_{\nu_1^\prime})\right)^2}\right] ,
		\end{align}
		where the change of order in the expectation follows from the independence of the draw of the set $S_2$ from $\nu_1 \sim Q_1$ and $\nu_1^\prime \sim Q_1$, i.e., $Q_1$ is $\mathcal{G}_1$-measurable and $\mathcal{G}_1 \indep \sigma(S_2)$.
		
		Now let 
		\begin{equation*}
		Z_i \triangleq  l(h_{\nu_1}(x_i), y_i) - l(h_{\nu_1^\prime}(x_i), y_i),
		\end{equation*}
		for all $1 \leq i \leq m_2$. Clearly, $Z_i \in [-1,1]$ and because of Eqs.~(\ref{eq:train_loss}) and of the definition of $\Delta$ in Eq. (\ref{eq:delta}),
		\begin{equation*}
		\Delta\left(h_{\nu_1}, h_{\nu_1^\prime}\right)  = \frac{1}{m_2} \sum_{i=1}^{m_2} \left(Z_i - \mathbb{E}[Z_i]\right).
		\end{equation*}
		Hoeffding's bound (Theorem \ref{thm:2}) implies therefore that for any $t \geq 0$,
		\begin{equation}\label{eq:deltau}
		\mathbb{P}_{S_2} \left(\lvert \Delta\left(h_{\nu_1}, h_{\nu_1^\prime}\right)\rvert \geq t \right) \leq 2 e^{- \frac{m_2}{2} t^2} .
		\end{equation}
		Denoting by $p(\Delta)$ the probability density function of $\lvert \Delta\left(h_{\nu_1}, h_{\nu_1^\prime}\right) \rvert$, inequality~(\ref{eq:deltau}) implies that for any $t \geq 0$,
		\begin{equation}\label{eq:upperp}
		\int_{t}^{\infty} p(\Delta) d \Delta \leq 2 e^{- \frac{m_2}{2} t^2} .
		\end{equation}
		The density $\tilde{p}(\Delta)$ that maximizes $\int_{0}^{\infty} e^{(\frac{m_2}{2}-1)\Delta^2} p(\Delta) d \Delta$ (the term in the first expectation of the upper bound of Eq.~(\ref{eq:22})
		), is the density achieving equality in~(\ref{eq:upperp}), which is ${\tilde{p}(\Delta) =  2 m_2 \Delta e^{-\frac{m_2}{2}\Delta^2}}$.
		As a result,
		\begin{align*}
		\mathbb{E}_{S_2}\left[e^{(\frac{m_2}{2}-1)\Delta^2}\right] &\leq \int_{0}^{\infty} e^{(\frac{m_2}{2}-1)\Delta^2} 2 m_2 \Delta e^{-\frac{m_2}{2}\Delta^2} d \Delta 
		\\&= \int_{0}^{\infty} 2 m_2 \Delta e^{-\Delta^2} d \Delta 
		= m_2 
		\end{align*}
		and consequently, inequality (\ref{eq:22}) becomes
		\begin{equation*}
		\mathbb{E}_{S_2}\left[e^{X_{S_2}}\right] \leq m_2 .
		\end{equation*}
		Applying Markov's inequality on $X_{S_2}$, we have therefore that for any $0 < \delta \leq 1$,
		\begin{align*}
		\mathbb{P}_{S_2} \left[X_{S_2} \geq \ln\frac{2m_2}{\delta} \right] = \mathbb{P}_{S_2}\left[e^{X_{S_2}} \geq \frac{2m_2}{\delta}\right] \leq \frac{\delta}{2m_2}\mathbb{E}_{S_2}\left[e^{X_{S_2}}\right] \leq \frac{\delta}{2} .
		\end{align*}
		Replacing $X_{S_2}$ by its expression defined in Eq.~(\ref{eq:def_f_s}), the previous inequality shows that with probability at least $1-\delta/2$
		\begin{equation*}
		\left(\frac{m_2}{2}-1\right) \; \mathbb{E}_{\nu_1\sim Q_1} \mathbb{E}_{\nu_2\sim Q_2} \left[ \left(\Delta(h_{\nu_1}, h_{\nu_2})\right)^2\right] - \text{KL}(Q_2||Q_1) \leq \ln{\frac{2m_2}{\delta}}  .
		\end{equation*}
		Using Jensen's inequality and the convexity of $\left(\Delta(h_{\nu_1}, h_{\nu_2})\right)^2$, and assuming that $m_2 > 2$, we therefore have that with probability at least $1 - \delta/2$,
		\begin{align*}
		\left(\mathbb{E}_{\nu_1\sim Q_1} \mathbb{E}_{\nu_2 \sim Q_2} [\Delta\left(h_{\nu_1}, h_{\nu_2}\right)]\right)^2  &\leq \mathbb{E}_{\nu_1\sim Q_1} \mathbb{E}_{\nu_2 \sim Q_2} \left[\left(\Delta\left(h_{\nu_1}, h_{\nu_2}\right)\right)^2\right] 
		\\ & \leq \frac{\text{KL}(Q_2||Q_1) + \ln{\frac{2m_2}{\delta}}}{\frac{m_2}{2}-1}  .
		\end{align*}
		Replacing $\Delta(h_{\nu_1}, h_{\nu_2})$ by its expression Eq.~(\ref{eq:delta}) in the above inequality, yields that with probability at least $1-\delta/2$ over the choice of the sample set $S_2$, 
		\begin{equation}\label{eq:1}
		\mathbb{E}_{\nu_1 \sim Q_1} \left[L_{S_2}(h_{\nu_1}) - L(h_{\nu_1}) \right] \\ \leq  \mathbb{E}_{\nu_2 \sim Q_2} \left[L_{S_2}(h_{\nu_2}) - L(h_{\nu_2}) \right] + \sqrt{\frac{2 \text{KL}(Q_2||Q_1) + 2 \ln{\frac{2m_2}{\delta}}}{m_2-2}}.
		\end{equation}
		Similar computations with $S_1$ and $S_2$ switched, and considering that $m_1 > 2$, yields that with probability at least $1-\delta/2$ over the choice of the sample set $S_1$,
		\begin{equation}\label{eq:2}
		\mathbb{E}_{\nu_2 \sim Q_2} \left[L_{S_1}(h_{\nu_2}) - L(h_{\nu_2}) \right] \\ \leq \mathbb{E}_{\nu_1 \sim Q_1} \left[L_{S_1}(h_{\nu_1}) - L(h_{\nu_1}) \right]  + \sqrt{\frac{2 \text{KL}(Q_1||Q_2) + 2 \ln{\frac{2m_1}{\delta}}}{m_1-2}}.
		\end{equation}
		The events in Eqs.~(\ref{eq:1}) and (\ref{eq:2}) jointly hold with probability at least $1-\delta$ over the choice of the sample sets $S_1$ and $S_2$ (using the union bound and De Morgan's law), and by adding the two inequalities we therefore have
		\begin{multline*}
		\mathbb{E}_{\nu_1 \sim Q_1} \left[L_{S_2}(h_{\nu_1})\right] + \mathbb{E}_{\nu_2 \sim Q_2} \left[L_{S_1}(h_{\nu_2})\right] 
		 \leq  \; \mathbb{E}_{\nu_2 \sim Q_2} \left[L_{S_2}(h_{\nu_2})\right] + \mathbb{E}_{\nu_1 \sim Q_1} \left[L_{S_1}(h_{\nu_1})\right] \\
		+ \sqrt{\frac{2\text{KL}(Q_2||Q_1) + 2\ln{\frac{2m_2}{\delta}}}{{m_2}-2}} 
		+ \sqrt{\frac{2\text{KL}(Q_1||Q_2) + 2\ln{\frac{2m_1}{\delta}}}{{m_1}-2}} ,
		\end{multline*}
		which concludes the proof.
	\end{proof}

	\section{A Simple Connection Between Generalization Penalty and \\ Gradient Disparity}\label{app:simple}
	In this section, we present an alternative and much simpler connection between the notions of generalization penalty and gradient disparity than the one presented in Sections~\ref{sec:pre} and \ref{sec:BGP}. Recall that each update step of the mini-batch gradient descent is written as $w_i = w - \gamma g_i$ for $i \in \{1,2\}$. By applying a first order Taylor expansion over the loss, we have
	\begin{equation*}
		L_{S_1} (h_{w_1}) = L_{S_1} (h_{w-\gamma g_1}) \approx L_{S_1}(h_w) - \gamma g_1 \cdot g_1 . 
	\end{equation*}
	The generalization penalties $\mathcal{R}_1$ and $\mathcal{R}_2$ would therefore be 
	\begin{align*}
	\mathcal{R}_1 =  L_{S_1} (h_{w_2}) - L_{S_1} (h_{w_1}) 
	\approx \gamma g_1 \cdot (g_1 - g_2),
	\end{align*}
	and
	\begin{align*}
	\mathcal{R}_2 =  L_{S_2} (h_{w_1}) - L_{S_2} (h_{w_2}) 
	\approx \gamma g_2 \cdot (g_2 - g_1),
	\end{align*}
	respectively. Consequently,
	\begin{align*}
	\mathcal{R}_1 + \mathcal{R}_2 \approx \gamma \norm{g_1 - g_2}_2^2 . 
	\end{align*}
	This derivation requires the loss function to be (approximately) linear near parameter vectors $w_1$ and $w_2$, which does not necessarily hold. Therefore, in the main paper we only focus on the connection between generalization penalty and gradient disparity via Theorem~\ref{THM1}, which does not require such an assumption. Nevertheless, it interesting to note that this simple derivation recovers the connection between generalization penalty and gradient disparity.
	
	\section{Common Experimental Details}\label{app:det}
	
	The training objective in our experiments is to minimize the cross-entropy loss, and both the cross entropy and the error percentage are displayed in the figures. The training error is computed using Eq.~(\ref{eq:train_loss}) over the training set. The empirical test error also follows Eq.~(\ref{eq:train_loss}) but it is computed over the test set. 
	The generalization loss (respectively, error) is the difference between the test and the training cross entropy losses (resp.,  classification errors).
	The batch size in our experiments is 128 unless otherwise stated. The SGD learning rate is $\gamma=0.01$ and no momentum is used (unless otherwise stated). All the experiments took at most few hours on one Nvidia Titan X Maxwell GPU. All the reported values throughout the paper are an average over at least 5 runs. 
	
	To present results throughout the training, in the x-axis of figures, both epoch and iteration are used: an epoch is the time spent to pass through the entire dataset, and an iteration is the time spent to pass through one batch of the dataset. Thus, each epoch has $B$ iterations, where $B$ is the number of batches. The convolutional neural network configurations we use are: AlexNet \cite{krizhevsky2012imagenet}, VGG \cite{simonyan2014very} and ResNet \cite{he2016deep}. In those experiments with varying width, we use a scaling factor to change both the number of channels and the number of hidden units in convolutional and fully connected layers, respectively. The default configuration is with scaling  factor $=1$. For the experiments with data augmentation (Fig.~\ref{fig:DA}) we use random crop with padding $= 4$ and random horizontal flip with probability~$= 0.5$.
	
	In experiments with a random labeled training set, we modify the dataset similarly to \cite{chatterjee2020coherent}. For a fraction of the training samples, which is the amount of noise ($0\%$, $25\%$, $50\%$, $75\%$, $100\%$), we choose the labels at random. For a classification dataset with a number $k$ of classes, if the label noise is $25\%$, then on average ${75\% + 25\% * 1/k}$ of the training points still have the correct label.

%% file: details.tex
\subsection{Re-scaling the Loss}\label{app:norm}
\addtocounter{footnote}{-1}
\begin{wrapfigure}[24]{R}{0.5\textwidth}
	\includegraphics[width=0.5\textwidth, height=0.28\textwidth]{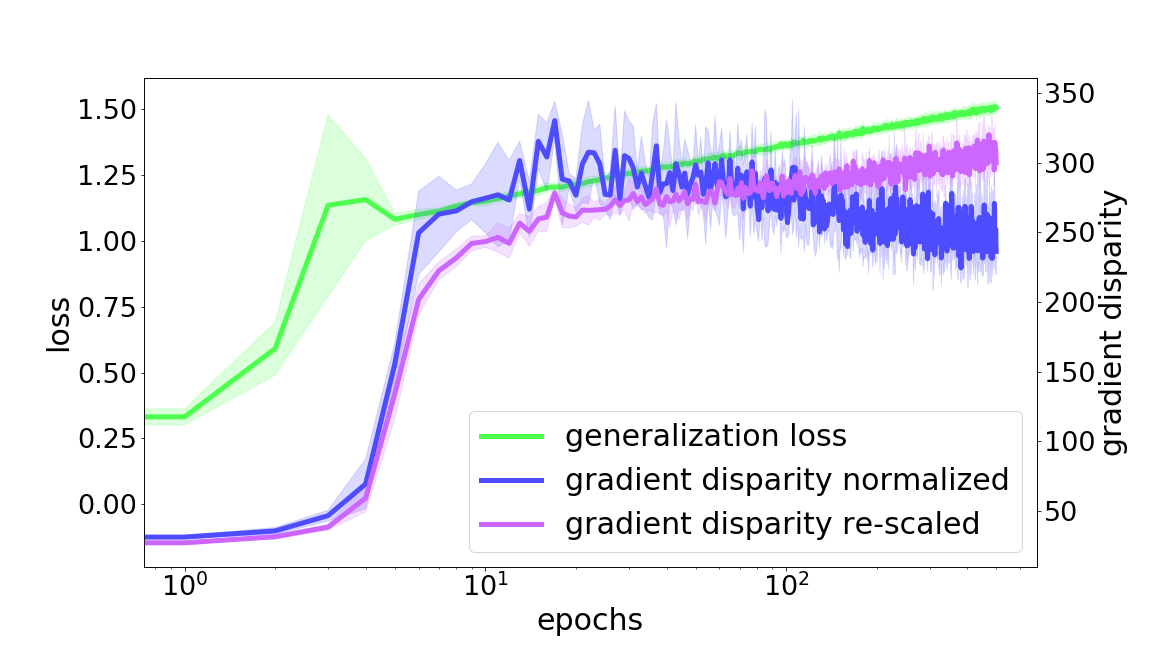}
	\caption{Normalizing versus re-scaling loss before computing average gradient disparity $\overline{\mathcal{D}}$ for a VGG-11 trained on 12.8 k points of the CIFAR-10 dataset\protect\footnotemark. For this experiment, Pearson's correlation coefficient between gradient disparity re-scaled (our chosen metric) and the test loss is $0.91$. If we would have instead normalized the loss values the correlation would be $0.88$. If we would have re-scaled with respect to the gradients (instead of the loss), the correlation would be $0.79$. 
	}\label{fig:norm}
\end{wrapfigure}

Let us track the evolution of gradient disparity (Eq.~(\ref{eq:bgd})) during training. As training progresses, the training losses of all the batches start to decrease when they get selected for the parameter update. Therefore, the value of gradient disparity might decrease, not necessarily because the distance between the two gradient vectors is decreasing, but because the value of each gradient vector is itself decreasing. To avoid this, a re-scaling or normalization is needed to compare gradient disparity at different stages of training.

If we perform a re-scaling with respect to the gradient vectors, then the gradient disparity between two batches $S_1$ and $S_2$ would be $\norm{g_1/\text{std}(g_1)  - g_2/\text{std}(g_2) }_2$, where $\text{std}(g_i)$ is the standard deviation of the gradients within batch $S_i$ for $i \in \{1,2\}$. However, such a re-scaling would also absorb the variations of $g_1$ and $g_2$ with respect to each other. That is, if after an iteration, $g_1$ is scaled by a factor $\alpha<1$, while $g_2$ remains unchanged, this re-scaling would leave the gradient disparity unchanged, although the performance of the network has improved only on $S_1$ and not on $S_2$, which might be a signal of overfitting.

\footnotetext{Note that in this figure, both the gradient disparity re-scaled and the generalization loss are increasing from the very first epoch. If we would use gradient disparity as an early stopping criterion, optimization would stop at epoch 5 and we would have a 0.36 drop in the test loss value, compared to the loss reached when the model achieves 0 training loss.}

We therefore propose to normalize the loss values instead, before computing gradient disparity, so that the initial losses of two different iterations would have the same scale. We can normalize the loss values by
\begin{equation*}
{L_{S_j} = \frac{1}{m_j} \sum_{i=1}^{m_j} \frac{l_i-\text{Min}_i\left(l_i\right)}{\text{Max}_i\left(l_i\right)-\text{Min}_i\left(l_i\right)}} ,
\end{equation*}
where with some abuse of notation, $l_i$ is the cross entropy loss for the data point $i$ in the batch $S_j$. 
However, this normalization might be sensitive to outliers, making the bulk of data end up in a very narrow range within 0 and 1, and degrading in turn the accuracy of the signal of overfitting.
Re-scaling is usually less sensitive to outliers in comparison with normalization, it leads to loss values that are given by
\begin{equation*}
{L_{S_j} = \frac{1}{m_j} \sum_{i=1}^{m_j} \frac{l_i}{\text{std}_i\left(l_i\right)}} .
\end{equation*}

We experimentally compare these two ways of computing gradient disparity in Fig.~\ref{fig:norm}. Both the re-scaled and normalized losses might get unbounded, if within each batch the loss values are very close to each other. However, in our experiments, we do not observe gradient disparity becoming unbounded either way. We observe that the correlation between gradient disparity and the test loss is the highest if we re-scale the loss values before computing gradient disparity. This is therefore how we compute gradient disparity in all experiments presented in the paper. Note that this re-scaling does not affect the training algorithm, since it is only used to compute the gradient disparity metric (we do not perform loss re-scaling before \verb|opt.step()|).

\subsection{The Hyper-parameter $s$}\label{app:s}
\begin{wrapfigure}[16]{R}{0.5\textwidth}
	\includegraphics[width=0.5\textwidth, height=0.28\textwidth]{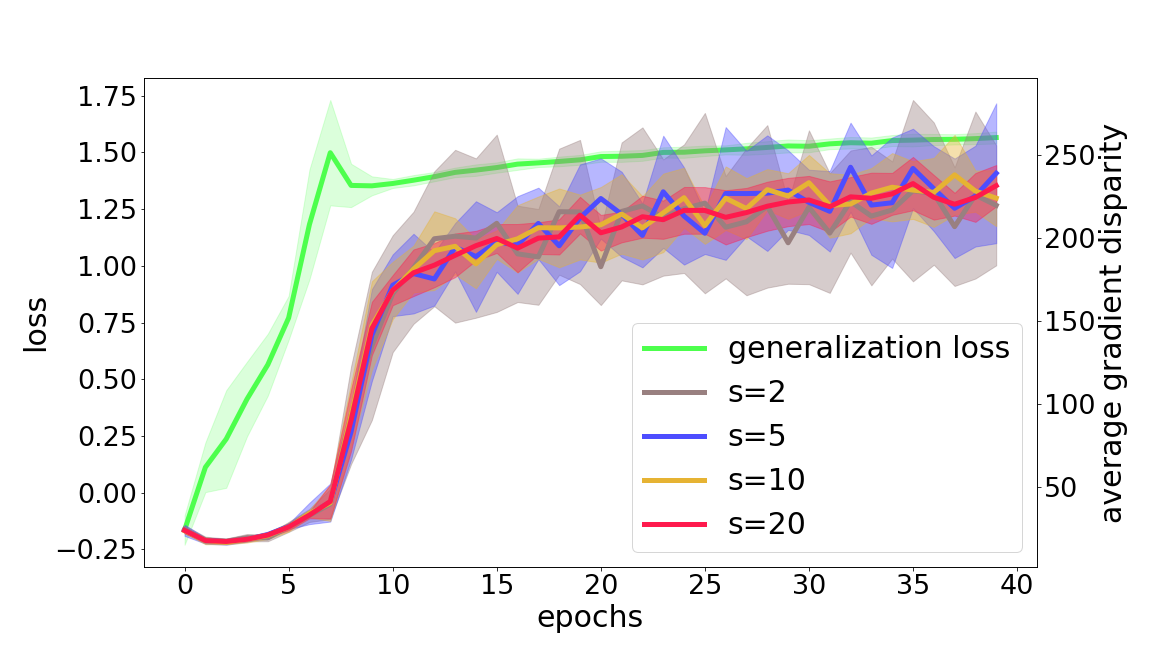}
	\caption{Average gradient disparity for different averaging parameter $s$ for a ResNet-18 that has been trained on 12.8k points of the CIFAR-10 dataset.}\label{fig:s}
\end{wrapfigure}

In this section, we briefly study the choice of the size $s$ of the subset of batches to compute the average gradient disparity $$\overline{\mathcal{D}} = \frac{1}{s(s-1)} \sum_{i=1}^{s}\sum\limits_{\substack{j=1 , j\neq i }}^s \mathcal{D}_{i,j} .$$ Fig.~\ref{fig:s} shows the average gradient disparity when averaged over number $s$ of batches\footnote{In the setting of Fig.~\ref{fig:s}, if we use gradient disparity as an early stopping criterion, optimization would stop at epoch 9 and we would have a 0.28 drop in the test loss value compared to the loss reached when the model achieves 0 training loss.}. When $s=2$, gradient disparity is the $\ell_2$ norm distance of the gradients of two randomly selected batches and has a quite high variance. Although with higher values of $s$ the results have lower variance, computing it with a large value of $s$ is more computationally expensive (refer to Appendix~\ref{app:kfold} for more details).
Therefore, we find the choice of $s=5$ to be sufficient enough to track overfitting; in all the experiments reported in this paper, we use $s=5$.

\clearpage
\subsection{The Surrogate Loss Function}\label{app:mse}

Cross entropy has been shown to be better suited for computer-vision classification tasks, compared to mean square error~\cite{kline2005revisiting,hui2020evaluation}. Hence, we choose the cross entropy criterion for all our experiments to avoid possible pitfalls of the mean square error, such as not tracking the confidence of the predictor. 

\cite{soudry2018implicit} argues that when using cross entropy, as training proceeds, the magnitude of the network parameters increases. This can potentially affect the value of gradient disparity. Therefore, we compute the magnitude of the network parameters over iterations in various settings.
We observe that this increase is very low both at the end of the training and, more importantly, at the time when gradient disparity signals overfitting (denoted by GD epoch in Table~\ref{tab:norm}). Therefore, it is unlikely that the increase in the magnitude of the network parameters affects the value of gradient disparity.

\begin{figure*}[h]
	\begin{subfigure}[b]{\textwidth}
		\includegraphics[width=0.33\textwidth, height=0.1856\textwidth]{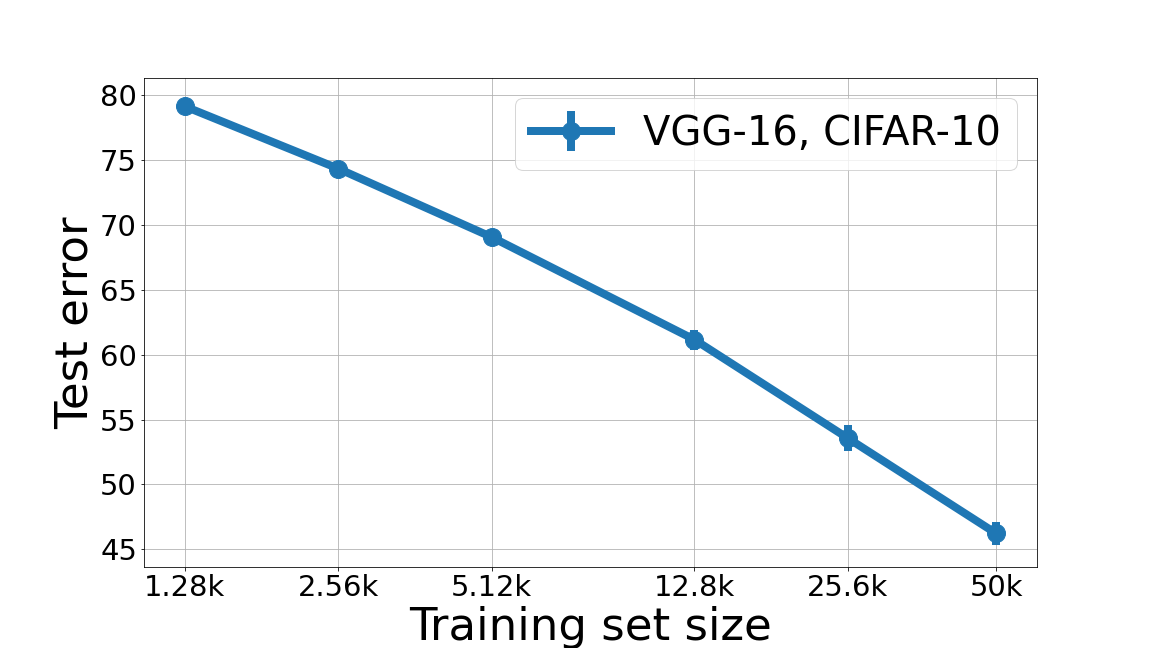}%
		\includegraphics[width=0.33\textwidth, height=0.1856\textwidth]{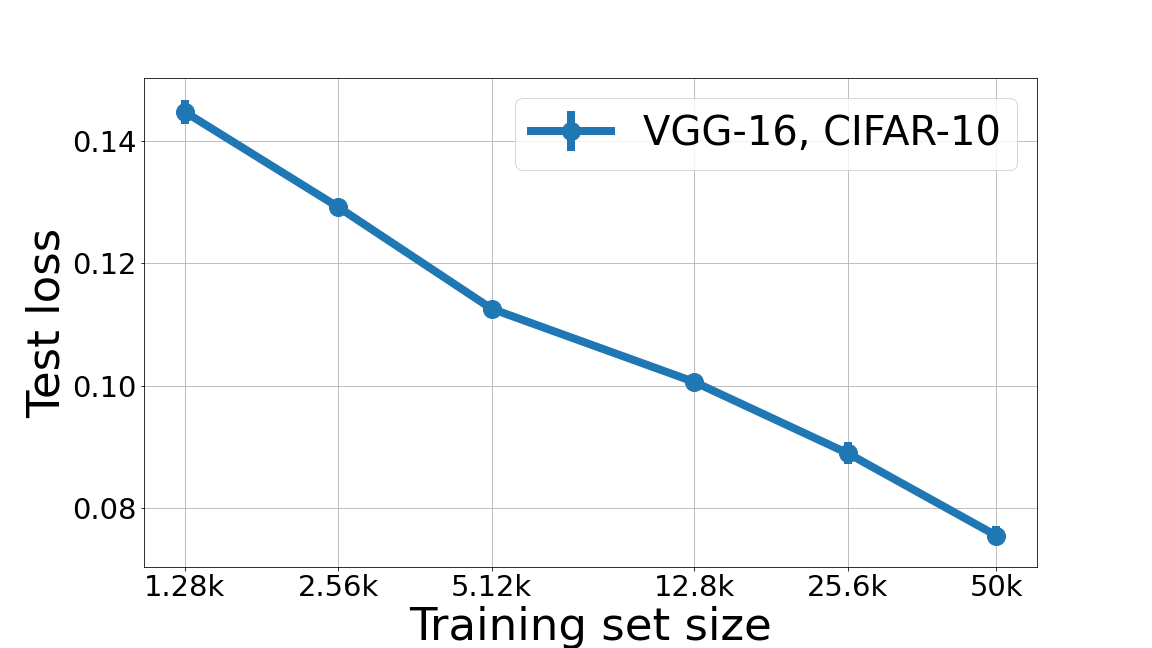}%
		\includegraphics[width=0.33\textwidth, height=0.1856\textwidth]{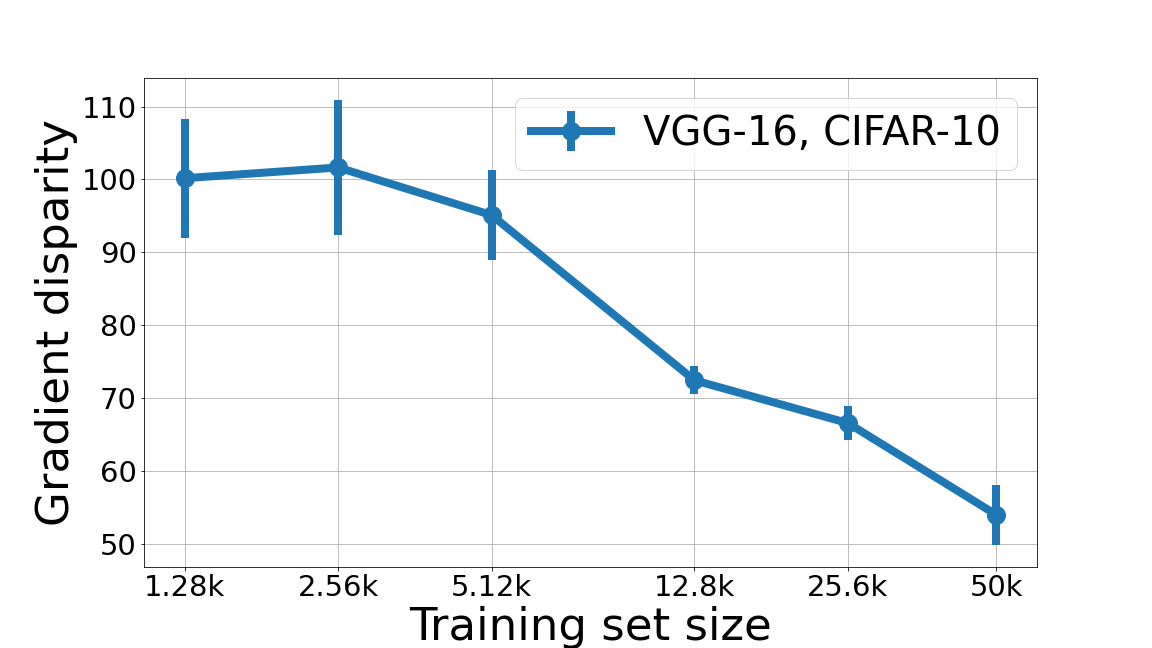}
	\end{subfigure}
	\caption{Test error (TE), test loss (TL), and gradient disparity ($\overline{\mathcal{D}}$) for VGG-16 trained with different training set sizes to minimize the mean square error criterion on the CIFAR-10 dataset. The Pearson correlation coefficient between TE and $\overline{\mathcal{D}}$ and between TL and $\overline{\mathcal{D}}$ are $\rho_{\overline{\mathcal{D}}, \text{TE}}=0.976$ and $\rho_{\overline{\mathcal{D}}, \text{TL}}=0.943$, respectively. 
	}\label{fig:train_size_mse}
\end{figure*}

Furthermore, we examine gradient disparity for models trained on the mean square error, instead of the cross entropy criterion. 
We observe a high correlation between gradient disparity and test error/loss (Fig.~\ref{fig:train_size_mse}), which is consistent with the results obtained using the cross entropy criterion. The applicability of gradient disparity as a generalization metric is therefore not limited to settings with the cross entropy criterion.

\begin{table}[h]
	\centering
	\caption{The ratio of the magnitude of the network parameter vector at epoch $t$ to the magnitude of the network parameter vector at epoch $0$, for $t \in \{0,\text{GD},200\}$, where GD stands for the epoch when gradient disparity signals to stop the training. Setting (1): AlexNet, MNIST, (2): AlexNet, MNIST, $50\%$ random, (3): VGG-16, CIFAR-10, and (4): VGG-16, CIFAR-10, $50\%$ random. }\label{tab:norm}
	\vspace*{1em}
	\begin{tabularx}{0.45\textwidth}{@{} l|C|C|C @{}}
		\toprule
		Setting & at epoch $0$ & at GD epoch & at epoch $200$  \\ 
		\midrule
		($1$) & $1$  & $1.00034$ & $1.00123$ \\
		\midrule
		($2$)    & $1$ & $1.00019$ & $1.00980$ \\ 
		\midrule
		($3$) & $1$ & $1.00107$ & $1.00127$  \\
		\midrule
		($4$) & $1$ & $1.00222$ & $1.00233$  \\ 
		\bottomrule
	\end{tabularx}
\end{table}

%% file: kfoldapp.tex
	\clearpage
	\newpage
	\section{$k$-fold Cross-Validation}\label{app:kfold}
	$k$-fold cross-validation (CV) splits the available dataset into $k$ sets, training on $k-1$ of them and validating on the remaining one. This is repeated $k$ times so that every set is used once as the validation set. Each experiment out of $k$ folds can itself be viewed as a setting where the available data is split into a training and a validation set. Early stopping can then be done by evaluating the network performance on the validation set and stopping the training as soon as there is an increase in the value of the validation loss. However, in practice, the validation loss curve is not necessarily smooth (as can be clearly observed in our experiments throughout the paper), and therefore as discussed in \cite{prechelt1998early,lodwich2009evaluation}, there is no obvious early stopping rule (threshold) to obtain the minimum value of the generalization error.

	\subsection{Early Stopping Threshold}\label{app:patthre}
	In this paper, we adapt two different early stopping thresholds: (t1) stop training when there are $p=5$ (consecutive or nonconsecutive) increases in the value of the early stopping metric (the hyper-parameter $p$ is commonly referred to as the “patience” parameter among practitioners), which is indicated by the gray vertical bars in Figs.~\ref{fig:kfoldl} and \ref{fig:kfoldn}, and (t2) stop training when there are $p$ consecutive increases in the value of the early stopping metric, which is indicated by the magenta vertical bars in Figs.~\ref{fig:kfoldl} and \ref{fig:kfoldn}. When there are either low variations or a sharp increase in the value of the metric, the two coincide (for instance, in Fig.~\ref{fig:kfoldl}~(b)~(middle left)). For $k$-fold CV, the early stopping metric is the validation loss, and for our proposed method, the early stopping criterion is gradient disparity (GD).
	
	Which exact patience parameter to choose as an early stopping threshold, or whether it should include non-consecutive increases or not, are indeed interesting questions \cite{prechelt1998early}, which do not have a definite answer to date even for $k$-fold CV. Tables~\ref{tab:robustness} and~\ref{tab:robustnessn} give the results obtained by 20 different early stopping thresholds for both $k$-fold CV (shown on the left tables) and GD (shown on the right tables). We also give the best, mean and standard deviation of test loss and test accuracy across all thresholds. In the following two paragraphs, we summarize the findings of these two tables.
	
\begin{figure}[h]
		\centering
		\begin{subfigure}[b]{1\textwidth}       
			\centering     
			\includegraphics[width=0.7\textwidth, height=0.392\textwidth]{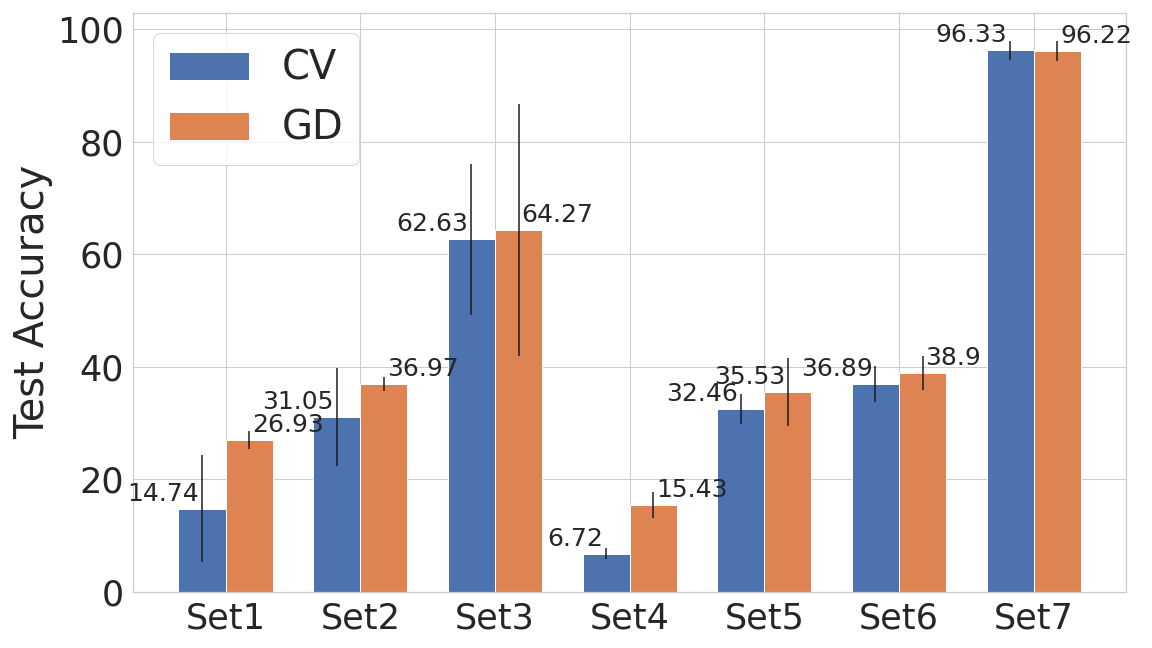}%
		\end{subfigure}
		\caption{Test Accuracy achieved by using GD and $k$-fold CV as early stopping methods in 7 experimental settings (indicated in the x-axis by Set1-7). The result is averaged over 20 choices of the early stopping threshold. The complete set of results are reported in Tables~\ref{tab:robustness} and~\ref{tab:robustnessn}. For the CIFAR-100 experiments (Set1, 4, and 5) the top-$5$ accuracy is reported.
		} \label{fig:sen}
	\end{figure}

	\paragraph{Performance} In Fig.~\ref{fig:sen}, we observe that the test accuracy (averaged over 20 thresholds) is higher when using GD as an early stopping criterion, than $k$-fold CV. Note that beforehand we do not have access to the test set to choose the best possible threshold, hence in Fig.~\ref{fig:sen} the average test accuracy is reported over all thresholds. However, even if we did have the test set to choose the best threshold, we can still observe from Tables~\ref{tab:robustness} and~\ref{tab:robustnessn} that GD either performs comparably to CV, or that it significantly outperforms CV.
	
	\paragraph{Sensitivity to Threshold} Ideally, we would like to have a robust metric that does not strongly depend on the early stopping threshold. 
	To compute the sensitivity of each method to the choice of the threshold, we compute 
	\begin{equation}\label{eq:sen}
		\text{Sensitivity to the threshold} = {\sum_{i=1}^{7} \text{std}(\text{Set}_i) / \text{Mean}(\text{Set}_i)} ,
	\end{equation}
	where $\text{Mean}(\text{Set}_i)$ and $\text{std}(\text{Set}_i)$ are the mean and standard deviation of the test accuracy/loss across different thresholds, respectively, of setting $i$ across the early stopping thresholds, which are reported in Tables~\ref{tab:robustness} and~\ref{tab:robustnessn}. The lower the sensitivity is, the more the method is robust to the choice of the early stopping threshold. In Table~\ref{tab:senmain}, we observe that GD is less sensitive and more robust to the choice of the early stopping threshold than $k$-fold CV, which is another advantage of GD over CV.
	This can also be observed from our figures: In most of the experiments (more precisely, in 5 out of 7 experiments of Figs.~\ref{fig:kfoldl} and \ref{fig:kfoldn}), GD is not sensitive to the choice of the threshold (see Figs. ~\ref{fig:kfoldl}~(a), ~\ref{fig:kfoldl}~(b), ~\ref{fig:kfoldn}~(a), ~\ref{fig:kfoldl}~(b) and ~\ref{fig:kfoldl}~(c) on the 2nd column, where the gray and magenta bars almost coincide). In contrast, $k$-fold CV is more sensitive (see for example the leftmost column of Figs. ~\ref{fig:kfoldl}~(a) and ~\ref{fig:kfoldl}~(b), where the gray and magenta bars are very far away when using $k$-fold CV). In the other 2 experiments (Figs.~\ref{fig:kfoldl}~(c) and ~\ref{fig:kfoldn}~(d), both with the MNIST dataset), the thresholds (t1) and (t2) do not coincide for neither GD nor $k$-fold CV. In Table~\ref{tab:patience}, we further study these two settings: we provide the test accuracy for experiments of Figs.~\ref{fig:kfoldl}~(c) and ~\ref{fig:kfoldn}~(d) (both with the MNIST dataset), for different values of $p$. 
	We again observe that even if we optimize $p$ for $k$-fold CV (reported in bold in Table~\ref{tab:patience}), GD still outperforms $k$-fold CV.

		\begin{table*}
		\centering
		\caption{The test accuracy, test loss, top-5 accuracy, and stopping epoch obtained by using $k$-fold cross-validation (CV) (left columns (a), (c), and (e)) and by using gradient disparity (GD) (right columns (b), (d), and (f)) as early stopping criteria for different patience values and for different thresholds (t1) and (t2). $(\text{t1}_p)$: training is stopped after $p$ increases in the value of the validation loss in $k$-fold CV, and of GD, respectively. $(\text{t2}_p)$: training is stopped after $p$ consecutive increases in the value of the validation loss in $k$-fold CV, and of GD, respectively. For the rest of the experiments we only report the best values, mean and standard deviation (std) over all thresholds. $-^{a}$: The epoch to have the best test loss does not coincide with the epoch to have the best test accuracy. $-^{b}$: The metric does not have $p$ consecutive increases during training.}\label{tab:robustness}
		\vspace*{1em}
		\resizebox{0.55\textheight}{!}{
		\begin{subtable}{0.48\linewidth}
				\centering
					\begin{tabularx}{\textwidth}{@{} l|C|C|C|C @{}}
						\toprule
						Threshold & Loss & ACC & top-5 ACC & Epoch   \\ 
						\midrule
						$(\text{t1}_1)$ & $4.63$ & $0.98$ & $5.02$ & $1$ \\
						$(\text{t1}_2)$  & $4.65$ & $0.98$ &  $5.0$ & $2$ \\
						$(\text{t1}_3)$ & $4.65$ &  $1.19$  &  $5.99$ & $3$ \\
						$(\text{t1}_4)$ & $4.25$ &  $6.73$  &  $21.47$ & $12$ \\
						$(\text{t1}_5)$ & $4.25$ &  $6.79$  &  $22.19$ & $14$ \\ 
						$(\text{t1}_6)$ & $4.23$ &  $7.46$  &  $22.48$ & $17$  \\
						$(\text{t1}_7)$ & $4.25$ &  $7.22$  &  $22.18$ & $18$  \\
						$(\text{t1}_8)$ & $4.21$ &  $7.69$  &  $23.50$ & $21$ \\
						$(\text{t1}_9)$ & $4.23$ &  $7.94$  &  $23.72$ & $22$ \\
						$(\text{t1}_{10})$  & $4.27$ &  $7.58$  &  $22.84$ & $24$ \\
						$(\text{t2}_1)$ & $4.63$  &  $0.98$  &  $5.02$ & $1$ \\
						$(\text{t2}_2)$ & $4.65$  &  $0.98$  &  $5.0$ & $2$ \\
						$(\text{t2}_3)$ & $4.65$  &  $1.19$  &  $5.99$ & $3$ \\
						$(\text{t2}_4)$ & $4.12 $ &  $9.50 $ & $ 26.71$ & $45$ \\
						$(\text{t2}_5)$ & $4.12 $ &  $9.45 $ & $ 26.32$ & $46$ \\
						$(\text{t2}_6)$ & $4.19 $ &  $9.51$  &  $26.43$ & $314$ \\
						$(\text{t2}_7)$ & $4.20$ &  $9.54$  &  $26.47$ & -\footnote[2]{}\\
						$(\text{t2}_8)$ &  - &  -  &  - & - \\
						$(\text{t2}_9)$ &  - &  -  &  - & -\\
						$(\text{t2}_{10})$ &  - &  -  &  - & - \\
						\midrule
						best & $4.11$ & $9.51$ & $26.71$ &  -\\
						mean & $4.42$ & $4.64$ & $14.74$ &  $38.33$ \\
						std & $0.23$ & $3.70$ & $9.56$ & $84.53$ \\
						range & $4.11-4.65$ & $0.98-9.51$ & $5.0-26.71$ & $1-314$ \\
						\bottomrule
			\end{tabularx}
			\caption{\centering CV, CIFAR-100, ResNet-34, limited dataset (Figure~\ref{fig:kfoldl} (a))}
			
		\end{subtable}\hspace{0.4em}%
	\begin{subtable}{0.48\linewidth}
		\centering
		\begin{tabularx}{\textwidth}{@{} l|C|C|C|C @{}}
			\toprule
			Threshold & Loss & ACC & top-5 ACC & Epoch   \\ 
			\midrule
			$(\text{t1}_1)$ & $4.27$ & $7.36$ & $23.58$ & $17$ \\
			$(\text{t1}_2)$  & $4.19$ & $7.88$ &  $24.28 $& $19$ \\
			$(\text{t1}_3)$ & $4.14$ &  $8.85$  &  $25.67 $& $22$ \\
			$(\text{t1}_4)$ & $4.13$ &  $9.08$  &  $25.91$ & $23$ \\
			$(\text{t1}_5)$ & $4.06$ &  $9.99$  &  $27.84$ & $24$ \\
			$(\text{t1}_6)$ & $4.12$ &  $9.32$  &  $26.61$ & $25$ \\ 
			$(\text{t1}_7)$ & $4.07$ &  $9.86$  &  $27.34$ & $26$ \\
			$(\text{t1}_8)$ & $4.06$ &  $10.04$  &  $27.89$ & $27$ \\
			$(\text{t1}_9)$ & $4.04$ &  $10.43$  &  $28.35$ & $28$ \\
			$(\text{t1}_{10})$  & $4.04$ &  $10.41$  &  $28.40$ & $29$ \\
			$(\text{t2}_1)$ & $4.27$  &  $7.36$  &  $23.58$ & $17$ \\
			$(\text{t2}_2)$ & $4.13$  &  $9.08$  &  $25.91$ & $23$ \\
			$(\text{t2}_3)$ & $4.06$  &  $9.99$  &  $27.84$ & $24$ \\ 
			$(\text{t2}_4)$ & $4.12$  &  $9.33$  &  $26.61$ & $25$ \\
			$(\text{t2}_5)$ & $4.07  $&  $9.86$  &  $27.34$ & $26$ \\
			$(\text{t2}_6)$ & $4.06  $&  $10.04$  &  $28.89$ & $27$ \\
			$(\text{t2}_7)$ & $4.04 $ &  $10.43$  &  $28.35$ & $28$ \\
			$(\text{t2}_8)$ & $4.04$  &  $10.41$  &  $28.40$ & $29$ \\
			$(\text{t2}_9)$ & $4.05$  &  $10.29$  &  $28.41$ & $30$ \\
			$(\text{t2}_{10})$ & $4.05$  &  $10.39$  &  $28.31$ & $31$ \\
			\midrule
			best & $4.04$ & $10.43$ & $28.41$ & -\footnote[1]{} \\
			mean & $\mathbf{4.1}$ & $\mathbf{9.52}$ & $\mathbf{26.93}$ &  $25$\\
			std & $0.07$ & $0.98$ & $1.57$ & $3.89$ \\
			range & $4.04-4.27$ & $7.36-10.43$ & $23.58-28.41$ & $17-31$ \\
			\bottomrule
		\end{tabularx}
		\caption{\centering GD, CIFAR-100, ResNet-34, limited dataset (Figure~\ref{fig:kfoldl} (a))}
	\end{subtable}}
\resizebox{0.55\textheight}{!}{
		\begin{subtable}{0.48\linewidth}
				\centering
				\begin{tabularx}{\textwidth}{@{} l|C|C|C @{}}
					\toprule
					Threshold & Loss & ACC &  Epoch   \\ 
					\midrule
					best & $1.84$ & $37.03$ & -  \\
					mean & $1.97$ & $31.05$  & $12.25$ \\
					std & $0.22$ & $8.80$  & $10.54$ \\
					range & $1.84-2.35$ & $15.84-37.03$  & $1-30$ \\
					\bottomrule
				\end{tabularx}
				\caption{\centering CV, CIFAR-10, VGG-13, limited dataset (Figure~\ref{fig:kfoldl} (b))}
		\end{subtable}\hspace{0.4em}%
		\begin{subtable}{0.48\linewidth}
			\centering
			\begin{tabularx}{\textwidth}{@{} l|C|C|C @{}}
				\toprule
				Threshold & Loss & ACC & Epoch   \\ 
				\midrule
				best & $1.79$ & $38.16$  & - \\
				mean & $\mathbf{1.80}$ & $\mathbf{36.97}$  & $6.5$ \\
				std & $0.02$ & $1.27$ &  $2.87$ \\
				range & $1.79-1.85$ & $33.71-38.16$ & $2-11$ \\
				\bottomrule
			\end{tabularx}
			\caption{\centering GD, CIFAR-10, VGG-13, limited dataset (Figure~\ref{fig:kfoldl} (b))}
		\end{subtable}}
	\resizebox{0.55\textheight}{!}{
		\begin{subtable}{0.48\linewidth}
				\centering
				\begin{tabularx}{\textwidth}{@{} l|C|C|C @{}}
					\toprule
					Threshold & Loss & ACC &  Epoch   \\ 
					\midrule
					best & $0.56$ & $81.39$ &  $34$ \\
					mean & $\mathbf{1.09}$ & $62.63$  & $21.71$ \\
					std & $0.36$ & $13.46$  & $9.59$ \\
					range & $0.63-1.64$& $41.15-81.39$  & $9-41$ \\
					\bottomrule
				\end{tabularx}
				\caption{\centering CV, MNIST, AlexNet, limited dataset (Figure~\ref{fig:kfoldl} (c))}
		\end{subtable}\hspace{0.4em}%
		\begin{subtable}{0.48\linewidth}
				
			\centering
			\begin{tabularx}{\textwidth}{@{} l|C|C|C @{}}
				\toprule
				Threshold & Loss & ACC & Epoch   \\ 
				\midrule
				best & $0.45$ & $86.39$  & $39$ \\
				mean & $1.13$ & $\mathbf{64.27}$  & $17.58$ \\
				std & $0.72$ & $22.38$ &  $13.91$ \\
				range & $0.45-2.18$ & $30.19-86.39$ & $1-40$ \\
				\bottomrule
			\end{tabularx}
			\caption{\centering GD, MNIST, AlexNet, limited dataset (Figure~\ref{fig:kfoldl} (c))}
			
		\end{subtable}}
		
	\end{table*}

	\begin{table*}
		\centering
		\caption{The test accuracy, test loss, top-5 accuracy, and stopping epoch obtained by using $k$-fold cross-validation (CV) (left columns (a), (c), (e), and (g)) and by using gradient disparity (GD) (right columns (b), (d), (f), and (h)) as early stopping criteria for different patience values and for different thresholds (t1) and (t2). $(\text{t1}_p)$: training is stopped after $p$ increases in the value of the validation loss in $k$-fold CV, and of GD, respectively. $(\text{t2}_p)$: training is stopped after $p$ consecutive increases in the value of the validation loss in $k$-fold CV, and of GD, respectively. We report the best values, mean and standard deviation (std) over 20 thresholds. }\label{tab:robustnessn}
		\vspace*{1em}
		\begin{subtable}{0.48\linewidth}
				\centering
				\begin{tabularx}{\textwidth}{@{} l|C|C|C|C @{}}
					\toprule
					Threshold & Loss & ACC & top-5 ACC & Epoch   \\ 
					\midrule
					best & $4.69$ & $2.00$ & $9.38$ & $19$ \\
					mean & $4.94$ & $1.60$ & $6.72$ &  $12.3$\\
					std & $0.1$ & $0.14$ & $0.98$ & $4.47$ \\
					range & $4.69-5.03$ & $1.42-2.00$ & $5.62-9.38$ & $6-20$ \\
					\bottomrule
				\end{tabularx}
				\caption{\centering CV, CIFAR-100, ResNet-18, noisy dataset (Figure~\ref{fig:kfoldn} (a))}
			
		\end{subtable}\hspace{0.2em}%
		\begin{subtable}{0.48\linewidth}
				
			\centering
			\begin{tabularx}{\textwidth}{@{} l|C|C|C|C @{}}
				\toprule
				Threshold & Loss & ACC & top-5 ACC & Epoch   \\ 
				\midrule
				best & $4.38$ & $4.38$ & $17.76$ & - \\
				mean & $\mathbf{4.47}$ & $\mathbf{3.79}$ & $\mathbf{15.43}$ &  $29.25$\\
				std & $0.08$ & $0.75$ & $2.31$ & $6.41$ \\
				range & $4.38-4.69$ & $2.13-4.38$ & $9.74-17.76$ & $18-41$ \\
				\bottomrule
			\end{tabularx}
			\caption{\centering GD, CIFAR-100, ResNet-18, noisy dataset (Figure~\ref{fig:kfoldn} (a))}
		\end{subtable}
		\begin{subtable}{0.48\linewidth}
				\centering
				\begin{tabularx}{\textwidth}{@{} l|C|C|C|C @{}}
					\toprule
					Threshold & Loss & ACC & top-5 ACC &  Epoch   \\ 
					\midrule
					best & $3.87$ & $12.14$ & $37.06$ & -  \\
					mean & $\mathbf{4.16}$ & $10.19$  & $32.46$ & $11.5$ \\
					std & $0.25$ & $1.27$  & $2.72$ & $2.87$ \\
					range & $3.87-4.5$ & $8.47-12.14$ & $27.80-37.06$ & $7-16$ \\
					\bottomrule
				\end{tabularx}
				\caption{\centering CV,  CIFAR-100, ResNet-34, noisy dataset (Figure~\ref{fig:kfoldn} (b))}
		\end{subtable}\hspace{0.2em}%
		\begin{subtable}{0.48\linewidth}
				
			\centering
			\begin{tabularx}{\textwidth}{@{} l|C|C|C|C @{}}
				\toprule
				Threshold & Loss & ACC & top-5 ACC & Epoch   \\ 
				\midrule
				best &  $3.82$ & $15.81$ & $40.97$  & - \\
				mean & $4.37$ & $\mathbf{12.82}$  & $\mathbf{35.53} $& $14.75$ \\
				std & $0.25$ & $1.76$ & $6.08$ &  $6.59$ \\
				range & $3.82-4.59$ & $10.41-15.81$ & $23.51-40.97$ & $3-23$ \\
				\bottomrule
			\end{tabularx}
			\caption{\centering GD, CIFAR-100, ResNet-34, noisy dataset (Figure~\ref{fig:kfoldn} (b))}
		\end{subtable}
		
		\begin{subtable}{0.48\linewidth}
				\centering
				\begin{tabularx}{\textwidth}{@{} l|C|C|C @{}}
					\toprule
					Threshold & Loss & ACC &  Epoch   \\ 
					\midrule
					best & $1.69$ & $43.02$ &  $2$ \\
					mean & $\mathbf{2.19}$ & $36.89$  & $6.5$ \\
					std & $0.35$ & $3.21$  & $2.87$ \\
					range & $1.69-2.66$ & $32.41-43.02$  & $2-11$ \\
					\bottomrule
				\end{tabularx}
				\caption{\centering CV, CIFAR-10, VGG-13, noisy dataset (Figure~\ref{fig:kfoldn} (c))}
		\end{subtable}\hspace{0.2em}%
		\begin{subtable}{0.48\linewidth}
				
			\centering
			\begin{tabularx}{\textwidth}{@{} l|C|C|C @{}}
				\toprule
				Threshold & Loss & ACC & Epoch   \\ 
				\midrule
				best & $1.77 $& $42.45$  & - \\
				mean & $2.35$ & $\mathbf{38.9}$ & $10.1$ \\
				std & $0.27$ & $2.98$ &  $3.59$ \\
				range & $1.77-2.59$ & $32.92-42.45$ & $4-16$ \\
				\bottomrule
			\end{tabularx}
			\caption{\centering GD, CIFAR-10, VGG-13, noisy dataset (Figure~\ref{fig:kfoldn} (c))}
		\end{subtable}
		\begin{subtable}{0.48\linewidth}
				\centering
				\begin{tabularx}{\textwidth}{@{} l|C|C|C @{}}
					\toprule
					Threshold & Loss & ACC &  Epoch   \\ 
					\midrule
					best & $0.59$ & $97.44$ &  - \\
					mean & $\mathbf{0.63}$ & $\mathbf{96.33}$  & $23.5$ \\
					std & $0.18$ & $1.68 $ & $13.11$ \\
					range & $0.59-0.69$ & $92.03-97.44$  & $7-49$ \\
					\bottomrule
				\end{tabularx}
				\caption{\centering CV, MNIST, AlexNet, noisy dataset (Figure~\ref{fig:kfoldn} (d))}
		\end{subtable}\hspace{0.2em}%
		\begin{subtable}{0.48\linewidth}
				
			\centering
			\begin{tabularx}{\textwidth}{@{} l|C|C|C @{}}
				\toprule
				Threshold & Loss & ACC & Epoch   \\ 
				\midrule
				best & $0.62$ & $97.49$  & - \\
				mean & $0.65$ & $96.22$ & $20.4$ \\
				std & $0.02$ & $1.81$ &  $13.81$ \\
				range & $0.62-0.66$& $92.58-97.49$ & $10-48$ \\
				\bottomrule
			\end{tabularx}
			\caption{\centering GD, MNIST, AlexNet, noisy dataset (Figure~\ref{fig:kfoldn} (d))}
		\end{subtable}
	\end{table*}

	\subsection{Image-classification Benchmark Datasets}
	
\begin{figure*}
		\centering
		\begin{subfigure}[b]{1\textwidth}            
			\includegraphics[width=0.23\textwidth, height=0.129\textwidth]{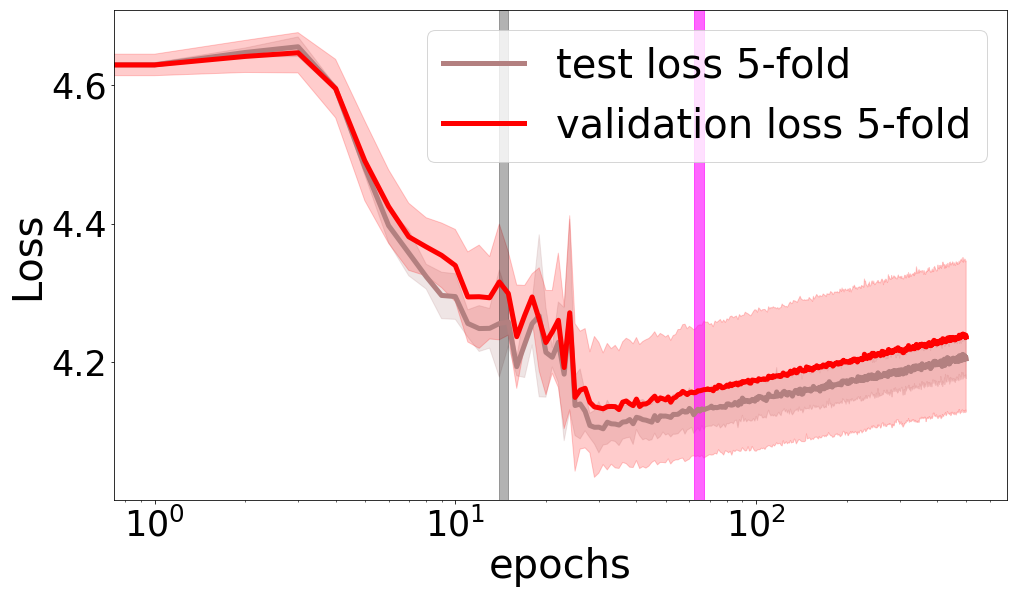}%
			\hspace{0.7em}%
			\includegraphics[width=0.23\textwidth, height=0.129\textwidth]{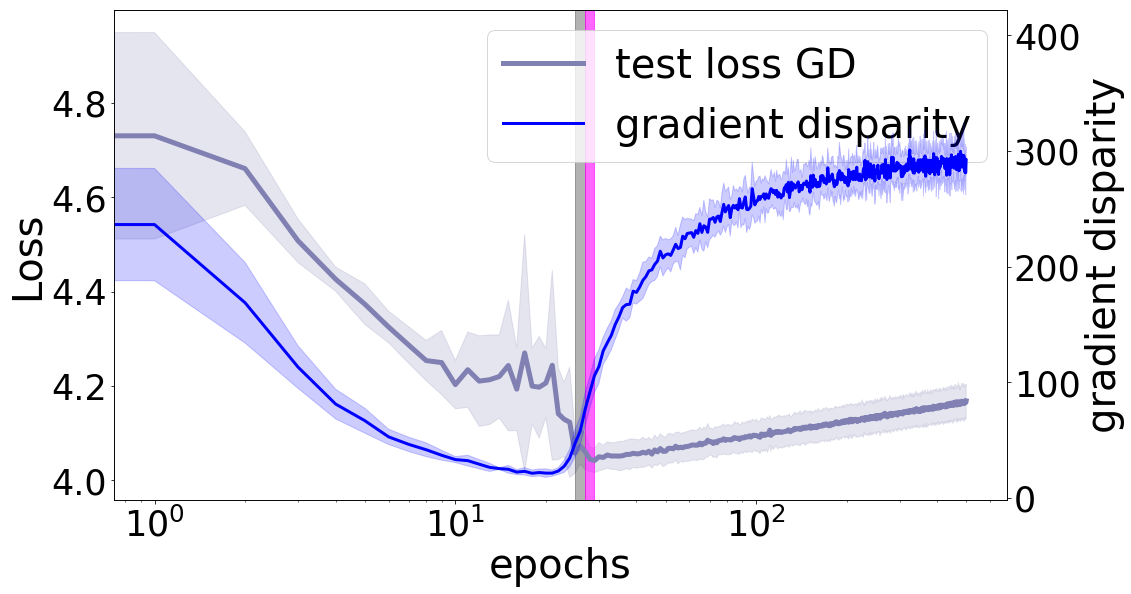}%
			\hspace{0.7em}%
			\includegraphics[width=0.23\textwidth, height=0.129\textwidth]{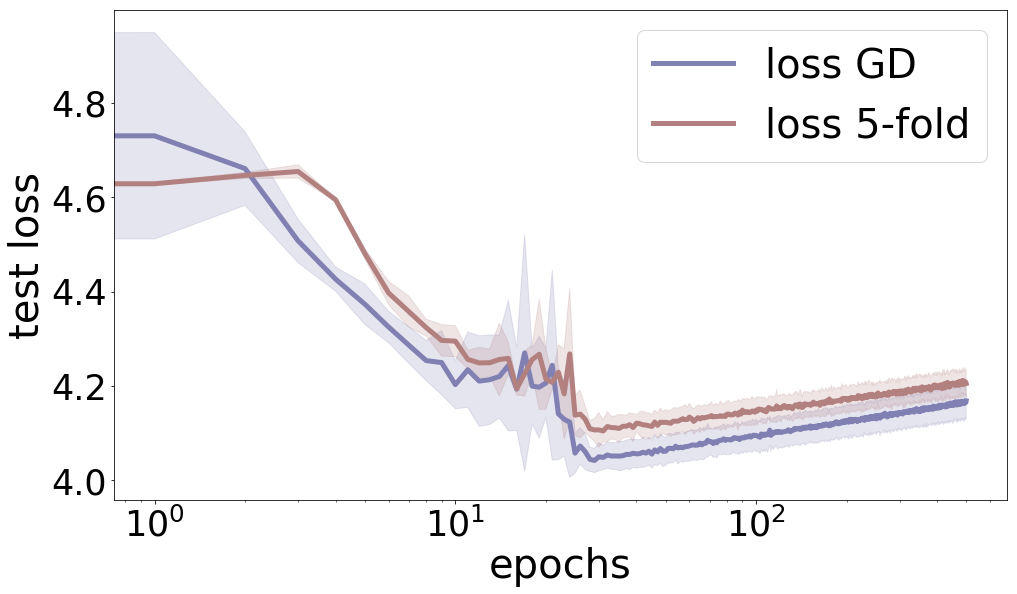}%
			\hspace{0.7em}%
			\includegraphics[width=0.23\textwidth, height=0.129\textwidth]{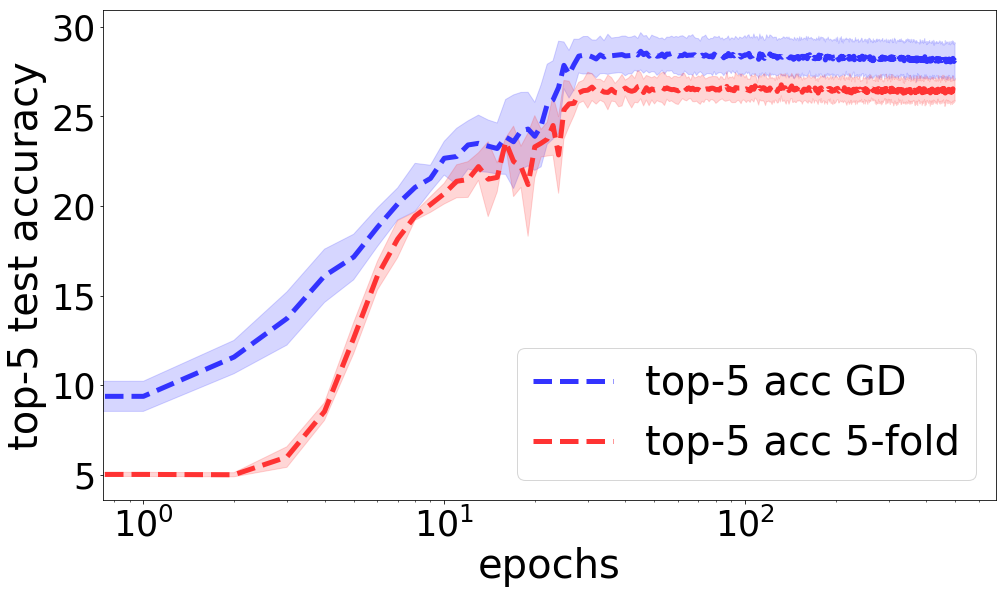}
			\caption{ResNet-34 trained on 1.28 k points of CIFAR-100}
		\end{subfigure}
		\begin{subfigure}[b]{1\textwidth}            
			\includegraphics[width=0.23\textwidth, height=0.129\textwidth]{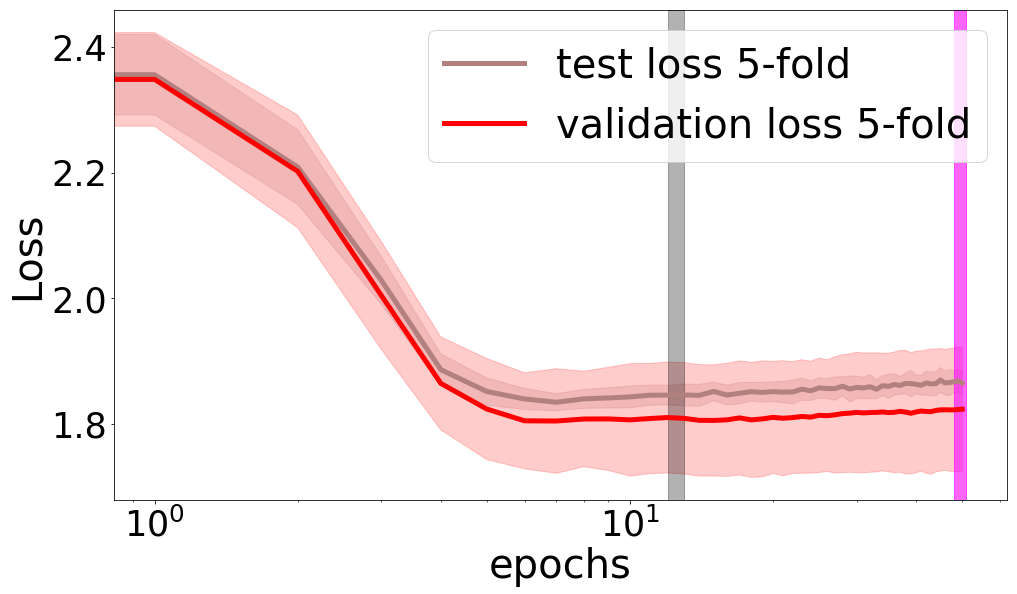}%
			\hspace{0.7em}%
			\includegraphics[width=0.23\textwidth, height=0.129\textwidth]{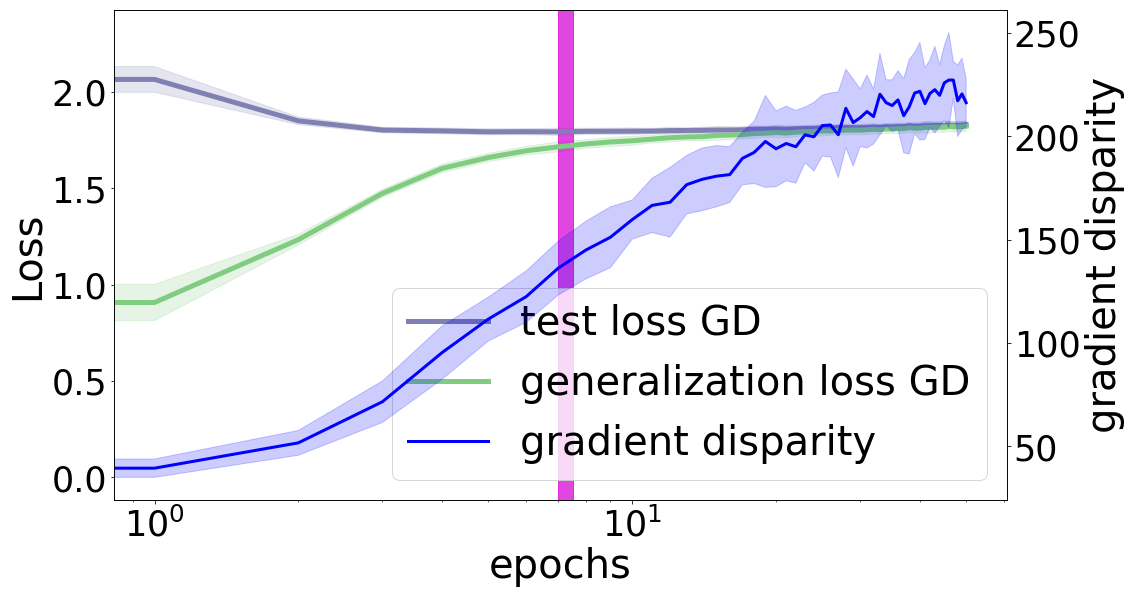}%
			\hspace{0.7em}%
			\includegraphics[width=0.23\textwidth, height=0.129\textwidth]{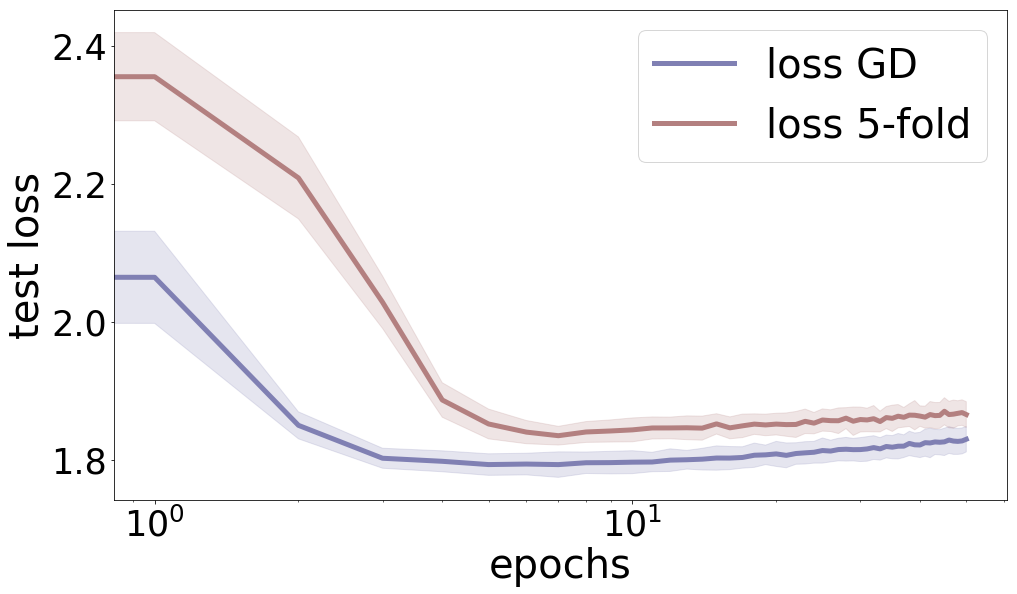}%
			\hspace{0.7em}%
			\includegraphics[width=0.23\textwidth, height=0.129\textwidth]{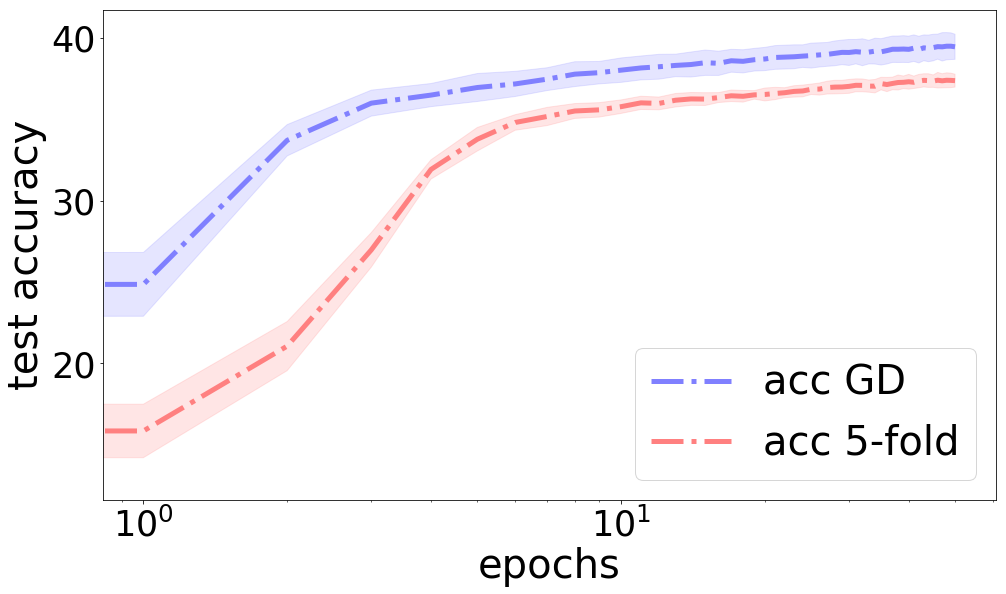}
			\caption{VGG-13 trained on 1.28 k points of CIFAR-10}
		\end{subfigure}
		\begin{subfigure}[b]{1\textwidth}            
			\includegraphics[width=0.23\textwidth, height=0.129\textwidth]{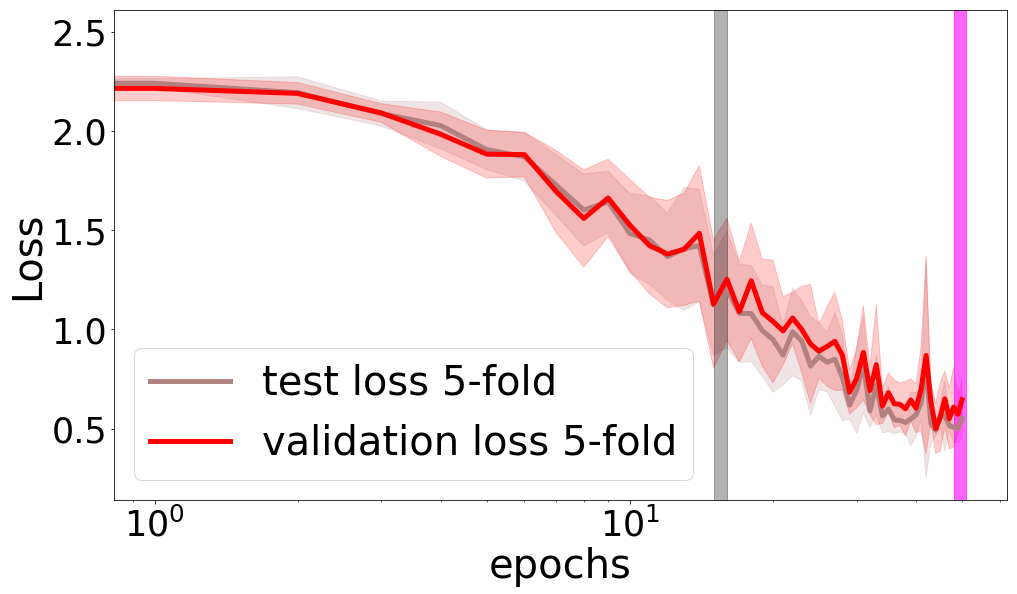}%
			\hspace{0.7em}%
			\includegraphics[width=0.23\textwidth, height=0.129\textwidth]{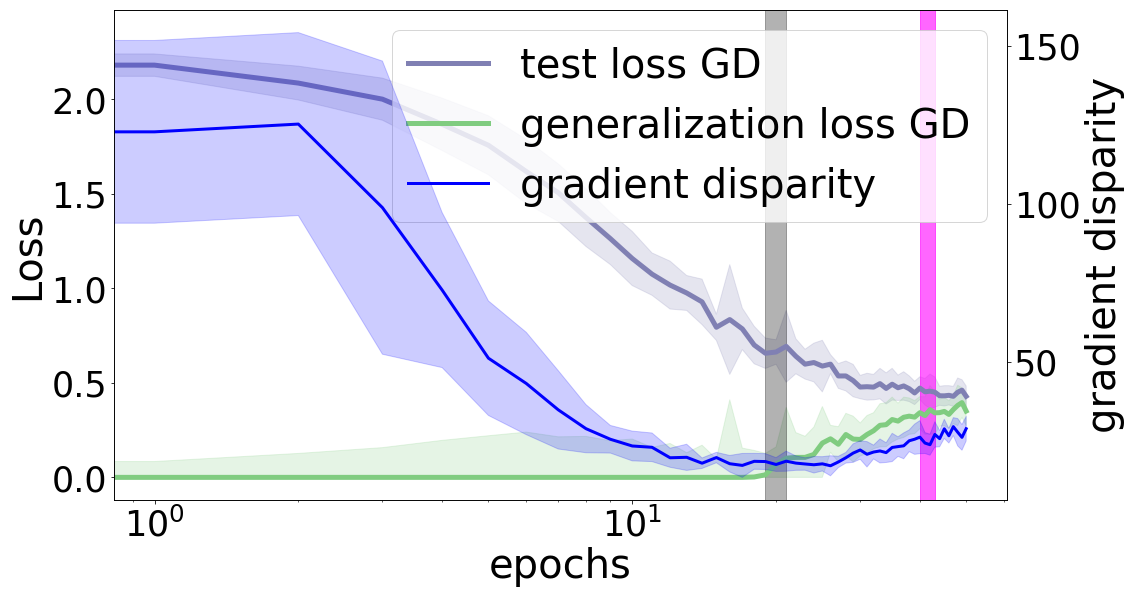}%
			\hspace{0.7em}%
			\includegraphics[width=0.23\textwidth, height=0.129\textwidth]{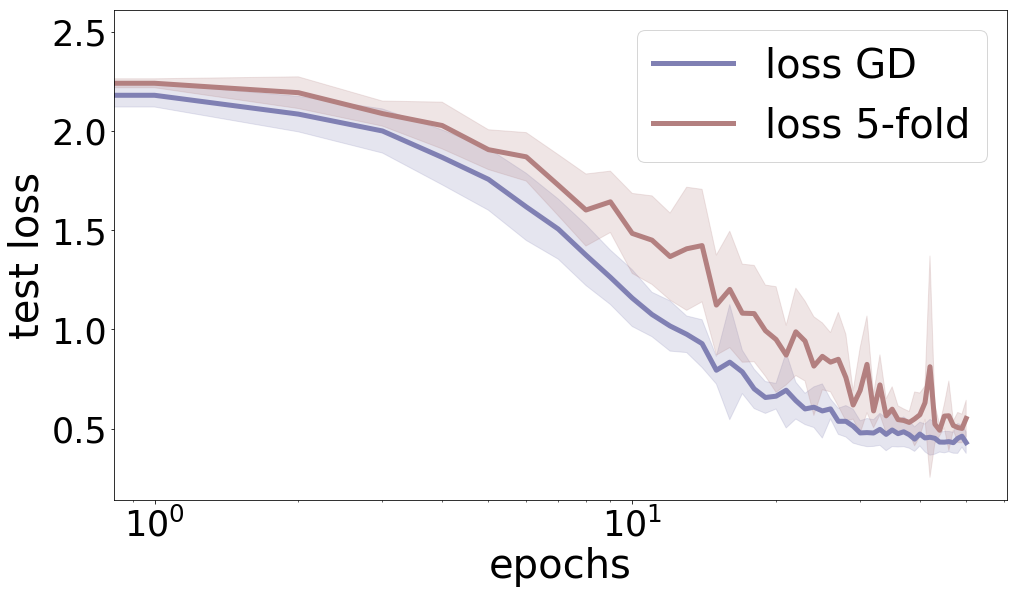}%
			\hspace{0.7em}%
			\includegraphics[width=0.23\textwidth, height=0.129\textwidth]{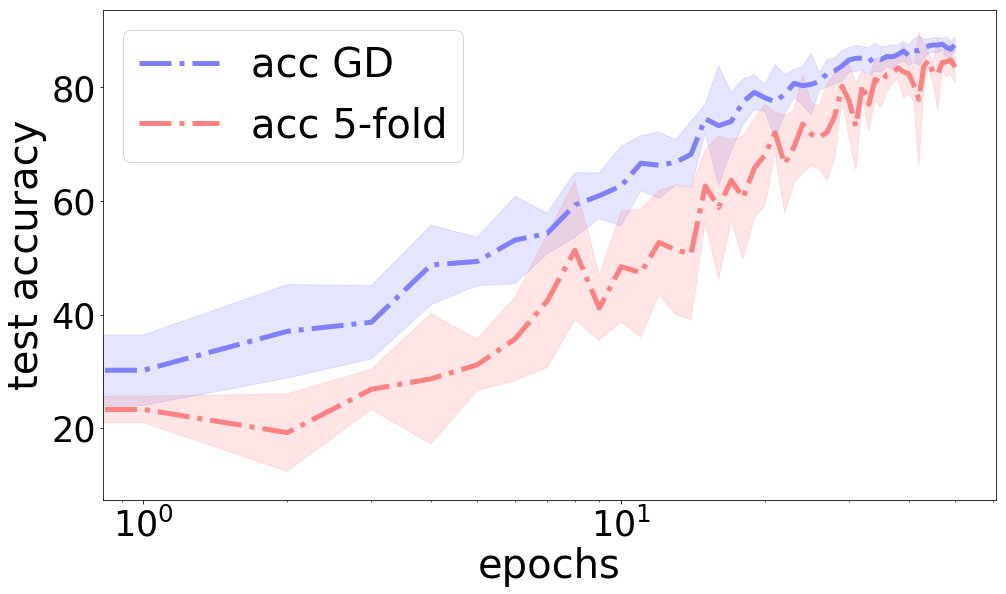}
			\caption{AlexNet trained on 256 points of MNIST}
		\end{subfigure}
		\caption{Comparing 5-fold cross-validation (CV) with gradient disparity (GD) as an early stopping criterion when the available dataset is limited. (left) Validation loss versus test loss in 5-fold cross-validation. (middle left) Gradient disparity versus test and generalization losses. (middle right and right) Performance on the unseen (test) data for GD versus $5$-fold CV. (a) The parameters are initialized by Xavier techniques with uniform distribution. (b, c) The parameters are initialized using He technique with normal distribution. (c) The batch size is 32. The gray and magenta vertical bars indicate the epoch in which the metric (the validation loss or gradient disparity) has increased for 5 epochs from the beginning of training and for 5 consecutive epochs, respectively. In (b) the middle left figure, these two bars meet each other. 
		} \label{fig:kfoldl}
	\end{figure*}

	\paragraph{Limited Data} Fig.~\ref{fig:kfoldl}  and Table~\ref{tab:kfoldl} show the results for MNIST, CIFAR-10 and CIFAR-100 datasets, where we simulate the limited data scenario by using a small subset of the training set. For the CIFAR-100 experiment (Figure~\ref{fig:kfoldl}~(a) and Table~\ref{tab:kfoldl}~(top row), we observe (from the left figure) that the validation loss predicts the test loss pretty well. We observe (from the middle left figure) that gradient disparity also predicts the test loss quite well. However, the main difference between the two settings is that when using cross-validation, $1/k$ of the data is set aside for validation and $1-1/k$ of the data is used for training. Whereas when using gradient disparity, all the data ($1-1/k+1/k =1$) is used for training. Hence, the test loss in the leftmost and middle left figures differ. The difference between the test accuracy (respectively, test loss) obtained in each setting is visible in the rightmost figure (resp., middle right figure). We observe that there is over $3\%$ improvement in the test accuracy when using gradient disparity as an early stopping criterion. This improvement is consistent for the MNIST and CIFAR-10 datasets (Figs.~\ref{fig:kfoldl}~(b) and (c) and Table~\ref{tab:kfoldl}). We conclude that in the absence of label noise, both $k$-fold cross-validation and gradient disparity predict the optimal early stopping moment, but the final test loss/error is much lower for the model trained with all the available data (thus, when gradient disparity is used), than the model trained with a $(1-1/k)$ portion of the data (thus when $k$-fold cross-validation is used). To further test on a dataset that is itself limited, a medical application with limited labeled data is empirically studied later in this section (Appendix~\ref{app:mrnet}). The same conclusion is made for this dataset.

\begin{figure*}
		\centering
		\begin{subfigure}[b]{1\textwidth}            
			\includegraphics[width=0.23\textwidth, height=0.129\textwidth]{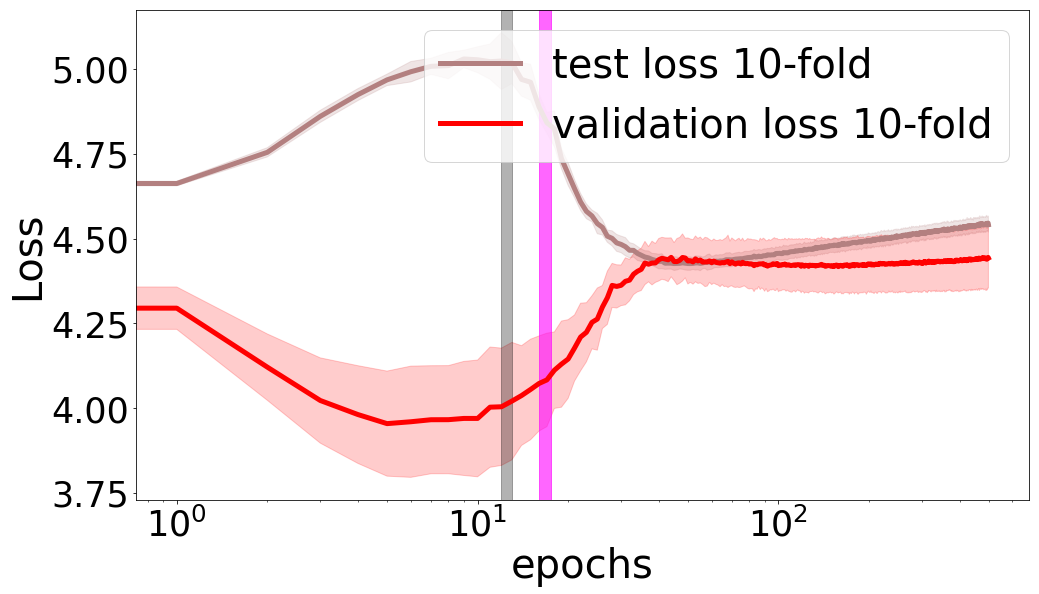}%
			\hspace{0.7em}%
			\includegraphics[width=0.23\textwidth, height=0.129\textwidth]{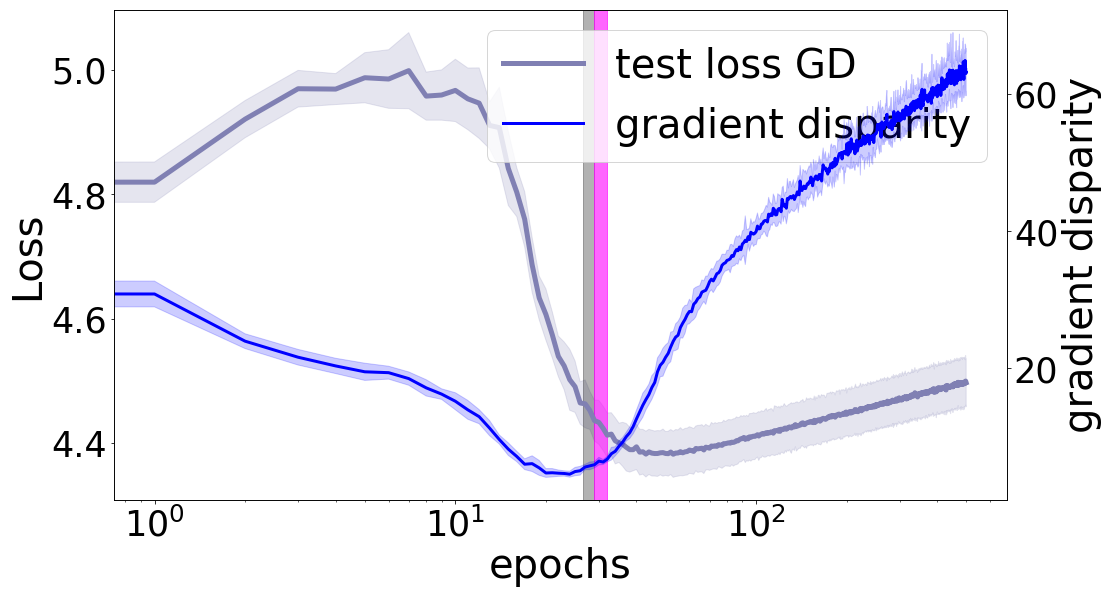}%
			\hspace{0.7em}%
			\includegraphics[width=0.23\textwidth, height=0.129\textwidth]{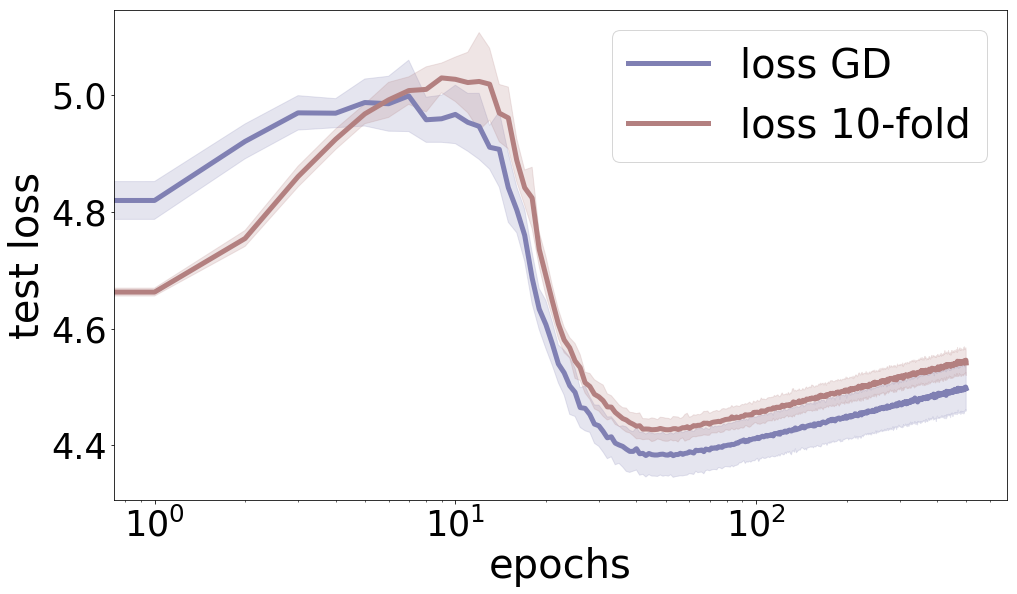}%
			\hspace{0.7em}%
			\includegraphics[width=0.23\textwidth, height=0.129\textwidth]{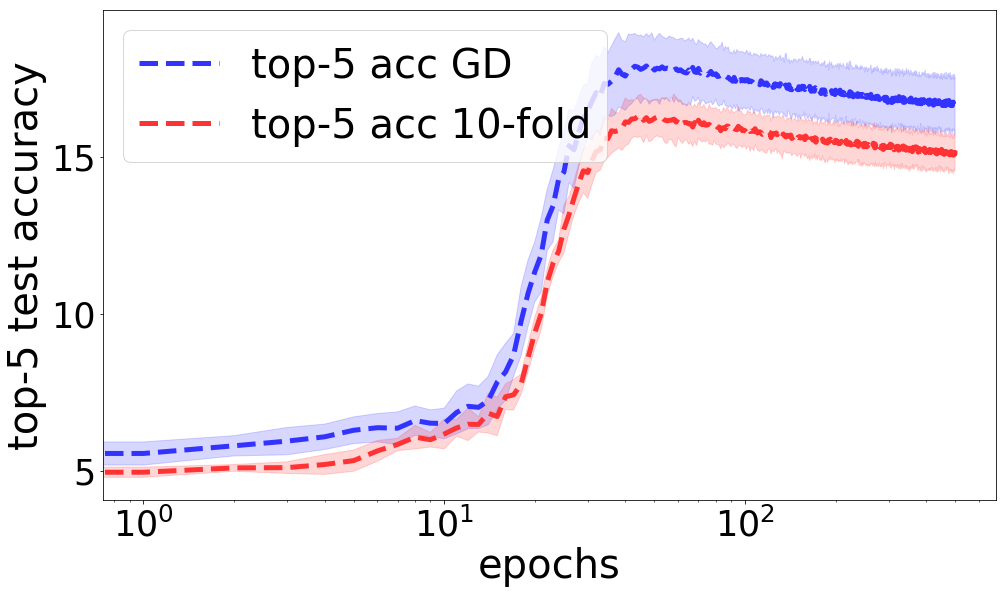}
			\caption{ResNet-18 trained on 1.28 k points of CIFAR-100 dataset with 50\% label noise}
		\end{subfigure}
		\begin{subfigure}[b]{1\textwidth}            
			\includegraphics[width=0.23\textwidth, height=0.129\textwidth]{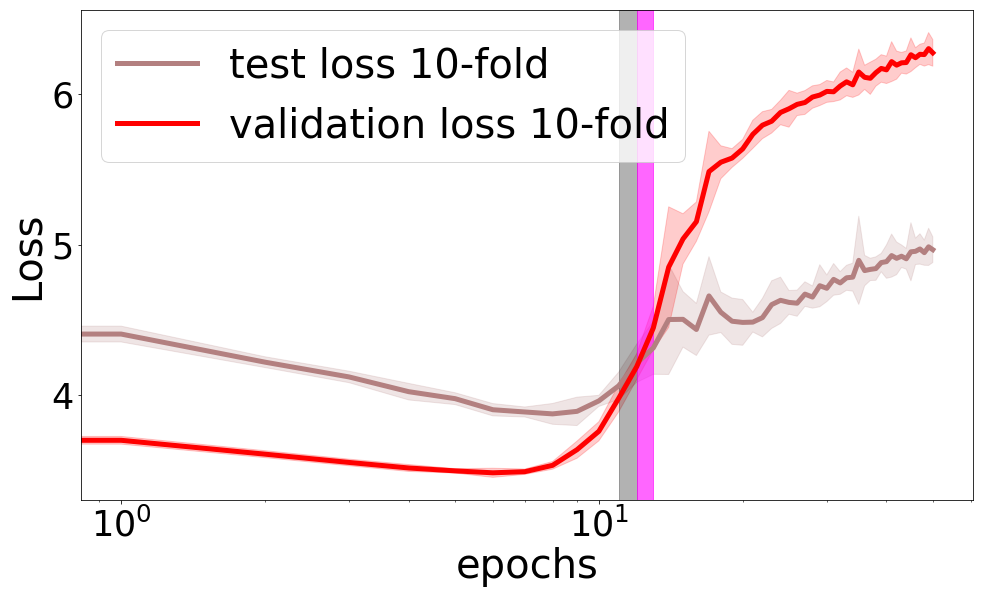}%
			\hspace{0.7em}%
			\includegraphics[width=0.23\textwidth, height=0.129\textwidth]{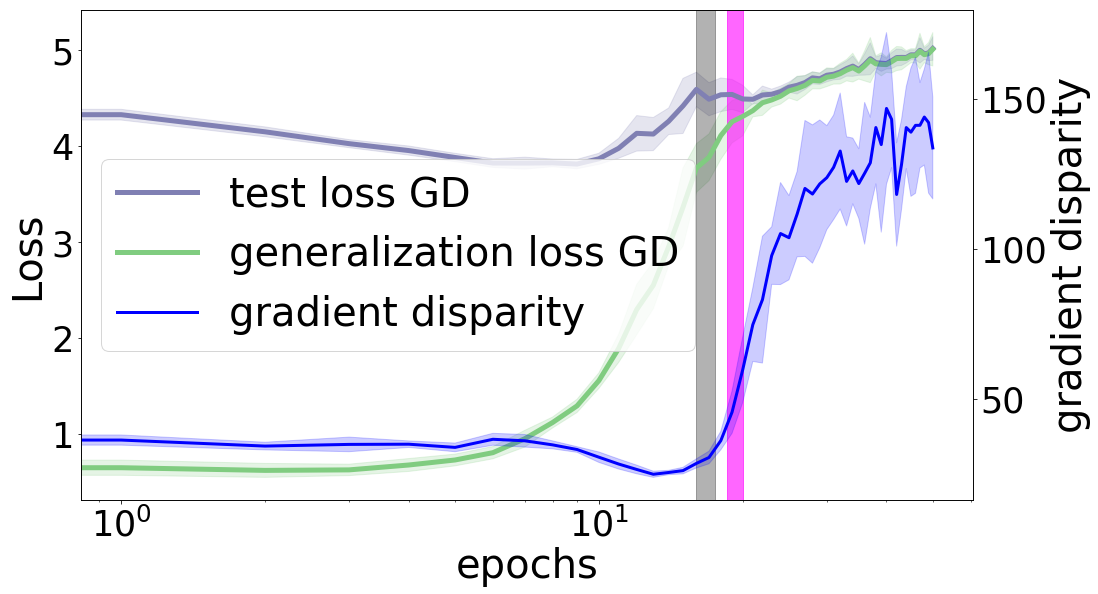}%
			\hspace{0.7em}%
			\includegraphics[width=0.23\textwidth, height=0.129\textwidth]{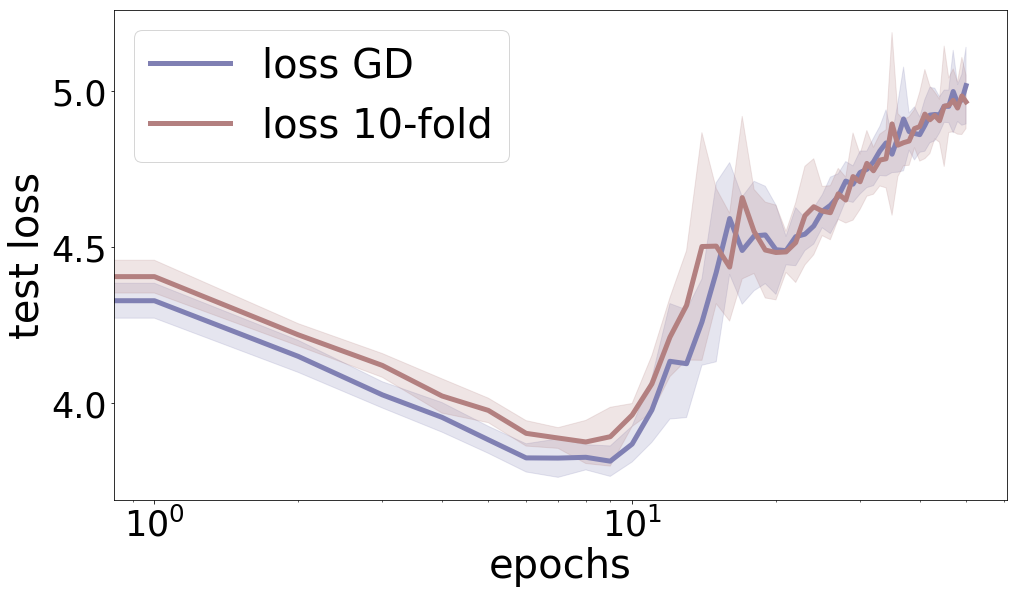}%
			\hspace{0.7em}%
			\includegraphics[width=0.23\textwidth, height=0.129\textwidth]{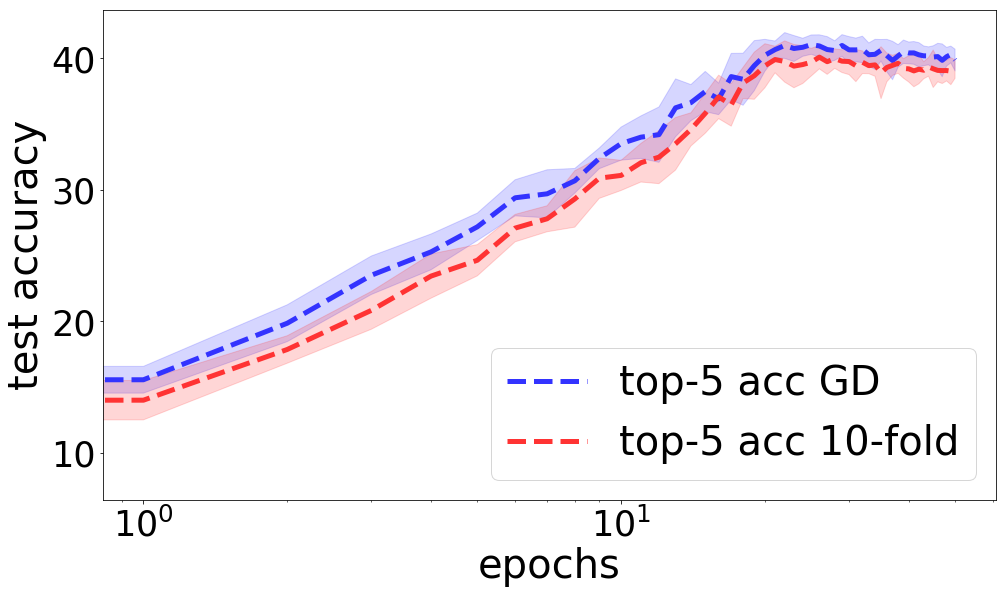}
			\caption{ResNet-34 trained on the entire CIFAR-100 dataset with 50\% label noise}
		\end{subfigure}
		\begin{subfigure}[b]{1\textwidth}            
			\includegraphics[width=0.23\textwidth, height=0.129\textwidth]{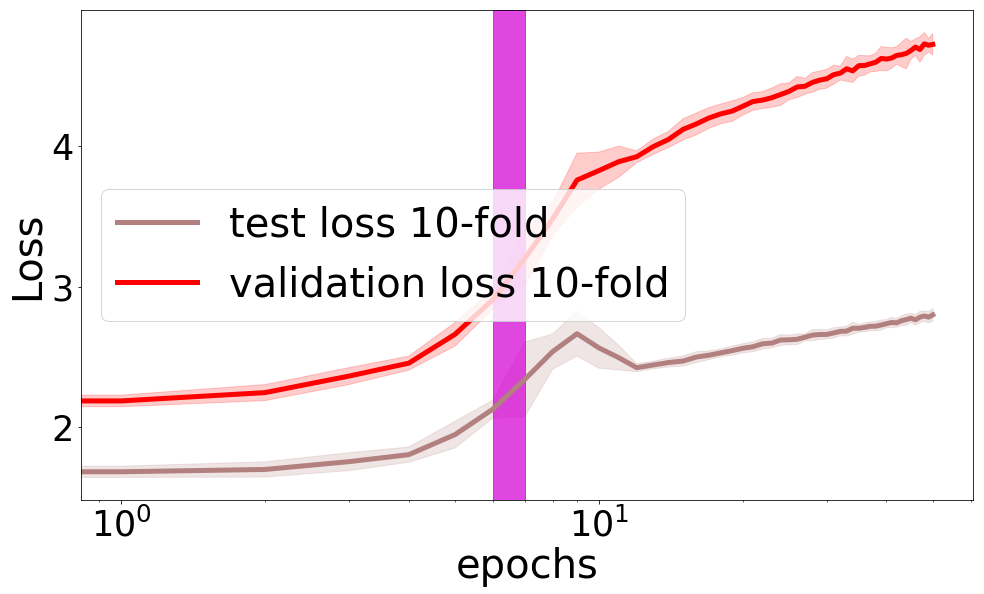}%
			\hspace{0.7em}%
			\includegraphics[width=0.23\textwidth, height=0.129\textwidth]{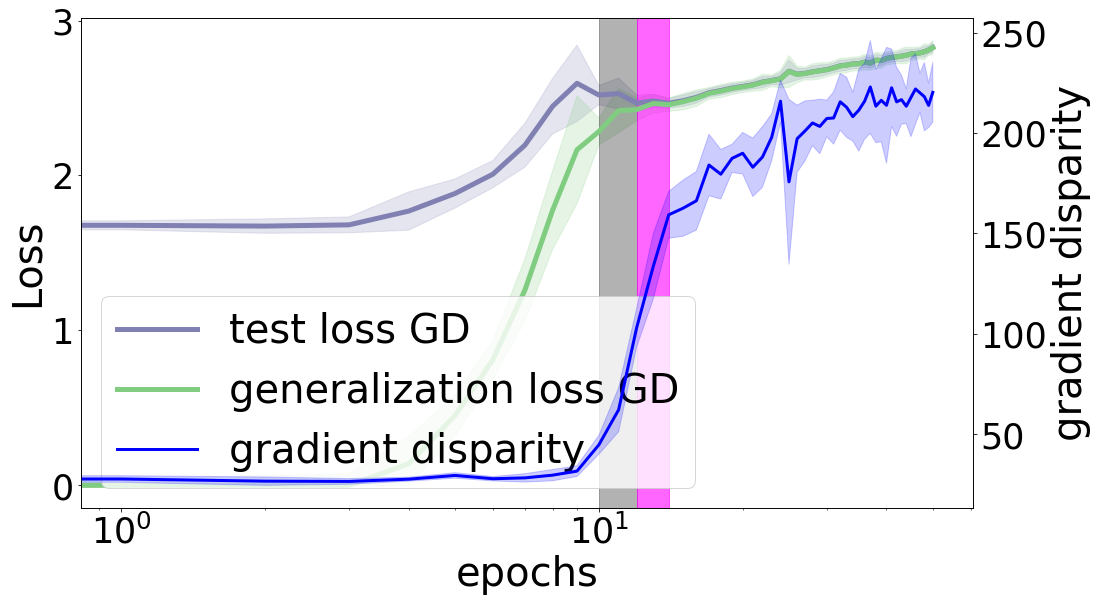}%
			\hspace{0.7em}%
			\includegraphics[width=0.23\textwidth, height=0.129\textwidth]{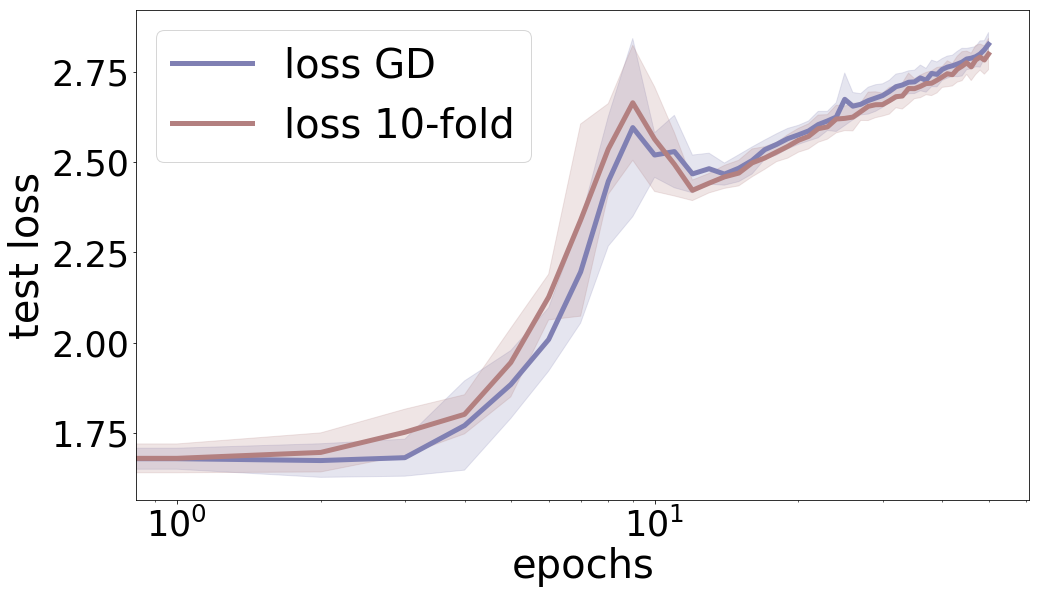}%
			\hspace{0.7em}%
			\includegraphics[width=0.23\textwidth, height=0.129\textwidth]{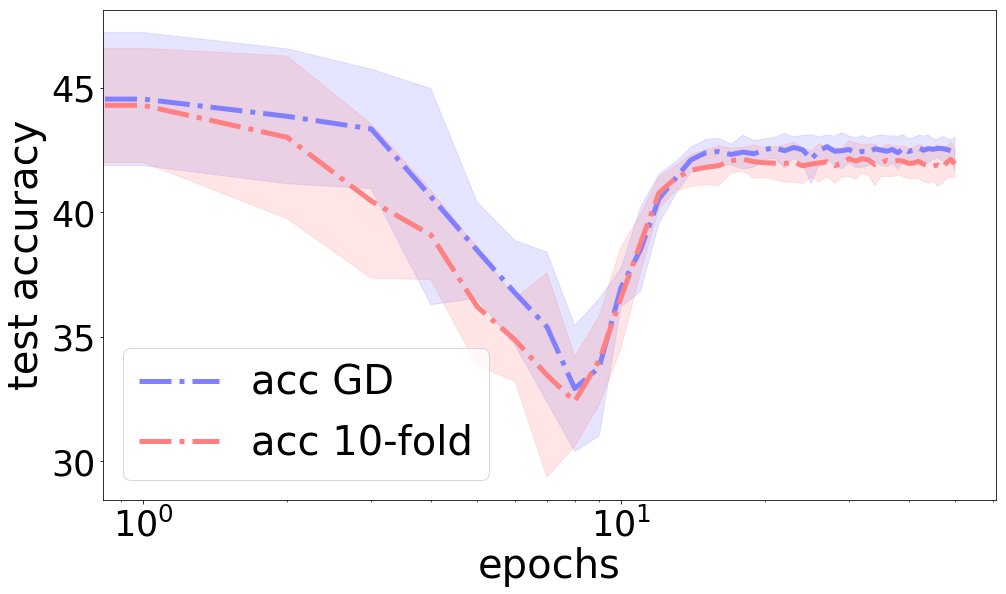}
			\caption{VGG-13 trained on the entire CIFAR-10 dataset with 50\% label noise}
		\end{subfigure}
		\begin{subfigure}[b]{1\textwidth}            
			\includegraphics[width=0.23\textwidth, height=0.129\textwidth]{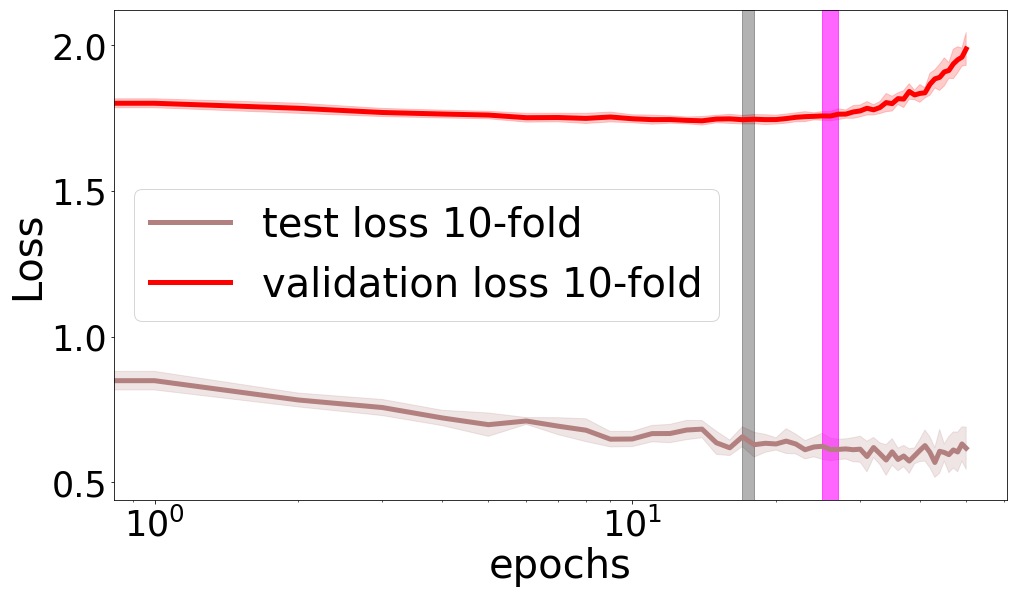}%
			\hspace{0.7em}%
			\includegraphics[width=0.23\textwidth, height=0.129\textwidth]{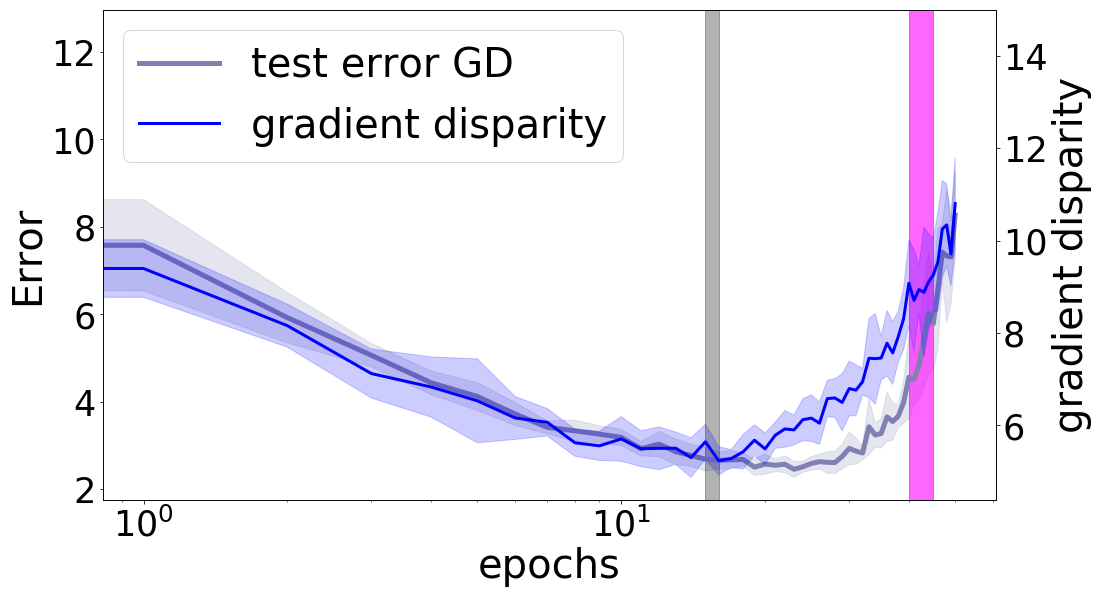}%
			\hspace{0.7em}%
			\includegraphics[width=0.23\textwidth, height=0.129\textwidth]{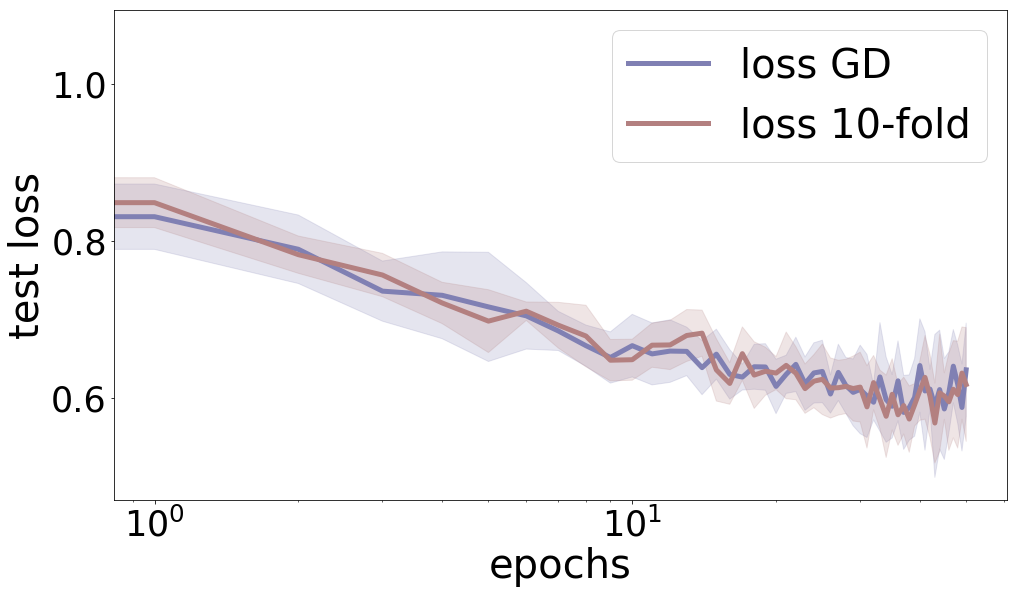}%
			\hspace{0.7em}%
			\includegraphics[width=0.23\textwidth, height=0.129\textwidth]{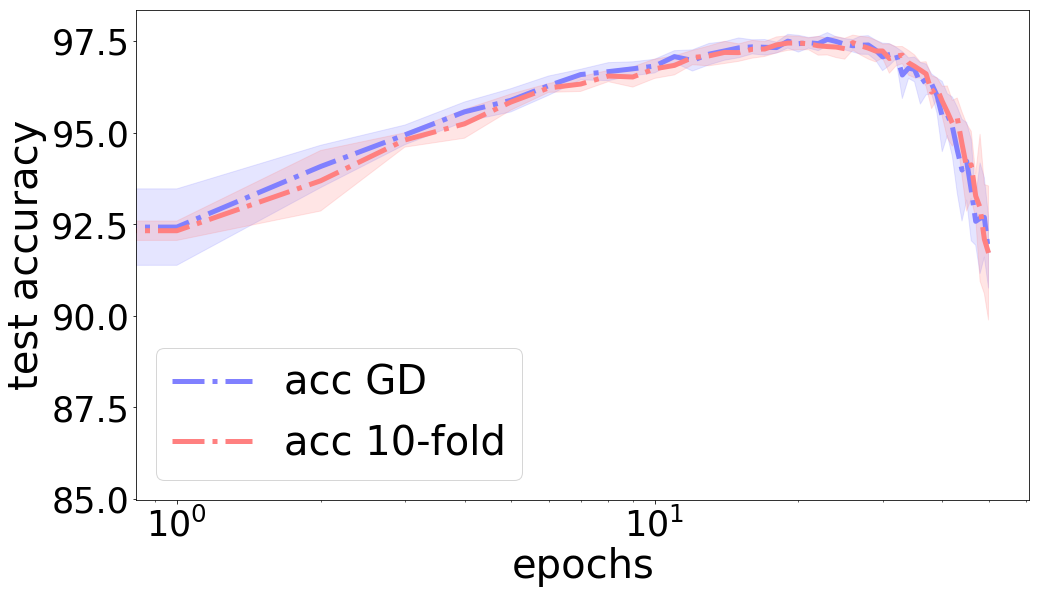}
			\caption{AlexNet trained on the entire MNIST dataset with 50\% label noise}
		\end{subfigure}
		\caption{Comparing 10-fold cross-validation with gradient disparity as early stopping criteria when the available dataset is noisy. (left) Validation loss versus test loss in 10-fold cross-validation. (middle left) Gradient disparity versus test and generalization losses. (middle right and right) Performance on the unseen (test) data for GD versus 10-fold CV. (a) The parameters are initialized by Xavier techniques with uniform distribution. (b, c, and d) The parameters are initialized using He technique with normal distribution. 
		}\label{fig:kfoldn}
	\end{figure*}

	\paragraph{Noisy Labeled Data} The results for datasets with noisy labels are shown in Fig.~\ref{fig:kfoldn} and Table~\ref{tab:kfoldn} for the MNIST, CIFAR-10 and CIFAR-100 datasets. We observe (from Fig.~\ref{fig:kfoldn}~(a)~(left)) that for the CIFAR-100 experiment, the validation loss does no longer predict the test loss. Nevertheless, although gradient disparity is computed on a training set that contains corrupted samples, it predicts the test loss quite well (Fig.~\ref{fig:kfoldn}~(a)~(middle left)). There is a $2\%$ improvement in the final test accuracy (for top-5 accuracy there is a $9\%$ improvement) (Table~\ref{tab:kfoldn}~(top two rows)) when using gradient disparity instead of a validation set as an early stopping criterion. This is also consistent for other configurations and datasets (Fig.~\ref{fig:kfoldn} and Table~\ref{tab:kfoldn}).
	We conclude that, in the presence of label noise, $k$-fold cross-validation does no longer predict the test loss and fails as an early stopping criterion, unlike gradient disparity. 
	
	\paragraph{Computational Cost} Denote the time, in seconds, to compute one gradient vector, to compute the $\ell_2$ norm between two gradient vectors, to take the update step for the network parameters
	, and to evaluate one batch (find its validation loss and error) by $t_1$, $t_2$, $t_3$ and $t_4$, respectively. Then, one epoch of $k$-fold cross-validation takes 
	\begin{equation*}
		\text{CV}_\text{epoch} = {k \times \left(\frac{k-1}{k}B(t_1+t_3) + \frac{B}{k} t_4\right)} 
	\end{equation*}
	 seconds, where $B$ is the number of batches. Performing one epoch of training and computing the gradient disparity takes
	 \begin{equation*}
	 	\text{GD}_\text{epoch} = {B(t_1+t_3) + s\left(t_1 + \frac{s-1}{2}t_2\right)} 
	 \end{equation*} 
	 seconds. In our experiments, we observe that $t_1\approx 5.1 t_2 \approx 100 t_3 \approx 3.4 t_4$, hence the approximate time to perform one epoch for each setting is
	 \begin{align*}
	 	\text{CV}_\text{epoch} &\approx (k-1) B t_1 , & &\text{and}  & \text{GD}_\text{epoch} &\approx (B+s) t_1.
	 \end{align*}
	 Therefore, as $s < B$, we have $\text{CV}_\text{epoch} \gg \text{GD}_\text{epoch} $.

	\begin{table*}
		\centering
		\caption{The test accuracies achieved by using $k$-fold cross-validation (CV) and by using gradient disparity (GD) as early stopping criteria for different patience values. For a given patience value of $p$, the training is stopped after $p$ increases in the value of the validation loss in $k$-fold CV (top rows) and of GD (bottom rows). Throughout the paper, we have chosen $p = 5$ as the default patience value for all methods without optimizing it even for GD. However, in this table (also in Tables~\ref{tab:robustness} and \ref{tab:robustnessn}), we observe that even if we tune the patience value for $k$-fold CV and for GD separately (which is indicated in bold), GD still outperforms $k$-fold CV. Moreover, as we discussed in Appendix~\ref{app:patthre}, even if we take an average over patience values and early stopping thresholds (to avoid the need to tune this parameter), GD again outperforms CV (Fig.~\ref{fig:sen}). }\label{tab:patience}
		\vspace*{1em}
		\begin{subtable}{\linewidth}
				\centering
				\begin{tabular}{c|c|c|c|c|c|c}
					\toprule
					\backslashbox{Method}{Patience}  & 1 & 5 & 10 & 15 & 20 & 25  \\ 
					\midrule
					5-fold CV  & $41.15_{\pm 5.68}$ & $62.62_{\pm 6.36}$ & $81.39 _{\pm 3.64}$ &$80.39_{\pm 2.88}$ & $\mathbf{84.84}_{\pm 2.53}$ & $83.55_{\pm 2.84}$ \\
					GD    & $30.19_{\pm 6.21}$ & $79.12_{\pm 3.04}$ & $84.82_{\pm 2.14}$ &$85.35_{\pm 2.09}$ & $\mathbf{87.28}_{\pm 1.24}$ & $86.69_{\pm 1.31}$ \\ 
					\bottomrule
			\end{tabular}
			\caption{MNIST, AlexNet, limited dataset (Fig.~\ref{fig:kfoldl} (c))}
		\end{subtable}
		\begin{subtable}{\linewidth}
				\centering
				\begin{tabular}{c|c|c|c|c|c|c}
					\toprule
					\backslashbox{Method}{Patience}   & 1 & 5 & 10 & 15 & 20 & 25  \\ 
					\midrule
					10-fold CV& $96.54_{\pm 0.15}$ & $97.28_{\pm 0.20}$ & $\mathbf{97.35}_{\pm 0.23}$ & $97.22_{\pm 0.19}$ & $96.60_{\pm 0.33}$ & $94.69_{\pm 0.87}$ \\
					GD &  $97.07_{\pm 0.16}$ & $97.32_{\pm 0.15}$ & $\mathbf{97.41}_{\pm 0.15}$ & $96.57_{\pm 0.64}$ & $95.44_{\pm 0.96}$ & $92.58_{\pm 0.65}$  \\ 
					\bottomrule
			\end{tabular}
			\caption{MNIST, AlexNet, noisy dataset (Fig.~\ref{fig:kfoldn} (d))}
		\end{subtable}

	\end{table*}

\begin{table*}
	\centering
	\caption{The loss and accuracy on the test set comparing 5-fold cross-validation and gradient disparity as early stopping criterion when the available dataset is limited. The corresponding curves during training are presented in Fig.~\ref{fig:kfoldl}. The results below are obtained by stopping the optimization when the metric (either validation loss or gradient disparity) has increased for five epochs from the beginning of training.}\label{tab:kfoldl}
	\vspace*{1em}
	\begin{tabular}{c|c|c|c}
		\toprule
		Setting & Method & Test loss & Test accuracy  \\ 
		\midrule
		\multirow{2}{*}{CIFAR-100, ResNet-34} & 5-fold CV  & $4.249_{\pm 0.028}$ & $6.79_{\pm 0.49}$ (top-5: $22.19_{\pm 0.77}$) \\
		& GD    & $\mathbf{4.057}_{\pm 0.043}$ & $\mathbf{9.99}_{\pm 0.92}$ (top-5: $\mathbf{27.84}_{\pm 1.30}$) \\ 
		\midrule
		\multirow{2}{*}{CIFAR-10, VGG-13} & 5-fold CV& $1.846_{\pm 0.016}$ & $35.982_{\pm 0.393}$  \\
		& GD &  $\mathbf{1.793}_{\pm 0.016}$ & $\mathbf{36.96}_{\pm 0.861}$  \\ 
		\midrule
		\multirow{2}{*}{MNIST, AlexNet} & 5-fold CV &  $1.123_{\pm 0.25}$ & $62.62_{\pm 6.36}$  \\ 
		& GD & $\mathbf{0.656}_{\pm 0.080}$ & $\mathbf{79.12}_{\pm 3.04}$  \\ 
		\bottomrule
	\end{tabular}

\end{table*}

	\begin{table*}
		\centering
		\caption{The loss and accuracy on the test set comparing 10-fold cross-validation and gradient disparity as early stopping criterion when the available dataset is noisy. In all the experiments, 50\% of the available data has random labels. The corresponding curves during training are shown in Fig.~\ref{fig:kfoldn}. The results below are obtained by stopping the optimization when the metric (either validation loss or gradient disparity) has increased for five epochs from the beginning of training. The last row in each setting, which we call 10$^{+}$-fold CV, refers to the test loss and accuracy reached at the epoch suggested by 10-fold CV, for a network trained on the entire set. In all these settings, using GD still results in a higher test accuracy. 
		}\label{tab:kfoldn}
	\vspace*{1em}
		\begin{tabular}{c|c|c|c}
			\toprule
			Setting & Method    & Test loss & Test accuracy  \\ 
			\midrule
			\multirow{3}{*}{CIFAR-100, ResNet-18} & 10-fold CV  & $5.023_{\pm 0.083}$ & $1.59_{\pm 0.15}$  (top-5: $6.47_{\pm 0.52}$) \\
			 & GD          & $\mathbf{4.463}_{\pm 0.038}$ & $\mathbf{3.68}_{\pm 0.52}$  (top-5: $\mathbf{15.22}_{\pm 1.24}$) \\
			 & 10$^{+}$-fold CV          & $4.964_{\pm 0.057}$ & $1.68_{\pm 0.24}$  (top-5: $7.05_{\pm 0.71}$) \\
			\midrule
			\multirow{3}{*}{CIFAR-100, ResNet-34} & 10-fold CV  & $4.062_{\pm 0.091}$ & $9.62_{\pm 1.08}$  (top-5: $32.06_{\pm 1.47}$) \\
			 & GD          & $4.592_{\pm 0.179}$ & $\mathbf{10.41}_{\pm 1.40}$  (top-5: $\mathbf{36.92}_{\pm 1.20}$) \\
			 & 10$^{+}$-fold CV         & $4.134_{\pm 0.185}$ & $10.11_{\pm 1.60}$  (top-5: $34.19_{\pm 2.10}$) \\
			\midrule
			\multirow{3}{*}{CIFAR-10, VGG-13} & 10-fold CV  & $2.126_{\pm 0.063}$ & $34.88_{\pm 1.66}$  \\
			 & GD  & $2.519_{\pm 0.062}$ & $\mathbf{36.98}_{\pm 0.77}$  \\
			 & 10$^{+}$-fold CV & $2.195_{\pm 0.142}$ & $35.40_{\pm 3.00}$  \\
			\midrule
			\multirow{3}{*}{MNIST, AlexNet} & 10-fold CV & $0.656_{\pm 0.034}$ & $97.28_{\pm 0.20}$  \\
			 & GD  & $0.654_{\pm 0.031}$ & $\mathbf{97.32}_{\pm 0.27}$  \\
			 & 10$^{+}$-fold CV  & $0.639_{\pm 0.029}$ & $97.31_{\pm 0.15}$  \\
			\bottomrule
		\end{tabular}
		
	\end{table*}

	\subsection{MRNet Dataset}\label{app:mrnet}
	So far, we have shown the improvement of gradient disparity over cross-validation for limited subsets of MNIST, CIFAR-10 and CIFAR-100 datasets. In this sub-section, we give the results for the MRNet dataset \cite{bien2018deep} used for diagnosis of knee injuries, which is by itself limited. The dataset contains 1370 magnetic resonance imaging (MRI) exams to study the presence of abnormality, anterior cruciate ligament (ACL) tears and meniscal tears. The labeled data in the MRNet dataset is therefore very limited. Each MRI scan is a set of $S$ slices of images stacked together. Note that, in this dataset, because slice $S$ changes from one patient to another, it is not possible to stack the data into batches, hence the batch size is 1, which may explain the fluctuations of both the validation loss and gradient disparity in this setting. Each patient (case) has three MRI scans: sagittal, coronal and axial. The MRNet dataset is split into training (1130 cases), validation (120 cases) and test sets (120 cases). The test set is not publicly available. We need however to set aside some data to evaluate both gradient disparity and $k$-fold cross-validation, hence, in our experiments, the validation set becomes the unseen (test) set. 
	To perform cross-validation, we split the set used for training in \cite{bien2018deep} into a first subset used for training in our experiments, and a second subset used as validation set. We use the SGD optimizer with the learning rate $10^{-4}$ for training the model. Each task in this dataset is a binary classification with an unbalanced set of samples, hence we report the area under the curve of the receiver operating characteristic (AUC score).

\begin{figure*}
		\centering
		
		\begin{subfigure}[b]{1\textwidth}            
			\includegraphics[width=0.31\textwidth, height=0.174\textwidth]{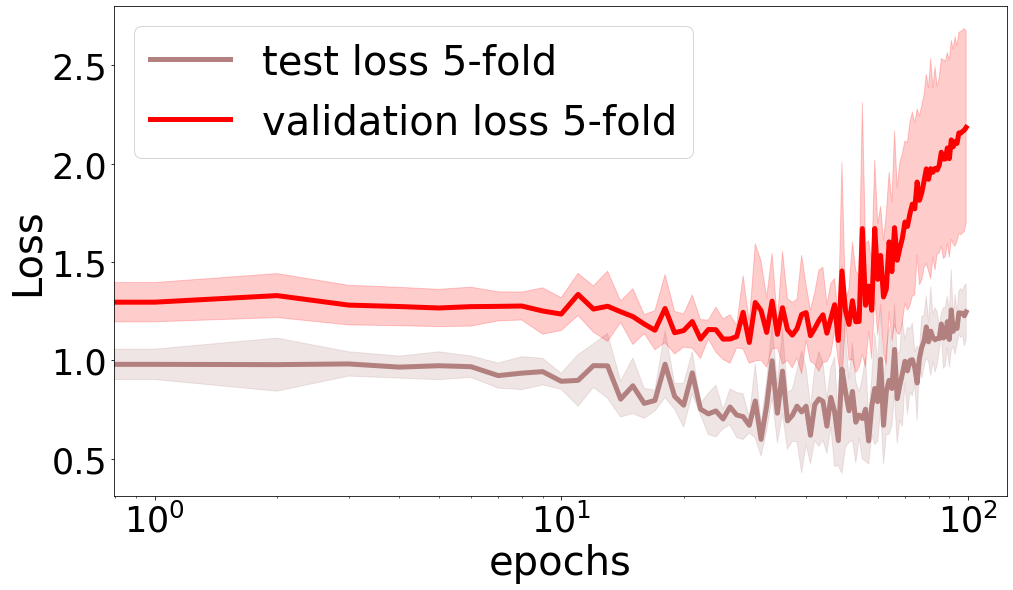}%
			\hspace{0.5em}%
			\includegraphics[width=0.31\textwidth, height=0.174\textwidth]{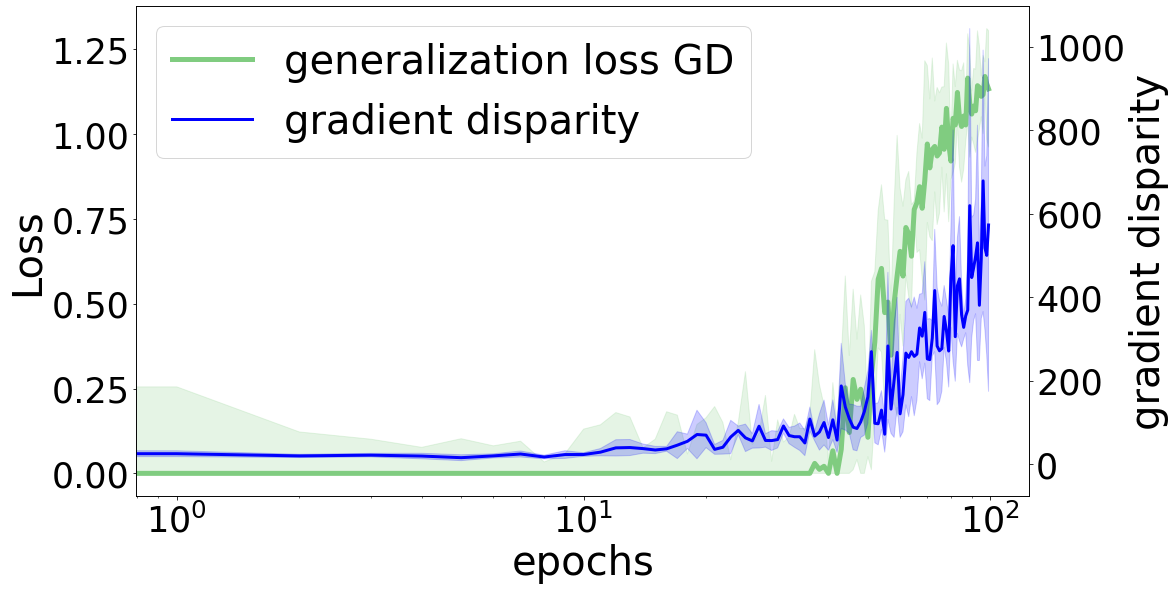}%
			\hspace{0.5em}%
			\includegraphics[width=0.31\textwidth, height=0.174\textwidth]{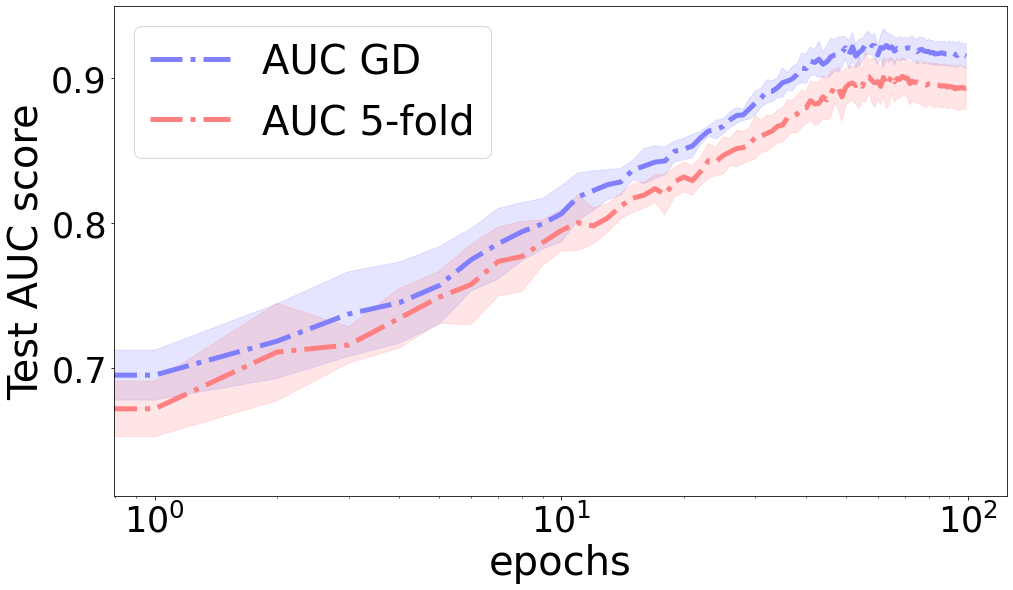}
			\caption{Task: detecting ACL tears}
		\end{subfigure}
		\begin{subfigure}[b]{1\textwidth}            
			\includegraphics[width=0.31\textwidth, height=0.174\textwidth]{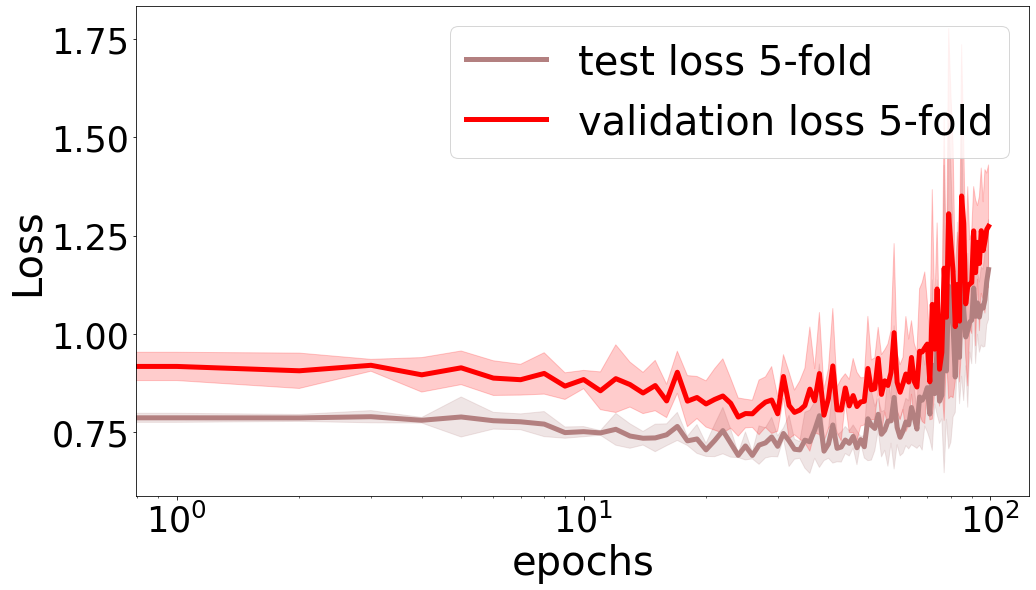}%
			\hspace{0.5em}%
			\includegraphics[width=0.31\textwidth, height=0.174\textwidth]{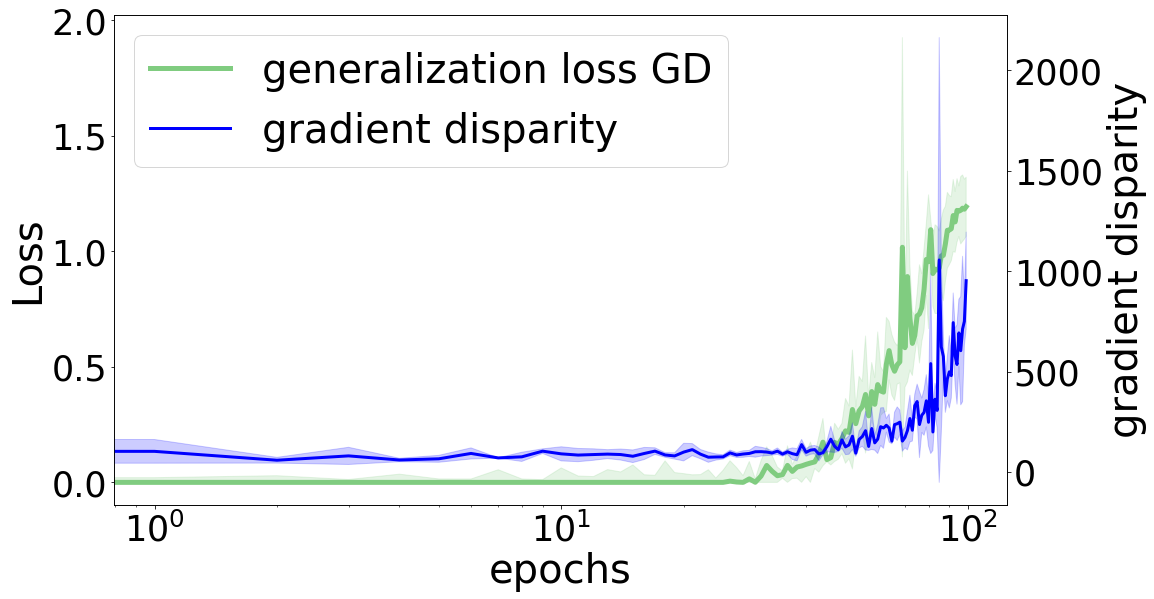}%
			\hspace{0.5em}%
			\includegraphics[width=0.31\textwidth, height=0.174\textwidth]{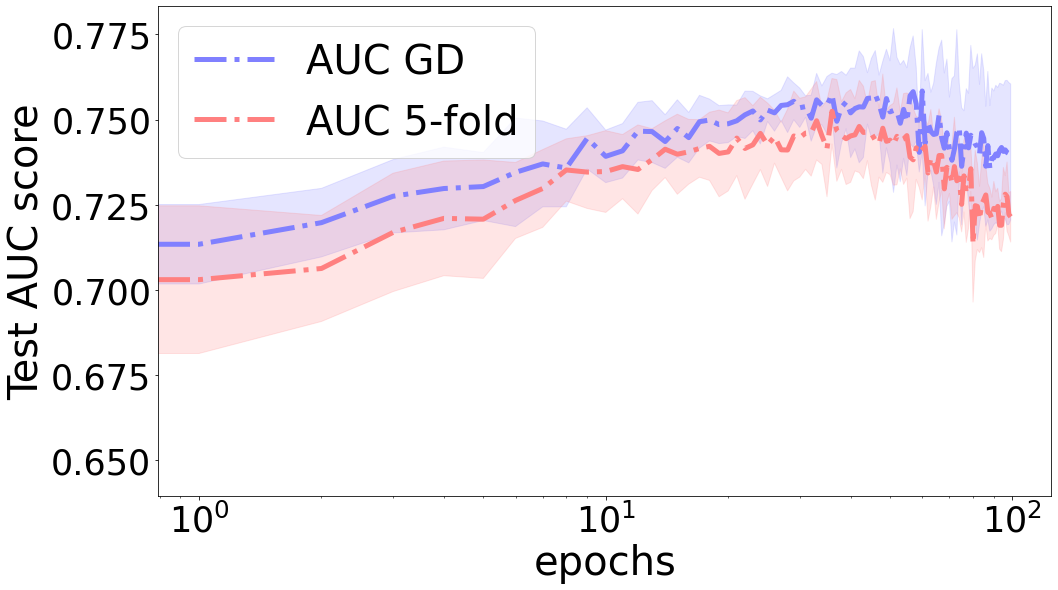}
			\caption{Task: detecting meniscal tears}
		\end{subfigure}
	\begin{subfigure}[b]{1\textwidth}            
		\includegraphics[width=0.31\textwidth, height=0.174\textwidth]{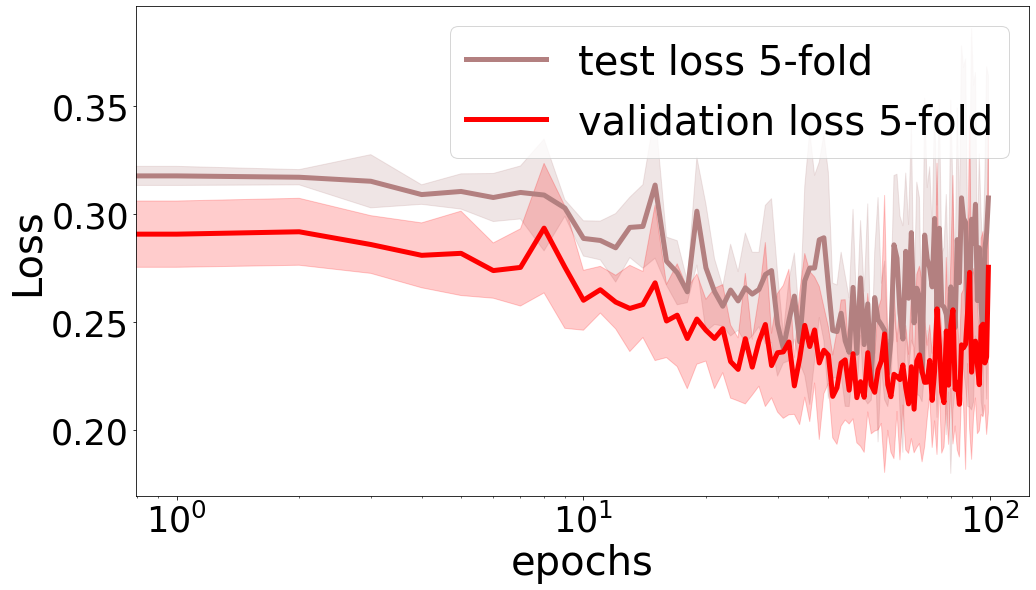}%
		\hspace{0.5em}%
		\includegraphics[width=0.31\textwidth, height=0.174\textwidth]{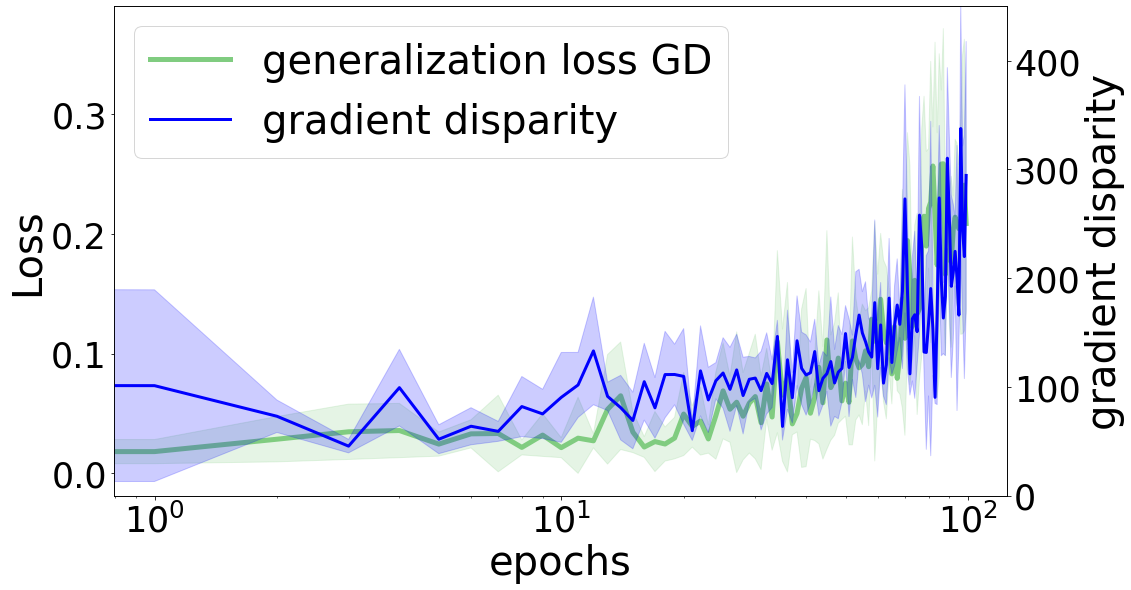}%
		\hspace{0.5em}%
		\includegraphics[width=0.31\textwidth, height=0.174\textwidth]{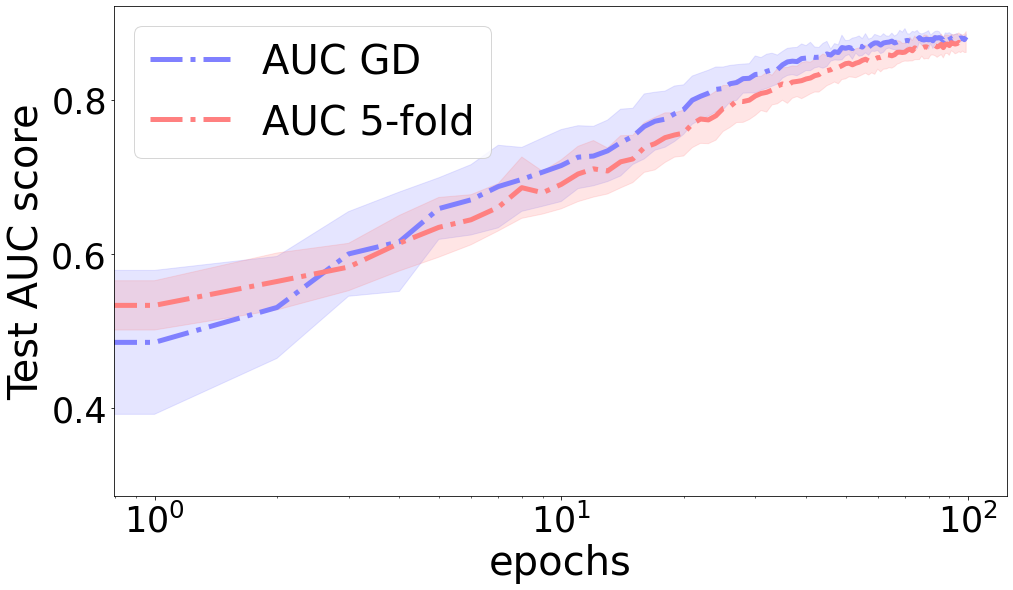}
		\caption{Task: detecting abnormality}
	\end{subfigure}
		\caption{Detecting three tasks from the MRNet dataset from the sagittal plane MRI scans. (left) Validation loss versus test loss in 5-fold cross-validation. (middle) Gradient disparity versus generalization loss. (right) Performance comparison on the final unseen data when applying 5-fold CV versus gradient disparity. For the results of applying early stopping refer to Table~\ref{tab:mrnet}.} \label{fig:mrnet}
	\end{figure*}

	The results for three tasks (detecting ACL tears, meniscal tears and abnormality) are shown in Fig.~\ref{fig:mrnet} and Table~\ref{tab:mrnet}. We can observe that both the validation loss (despite a small bias) and the gradient disparity predict the generalization loss quite well.
	Yet, when using gradient disparity, the final test AUC score is higher (Fig.~\ref{fig:mrnet}~(right)). 
	As mentioned above, for this dataset, both the validation loss and gradient disparity vary a lot. Hence, in Table~\ref{tab:mrnet}, we show the results of early stopping, both when the metric has increased for 5 epochs from the beginning of training, and between parenthesis when the metric has increased for 5 consecutive epochs.
	We conclude that with both approaches, the use of gradient disparity as an early stopping criterion results in more than $1\%$ improvement in the test AUC score. 
	Because the test set used in \cite{bien2018deep} is not publicly available, it is not possible to compare our predictive results with~\cite{bien2018deep}. 
	Nevertheless, we can take as a baseline the results presented in the work given at \url{https://github.com/ahmedbesbes/mrnet}, which report a test AUC score of $88.5\%$ for the task of detecting ACL tears.  
	We observe in Table~\ref{tab:mrnet} that stopping training after $5$ consecutive increases in gradient disparity leads to $91.52\%$ test AUC score for this task. With further tuning, and combining the predictions found on two other MRI planes of each patient (axial and coronal), our final prediction results could even be improved.
	
\clearpage

%% file: add.tex
	\clearpage
	\section{Additional Experiments}\label{app:more}

	In this section, we provide additional experiments on benchmark image-classification datasets.
	
	\vspace{-0.5em}
	\subsection{MNIST Experiments}\label{sec:mnist}
	Fig. \ref{fig:random} shows the test error for networks trained with different amounts of label noise. Interestingly, observe that for this setting the test error for the network trained with $75\%$ label noise remains relatively small, indicating a good resistance of the model against memorization of corrupted samples.
	As suggested both from the test error (Fig.~\ref{fig:random}~(a)) and gradient disparity (Fig.~\ref{fig:random}~(c)), there is no proper early stopping time for these experiments\footnote{According to Table~\ref{tab:patience}, for the noisy MNIST dataset, the patience value of $p=10$ is preferred for both GD and CV. Therefore, in Fig.~\ref{fig:random}~(c), even for the setting with $75\%$ label noise, gradient disparity does not increase for $p=10$ consecutive iterations, and would therefore not signal overfitting throughout training. }. The generalization error (Fig.~\ref{fig:random}~(b)) 
	remains close to zero, regardless of the level of label noise, and hence fails to account for label noise. In contrast,  gradient disparity is very sensitive to the label noise level in \emph{all} stages of training, even at early stages of training, as desired for a metric measuring generalization.  
	
	Fig.~\ref{fig:BGP_vs_gen_m} shows the results for an AlexNet \cite{krizhevsky2012imagenet} trained on the MNIST dataset\footnote{\url{http://yann.lecun.com/exdb/mnist/}}. This model generalizes quite well for this dataset. We observe that, throughout the training, the test curves are even below the training curves, which is due to the dropout regularization technique \cite{srivastava2014dropout} being applied during training and not during testing. The generalization loss/error is almost zero, until around iteration 1100 (indicated in the figure by the gray vertical bar), which is when overfitting starts and the generalization error becomes non-zero, and when gradient disparity signals to stop training.
	
	Fig. \ref{fig:fc_mnist} shows the results for a 4-layer fully connected neural network trained on the entire MNIST training set. Figs.~\ref{fig:fc_mnist} (e) and (f) show the generalization losses. We observe that at the early stages of training, generalization losses do not distinguish between different label noise levels, whereas gradient disparity does so from the beginning (Figs.~\ref{fig:fc_mnist}~(g) and (h)). At the middle stages of training we can observe that, surprisingly in this setting, the network with 0\% label noise has higher generalization loss than the networks trained with 25\%, 50\% and 75\% noise, and this is also captured by gradient disparity. The final gradient disparity values for the networks trained with higher label noise level are also larger. For the network trained with 0\% label noise we show the results with more details in Fig.~\ref{fig:fc_mnist0},
	and observe again how gradient disparity is well aligned with the generalization loss/error. In this experiment, the early stopping time suggested by gradient disparity is epoch 9, which is the exact same time when the training and test losses/errors start to diverge, and signals therefore the start of overfitting.

	\subsection{CIFAR-10 Experiments}\label{sec:cifar10}
	
\begin{figure*}[t]
			\centering
			\begin{subfigure}[]{0.32\textwidth}            
				\includegraphics[width=\textwidth, height=0.5625\textwidth]{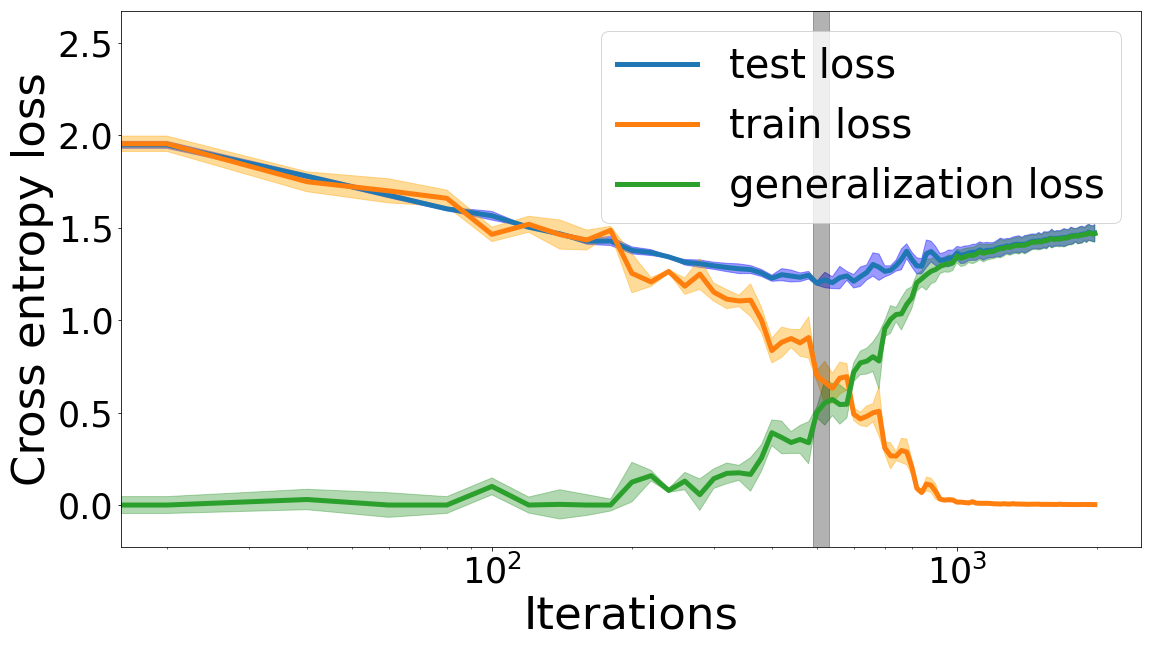}
				\caption{Cross entropy loss}
			\end{subfigure}\hspace{0.9em}%
			\begin{subfigure}[]{0.32\textwidth}            
				\includegraphics[width=\textwidth, height=0.5625\textwidth]{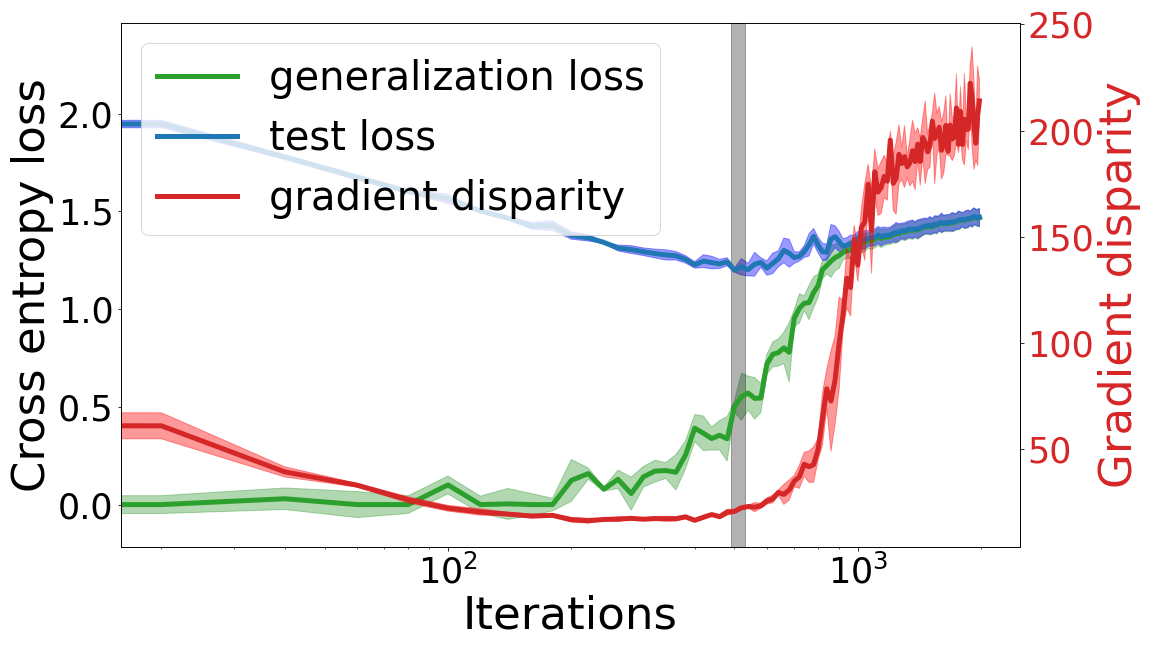}
				\caption{$\overline{\mathcal{D}}$ vs. loss}
			\end{subfigure}\hspace{0.3em}%
			\begin{subfigure}[]{0.32\textwidth}
				\centering
				\includegraphics[width=\textwidth, height=0.5625\textwidth]{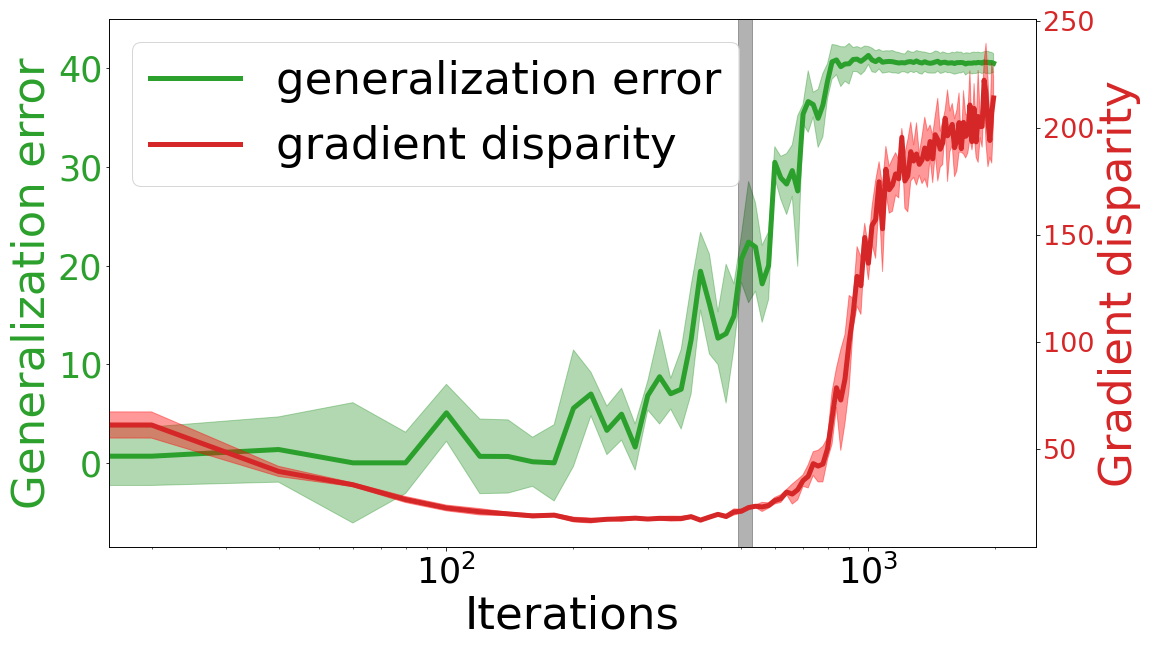}
				\caption{$\overline{\mathcal{D}}$ vs. error}
			\end{subfigure}
			\caption{The cross entropy loss, the error percentage and the average gradient disparity during training for a ResNet-18 trained on a subset of 12.8 k points of the CIFAR-10 training set (the parameter initialization is Xavier \cite{glorot2010understanding}). Pearson's correlation coefficient $\rho$ between $\overline{\mathcal{D}}$ and generalization loss/error over all the training iterations are $\rho_{\overline{\mathcal{D}}, \text{gen loss}} = 0.755$ and $\rho_{\overline{\mathcal{D}}, \text{gen error}} = 0.846$. }\label{fig:BGP_vs_gen}
		\end{figure*}

	Fig.~\ref{fig:BGP_vs_gen} shows the results for a ResNet-18 \cite{he2016deep} trained on the CIFAR-10 dataset\footnote{\url{https://www.cs.toronto.edu/~kriz/cifar.html}}.
	Around iteration~500 (which is indicated by a thick gray vertical bar in the figures), the training and test losses (and errors) start to diverge, and the test loss reaches its minimum. This is indeed when gradient disparity increases and signals overfitting. 
	
	To compare models with a different number of parameters using gradient disparity, we need to normalize it. The dimension of a gradient vector is the number $d$ of parameters of the model. Gradient disparity being the $\ell_2$-norm of the difference of gradient vectors will thus grow proportionally to $\sqrt{d}$, hence to compare different architectures, we propose to use the normalized gradient disparity $\tilde{\mathcal{D}} = \overline{\mathcal{D}}/{\sqrt{d}}$.
	We observe in Fig.~\ref{fig:width2} that both the normalized\footnote{Note that the normalization with respect to the number of parameters is different than the normalization mentioned in Section~\ref{app:norm} which was with respect to the loss values. The value of gradient disparity reported everywhere is the re-scaled gradient disparity; further if comparison between two different architectures is taking place the normalization with respect to dimensionality will also take place.} gradient disparity and test error decrease with the network width (the \emph{scale} is a hyper-parameter used to change both the number of channels and hidden units in each configuration).

	Fig. \ref{fig:fc_cifar10} shows the results for a 4-layer fully connected neural network, which is trained on the entire CIFAR-10 training set. We observe that gradient disparity reflects the test error at the early stages of training quite well. In the later stages of training we observe that the ranking of gradient disparity values for different label noise levels matches with the ranking of generalization losses and errors.
	In all experiments the gradient disparity is indeed very informative about the test error. 
	
	The test error decreases with the size of the training set (Fig.~\ref{fig:DA} (bottom)) and a reliable signal of overfitting should therefore reflect this property. Many of the previous metrics fail to do so, as shown by \cite{neyshabur2017exploring,nagarajan2019uniform}. 
	In contrast, gradient disparity indeed decreases with the training set size, as shown in Fig.~\ref{fig:DA}~(top) and Fig.~\ref{fig:train_size2} in Appendix~\ref{app:more}.
	In Figure~\ref{fig:DA}, we study the effect of data augmentation \cite{shorten2019survey}, which is one of the popular techniques used to reduce overfitting given limited labeled data. 
	Consistently with the rest of the paper, we observe a strong positive correlation ($\rho=0.979$) between the test error and gradient disparity for networks that are trained with data augmentation. Moreover, we observe that applying data augmentation decreases the values of both gradient disparity and the test error.

	Fig.~\ref{fig:train_size2} shows the test error and gradient disparity for networks that are trained with different training set sizes. In Fig.~\ref{fig:batch_size2}, we observe that, as discussed in Section~\ref{sec:gen}, gradient disparity, similarly to the test error, increases with the batch size for not too large batch sizes. As expected, when the batch size is very large (512 for the CIFAR-10 experiment and 256 for the CIFAR-100 experiments) gradient disparity starts to decrease, because gradient vectors are averaged over a large batch. 
	Note that even with such large batch sizes, gradient disparity correctly detects the early stopping time, although it can no longer be compared to the value of gradient disparity found with other batch sizes.

	\vspace{-0.5em}
	\subsection{CIFAR-100 Experiments}\label{sec:cifar100}
	Fig. \ref{fig:resnet_cifar100} shows the results for a ResNet-18 that is trained on the CIFAR-100 training set\footnote{\url{https://www.cs.toronto.edu/~kriz/cifar.html}}. Clearly, the model is not sufficient to learn the complexity of the CIFAR-100 dataset: It has $99\%$ error for the network with $0\%$ label noise, as if it had not learned anything about the dataset and is just making a random guess for classification (because there are 100 classes, random guessing would give $99\%$ error on average). We observe from Fig. \ref{fig:resnet_cifar100} (f) that as training progresses, the network overfits more, and the generalization error increases. Although the test error is high (above $90\%$), very surprisingly for this example, the networks with higher label noise level have a lower test loss and error (Figs.~\ref{fig:resnet_cifar100} (b) and (d)). Quite interestingly, gradient disparity (Fig.~\ref{fig:resnet_cifar100} (g)) captures also this surprising trend as well.

\input{mnist.tex}

\input{cifar10.tex}

\input{cifar100.tex}

%% file: mnist.tex
	\begin{figure*}[t]
		\begin{subfigure}[]{0.32\textwidth}            
			\includegraphics[width=\textwidth, height=0.5625\textwidth]{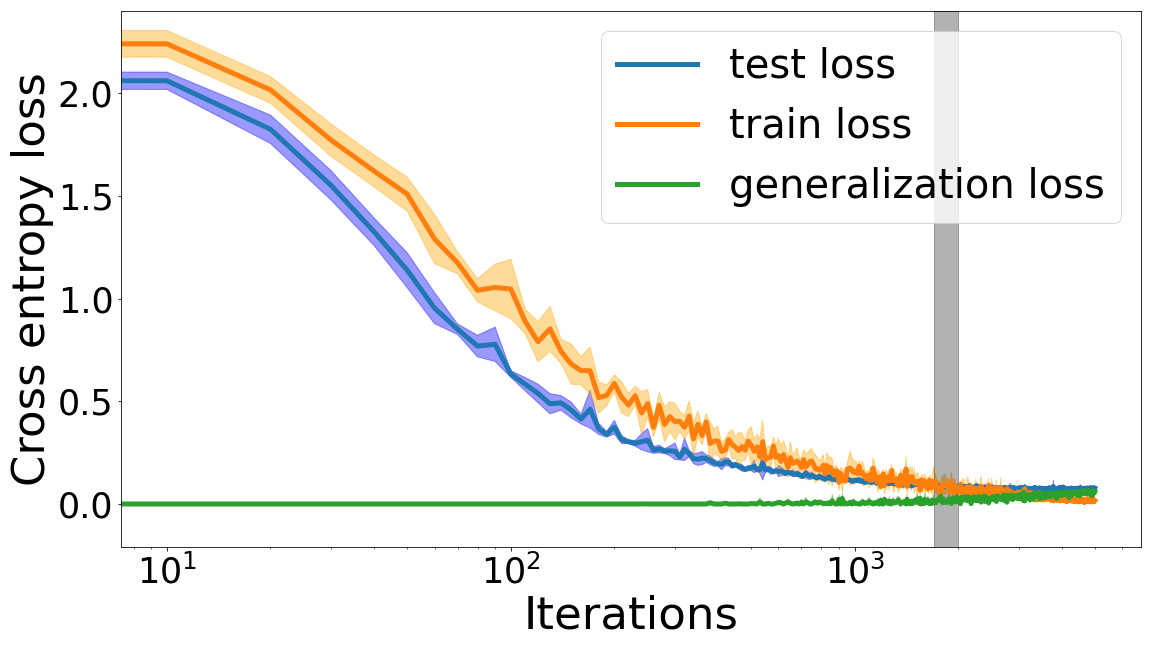}
			\caption{Cross entropy loss}
		\end{subfigure}\hspace{0.9em}%
		\begin{subfigure}[]{0.32\textwidth}            
			\includegraphics[width=\textwidth, height=0.5625\textwidth]{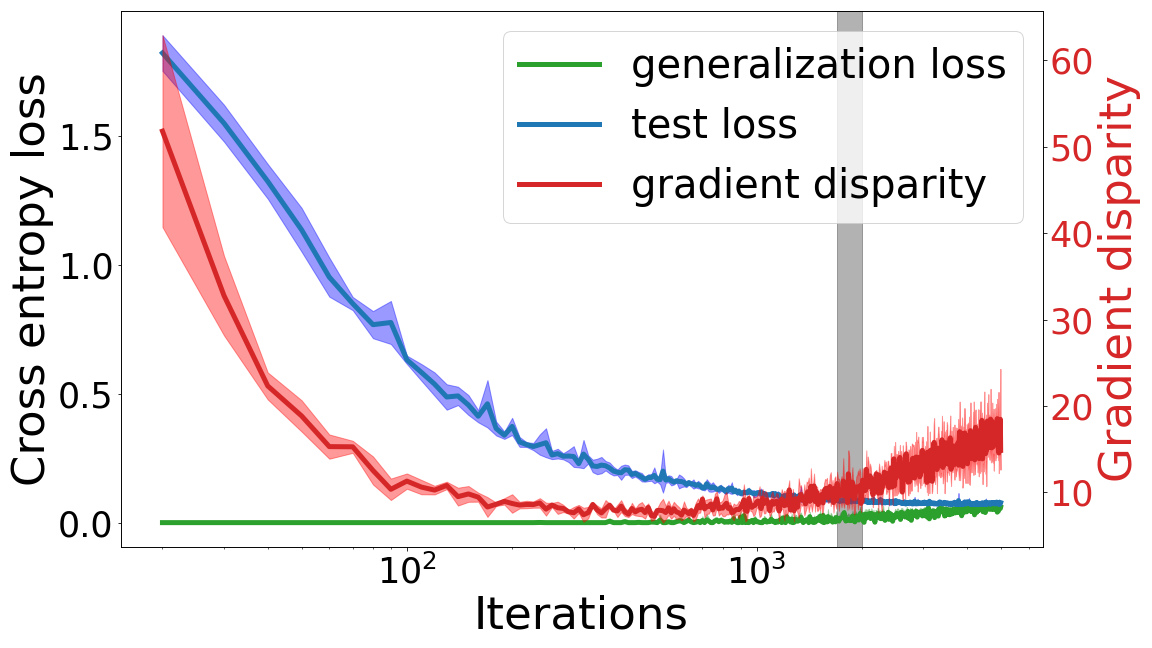}
			\caption{$\overline{\mathcal{D}}$ vs. loss}
		\end{subfigure}\hspace{0.3em}%
		\begin{subfigure}[]{0.32\textwidth}
			\centering
			\includegraphics[width=\textwidth, height=0.5625\textwidth]{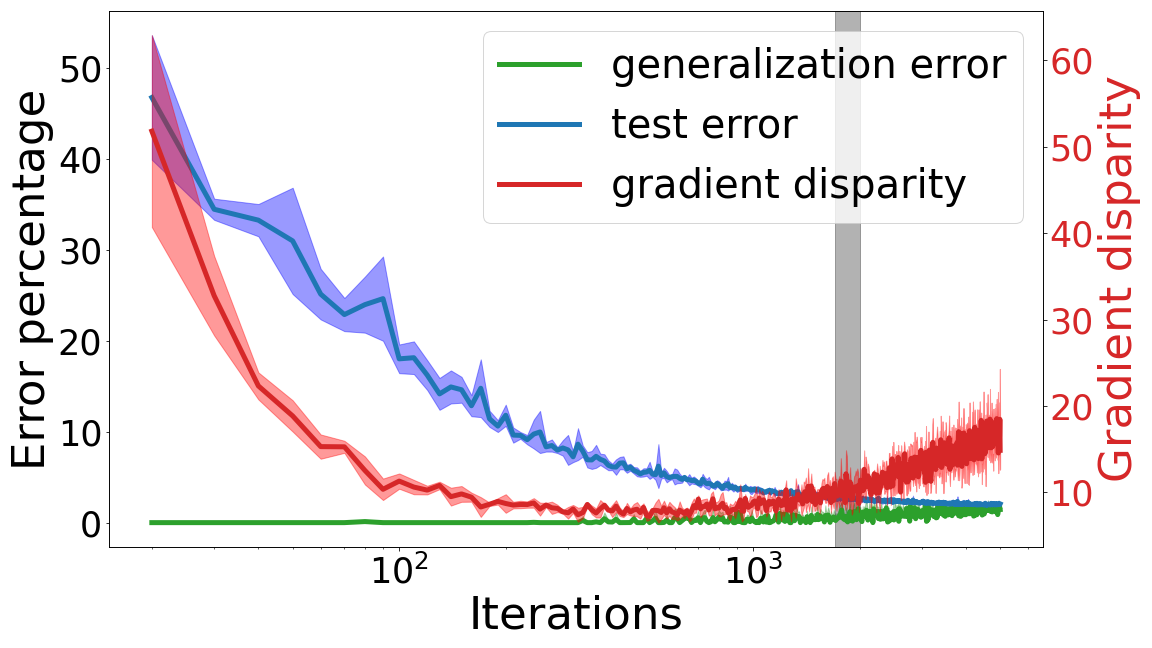}
			\caption{$\overline{\mathcal{D}}$ vs. error}
		\end{subfigure}
		\caption{The cross entropy loss, the error percentage and the average gradient disparity during training for an AlexNet trained on a subset of 12.8 k points of the MNIST training set (the parameters are initialized according to the He \cite{he2015delving} method with Normal distributions). For this experiment, $\rho_{\overline{\mathcal{D}}, \text{gen loss}} = 0.465$ and $\rho_{\overline{\mathcal{D}}, \text{gen error}} = 0.457$. The blue, orange, green, and red curves are the test loss/error, train loss/error, generalization loss/error, and the average gradient disparity $\overline{\mathcal{D}}$, respectively.}\label{fig:BGP_vs_gen_m}
	\end{figure*}

	\begin{figure*}
		\centering
		\begin{subfigure}[b]{0.24\textwidth}            
			\includegraphics[width=\textwidth, height=0.5625\textwidth]{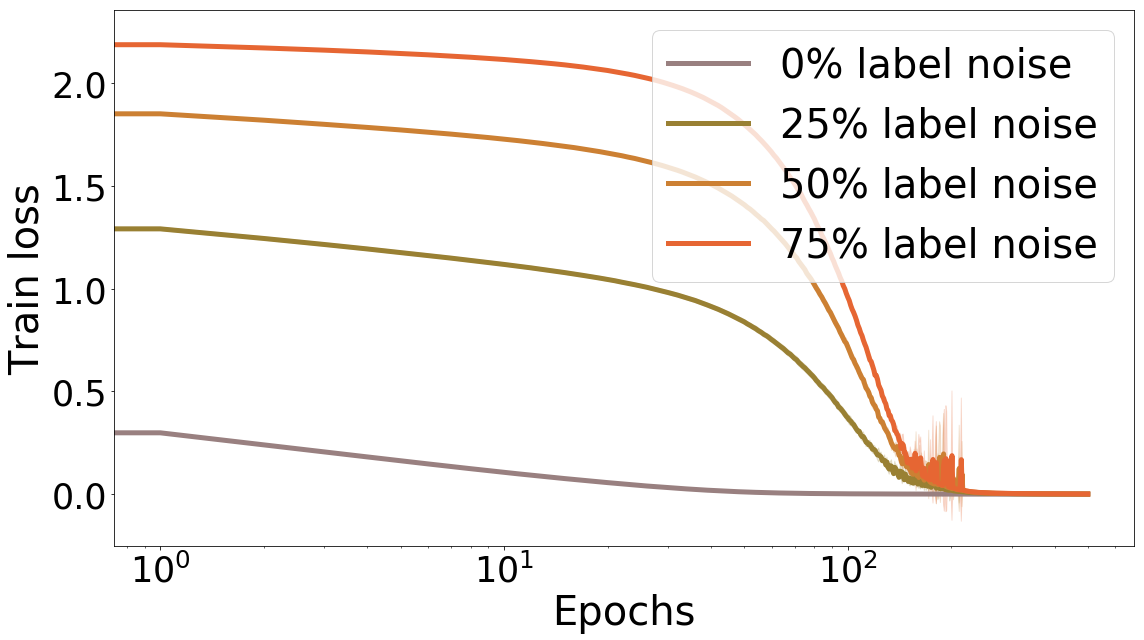}
			\caption{Training loss}
		\end{subfigure}%
		\hspace{0.5em}%
		\begin{subfigure}[b]{0.24\textwidth}
			\centering
			\includegraphics[width=\textwidth, height=0.5625\textwidth]{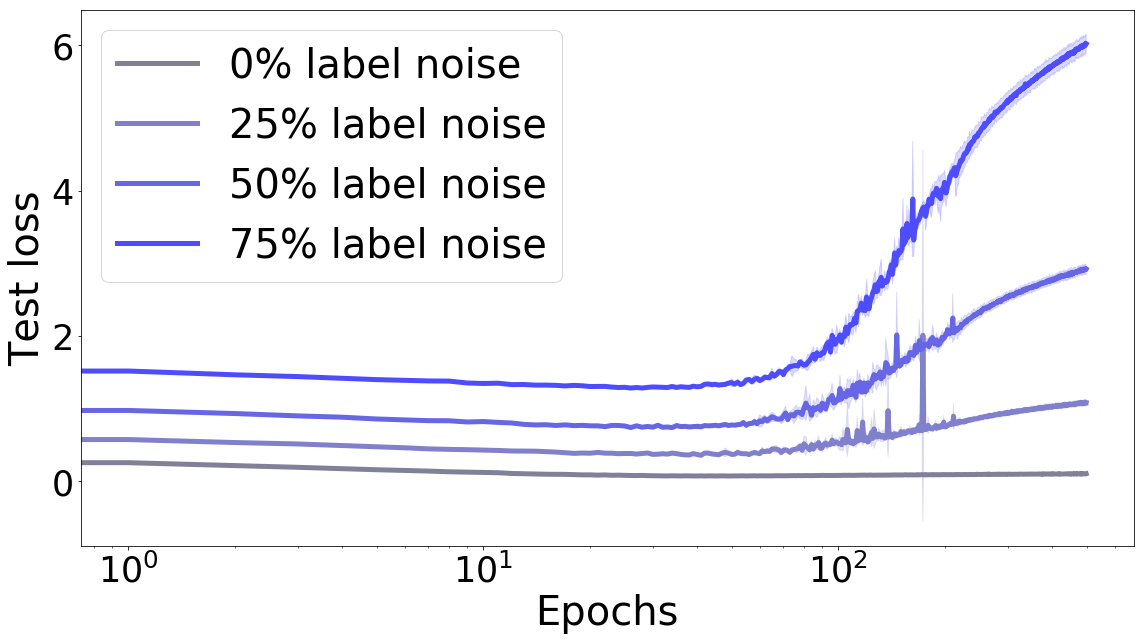}
			\caption{Test loss}
		\end{subfigure}%
		\hspace{0.5em}%
		\begin{subfigure}[b]{0.24\textwidth}            
			\includegraphics[width=\textwidth, height=0.5625\textwidth]{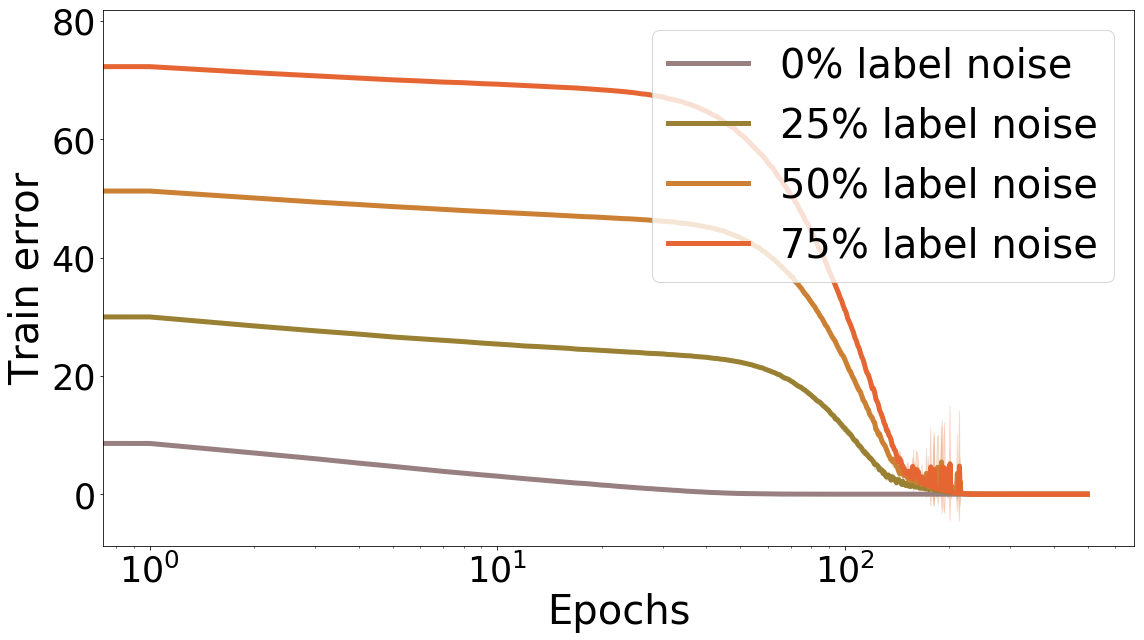}
			\caption{Training error}
		\end{subfigure}%
		\hspace{0.5em}%
		\begin{subfigure}[b]{0.24\textwidth}
			\centering
			\includegraphics[width=\textwidth, height=0.5625\textwidth]{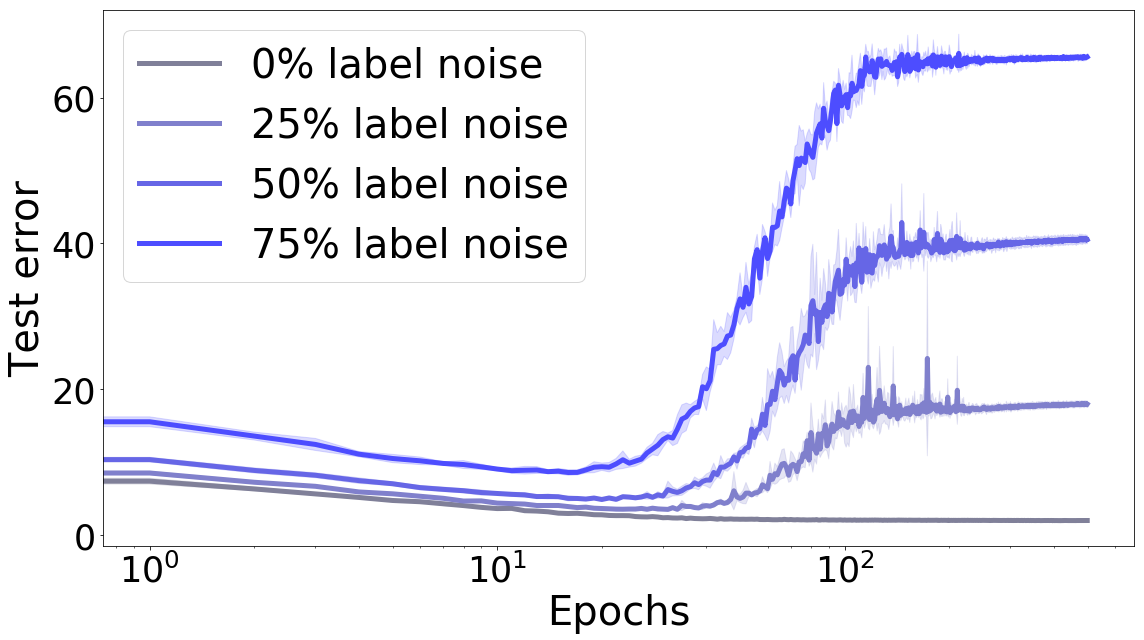}
			\caption{Test error}
		\end{subfigure}\\
		\begin{subfigure}[b]{0.35\textwidth}            
			\includegraphics[width=\textwidth, height=0.5625\textwidth]{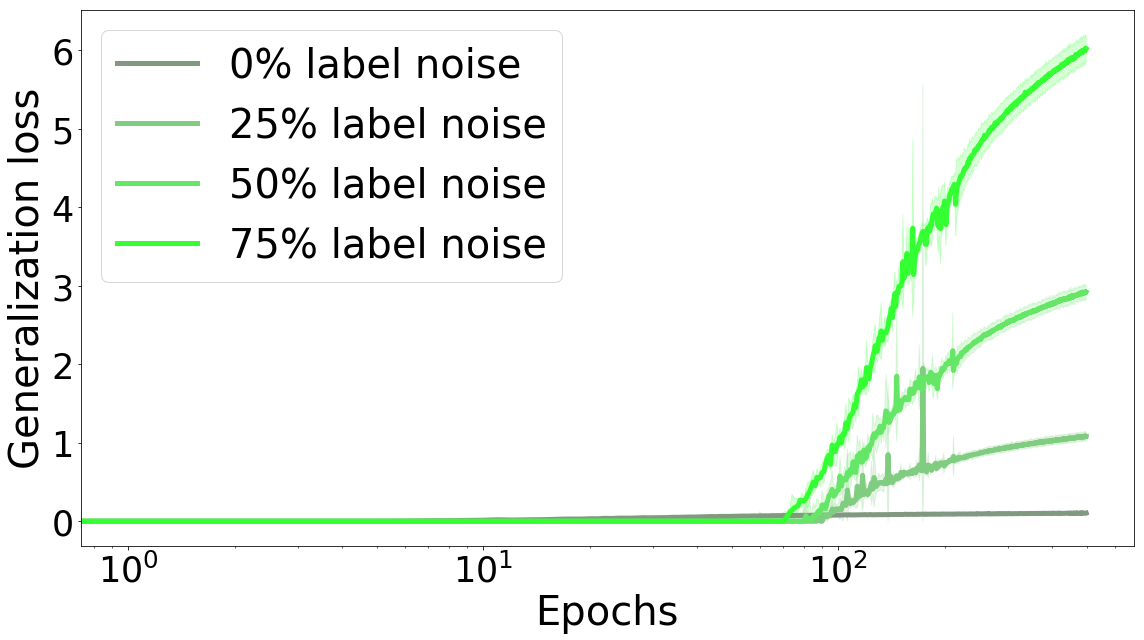}
			\caption{Generalization loss}
		\end{subfigure}%
		\hspace{1.5em}%
		\begin{subfigure}[b]{0.35\textwidth}            
			\includegraphics[width=\textwidth, height=0.5625\textwidth]{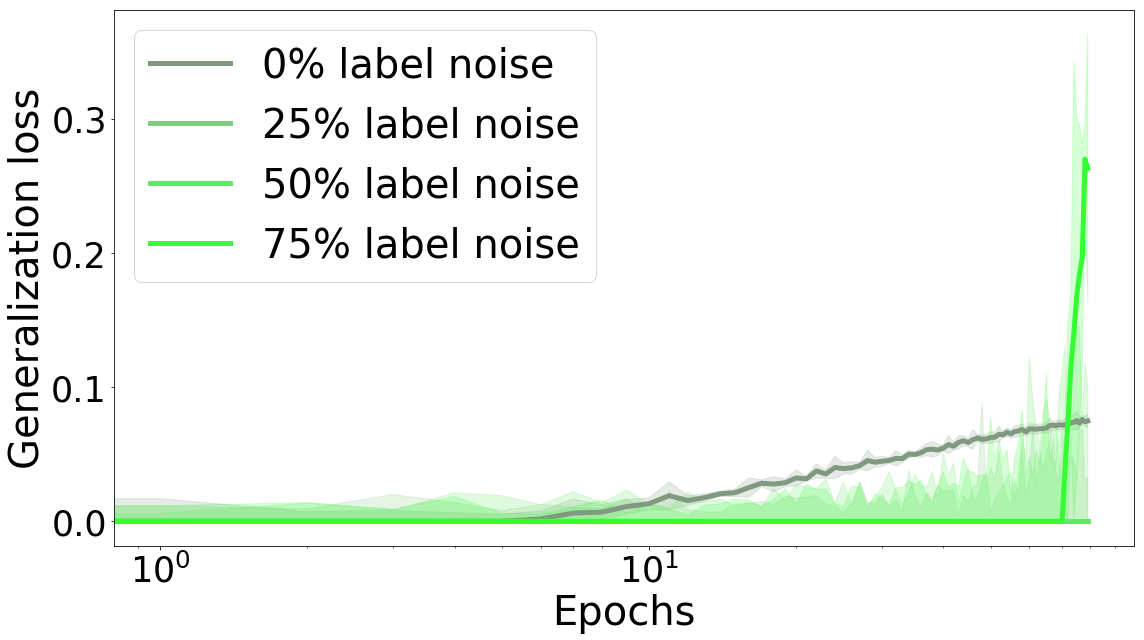}%
			\caption{Generalization loss for epoch $< 80$}
		\end{subfigure}
		\begin{subfigure}[b]{0.35\textwidth}
			\centering
			\includegraphics[width=\textwidth, height=0.5625\textwidth]{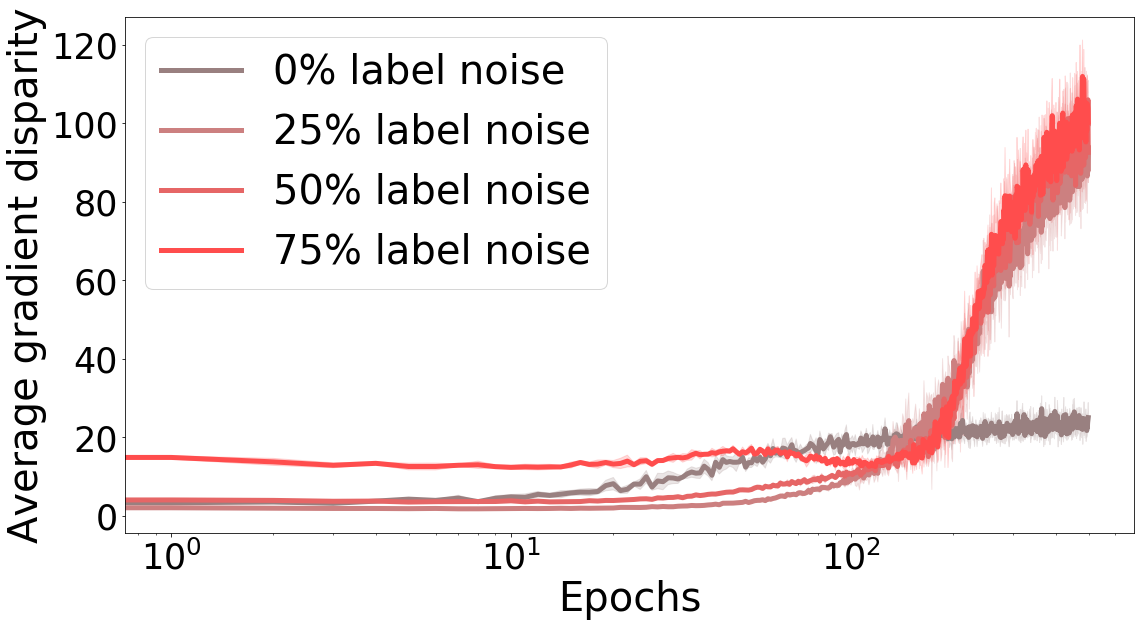}
			\caption{$\overline{\mathcal{D}}$}
		\end{subfigure}%
		\hspace{1.5em}%
		\begin{subfigure}[b]{0.35\textwidth}
			\centering
			\includegraphics[width=\textwidth, height=0.5625\textwidth]{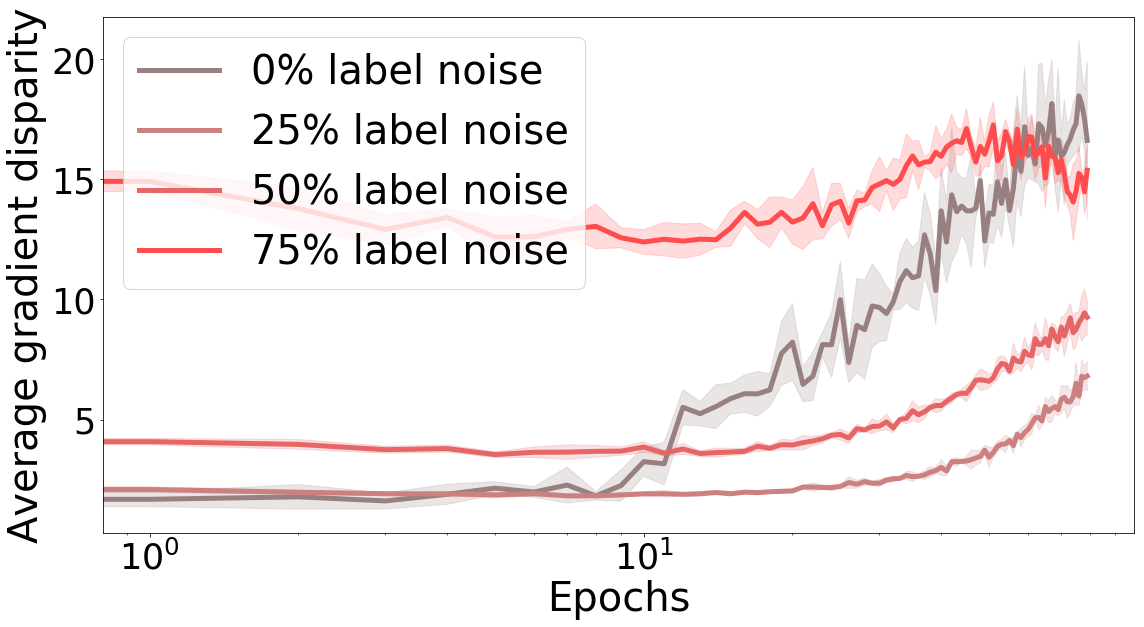}
			\caption{$\overline{\mathcal{D}}$ for epoch $< 80$}
		\end{subfigure}
		\caption{The cross entropy loss, error percentage, and average gradient disparity during training with different amounts of randomness in the training labels for a 4-layer fully connected neural network with 500 hidden units trained on the entire MNIST dataset. The parameter initialization is the He initialization with normal distribution. }\label{fig:fc_mnist}
	\end{figure*}
	
	\begin{figure*}
		\centering
		\begin{subfigure}[b]{0.5\textwidth}            
			\includegraphics[width=0.97\textwidth, height=0.5456\textwidth]{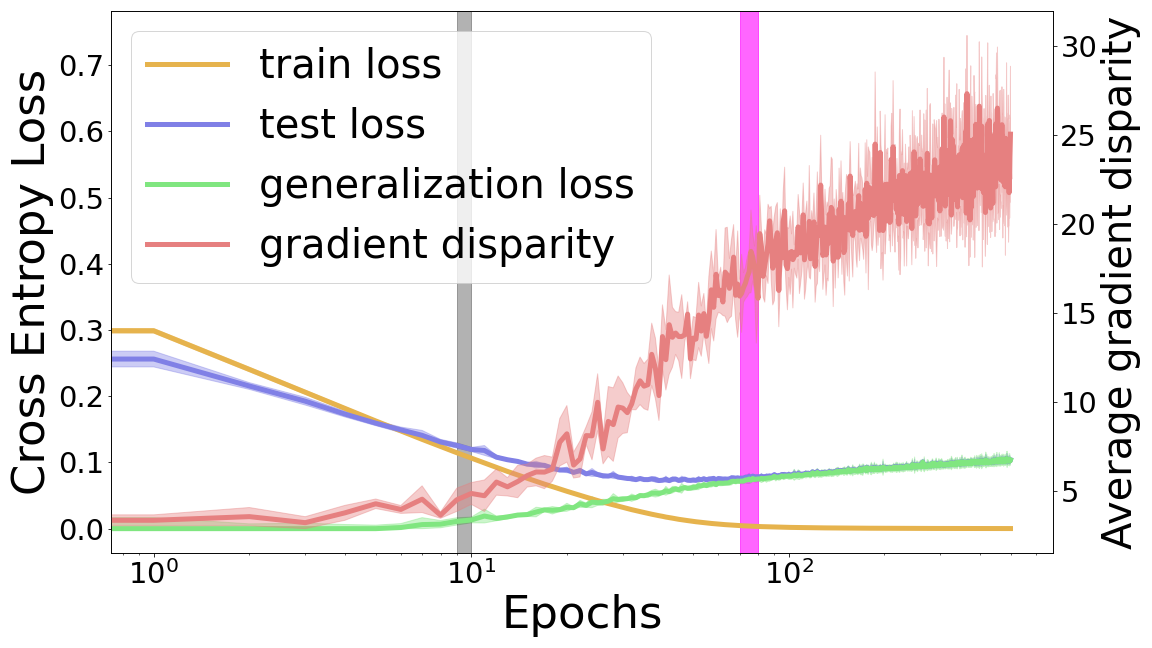}
			\caption{Loss vs gradient disparity}
		\end{subfigure}%
		\begin{subfigure}[b]{0.5\textwidth}
			\centering
			\includegraphics[width=0.97\textwidth, height=0.5456\textwidth]{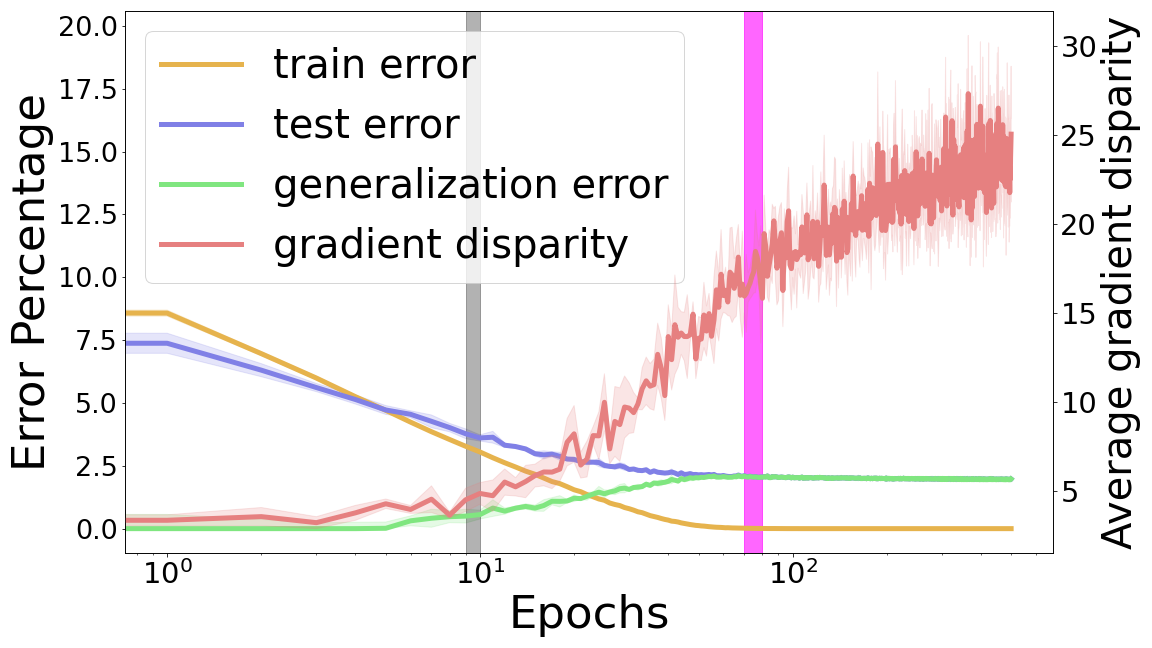}
			\caption{Error vs gradient disparity}
		\end{subfigure}%
		\caption{The cross entropy loss, error percentage, and average gradient disparity during training for a 4-layer fully connected neural network with 500 hidden units trained on the entire MNIST dataset with 0\% label noise. The parameter initialization is the He initialization with normal distribution. Pearson's correlation coefficient $\rho$ between $\overline{\mathcal{D}}$ and generalization loss/error over all the training iterations are $\rho_{\overline{\mathcal{D}}, \text{gen loss}} = 0.967$ and $\rho_{\overline{\mathcal{D}}, \text{gen error}} = 0.734$. The gray vertical bar indicates when GD increases for 5 epochs from the beginning of training. The magenta vertical bar indicates when GD increases for 5 \emph{consecutive} epochs. We observe that the gray bar signals when overfitting is starting, which is when the training and testing curves are starting to diverge. The magenta bar would be a good stopping time, because if we train beyond this point, although the test error remains the same, the test loss would increase, which would result in overconfidence on wrong predictions.}\label{fig:fc_mnist0}
	\end{figure*}

%% file: cifar10.tex
	\begin{figure*}[t]
		\begin{subfigure}[b]{1\textwidth} 
			\includegraphics[width=0.5\textwidth, height=0.28\textwidth]{scale_3.png}%
			\includegraphics[width=0.5\textwidth, height=0.28\textwidth]{scale_1.png}
			\caption{${\rho_{\tilde{\mathcal{D}}, \text{TL}} = 0.970}$ and ${\rho_{\tilde{\mathcal{D}}, \text{TE}} = 0.939}$}
		\end{subfigure}
		\begin{subfigure}[b]{1\textwidth} 
			\includegraphics[width=0.5\textwidth, height=0.28\textwidth]{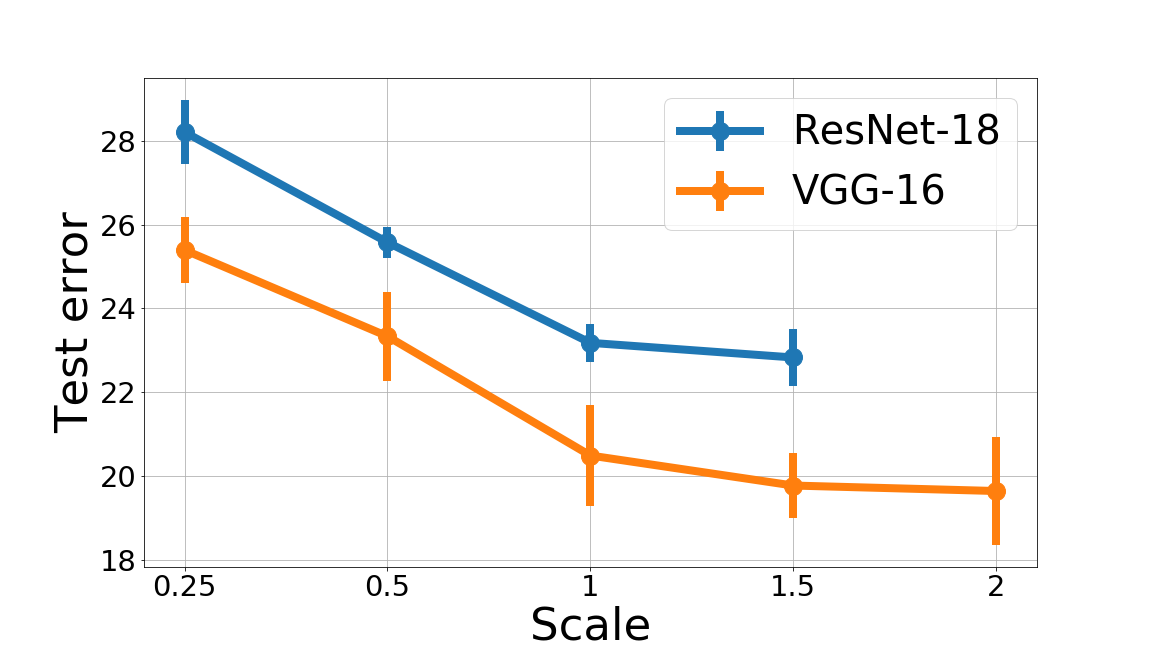}%
			\includegraphics[width=0.5\textwidth, height=0.28\textwidth]{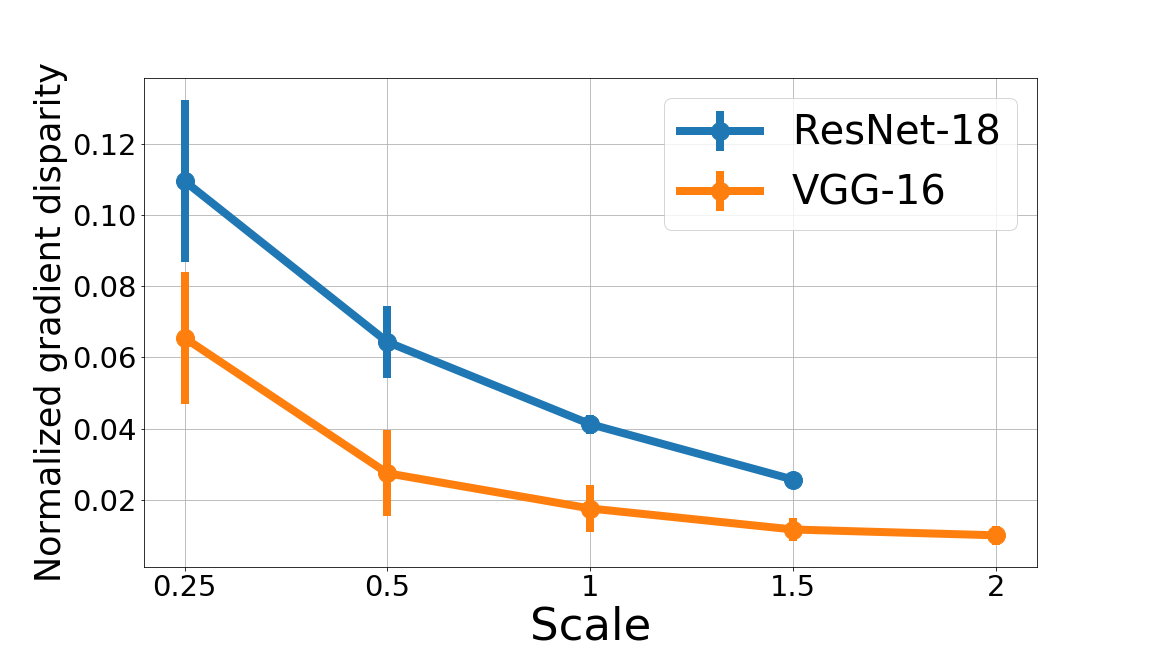}%
			
			\caption{$\rho_{\tilde{\mathcal{D}}, \text{TL}}=0.655, \rho_{\tilde{\mathcal{D}}, \text{TE}}=0.958$}  
		\end{subfigure}
	\begin{subfigure}[b]{1\textwidth} 
		\includegraphics[width=0.5\textwidth, height=0.28\textwidth]{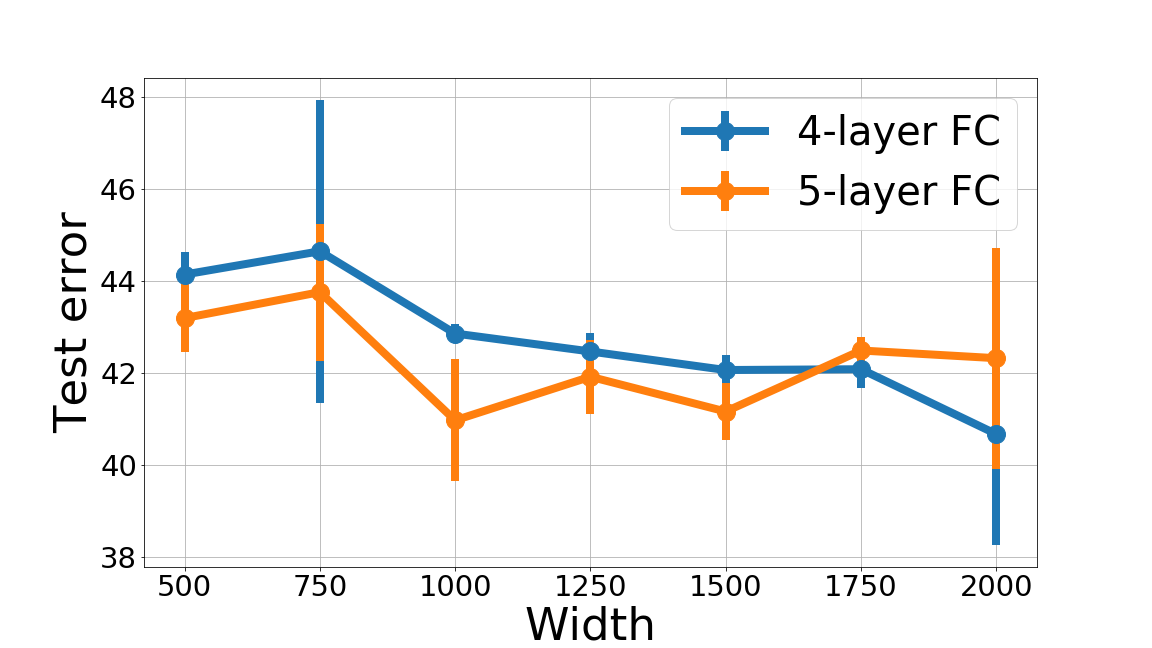}%
		\includegraphics[width=0.5\textwidth, height=0.28\textwidth]{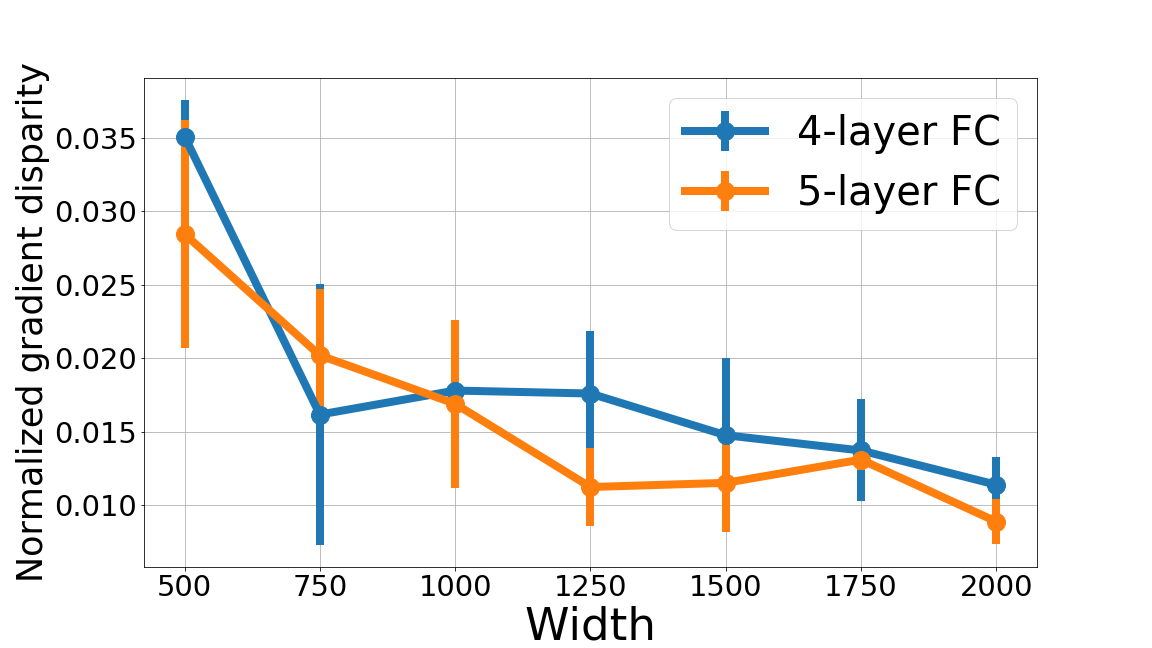}
		\caption{$\rho_{\tilde{\mathcal{D}}, \text{TL}}=0.771, \rho_{\tilde{\mathcal{D}}, \text{TE}}=0.601$}
	\end{subfigure}
		\caption{Test error and normalized gradient disparity for networks trained 
			on the CIFAR-10 dataset with different number of channels and hidden units for convolutional neural networks (CNN) (scale = 1 recovers the original configurations) and fully connected neural networks (FC). The correlation between normalized gradient disparity and test loss $\rho_{\tilde{\mathcal{D}}, \text{TL}}$ and between normalized gradient disparity and test error $\rho_{\tilde{\mathcal{D}}, \text{TE}}$ are reported in the captions.}\label{fig:width2}
	\end{figure*}

	\begin{figure*}[t]
		\centering
		\begin{subfigure}[b]{0.24\textwidth}            
			\includegraphics[width=\textwidth, height=0.5625\textwidth]{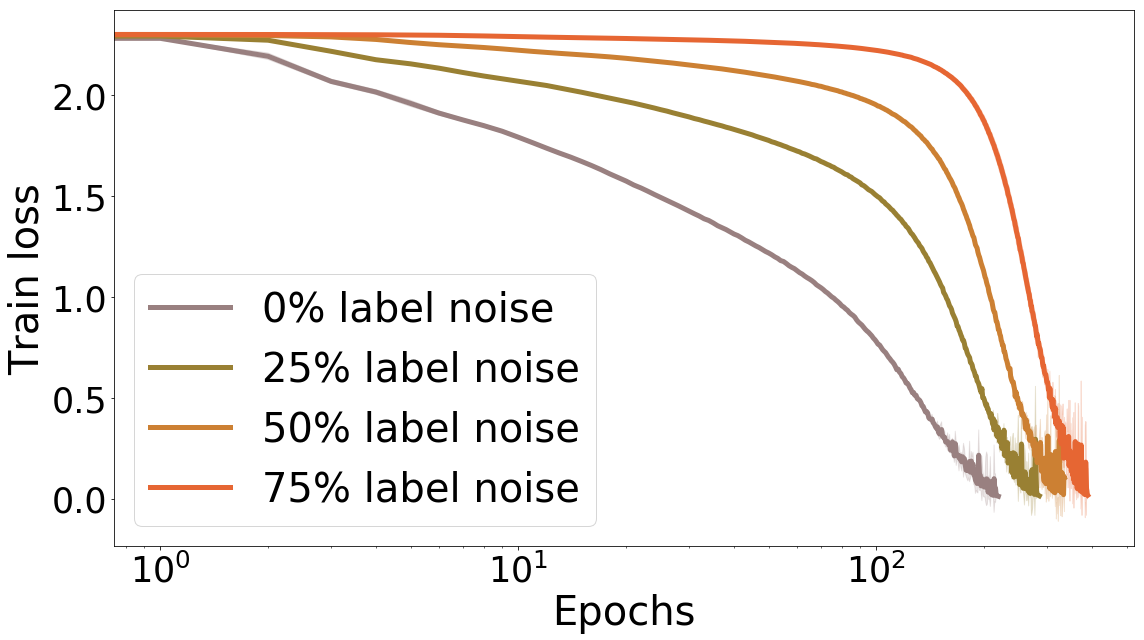}
			\caption{Training loss}
		\end{subfigure}%
	\hspace{0.5em}%
		\begin{subfigure}[b]{0.24\textwidth}
			\centering
			\includegraphics[width=\textwidth, height=0.5625\textwidth]{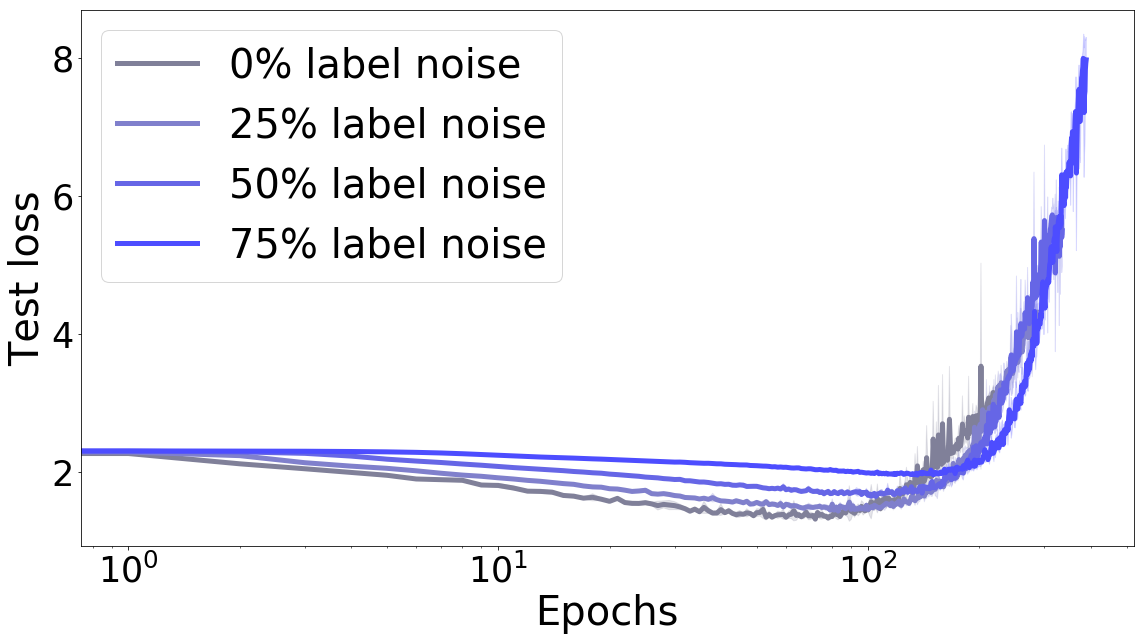}
			\caption{Test loss}
		\end{subfigure}%
	\hspace{0.5em}%
		\begin{subfigure}[b]{0.24\textwidth}            
			\includegraphics[width=\textwidth, height=0.5625\textwidth]{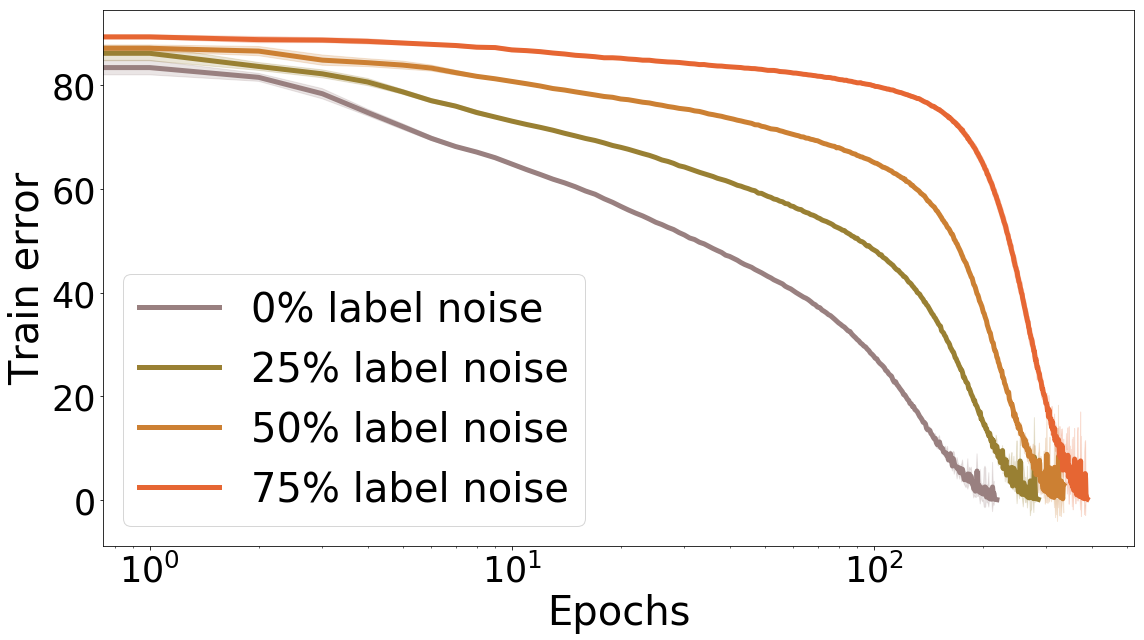}
			\caption{Training error}
		\end{subfigure}%
	\hspace{0.5em}%
		\begin{subfigure}[b]{0.24\textwidth}
			\centering
			\includegraphics[width=\textwidth, height=0.5625\textwidth]{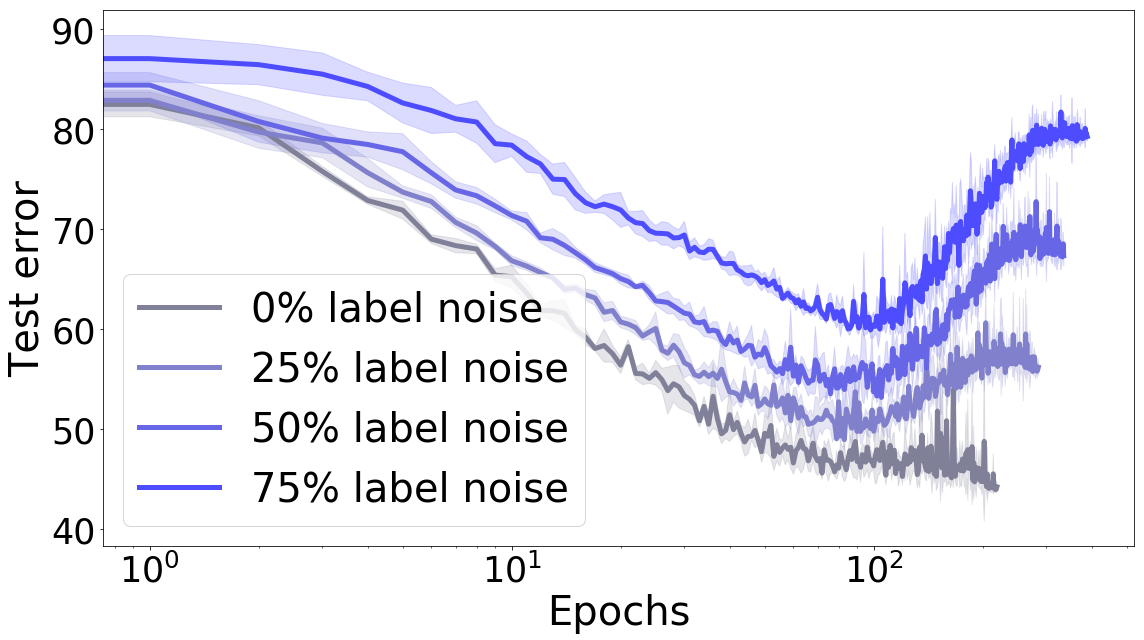}
			\caption{Test error}
		\end{subfigure}\\
		\begin{subfigure}[b]{0.35\textwidth}            
			\includegraphics[width=\textwidth, height=0.5625\textwidth]{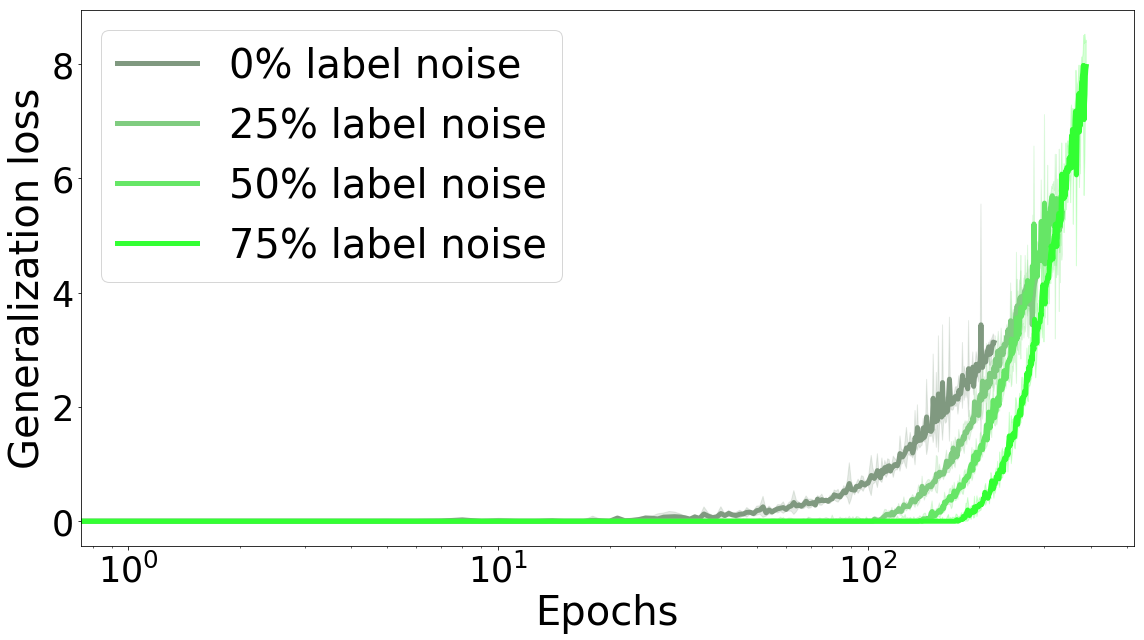}
			\caption{Generalization loss}
		\end{subfigure}%
	\hspace{1.5em}%
		\begin{subfigure}[b]{0.35\textwidth}            
			\includegraphics[width=\textwidth, height=0.5625\textwidth]{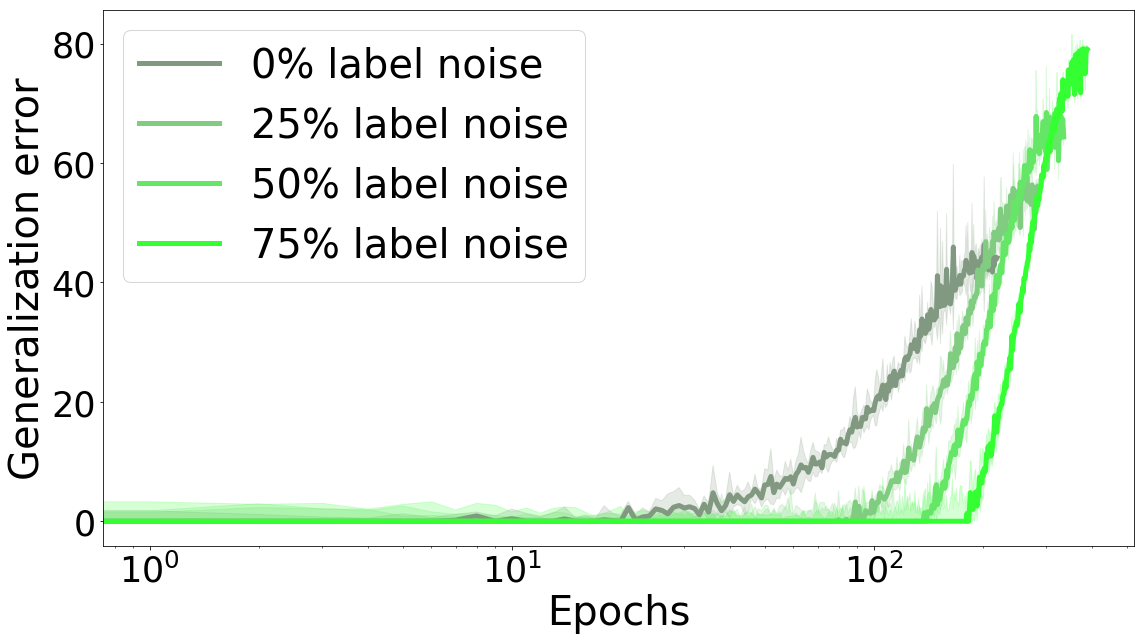}
			\caption{Generalization error}
		\end{subfigure}
	
		\begin{subfigure}[b]{0.35\textwidth}
			\centering
			\hspace{-1em}
			\includegraphics[width=1.05\textwidth, height=0.5625\textwidth]{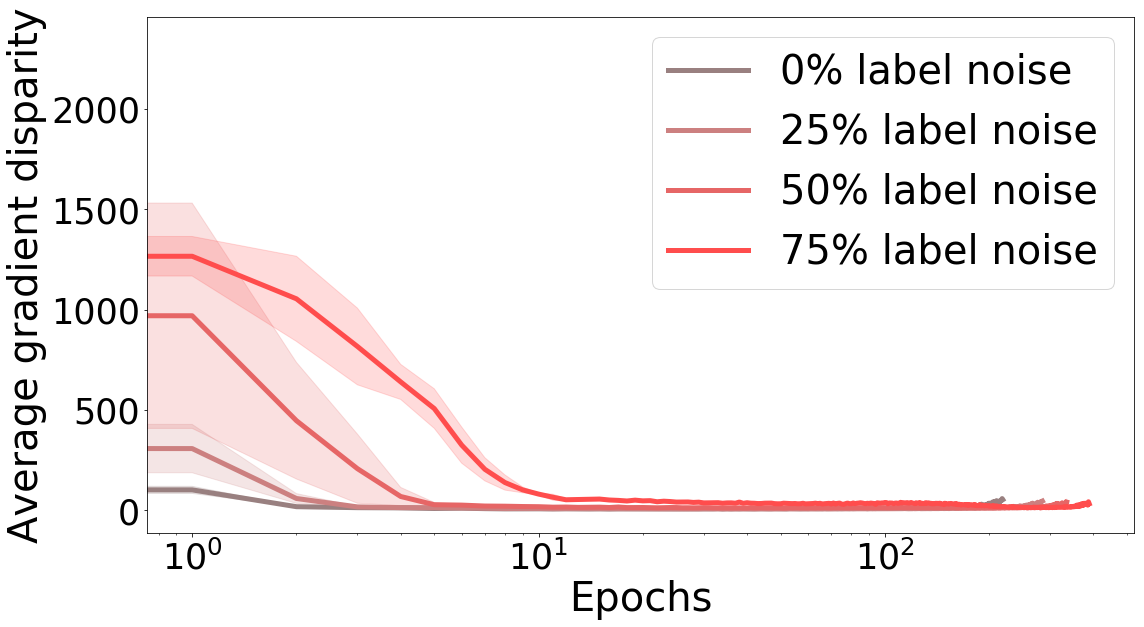}
			\caption{$\overline{\mathcal{D}}$}
		\end{subfigure}%
	\hspace{1.5em}%
	\begin{subfigure}[b]{0.35\textwidth}
		\centering
		\includegraphics[width=1\textwidth, height=0.5625\textwidth]{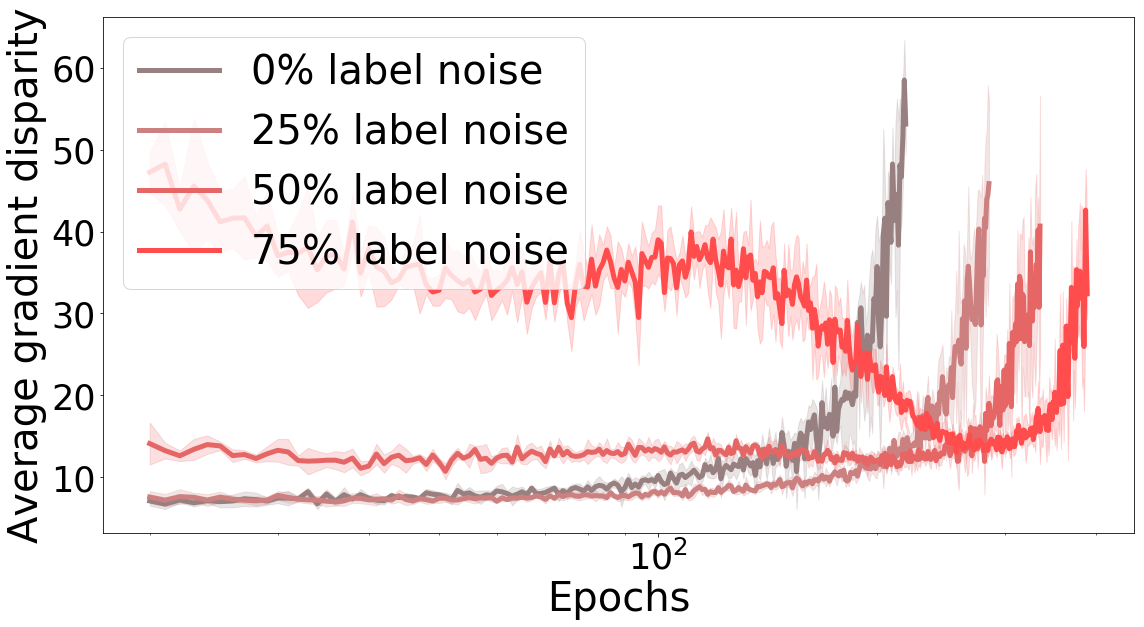}
		\caption{$\overline{\mathcal{D}}$ for epoch $>$ 20}
	\end{subfigure}
		\caption{The cross entropy loss, error percentage, and average gradient disparity during training with different amounts of randomness in the training labels for a 4-layer fully connected neural network with 500 hidden units trained on the entire CIFAR-10 dataset. The parameter initialization is the Xavier initialization with uniform distribution. 
		}\label{fig:fc_cifar10}
	\end{figure*}

\begin{figure*}[t]
	\begin{subfigure}[b]{1\textwidth}
	\includegraphics[width=0.5\textwidth, height=0.28\textwidth]{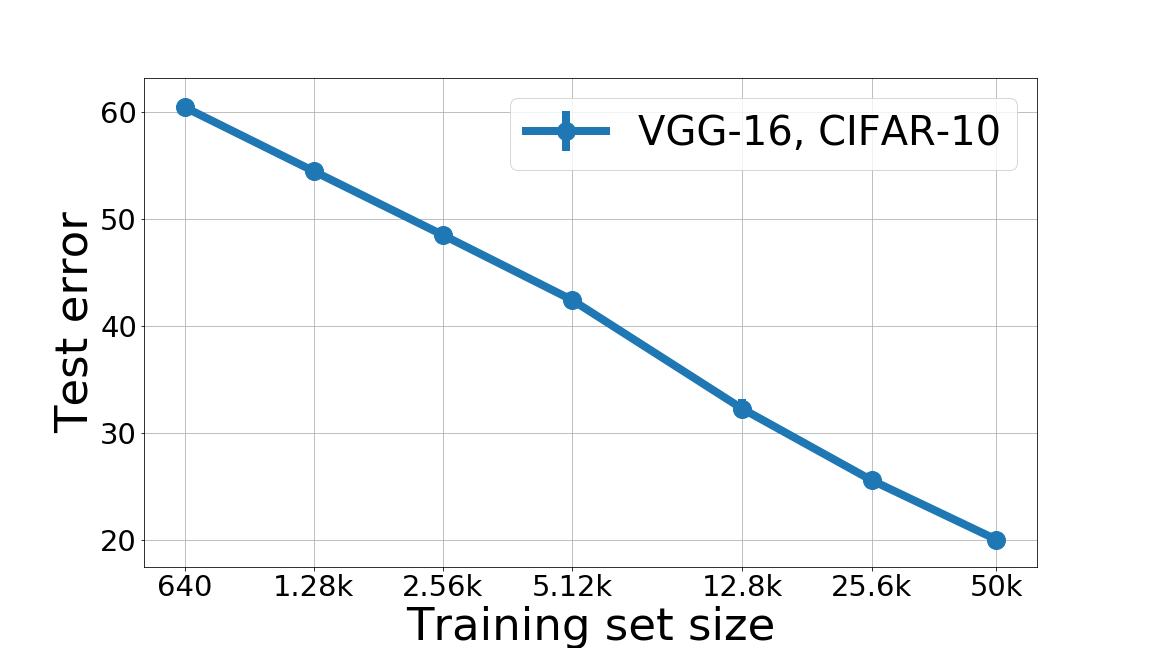}%
	\includegraphics[width=0.5\textwidth, height=0.28\textwidth]{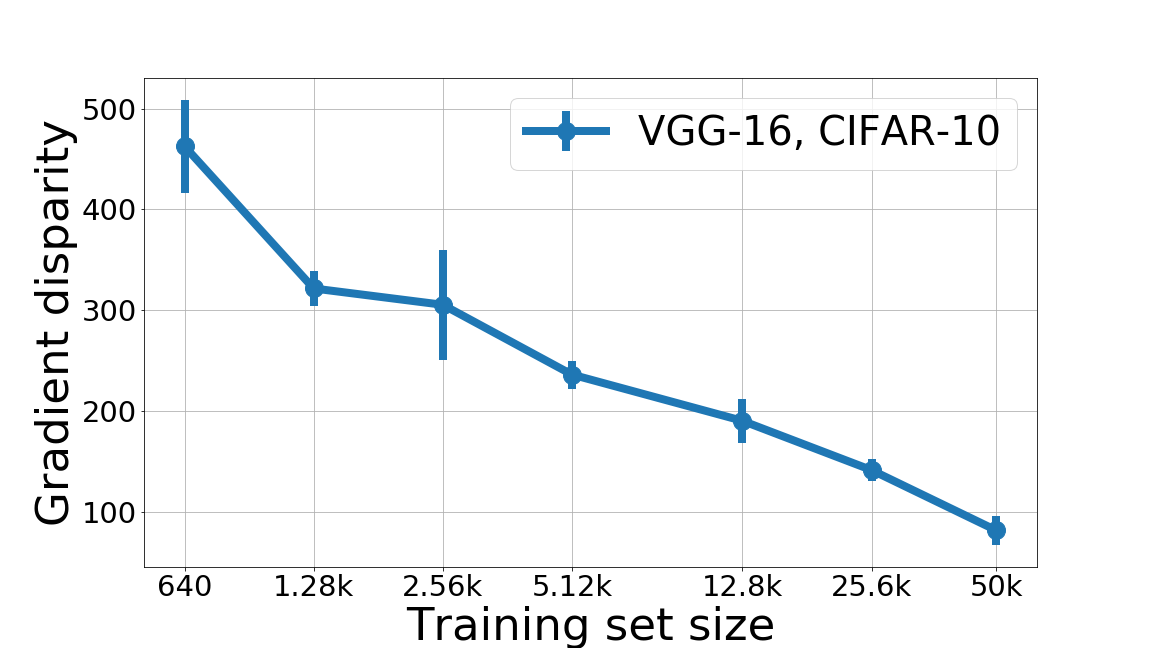}
	\caption{VGG-16, CIFAR-10, $\rho_{\overline{\mathcal{D}}, \text{TE}}=0.972$}\label{fig:train}
	\end{subfigure}
\begin{subfigure}[b]{1\textwidth}
	\includegraphics[width=0.5\textwidth, height=0.28\textwidth]{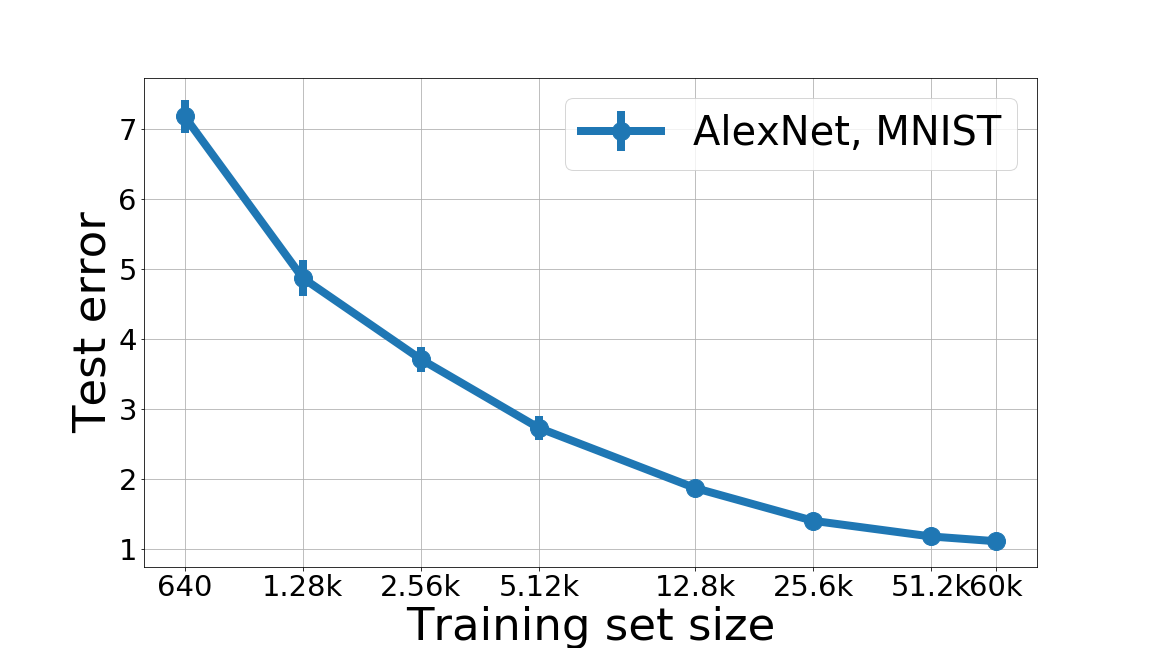}%
	\includegraphics[width=0.5\textwidth, height=0.28\textwidth]{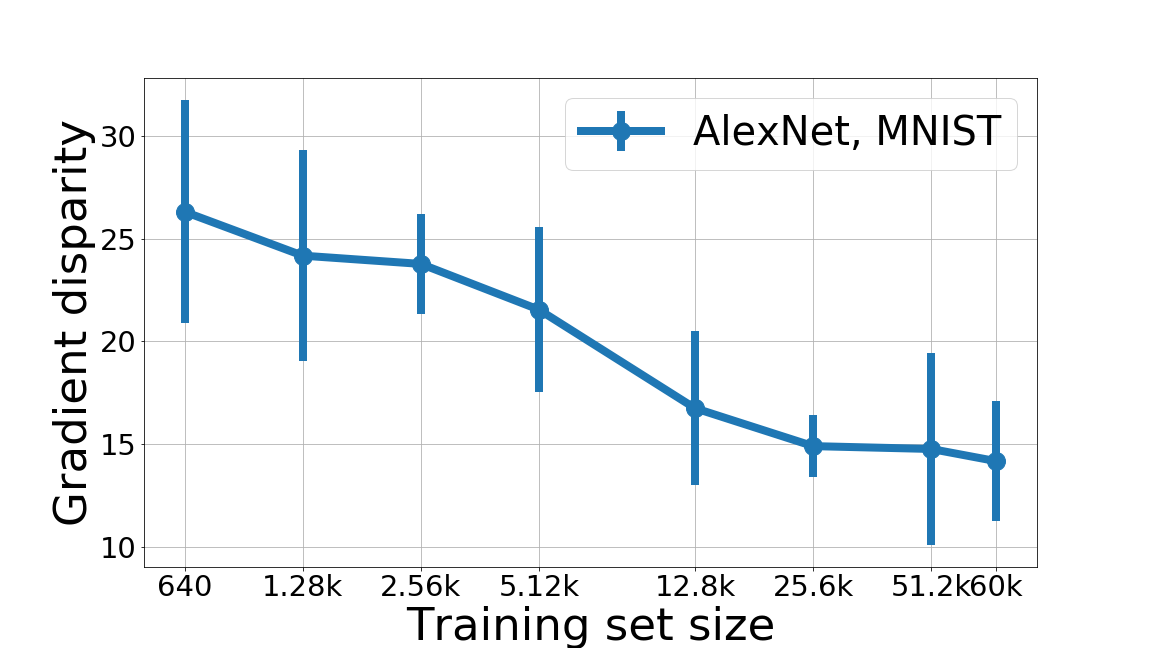}
	\caption{ALexNet, MNIST, $\rho_{\overline{\mathcal{D}}, \text{TE}}=0.929$}\label{fig:train2}
\end{subfigure}
\caption{Test error and gradient disparity for networks that are trained with different training set sizes. The training is stopped when the training loss is below $0.01$.}\label{fig:train_size2}
\end{figure*}

\begin{figure*}[h]
	\begin{subfigure}[b]{1\textwidth}
		\includegraphics[width=0.45\textwidth, height=0.252\textwidth]{batch_size_5.png}%
		\includegraphics[width=0.45\textwidth, height=0.252\textwidth]{batch_size_6.png}
		\caption{CIFAR-10, $\rho_{\overline{\mathcal{D}},\; \text{TE}} = 0.893$}
	\end{subfigure}
	\begin{subfigure}[b]{1\textwidth}
		\includegraphics[width=0.5\textwidth, height=0.28\textwidth]{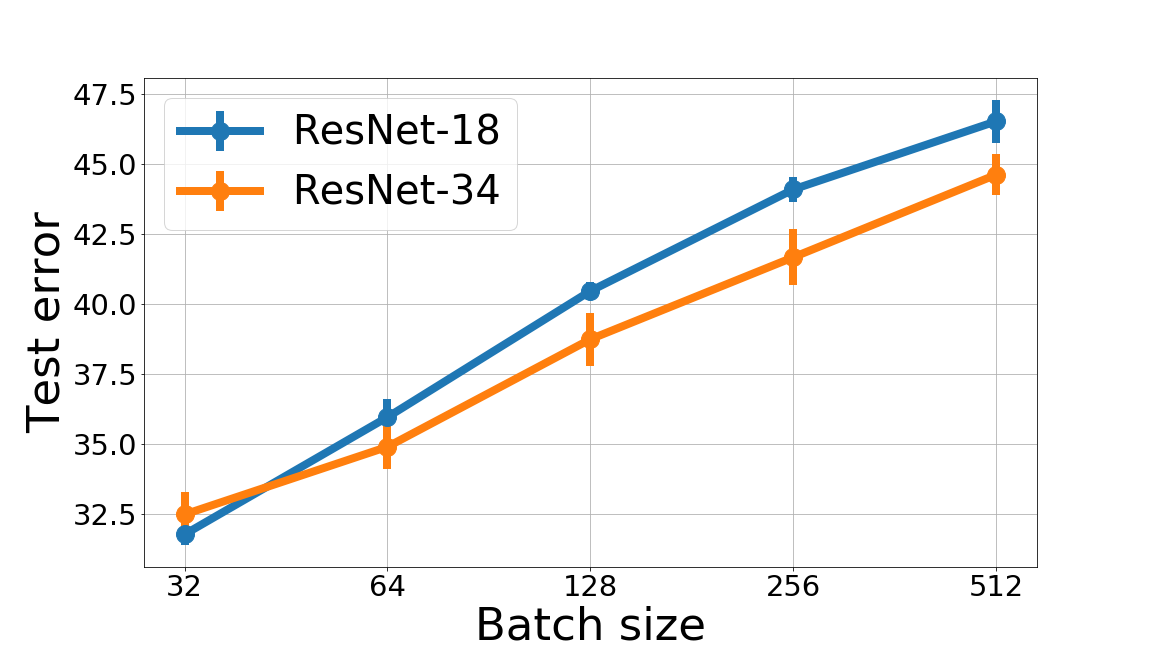}%
		\includegraphics[width=0.5\textwidth, height=0.28\textwidth]{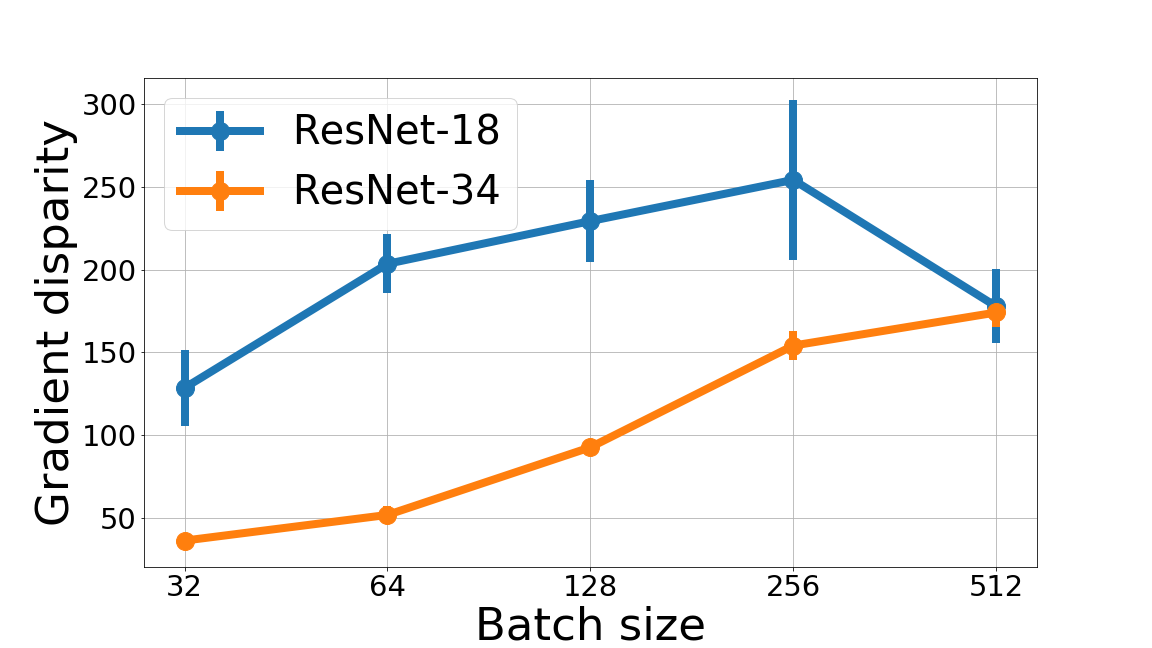}
		\caption{CIFAR-10, $\rho_{\overline{\mathcal{D}}, \text{TE}}=0.631$}
	\end{subfigure}
	\begin{subfigure}[b]{1\textwidth}
		\includegraphics[width=0.5\textwidth, height=0.28\textwidth]{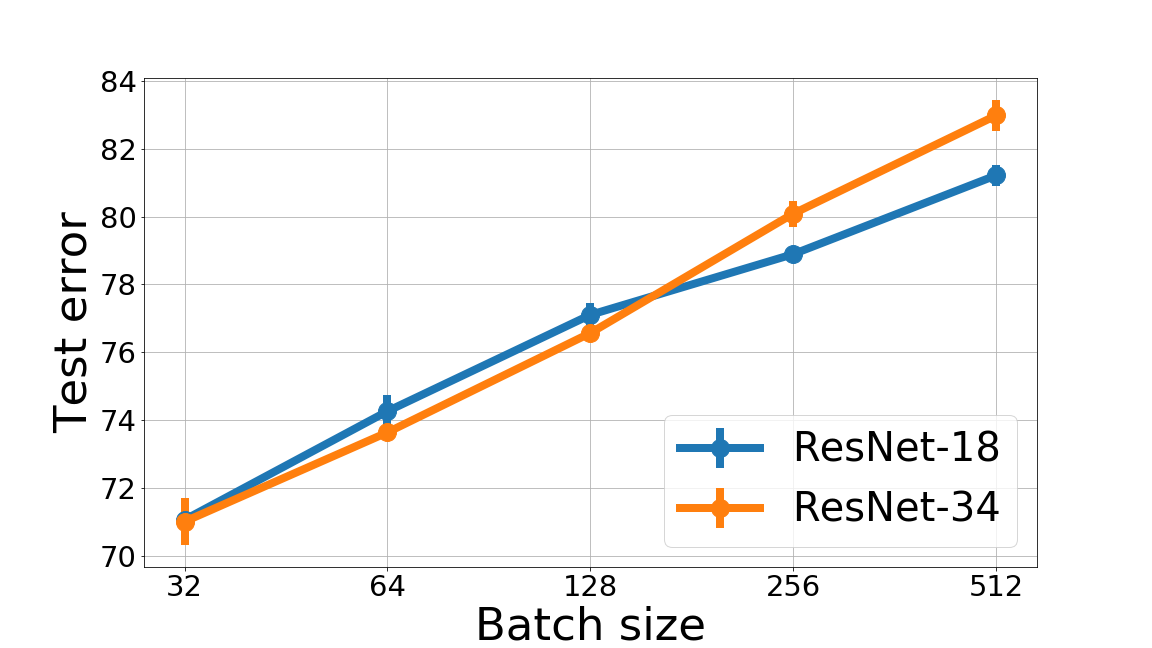}%
		\includegraphics[width=0.5\textwidth, height=0.28\textwidth]{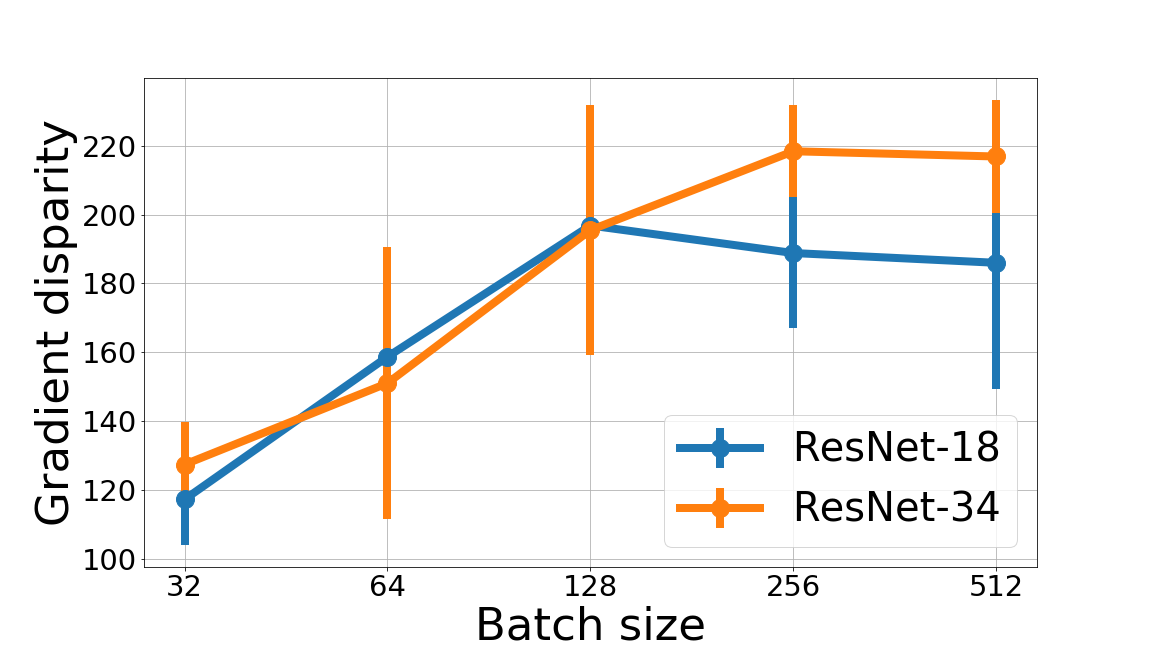}
		\caption{CIFAR-100, $\rho_{\overline{\mathcal{D}}, \text{TE}}=0.909$}
	\end{subfigure}
	\caption{Test error and gradient disparity for networks that are trained with different batch sizes trained on 12.8 k points of the CIFAR-10 and CIFAR-100 datasets. The training is stopped when the training loss is below $0.01$.}\label{fig:batch_size2}
\end{figure*}

%% file: cifar100.tex
	\begin{figure*}[h]
		\centering
		\begin{subfigure}[b]{0.24\textwidth}            
			\includegraphics[width=\textwidth, height=0.5625\textwidth]{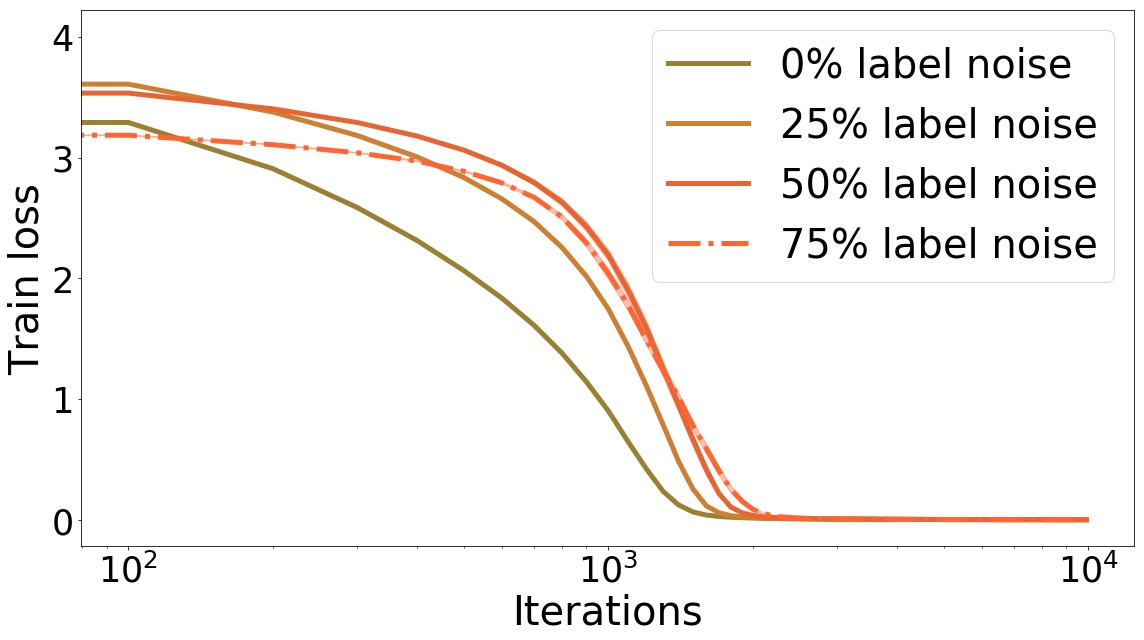}
			\caption{Training loss}
		\end{subfigure}%
	\hspace{0.5em}%
		\begin{subfigure}[b]{0.24\textwidth}
			\centering
			\includegraphics[width=\textwidth, height=0.5625\textwidth]{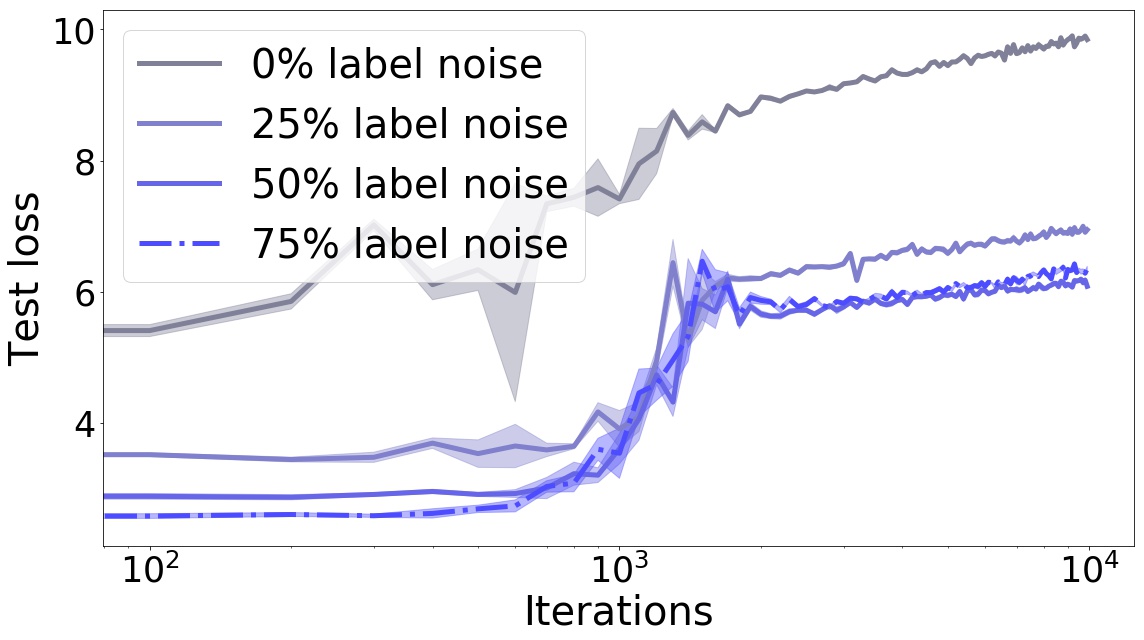}
			\caption{Test loss}
		\end{subfigure}%
		\hspace{0.5em}%
		\begin{subfigure}[b]{0.24\textwidth}            
			\includegraphics[width=\textwidth, height=0.5625\textwidth]{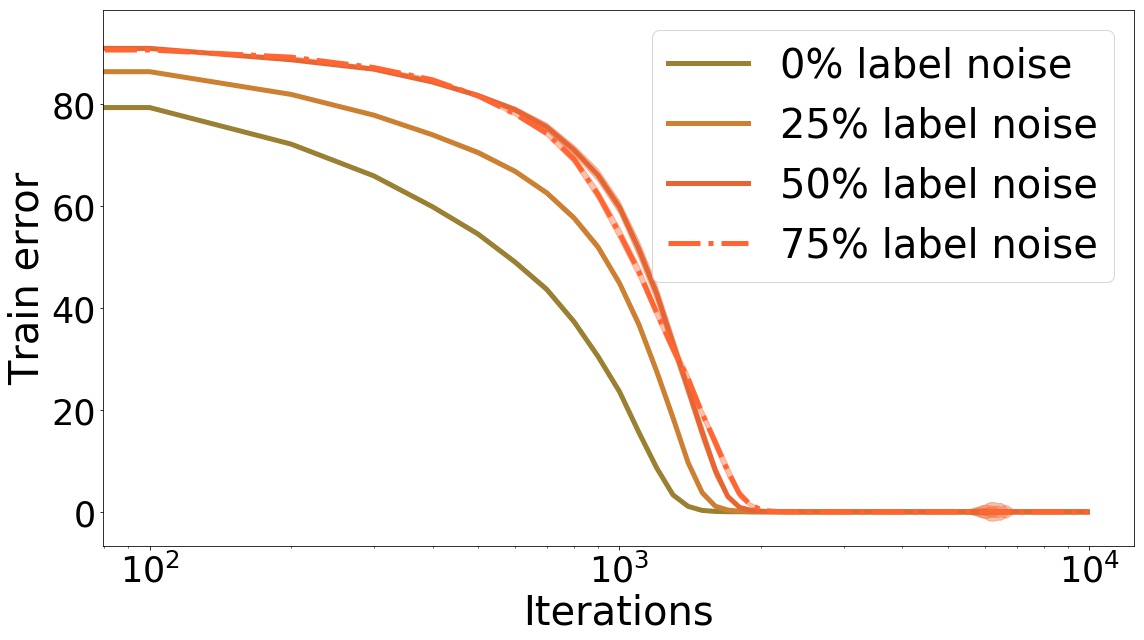}
			\caption{Training error}
		\end{subfigure}%
		\hspace{0.5em}%
		\begin{subfigure}[b]{0.24\textwidth}
			\centering
			\includegraphics[width=\textwidth, height=0.5625\textwidth]{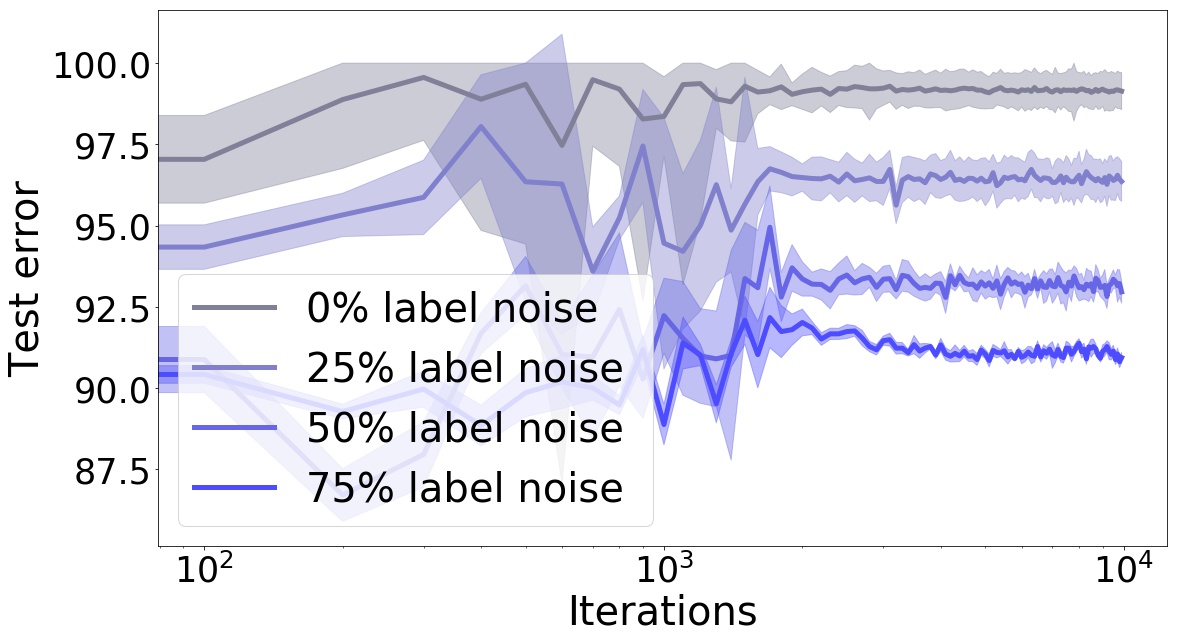}
			\caption{Test error}
		\end{subfigure}\\
		\begin{subfigure}[b]{0.31\textwidth}            
			\includegraphics[width=\textwidth, height=0.5625\textwidth]{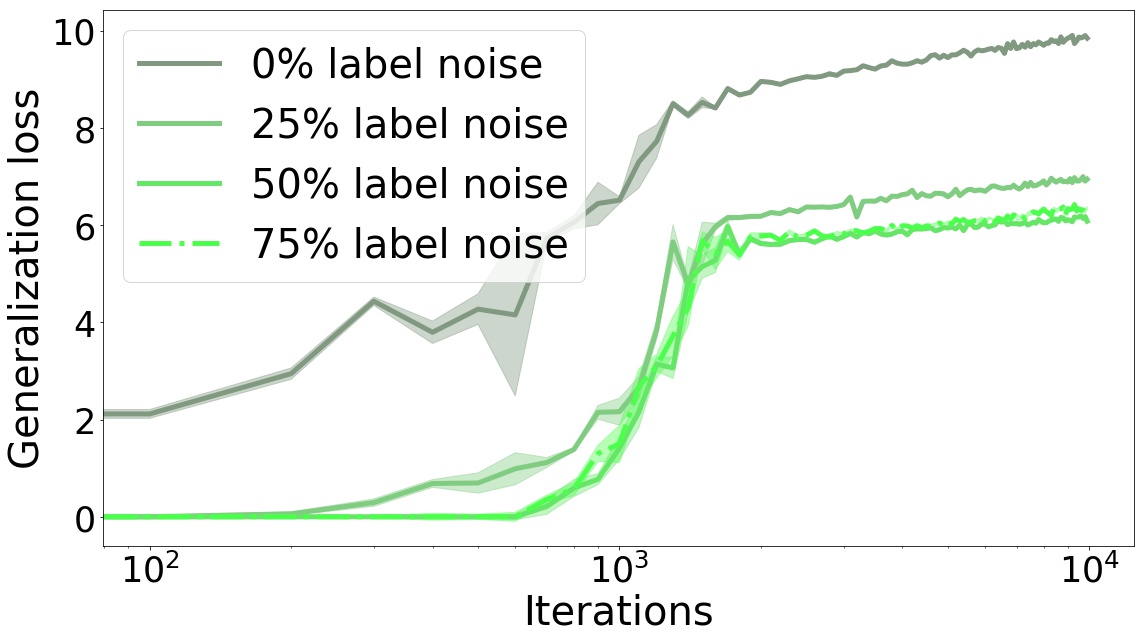}
			\caption{Generalization loss}
		\end{subfigure}%
		\hspace{0.5em}%
		\begin{subfigure}[b]{0.31\textwidth}            
			\includegraphics[width=\textwidth, height=0.5625\textwidth]{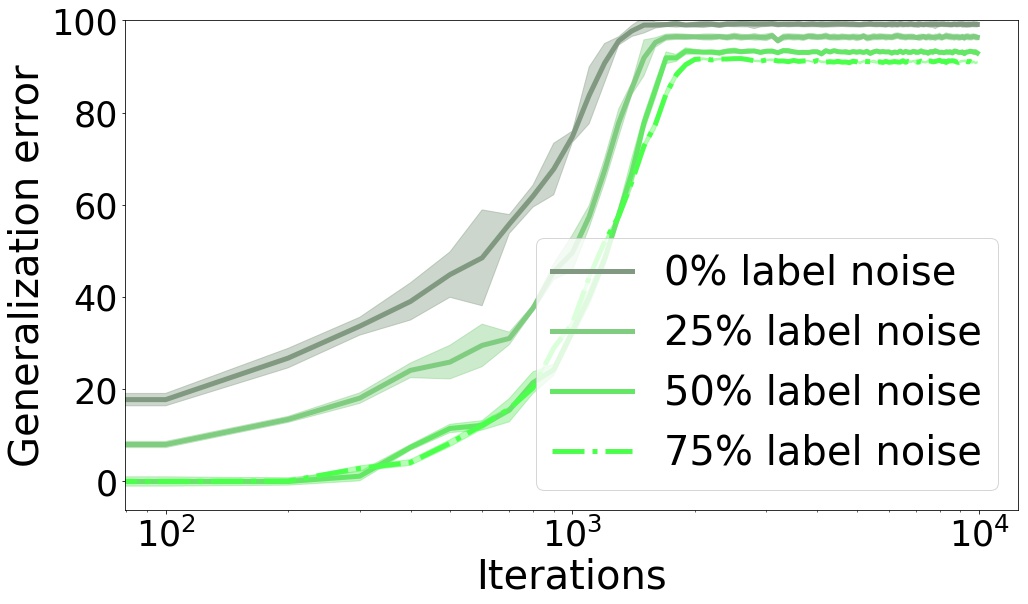}
			\caption{Generalization error}
		\end{subfigure}%
		\hspace{0.5em}%
		\begin{subfigure}[b]{0.31\textwidth}
			\centering
			\includegraphics[width=\textwidth, height=0.5625\textwidth]{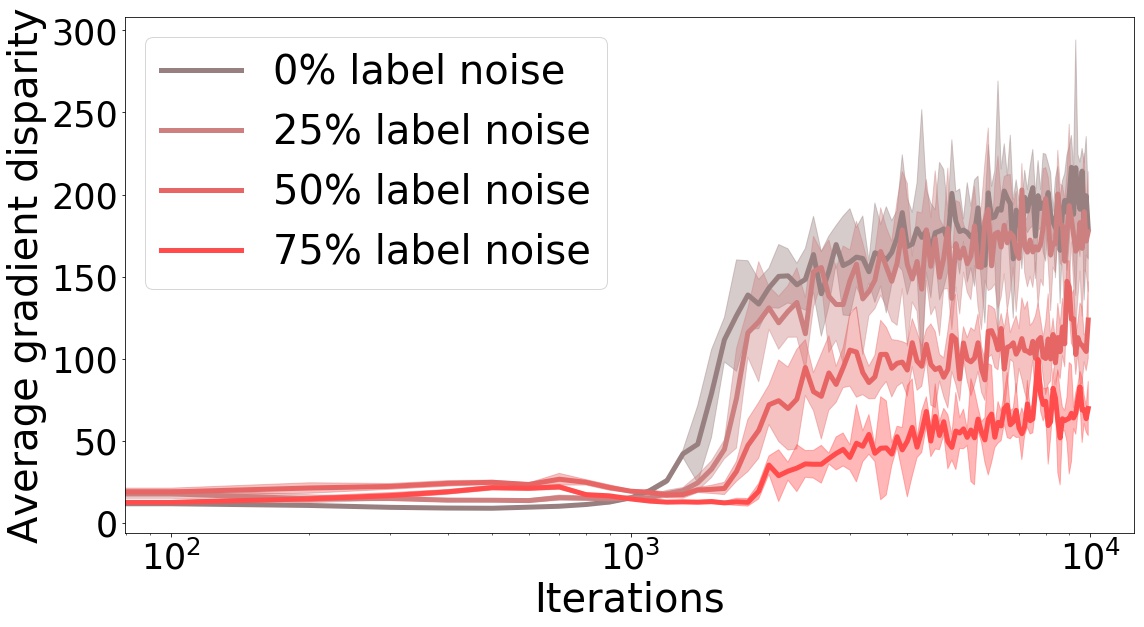}
			\caption{$\overline{\mathcal{D}}$}
		\end{subfigure}
		\caption{The cross entropy loss, error percentage, and average gradient disparity during training with different amounts of randomness in the training labels for a ResNet-18 trained on the CIFAR-100 training set. The parameter initialization is the Xavier initialization.}\label{fig:resnet_cifar100}
	\end{figure*}

%% file: beyond.tex
	\clearpage

\input{opt_analytical.tex}
	\subsection{Experiments}
\begin{figure*}[h]
		\centering
		\begin{subfigure}[b]{0.5\textwidth}  
			\centering
			\includegraphics[width=1\textwidth, height=0.5625\textwidth]{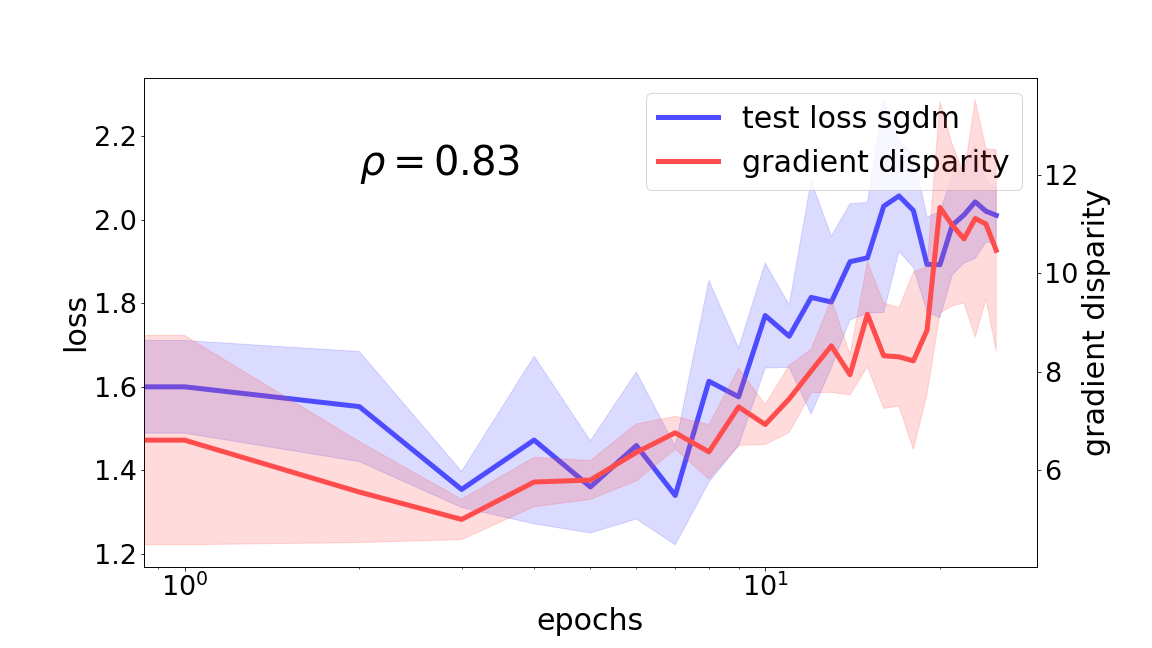}%
			\caption{\centering SGD with Momentum, test loss \newline epoch: 11, GD epoch: 10}
		\end{subfigure}%
		\begin{subfigure}[b]{0.5\textwidth}  
			\centering
			\includegraphics[width=1\textwidth, height=0.5625\textwidth]{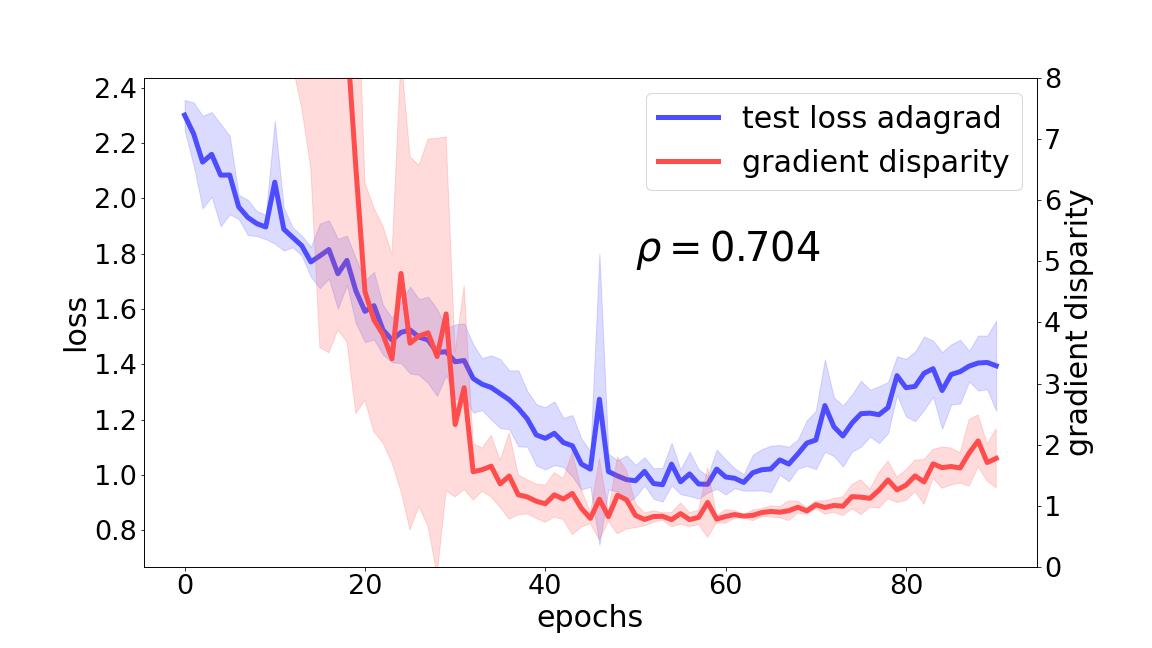}%
			\caption{\centering Adagrad, test loss epoch: 19, \newline GD epoch: 18}
		\end{subfigure}
		\begin{subfigure}[b]{0.5\textwidth}  
			\centering          
			\includegraphics[width=1\textwidth, height=0.5625\textwidth]{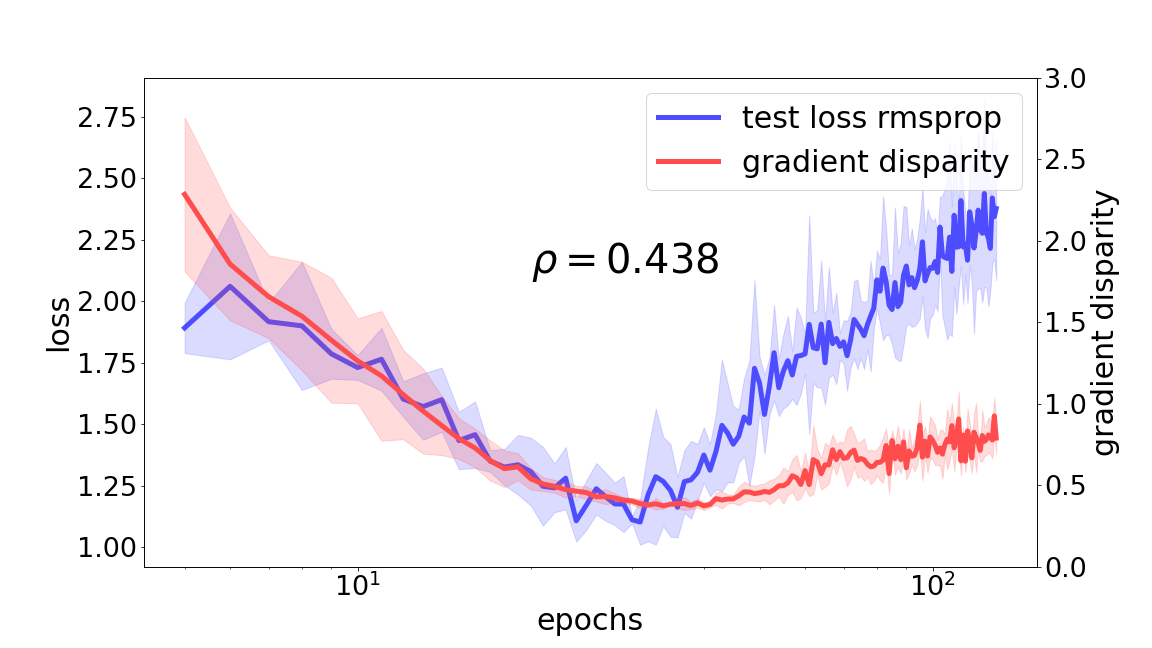}%
			\caption{\centering RmsProp, test loss epoch: 15  \newline (err: 54\%), GD epoch: 36 (err: \textbf{31\%}) }
		\end{subfigure}%
		\begin{subfigure}[b]{0.5\textwidth}   
			\centering   
			\includegraphics[width=1\textwidth, height=0.5625\textwidth]{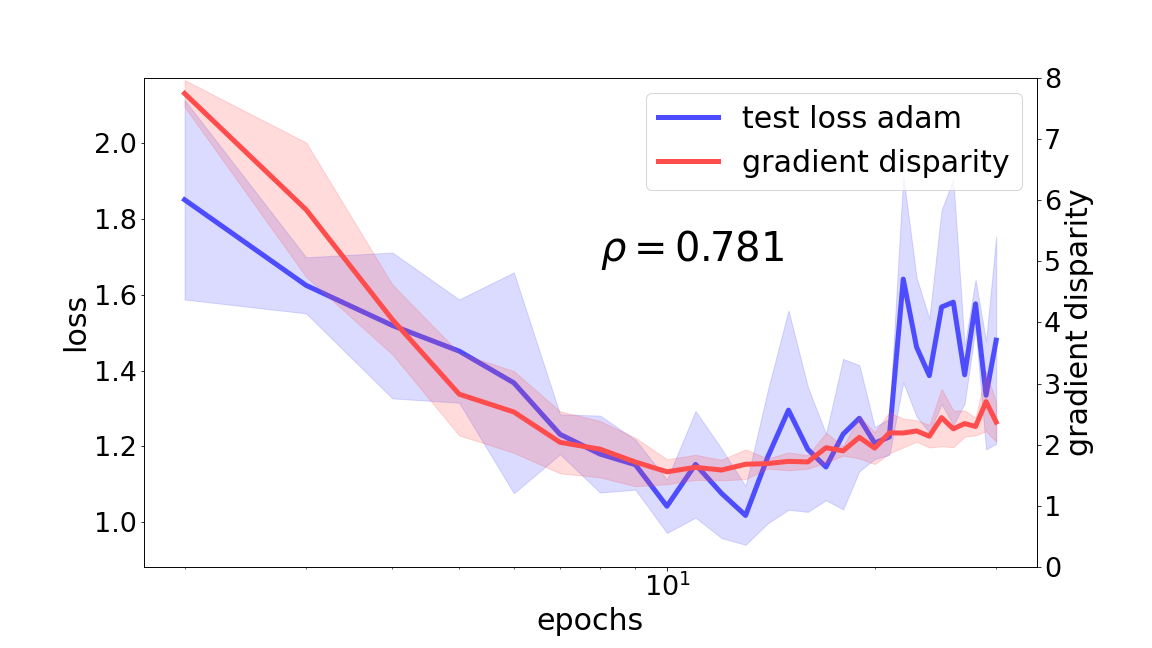}%
			\caption{\centering Adam, test loss epoch: 19, \newline GD epoch: 20}
		\end{subfigure}
		\begin{subfigure}[b]{0.5\textwidth}  
			\centering
			\includegraphics[width=1\textwidth, height=0.5625\textwidth]{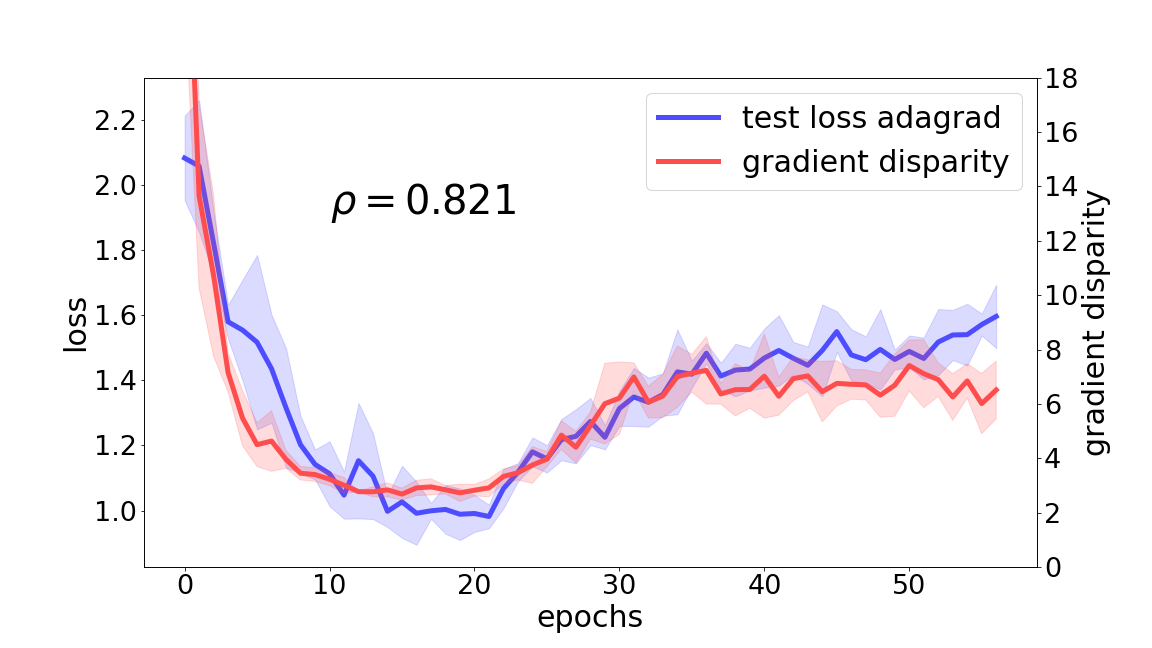}%
			\caption{\centering Adagrad, test loss epoch: 21, \newline GD epoch: 21}
		\end{subfigure}%
		\begin{subfigure}[b]{0.5\textwidth}    
			\centering        
			\includegraphics[width=1\textwidth, height=0.5625\textwidth]{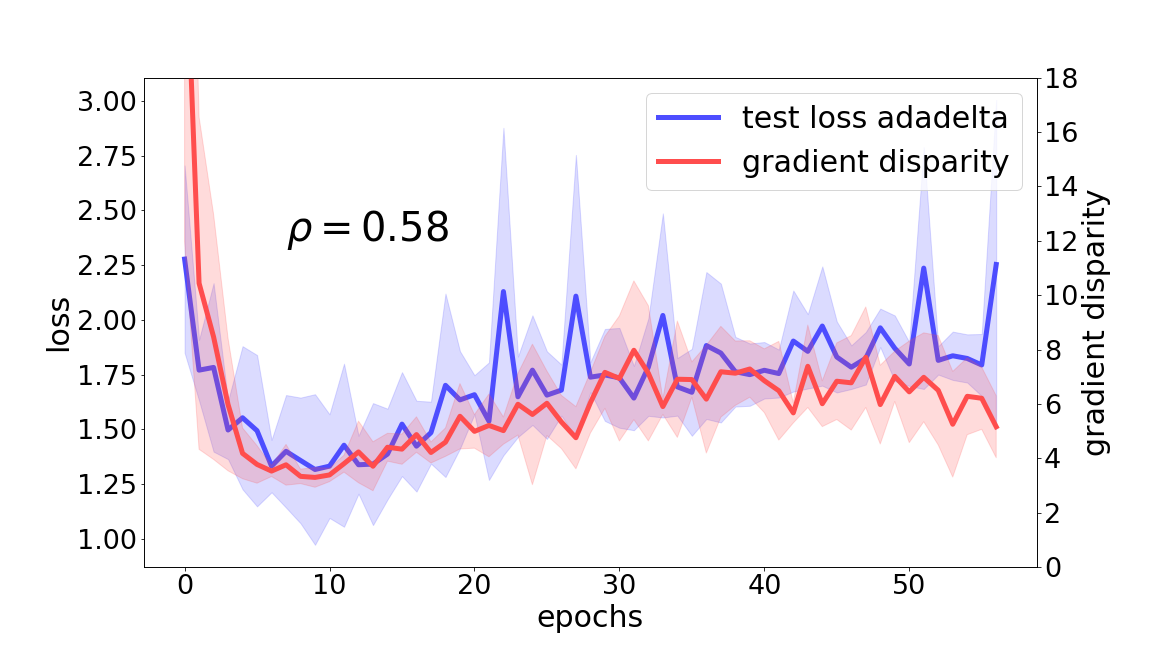}%
			\caption{\centering Adadelta, test loss epoch: 12, \newline GD epoch: 15}
		\end{subfigure}
		\begin{subfigure}[b]{0.5\textwidth}
			\centering            
			\includegraphics[width=1\textwidth, height=0.5625\textwidth]{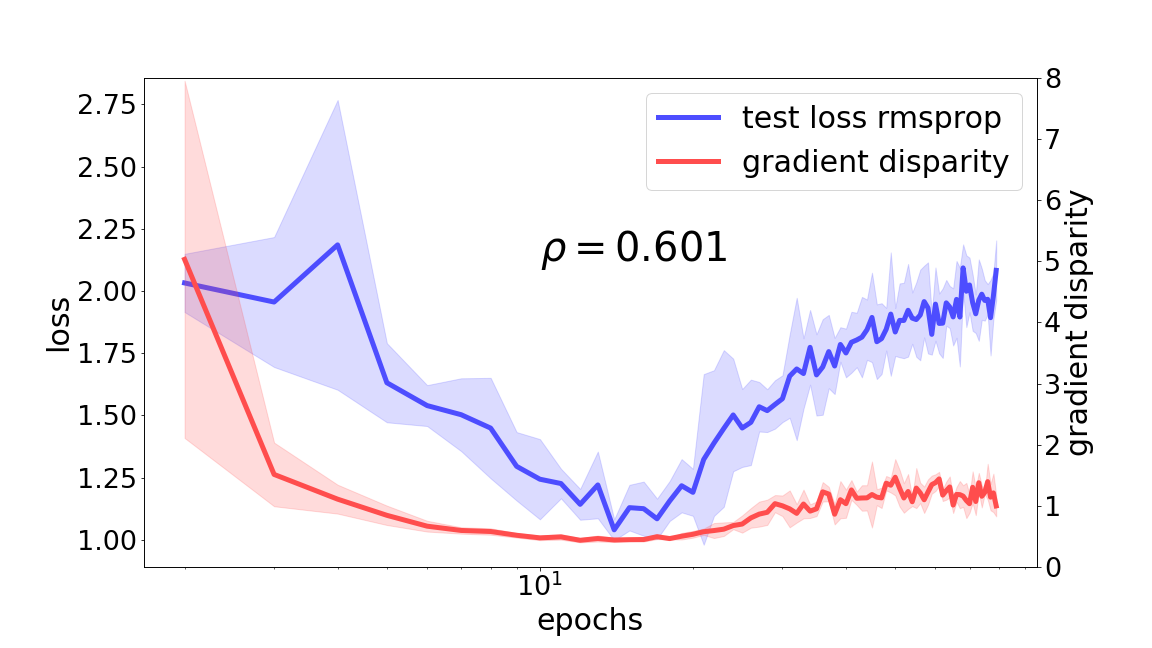}%
			\caption{\centering RmsProp, test loss epoch: 20, \newline GD epoch: 18}
		\end{subfigure}%
		\begin{subfigure}[b]{0.5\textwidth}  
			\centering    
			\includegraphics[width=1\textwidth, height=0.5625\textwidth]{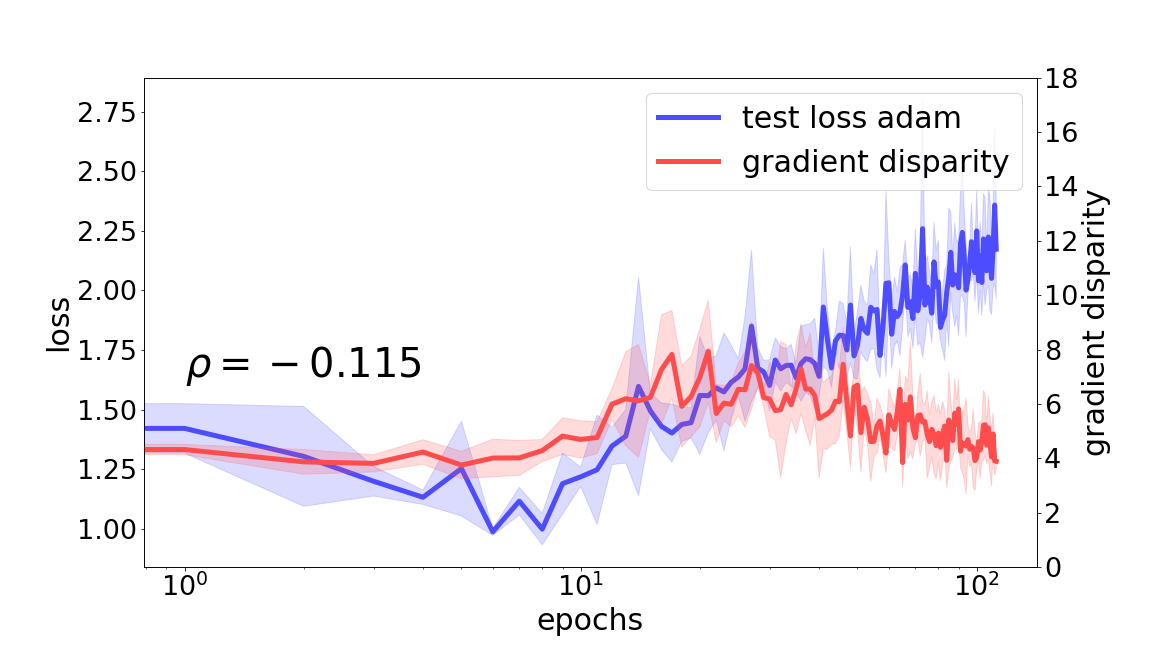}%
			\caption{\centering Adam, test loss epoch: 12,  \newline GD epoch: 10}
		\end{subfigure}
		\caption{(a-d) VGG-19 configuration trained on 12.8 k training points of CIFAR-10 dataset. (e-h) VGG-11 configuration trained on 12.8 k points of the CIFAR-10 dataset. 
			The training is stopped when the training loss gets below $0.01$. The presented results are an average over 5 runs. The captions below each figure give the epoch number where test loss and gradient disparity have respectively been increased for 5 epochs from the beginning of training. 
		}\label{fig:opt}
	\end{figure*}

	Fig. \ref{fig:opt} shows gradient disparity and the test loss curves during the course of training for adaptive optimizers. The epoch in which the fifth increase in the value of the test loss and gradient disparity has happened is shown in the caption of each experiment. We observe that the two suggested epochs for stopping the optimization (the one suggested by gradient disparity (GD) and the other one suggested by test loss) are extremely close to each other, except in Fig.~\ref{fig:opt}~(c) where the fifth epoch with an increase in the value of gradient disparity is much later than the epoch with the fifth increase in the value of test loss. However, in this experiment, there is a 23\% improvement in the test accuracy if the optimization is stopped according to GD compared to test loss, due to many variations of the test loss compared to gradient disparity.
	
	As an early stopping criterion, the increase in the value of gradient disparity coincides with the increase in the test loss in all our experiments presented in Fig.~\ref{fig:opt}. In Fig.~\ref{fig:opt}~(h), for the Adam optimizer, we observe that after around 20 epochs, the value of gradient disparity starts to decrease, whereas the test loss continues to increase. This mismatch between test loss and gradient disparity might result from other factors that appear in Eq.~(\ref{eq:adam}). Nevertheless, even in this experiment, the increase in the test loss and gradient disparity coincide, and hence gradient disparity can correctly detect early stopping time. 
	These experiments are a first indication that gradient disparity can be used as an early stopping criterion for optimizers other than SGD.
	
%% file: opt_analytical.tex
	\clearpage
	\section{Beyond SGD}\label{app:opt}
	In the following, we discuss how the analysis of Section~\ref{sec:BGP} can be extended to other optimizers (refer to \cite{ruder2016overview} for an overview on popular optimizers).
	
	\subsection{SGD with Momentum}\label{app:sgdm}
	The momentum method \cite{qian1999momentum} is a variation of SGD which adds a fraction of the update vector of the previous step to the current update vector to accelerate SGD:
	\setlength\abovedisplayskip{8pt}
	\setlength\belowdisplayskip{8pt}
	\begin{align*}
		\vspace{-5em}
		\upsilon^{(t+1)} &= \eta \upsilon^{(t)} + \gamma g^{(t)}, \\
		w^{(t+1)} &= w^{(t)} - \upsilon^{(t+1)} ,
		\vspace{-5em}
	\end{align*}
	where $g^{(t)}$ is either $g_1$ or $g_2$ depending on the selection of the batch $S_1$ or $S_2$ for the current update step. As $\upsilon^{(t)}$ remains the same for either choice, the KL-divergence between $Q_1$ and $Q_2$ for SGD with momentum, is the same as Eq.~(\ref{eq:kl}).
	\subsection{Adagrad}
	Adagrad \cite{duchi2011adaptive} performs update steps with a different learning rate for each individual parameter. By denoting each coordinate of the parameter vector $w$ by $d$, one update step of the Adagrad algorithm is
	\begin{align}\label{eq:adagrad}
	w_d^{(t+1)} = w_d^{(t)} - \frac{\gamma}{\sqrt{G_{dd}^{(t)}+\epsilon}} g_d^{(t)}, 
	\end{align}
	where the vector $g^{(t)}$ is either $g_1$ or $g_2$ depending on the selection of the batch for the current update step, and $G_{dd}^{(t)}$ is the accumulative squared norm of the gradients up until iteration~$t$. 
	Hence, for Adagrad, Eq.~(\ref{eq:kl}) is replaced by
	\begin{align}\label{eq:adag}
	\text{KL}(Q_1 || Q_2) = \frac{1}{2} \frac{\gamma^2}{\sigma^2} \norm{\frac{1}{G^{(t)}+\epsilon} \odot \left(g_1 - g_2\right)}_2^2 \nonumber \\ \leq \frac{1}{2} \frac{\gamma^2}{\sigma^2} \norm{\frac{1}{G^{(t)}+\epsilon}}_2^2 \norm{g_1 - g_2}_2^2,
	\end{align}
	where $\odot$ denotes the element-wise product of two vectors, where division is also taken element-wise and where $\epsilon$ is a small positive constant that avoids a possible division by $0$. 
	To compare the upper bound in Theorem \ref{THM1} from one iteration to the next one (as needed to determine the early stopping moment in Section~\ref{sec:early}), gradient disparity is not the only factor in Eq.~(\ref{eq:adag}) that evolves over time. Indeed $G^{(t)}$ is an increasing function of $t$. However, after a few iterations when the gradients become small, this value becomes approximately constant (the initial gradient values dominate the sum in $G^{(t)}$). Then the right hand side of Eq.~(\ref{eq:adag}) varies mostly as a function of gradient disparity, and therefore gradient disparity approximately tracks down the generalization penalty upper bound. 

	\subsection{Adadelta and RmsProp}
	Adadelta \cite{zeiler2012adadelta} is an extension of Adagrad, which computes a decaying average of the past gradient vectors instead of the accumulative squared norm of the gradients during the previous update steps.
	$G_{dd}^{(t)}$ in Eq.~(\ref{eq:adagrad}) is then replaced by $\upsilon_d^{(t+1)}$ where ${\upsilon_d^{(t+1)}=\eta \upsilon_d^{(t)} + (1-\eta) (g_d^{(t)})^2}$.
	As training proceeds, the gradient magnitude decreases. Also, $\eta$ is usually close to $1$. Therefore, the dominant term in $\upsilon_d^{(t+1)}$ becomes $\eta \upsilon_d^{(t)}$. Then, if we approximate $\upsilon_1^{(t+1)} = \eta \upsilon^{(t)} + (1-\eta) \left(g_1\right)^2 \approx  \eta \upsilon^{(t)} + (1-\eta) \left(g_2\right)^2 = \upsilon_2^{(t+1)}$ (squares are done element-wise), then for Adadelta we have
	\begin{align}\label{eq:adad}
	\text{KL}(Q_1 || Q_2)  \leq \frac{1}{2} \frac{\gamma^2}{\sigma^2} \norm{\frac{1}{\upsilon^{(t+1)}+\epsilon}}_2^2 \norm{g_1 - g_2}_2^2 ,
	\end{align}
	where again the division is done element-wise.
	The denominator in Eq.~(\ref{eq:adad}) is smaller than the denominator in Eq.~(\ref{eq:adag}). In both equations, the first non-constant factor in the upper bound of $\text{KL}(Q_1 || Q_2) $ decreases as a function of $t$, and therefore an increase in the value of $\text{KL}(Q_1 || Q_2) $ should be accounted for by an increase in the value of gradient disparity. Moreover, as training proceeds, gradient magnitudes decrease and the first factor on the upper bound of Eqs.~(\ref{eq:adag}) and (\ref{eq:adad}) becomes closer to a constant. Therefore, an upper bound on the generalization penalties can be tracked by gradient disparity.
	The update rule of RmsProp\footnote{\url{https://www.cs.toronto.edu/~tijmen/csc321/slides/lecture_slides_lec6.pdf}} is very similar to Adadelta, and the same conclusions can be made.

	\subsection{Adam}\label{app:adam}
	\allowdisplaybreaks
	Adam \cite{kingma2014adam} combines Adadelta and momentum by storing an exponentially decaying average of the previous gradients and squared gradients:
	\begin{align*}
		m^{(t+1)} &= \beta_1 m^{(t)} + (1-\beta_1) g^{(t)},  \qquad  \upsilon^{(t+1)} = \beta_2 \upsilon^{(t)} + (1-\beta_2) \left({g^{(t)}}\right)^2   ,\\
		\hat{m}^{(t+1)} &=\frac{m^{(t+1)}}{1-\left(\beta_1\right)^t}, \qquad \hat{\upsilon}^{(t+1)} =\frac{\upsilon^{(t+1)}}{1-\left(\beta_2\right)^t} , \qquad
		\\ & w^{(t+1)} = w^{(t)} - \frac{\gamma}{\sqrt{\hat{\upsilon}^{(t+1)}}+\epsilon} \hat{m}^{(t+1)}.
	\end{align*}
	\allowdisplaybreaks[0]
	All the operations in the above equations are done element-wise. As $\beta_2$ is usually very close to $1$ (around 0.999), and as squared gradient vectors at the current update step are much smaller than the accumulated values during the previous steps, we approximate: $\upsilon_1^{(t+1)} = \beta_2 \upsilon^{(t)} + (1-\beta_2) \left(g_1\right)^2 \approx  \beta_2 \upsilon^{(t)} + (1-\beta_2) \left(g_2\right)^2 = \upsilon_2^{(t+1)}$ (squares are done element-wise). Hence, Eq.~(\ref{eq:kl}) becomes
	 \begin{equation}\label{eq:adam}
	 \text{KL}(Q_1 || Q_2)  \leq \frac{1}{2} \frac{\gamma^2}{\sigma^2} \frac{1-\beta_1}{1-(\beta_1)^t} \norm{\frac{1}{\sqrt{\hat{\upsilon}^{(t+1)}}+\epsilon}}_2^2 \norm{g_1 - g_2}_2^2 .
	 \end{equation}
	 The first non-constant factor in the equation above decreases with $t$ (because $\beta_1 < 1$).
	 However, it is not clear how the second factor varies as training proceeds. Therefore, unlike previous optimizers, it is more hazardous to claim that the factors other than gradient disparity in Eq.~(\ref{eq:adam}) become constant as training proceeds. Hence, tracking only gradient disparity for the Adam optimizer may be insufficient. This is empirically investigated in the next sub-section. 

%% file: inner_product.tex
	\clearpage
	\section{Comparison to Related Work}\label{app:var}
	
	\begin{table*}
		\caption{Test error (TE) and test loss (TL) achieved by using various metrics as early stopping criteria. On the leftmost column, the minimum values of TE and TL over all the iterations are reported (which is not accessible during training). The results of 5-fold cross validation are reported on the right, which serve as a baseline. For each experiment, we have underlined those metrics that result in a better performance than 5-fold cross-validation. We observe that gradient disparity (GD) and variance of gradients (Var) consistently outperform $k$-fold cross-validation, unlike other metrics. On the rightmost column (No ES) we report the results without performing early stopping (ES) (training is continued until the training loss is below $0.01$).}\label{tab:RW}  
		\vspace*{1em}
		\begin{subtable}{\linewidth}{
				\resizebox{\textwidth}{!}{
				\begin{tabular}{c|c|cccccccc|c|c}
					\toprule
					& Min    & GD/Var   \hspace*{-0.7em}  & EB   \hspace*{-0.7em}   & GSNR  & $g_i\cdot g_j$ & $\text{sign}(g_i\cdot g_j)$   & $\cos(g_i\cdot g_j)$  & $\Omega_c$ \hspace*{-0.7em} & OV & $k$-fold& No ES\\ 
					\midrule
					\multicolumn{1}{l|}{TE} &  $4.84$ $\;$ & \ul{$\mathbf{4.84}$} $\;$& $\;$\ul{$\mathbf{4.84}$} $\;$ & $12.82$ & $22.30$   & $12.82$              
					& $18.31$             & $8.30$  $\;$ & $11.79$& $4.84$ & $4.96$\\ 
					TL                      & $0.18$  & \ul{$\mathbf{0.18}$}  & \ul{$\mathbf{0.18}$}  & $0.46$  & $0.82$    & $0.46$              &
					$0.69$            & $0.32$ & $0.38$  & $0.18$   & $0.22$\\
					\bottomrule
			\end{tabular}}}
			\caption{MNIST, AlexNet}
		\end{subtable}
		\begin{subtable}{\linewidth}{
				\resizebox{\textwidth}{!}{
				\begin{tabular}{c|c|cccccccc|c|c}
					\toprule
					& Min    & GD/Var   \hspace*{-0.7em}  & EB   \hspace*{-0.7em}   & GSNR  & $g_i\cdot g_j$ & $\text{sign}(g_i\cdot g_j)$   & $\cos(g_i\cdot g_j)$  & $\Omega_c$ \hspace*{-0.7em} & OV & $k$-fold& No ES\\ 
					\midrule
					\multicolumn{1}{l|}{TE} &  $13.76$  & \ul{$\mathbf{16.66}$} & $24.63$  & $35.68$   & $37.92$              & $24.63$ & 
					$35.68$         & $29.40$ & $34.36$ &  $17.86$ & $25.72$\\ 
					TL                      & $0.75$  & \ul{$1.08$}  & \ul{$\mathbf{0.86}$} & $1.68$  & $1.82$    & \ul{$\mathbf{0.86}$}              
					& $1.68$            & $1.46$  & $1.65$&  $1.09$ & $0.91$\\
					\bottomrule
			\end{tabular}}}
			\caption{MNIST, AlexNet, $50\%$ random}
		\end{subtable}
		\begin{subtable}{\linewidth}{
				\resizebox{\textwidth}{!}{
				\begin{tabular}{c|c|cccccccc|c|c}
					\toprule
					& Min    & GD/Var   \hspace*{-0.7em}  & EB   \hspace*{-0.7em}   & GSNR  & $g_i\cdot g_j$ & $\text{sign}(g_i\cdot g_j)$   & $\cos(g_i\cdot g_j)$  & $\Omega_c$ \hspace*{-0.7em} & OV & $k$-fold& No ES\\ 
					\midrule
					\multicolumn{1}{l|}{TE} & $45.54$ & \ul{$\mathbf{45.95}$} & $61.76$ & $70.46$ & $70.46$         & $55.84$               
					& $67.09$              & $67.37$ & $70.61$  & $51.64$  & $64.19$\\ 
					TL                      & $1.32$  & \ul{$\mathbf{1.45}$}  & $1.68$  & $1.92$  & $1.92$          & $1.52$                
					& $1.83$               & $1.85$  & $1.92$ & $1.49$ & $1.98$ \\
					\bottomrule
				\end{tabular}}
			}
			\caption{CIFAR-10, ResNet-18}
		\end{subtable}
		\begin{subtable}{\linewidth}{
				\resizebox{\textwidth}{!}{
				\begin{tabular}{c|c|cccccccc|c|c}
					\toprule
					& Min    & GD/Var   \hspace*{-0.7em}  & EB   \hspace*{-0.7em}   & GSNR  & $g_i\cdot g_j$ & $\text{sign}(g_i\cdot g_j)$   & $\cos(g_i\cdot g_j)$  & $\Omega_c$ \hspace*{-0.7em} & OV & $k$-fold& No ES\\ 
					\midrule
					\multicolumn{1}{l|}{TE} & $59.77$ & \ul{$71.97$} & $73.17$ & $77.08$ & $75.91$   & \ul{$\mathbf{65.80}$}               
					& $75.43$              & $77.71$  & $76.65$  & $72.56$ & $75.96$\\ 
					TL                      & $1.75$  & \ul{$2.00$}  & $2.03$  & $2.12$  & $2.13$      & \ul{$\mathbf{1.93}$}               
					& $2.07$              & $2.13$ &  $2.10$  & $2.02$  & $2.30$\\
					\bottomrule
				\end{tabular}
			}}
			\caption{CIFAR-10, ResNet-18, $50\%$ random } 
		\end{subtable}
		
	\end{table*}
	
	In Table~\ref{tab:RW}, we compare gradient disparity (GD) to a number of metrics that were proposed  either directly as an early stopping criterion, or as a generalization metric. For those metrics that were not originally proposed as early stopping criteria, we choose a similar method for early stopping as the one we use for gradient disparity. 
	We consider two datasets (MNIST and CIFAR-10), and two levels of label noise ($0\%$ and $50\%$). 
	Here is a list of the metrics that we compute in each setting (see Section~\ref{sec:rel} of the main paper where we introduce each metric):
	\begin{enumerate}
		\vspace{-0.7em}
		\item Gradient disparity (GD) (ours): we report the error and loss values at the time when the value of GD increases for the 5th time (from the beginning of the training).
		\item The EB-criterion \cite{mahsereci2017early}: we report the error and loss values when EB becomes positive. 
		\item Gradient signal to noise ratio (GSNR) \cite{Liu2020Understanding}: we report the error and loss values when the value of GSNR decreases for the 5th time (from the beginning of the training).
		\item Gradient inner product, $g_i\cdot g_j$ \cite{fort2019stiffness}: we report the error and loss values when the value of $g_i\cdot g_j$ decreases for the 5th time (from the beginning of the training).
		\item Sign of the gradient inner product, $\text{sign}(g_i\cdot g_j)$ \cite{fort2019stiffness}: we report the error and loss values when the value of $\text{sign}(g_i\cdot g_j)$ decreases for the 5th time (from the beginning of the training).
		\item Cosine similarity between gradient vectors, $\cos(g_i\cdot g_j)$ \cite{fort2019stiffness}: we report the error and loss values when the value of $\cos(g_i\cdot g_j)$ decreases for the 5th time (from the beginning of the training).
		\item Variance of gradients (Var) \cite{negrea2019information}: we report the error and loss values when the value of Var increases for the 5th time (from the beginning of the training). Variance is computed over the same number of batches used to compute gradient disparity, in order to compare metrics given the same computational budget.
		\item Average gradient alignment within the class $\Omega_c$ \cite{mehta2020extreme}: we report the error and loss values when the value of $\Omega_c$ decreases for the 5th time (from the beginning of the training).
		\item Optimization variance (OV) \cite{zhang2021optimization}: we report the error and loss values when the value of OV increases for the 5th time (from the beginning of the training).
	\end{enumerate}

	\vspace{-0.7em}
	On the leftmost column of Table~\ref{tab:RW}, we report the minimum values of the test error and the test loss over all the iterations, which may not necessarily coincide. For instance, in setting~(c), the test error is minimized at iteration 196, whereas the test loss is minimized at iteration 126. On the rightmost column of Table~\ref{tab:RW} we report the values of the test error and loss when no early stopping is applied and the training is continued until training loss value below $0.01$. Next to this, we report the values of the test error and the test loss when using $5$-fold cross-validation, which serves as a baseline. We have underlined in red the metrics that outperform $k$-fold cross validation. We observe that the only metrics that consistently outperform $k$-fold CV are GD and Variance of gradients.
	
	The EB-criterion, $\text{sign}(g_i\cdot g_j)$, and $\cos(g_i\cdot g_j)$ are metrics that perform quite well as early stopping criteria, although not as well as GD and Var. In Section~\ref{sec:cos}, we observe that these metrics are not informative of the label noise level, contrary to gradient disparity. 
	
	It is interesting to observe that gradient disparity and variance of gradients produce the exact same results when used as early stopping criteria (Table~\ref{tab:RW}). Moreover, these two are the only metrics that consistently outperform $k$-fold cross-validation. 
	However, in Section~\ref{app:var2}, we observe that the correlation between gradient disparity and the test loss is in general larger than the correlation between variance of gradients and the test loss.

	\subsection{Capturing Label Noise Level}\label{sec:cos}
	In this section, we show in particular three metrics that even though perform relatively well as early stopping criteria, fail to account for the level of label noise, contrary to gradient disparity.
	
\begin{figure*}[h]
		\centering
		\begin{subfigure}[b]{\textwidth}            
			\includegraphics[width=0.5\textwidth, height=0.28\textwidth]{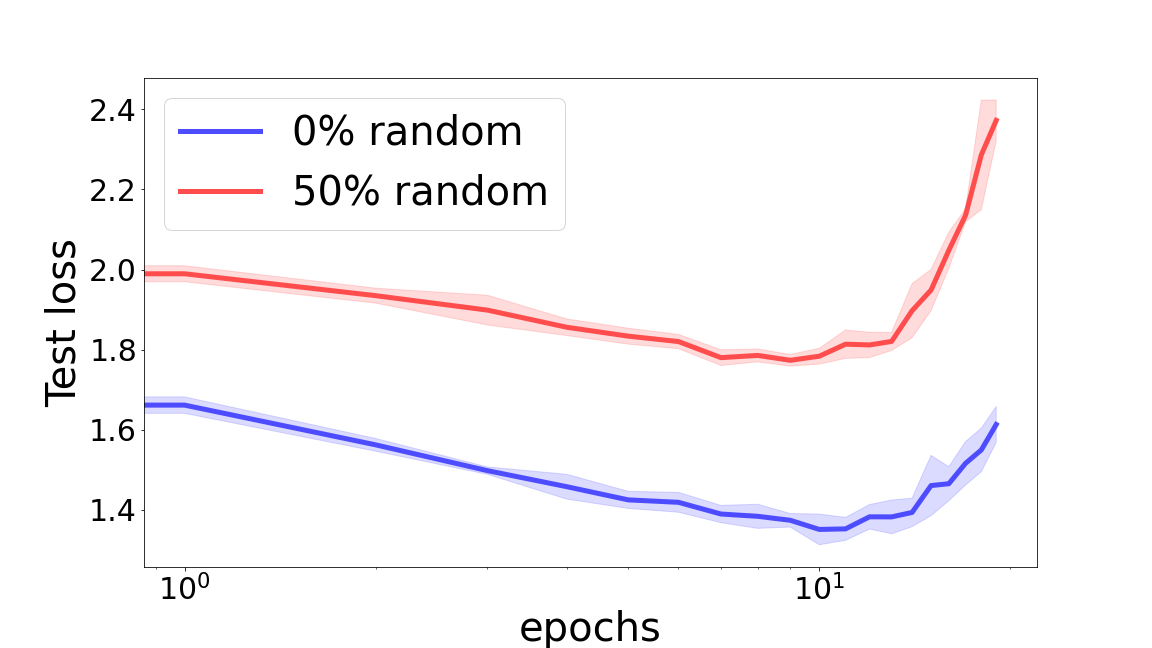}
			\includegraphics[width=0.5\textwidth, height=0.28\textwidth]{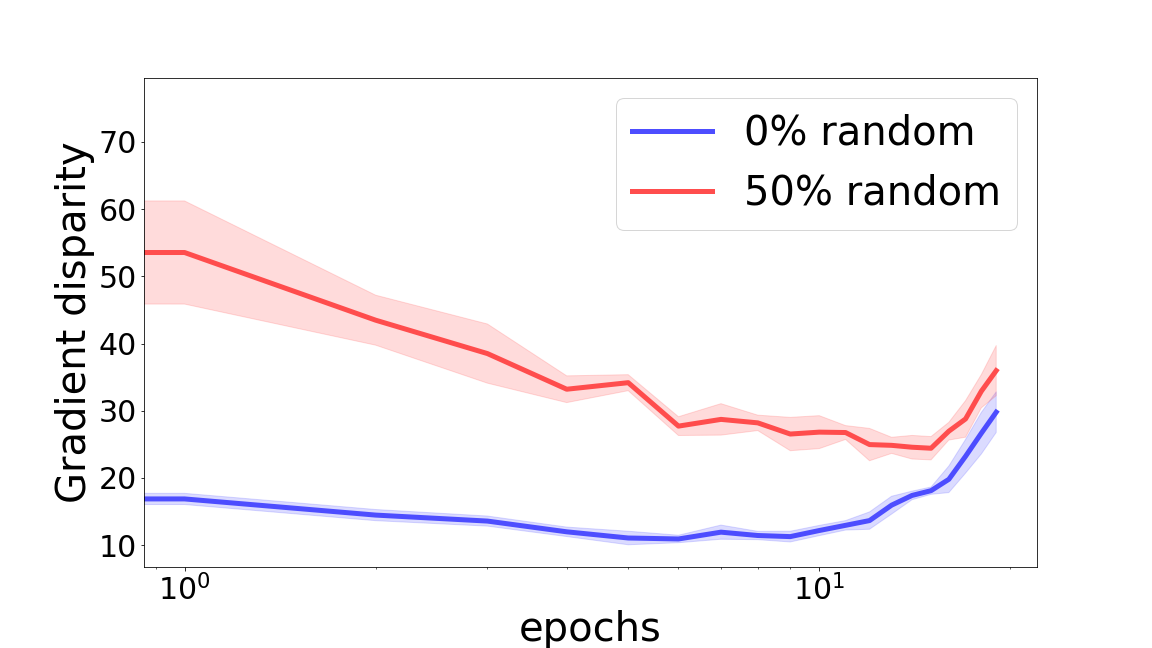}\\
			\includegraphics[width=0.5\textwidth, height=0.28\textwidth]{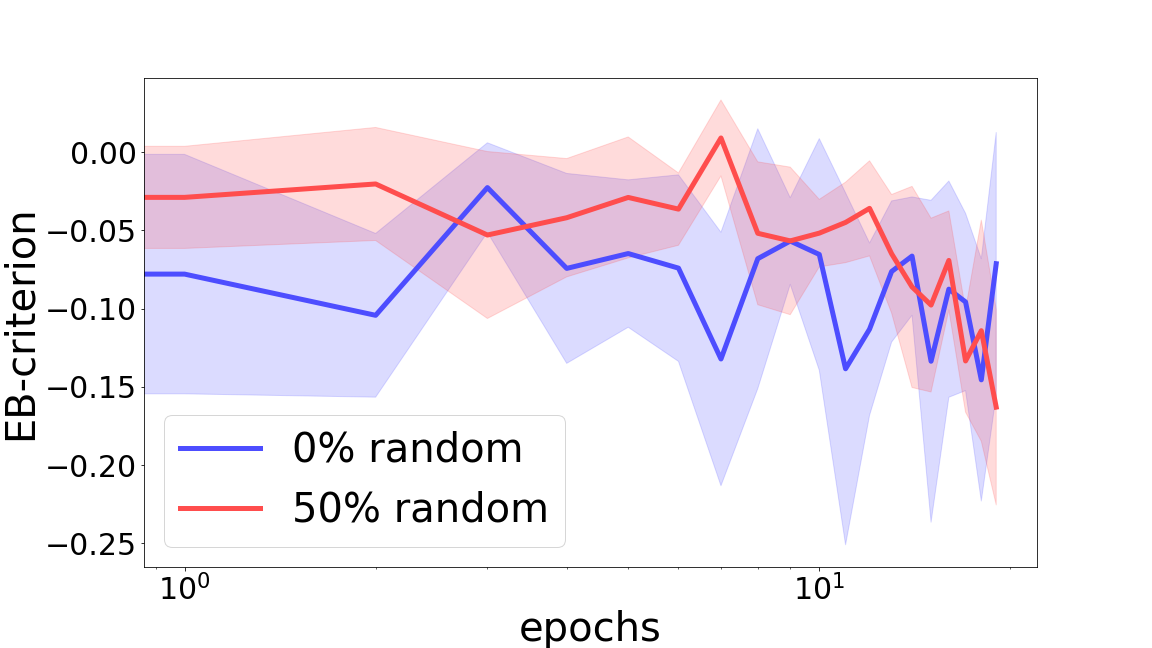}
			\includegraphics[width=0.5\textwidth, height=0.28\textwidth]{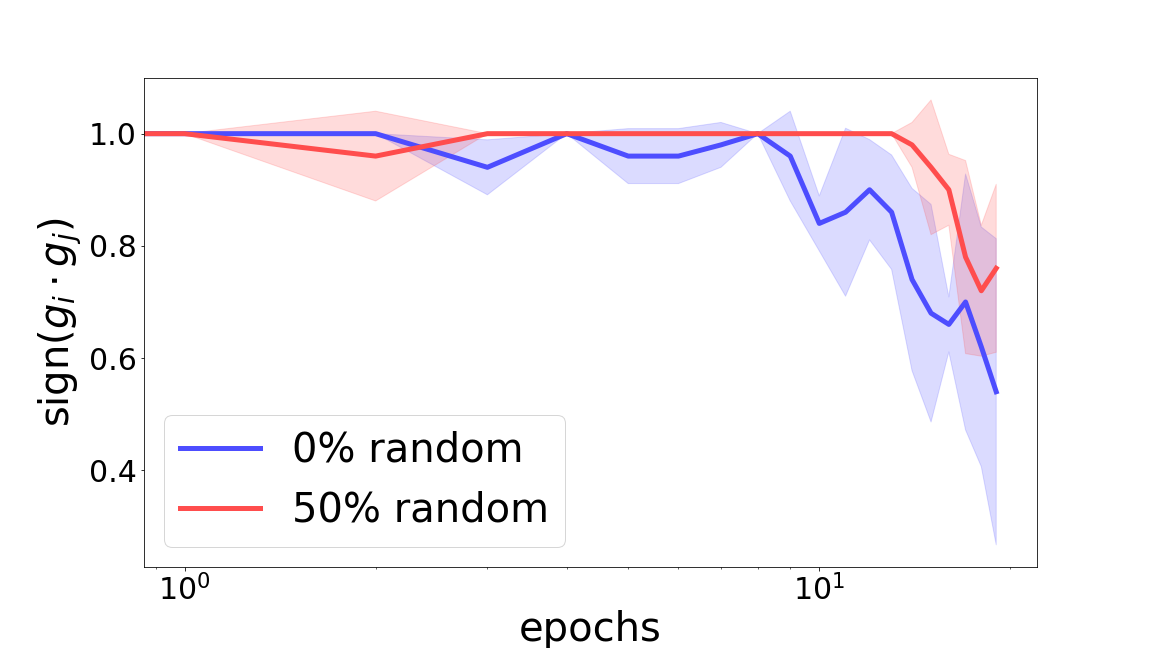}
		\end{subfigure}
		
		\caption{Test loss, gradient disparity, EB-criterion \cite{mahsereci2017early}, and $\text{sign}(g_i\cdot g_j)$ for a ResNet-18 trained on the CIFAR-10 dataset, with $0\%$ and $50\%$ random labels. Gradient disparity, contrary to EB-criterion and $\text{sign}(g_i\cdot g_j)$, clearly distinguishes the setting with correct labels from the setting with random labels.}\label{fig:eb}
	\end{figure*}

\begin{figure*}[h]
		\centering
		\begin{subfigure}[b]{\textwidth}            
			\includegraphics[width=0.33\textwidth, height=0.1856\textwidth]{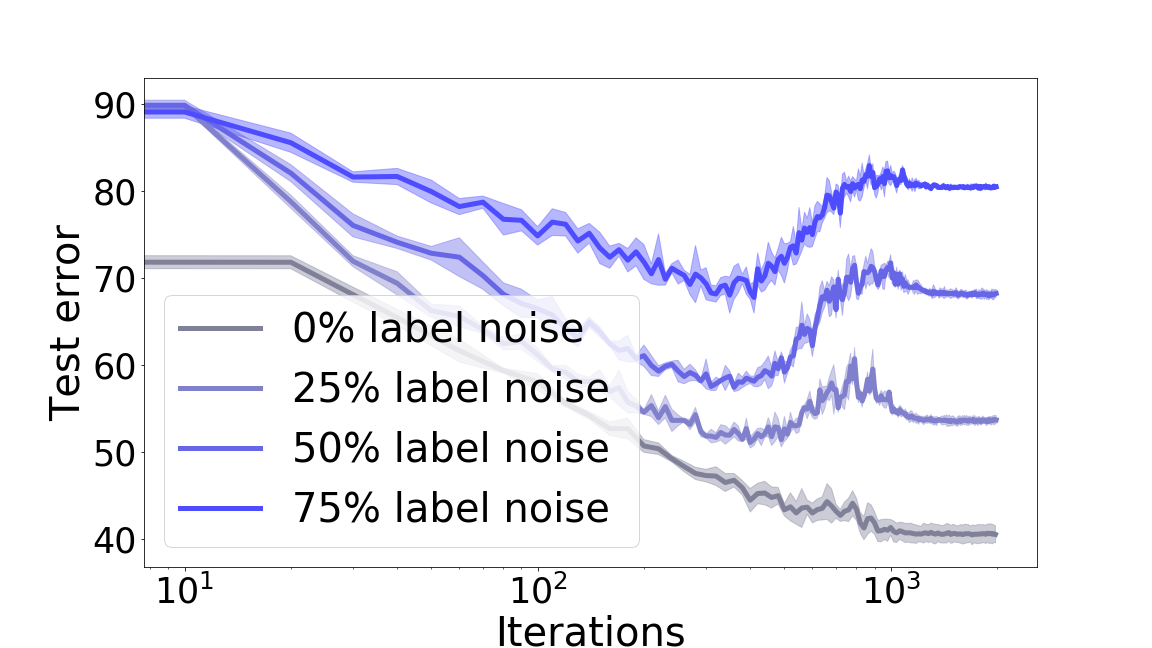}%
			\includegraphics[width=0.33\textwidth, height=0.1856\textwidth]{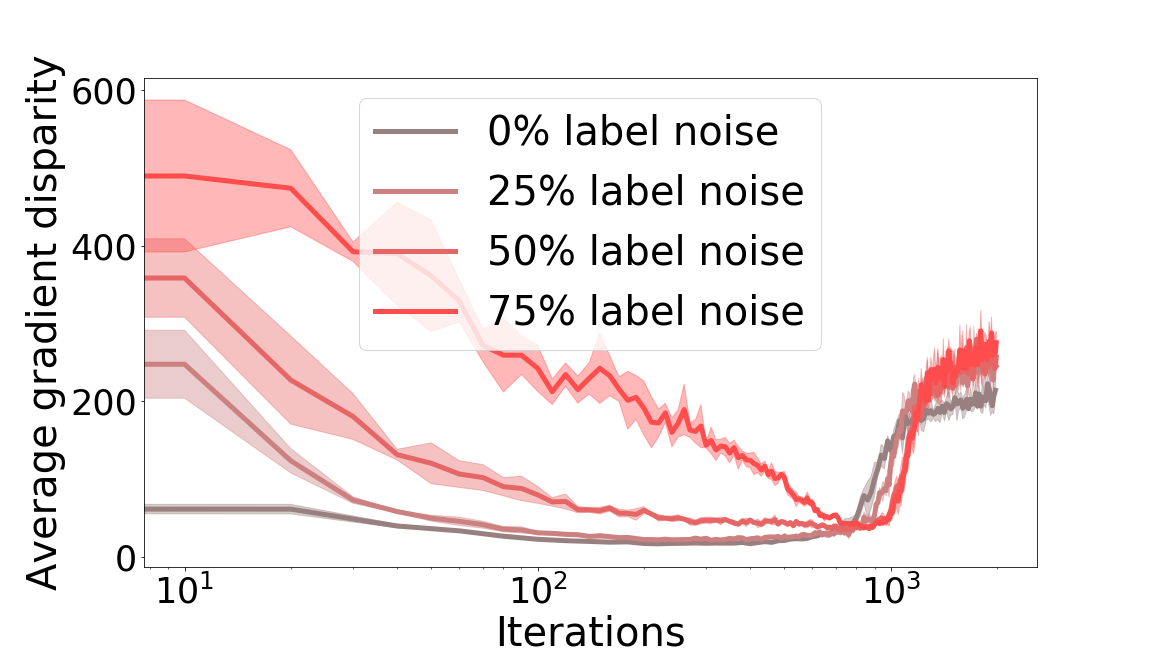}%
			\includegraphics[width=0.33\textwidth, height=0.1856\textwidth]{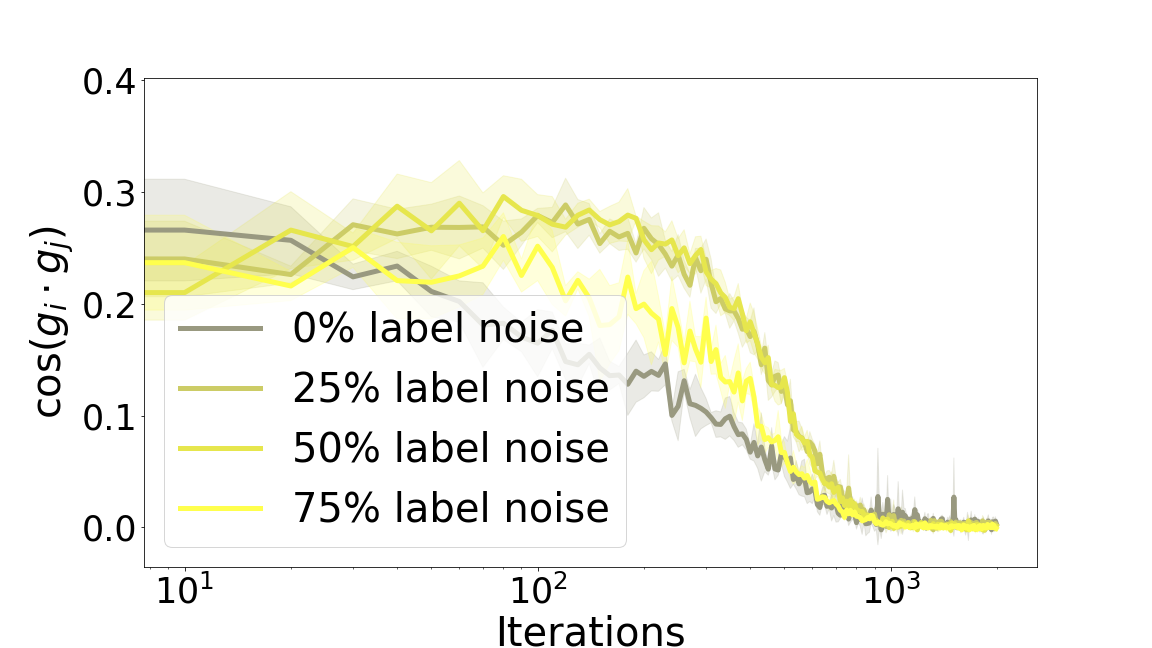}
			\caption{CIFAR-10, ResNet-18}
		\end{subfigure}
		\begin{subfigure}[b]{\textwidth}            
			\includegraphics[width=0.33\textwidth, height=0.1856\textwidth]{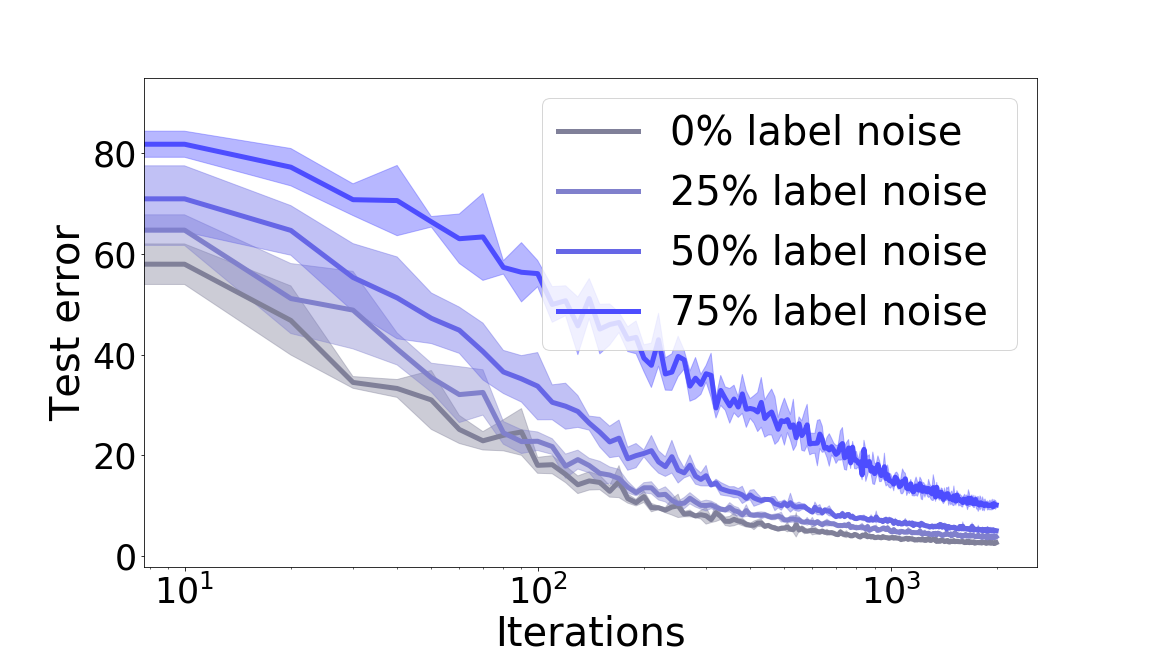}%
			\includegraphics[width=0.33\textwidth, height=0.1856\textwidth]{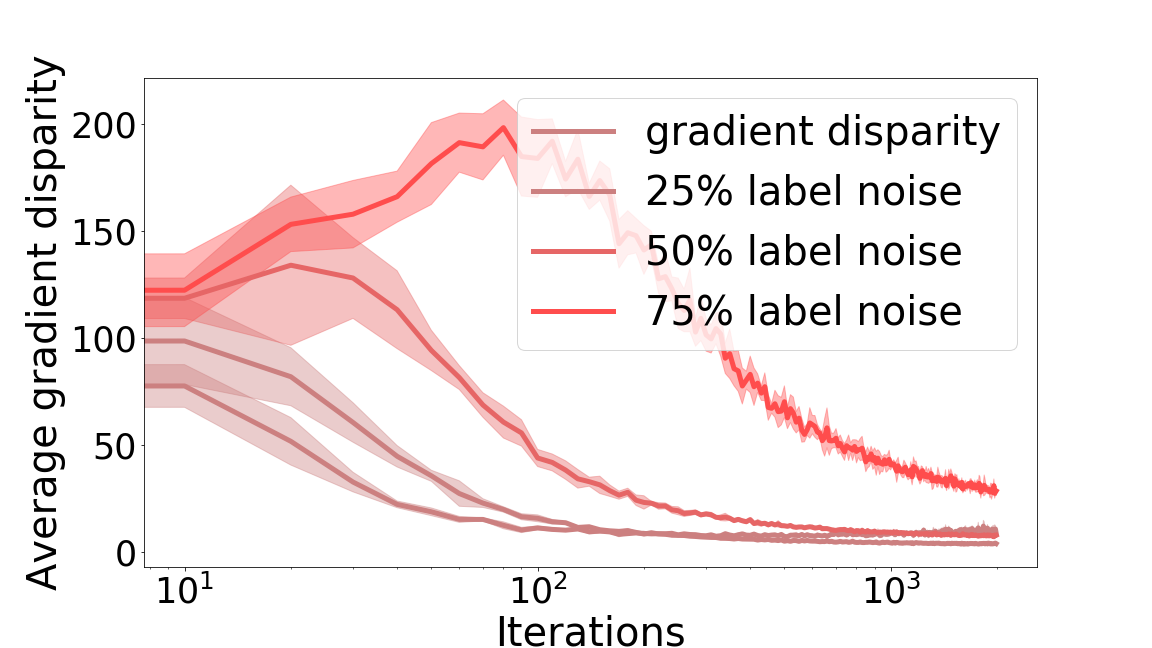}%
			\includegraphics[width=0.33\textwidth, height=0.1856\textwidth]{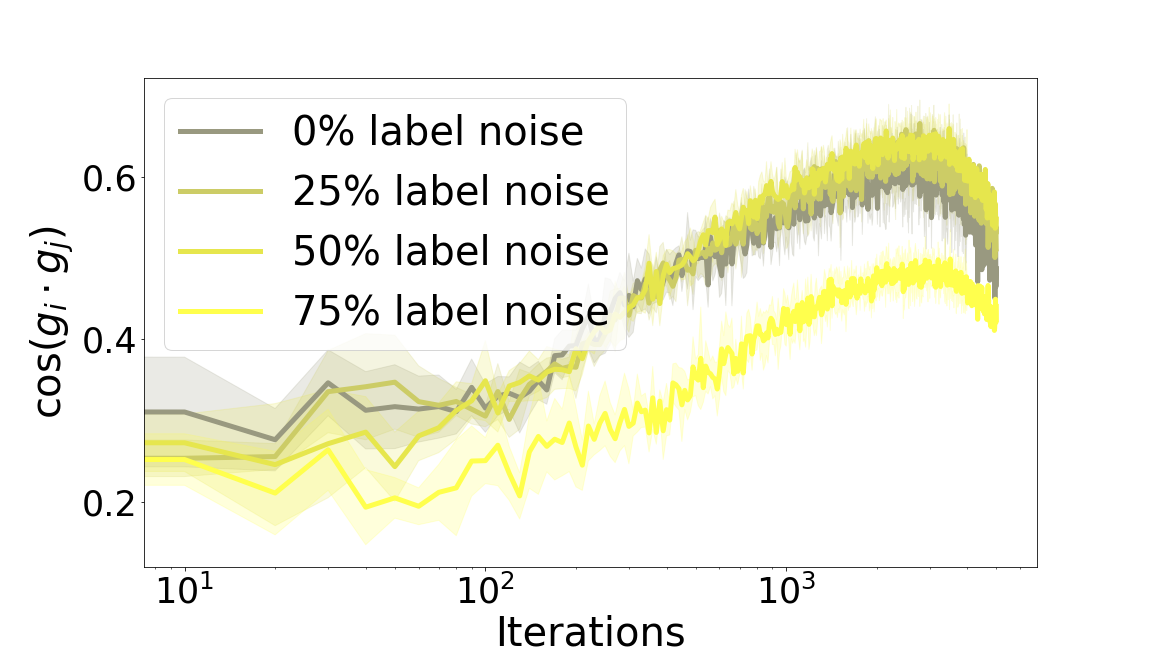}
			\caption{MNIST, AlexNet}
		\end{subfigure}
		
	\caption{The test error (TE), average gradient disparity ($\overline{\mathcal{D}}$), and cosine similarity ($\cos (g_i \cdot g_j)$) during training with different amounts of randomness in the training labels for two sets of experiments. 
		(a) ResNet-18 trained on 12.8k points of the CIFAR-10 training set. The Pearson correlation coefficient between test accuracy (TA) and cosine similarity (cos) over all levels of randomness and over all the iterations is $\rho_{\cos, \text{TA}}= -0.0088$, whereas the correlation between test error/generalization error and gradient disparity is $\rho_{\overline{\mathcal{D}}, \text{TE}}=0.2029$ and $\rho_{\overline{\mathcal{D}}, \text{GE}}=0.5268$, respectively. (b) AlexNet configuration trained on 12.8k points of the MNIST dataset. The correlation between the test accuracy and cosine similarity is $\rho_{\cos, \text{TA}}= 0.7521$, which is positive and relatively high for this experiment. Yet, it is still lower than the correlation between test error and gradient disparity which is $\rho_{\overline{\mathcal{D}}, \text{TE}}=0.8019$.}\label{fig:inner_prod}
\end{figure*}

	\begin{itemize}
		\item The sign of the gradient inner product, $\text{sign}(g_i\cdot g_j)$, should be inversely related to the test loss; it should decrease when overfitting increases. However, we observe that the value of $\text{sign}(g_i\cdot g_j)$ is larger for the setting with the higher label noise level; it incorrectly detects the setting with the higher label noise level as the setting with the better generalization performance (see Fig.~\ref{fig:eb}).
		\item The EB-criterion should be larger for settings with more overfitting. 
		In most stages of training, the EB-criterion does not distinguish between settings with different label noise levels, contrary to gradient disparity (see Fig.~\ref{fig:eb}). At the end of the training, the EB-criterion even mistakenly signals the setting with the higher label noise level as the setting with the better generalization performance.
		\item The cosine similarity between gradient vectors, $\cos (g_i \cdot g_j)$, should decrease when overfitting increases and therefore with the level of label noise in the training data. But $\cos (g_i \cdot g_j)$ does not appear to be sensitive to the label noise level, and in some cases (Fig.~\ref{fig:inner_prod} (a)) it even increases with the noise level. Gradient disparity is much more informative of the label noise level compared to cosine similarity and the correlation between gradient disparity and the test error is larger than the correlation between cosine similarity and the test accuracy (see Fig.~\ref{fig:inner_prod}).
	\end{itemize}
	
	\subsection{Gradient Disparity versus Variance of Gradients}\label{app:var2}
	
	It has been shown that generalization is related to gradient alignment experimentally in \cite{fort2019stiffness}, and to variance of gradients theoretically in \cite{negrea2019information}. 
	Gradient disparity can be viewed as bringing the two together. Indeed, one can check that ${\mathbb{E}\left[\mathcal{D}_{i,j}^2\right] = 2 \sigma_g^2 + 2 \mu_g^T \mu_g - 2 \mathbb{E} \left[g_i^Tg_j\right] }$, given that ${\mu_g = \mathbb{E}[g_i] = \mathbb{E}[g_j]}$ and ${\sigma_g^2 = \text{tr}\left(\text{Cov}\left[g_i\right]\right) = \text{tr}\left(\text{Cov}\left[g_j\right]\right)}$. 
	This shows that gradient variance $\sigma_g^2$ and gradient alignment $g_i^Tg_j$ both appear as components of gradient disparity. 
	We conjecture that the dominant term in gradient disparity is the variance of gradients, hence as early stopping criteria these two metrics almost always signal overfitting simultaneously. This is indeed what our experiments show; we show that variance of gradients is also a very promising early stopping criterion (Table~\ref{tab:RW}).
	However, because of the additional term in gradient disparity (the gradients inner product), gradient disparity emphasizes the alignment or misalignment of the gradient vectors. This could be the reason why gradient disparity in general outperforms variance of gradients in tracking the value of the generalization loss; the positive correlation between gradient disparity and the test loss is often larger than the positive correlation between variance of gradients and the test loss (Table~\ref{tab:var}). 
	\begin{table}[h]
		
		\centering
		\caption{Pearson's correlation coefficient between gradient disparity ($\overline{\mathcal{D}}$) and test loss (TL) over the training iterations is compared to the correlation between variance of gradients (Var) and test loss.}\label{tab:var}
		\vspace*{1em}
			\begin{tabular}{l|l|l}
				\toprule
				Setting & $\rho_{\overline{\mathcal{D}}, \text{TL}}$    & $\rho_{\text{Var}, \text{TL}}$ \\
				\midrule
				AlexNet, MNIST & $\mathbf{0.433}$  & $0.169$ \\
				\midrule
				AlexNet, MNIST, $50\%$ random labels & $\mathbf{0.535}$ & $0.161$ \\
				\midrule
				VGG-16, CIFAR-10 & $0.190$ & $\mathbf{0.324}$ \\
				\midrule
				VGG-16, CIFAR-10, $50\%$ random labels & $\mathbf{0.634}$ & $0.623$ \\
				\midrule
				VGG-19, CIFAR-10 & $\mathbf{0.685}$ & $0.508$ \\
				\midrule
				VGG-19, CIFAR-10, $50\%$ random labels & $\mathbf{0.748}$ & $0.735$ \\
				\midrule
				ResNet-18, CIFAR-10 & $\mathbf{0.975}$ & $0.958$ \\
				\midrule
				ResNet-18, CIFAR-10, $50\%$ random labels & $\mathbf{0.471}$ & $0.457$ \\
				\bottomrule
		\end{tabular}
	\end{table}
	
